\documentclass[twocolumn,twoside]{IEEEtran}    

\usepackage{amsfonts}
\usepackage{amssymb}
\usepackage{amsbsy} 
\usepackage{amsmath}
\usepackage{subfigure}
\usepackage[dvips]{graphicx}
\usepackage[T1]{fontenc}
\usepackage[latin1]{inputenc}
\usepackage{color}
\usepackage{array}
\usepackage{multirow}

\usepackage{hyperref}
\hypersetup{
    bookmarks=true,         
    unicode=false,          
    pdftoolbar=true,        
    pdfmenubar=true,        
    pdffitwindow=false,     
    pdfstartview={FitH},    
    pdftitle={Landmarks},   
    pdfauthor={Perez-Suay, A.},   
    pdfsubject={Landmarks},       
    pdfcreator={Perez-Suay, A.},  
    pdfproducer={Perez-Suay, A.}, 
    pdfkeywords={keyword1} {key2} {key3}, 
    pdfnewwindow=true,     
    colorlinks=true,       
    linkcolor=blue,        
    citecolor=blue,        
    filecolor=blue,        
    urlcolor=blue          
}

\usepackage{multicol}
\usepackage{float}
\usepackage{lscape}
\usepackage{cite}
\usepackage{tikz}
\usetikzlibrary{arrows}

\newcommand{\red}[1]{\textcolor[rgb]{0,0,0}{#1}}          
        
\newcommand{\blue}[1]{\textcolor[rgb]{0,0,0}{#1}} 
 
\definecolor{mycolor}{rgb}{0.1, 0.5, 0.2}

\begin{document}

\title{Pattern Recognition Scheme for Large-Scale\\
Cloud Detection over Landmarks}

\author{Adri{\'a}n P{\'e}rez-Suay, Julia Amor{\'o}s-L{\'o}pez, Luis G{\'o}mez-Chova, \\
Jordi Mu{\~n}oz-Mar{\'i}, Dieter Just, Gustau Camps-Valls
\thanks{APS, JAL, LGC, JMM and GCV are with the Image Processing Laboratory (IPL). Universitat de Val{\`{e}}ncia. C/ Catedr{\'a}tico Escardino, Paterna (Val{\`e}ncia) Spain. 
Web: http://isp.uv.es. E-mail: \{adrian.perez,jucaralo,jordi,chovago,gcamps\}@uv.es.\newline
DJ is with EUMETSAT, Darmstadt, Germany.}
\thanks{This paper has been partially supported by the Spanish Ministry of Economy and Competitiveness (MINECO-ERDF, TIN2015-64210-R and TEC2016-77741-R projects) and by the EUMETSAT Contract No. EUM/RSP/SOW/14/762293 and by the European Research Council (ERC) under the ERC-CoG-2014 SEDAL under grant agreement 647423.}
}

\maketitle


\begin{abstract} 
Landmark recognition and matching is a critical step in many Image Navigation and Registration (INR) models for geostationary satellite services, as well as to maintain the geometric quality assessment (GQA) in the instrument data processing chain of Earth observation satellites. Matching the landmark accurately is of paramount relevance, and the process can be strongly impacted by the cloud contamination of a given landmark. This paper introduces a complete pattern recognition methodology able to detect the presence of clouds over landmarks using Meteosat Second Generation (MSG) data. The methodology is based on the ensemble combination of dedicated support vector machines (SVMs) dependent on the particular landmark and illumination conditions. \blue{This divide-and-conquer strategy is motivated by the data complexity and follows a physically-based strategy that considers variability both in seasonality and illumination conditions along the day to split observations. In addition, it allows training the classification scheme with millions of samples at an affordable computational costs.} The image archive was composed of 200 landmark test sites with near 7 million multispectral images that correspond to MSG acquisitions during 2010. Results are analyzed in terms of cloud detection accuracy and computational cost. We provide illustrative source code and a portion of the huge training data to the community.
\end{abstract}


\section{Introduction}\label{sec:intro}

The Meteosat Second Generation (MSG) satellites constitute a fundamental tool for Remote Sensing (RS) in general and weather forecasting in particular. Actually its usage has important implications on effective agriculture, industry and transportation~\cite{ESAMSG99}. The advanced data and images provided by the Meteosat series span a wide range of applications: from the above mentioned weather forecasting to applications in hydrology, agriculture, environmental studies as well as risk prevention and disaster warnings. The data collected by the MSG constellation are routinely used for the study of meteorology and climate change.

Even though MSG contains several major improvements with respect to the first generation in terms of performance, there are important and critical steps before deploying high-quality data products\footnote{A `quality product' can be defined as one that meets the customers' requirements, in terms of performance, reliability, durability and usability.}. One important bottleneck is assessing the image geometric quality of the data. In order to assess such quality, the Image-Quality Ground Support Equipment (IQGSE) for MSG was developed: the IQGSE is a computer system for the processing and quality measurement of MSG images. The IQGSE is thus used for two different purposes: first, to qualify on-ground the geometric image-quality performance of the MSG satellite system, and second, to verify in flight the geometric image-quality performance of the MSG satellite system during the commissioning phase and other periods of the satellite's design lifetime~\cite{Hanson03}. The IQGSE architecture operates the pre-processing, the navigation filter, the image rectification, and the landmark processing function.

\begin{figure}[t!]
\centerline{\includegraphics[width=8.6cm]{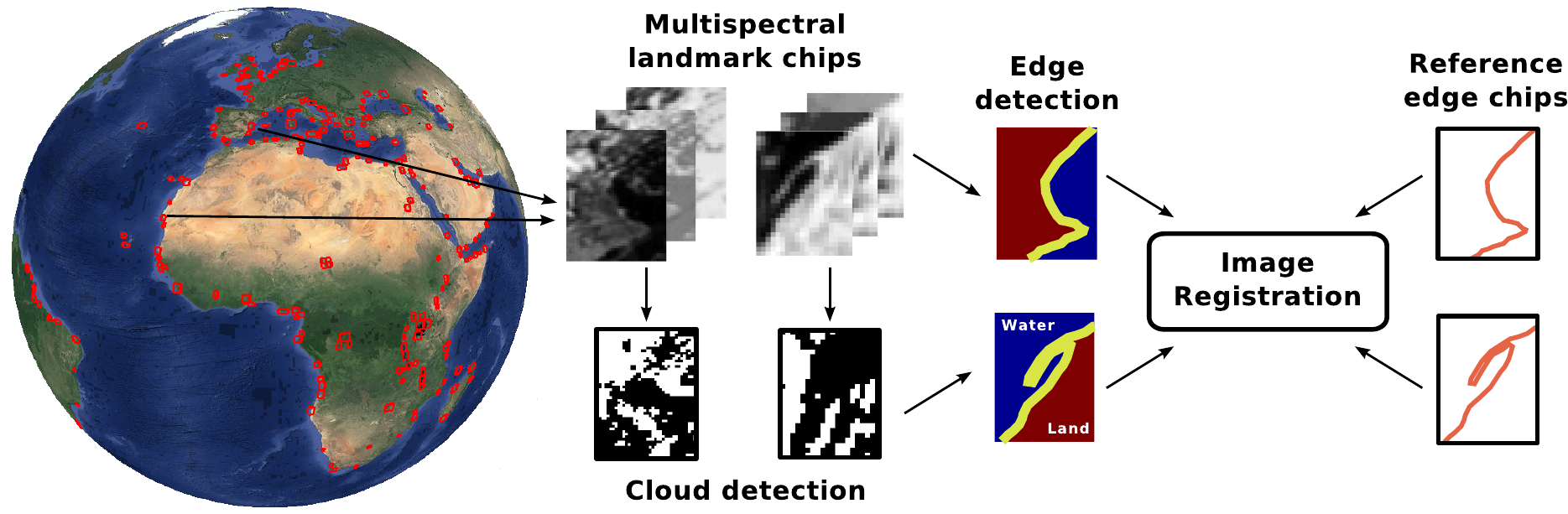}}
\vspace{-0.25cm}
\caption{Landmarks are essential in image registration and geometric quality assessment. Matching the landmark accurately is crucial, and this process can be strongly impacted if the landmark is contaminated by clouds.}
\label{fig:motivation}
\end{figure}

In this paper, we are concerned about the improvement of this last step of landmark processing. Actually, landmark recognition and matching is a critical step in Image Navigation and Registration (INR) models, as well as to maintain the geometric quality assessment (GQA) in the instrument data processing~\cite{Hanson04}. Matching the landmark accurately is of paramount relevance, and the process can be strongly impacted by the undetected cloud contamination of a landmark (see Fig.~\ref{fig:motivation}). This paper presents a general pattern recognition scheme for cloud detection over landmarks. We will pursue the implementation of automatic algorithms able to detect presence of clouds over landmarks, which will be rooted on machine learning, tailored to this particular RS application.

\begin{table*}[t!]
\begin{center}
\caption{Variables stored in the database for each landmark chip (e.g. landmark 0, Ad Dakhla, Morocco).} \label{tab:DBchip} 
\begin{tabular}{l l l l l}
\hline
Name & Size & Bytes & Class & Description\\
\hline
H & e.g. $78\times60$ & 37440 & double & High resolution visible SEVIRI channel \\
X & e.g. $26\times20\times11$ & 45760 & double & Hypercube with optical and thermal SEVIRI channels \\
M & e.g. $26\times20$ & 4160 & double & MSG level 2 cloud mask \\          
channels & $1\times13$ & 104 & double & SEVIRI channel numbers  \\
id & $1\times1$ & 8 & double    & EUMETSAT landmark identifier \\  
name & $1\times9$ & 18 & char   & Name assigned to the landmark geographic location \\
num & $1\times1$ & 8 & double & Sequential landmark number (0-199) \\
centre & $1\times2$ & 16 & double & Pixel coordinates of the landmark center \\
latlon & $1\times2$ & 16 & double & Geographic coordinates of the landmark center \\    
time & $1\times14$ & 28 & char & UTC acquisition time (format: YYYYMMDDhhmmss)\\
\hline
\end{tabular}
\end{center}
\end{table*}

In a wide range of RS applications, accurate and automatic detection of clouds in satellite images is a key issue. In particular, with no accurate cloud masking, undetected clouds are one of the most significant sources of error when using landmarks for accurately maintaining INR models and GQA. In the literature, cloud-detection approaches are generally based on the assumption that clouds show some features that can be used for their identification: clouds are usually brighter and colder than the underlying surface, the spectral response is different from that of the surface covers, and cloud height produces a shorter optical path thus lowering atmospheric absorption~\cite{GomezChovaTGARS07}.

The simplest approach to cloud detection in a scene is the use of a set of static thresholds (e.g. over radiance or brightness temperature) applied over pixels or over second-order moments of a patch/region. This strategy has been followed in both multispectral and hyperspectral sensors~\cite{Zhu12,Zhu15, Hagolle10}, and it is the current approach for MSG~\cite{Derrien05,Hocking11}. These procedures cast the problem as a binary classification task and provide a binary cloud label for the presence or absence of cloud. 
The problem of cloud identification over landmarks is also a binary one: classifiers only try to detect the presence or absence of clouds. While one could follow simple threshold-based strategies for detection as well, it is generally acknowledged that the problem is nonlinear in the representation space~\cite{GomezChovaTGARS07,Mei17}. 
Many machine learning approaches have been introduced for this problem: from neural networks~\cite{tian99,torres-arriaza03}, to kernel methods under supervised or semisupervised settings~\cite{GomezChovaGRSL08,GomezChovaTGARS10}, change detection schemes ~\cite{CampsVallsTGARS08,GomezChovaIGARSS13kcda}, and maximum entropy principles of image representation~\cite{GomezChova11ORSbook, Romero16deepcnn}.  
Last studies try to exploit also the multitemporal domain to detect clouds~\cite{Melgani06,Hagolle10,Zhu14,Cerra16,GomezChovaJARS17,Chen17}. In all these approaches, however, illustration is limited to very few images being non-operational. 
The previous machine learning approaches have been, however, rarely exploited in real-life large-scale scenarios of cloud detection. 
With the growing availability of massive and complex remote sensing data, developing automatic, robust and scalable cloud detection classifiers for real-time processing is an urgent need. This problem imposes some specific questions when it comes to MSG SEVIRI, which acquires the Earth full disk (3712$\times$3712 pixels) in 12 spectral channels every 15 minutes. The requirements of the problem imply that this preliminary cloud detection should be only done at a fixed number of 200 landmarks positions of variable chip size. In order to develop accurate and robust classifiers, we use data covering all seasonal and possible illumination variations. The problem thus gives rise to a very large data volume. Summarizing, 
the objectives of this work are: (1) Develop an automatic machine learning based scheme for the classification of clouds over landmarks; and (2) Study the robustness and accuracy of the classification scheme, as well as to assess its complexity, scalability, and parallelization possibilities. 

This paper introduces a complete pattern recognition processing chain able to detect the presence of clouds over landmarks using MSG SEVIRI data from the whole 2010 year. The methodology is based on an ensemble combination of dedicated support vector machines (SVMs), which are configured in a divide-and-conquer classification strategy: we develop specific SVMs per landmark, illumination conditions, and land covers. This strategy allows to train statistical classifiers with millions of examples at computationally affordable times. The archive was composed of 200 landmarks with more than 7 million of multispectral MSG image chips for training, corresponding to MSG acquisitions during 2010. This real problem, and hence the proposed scheme for cloud detection/classification over landmarks, will require intensive data pre-processing and characterization, perform a proper feature extraction of the existing landmarks, evaluate classifier combination strategies over land and sea, as well as the evaluation of state-of-the-art non-linear classification techniques. 

The remainder of the paper is organized as follows. Section~\ref{sec:archive} describes the landmark archive used in this paper and the specifics for pre-processing and data characterization. Section~\ref{sec:scheme} introduces the proposed pattern recognition scheme, paying special attention to the techniques used for feature extraction (both statistical and physically-based) and the proposed classifiers. Section~\ref{sec:results} shows the experimental results, and evaluates the proposed scheme in terms of accuracy, computational cost, robustness and scalability. Finally, we conclude in Section~\ref{sec:conclusions} with some final remarks and further work.

\section{Landmarks Archive}\label{sec:archive}

The landmark archive provided by EUMETSAT contains MSG-SEVIRI level 1.5 acquisitions for 200 landmarks of variable size for the whole year 2010, which are mainly located over the coastline, islands, or inland waters. Acquisition frequency is every 15 minutes which produces 96 images (full disk) per day, which resulted in 35040 images (or chips) per landmark in 2010. Additionally, the MSG Level 2 cloud products were used as 'ground truth' for each landmark observation.

Each landmark chip is stored in the database with the variables showed in Table \ref{tab:DBchip}, which includes the optical and thermal MSG channels, the MSG level 2 mask, and some additional information about the landmark location and acquisition time.
Furthermore, in order to take into account the daily and seasonal variability and the different day/night conditions, the solar elevation must be known for each acquisition. The sun position (zenith and azimuth angles at the landmark location) was computed for the center of the landmark as a function of the landmark local time and geographic position~\cite{Reda03}. The sun zenith angle (SZA) is also stored for each landmark chip and allows us to easily select the landmark chips to be analyzed attending to their solar illumination or to their actual local time acquisition.

\subsection{Pre-processing: Conversion from radiance to reflectance and brightness temperature} 

The provided MSG SEVIRI Level 1.5 images are corrected for non-linearities in the detector response, for differences in the detector response within a given channel, and are represented as 10 bit data with no physical units but a direct relation to the corresponding radiometric units. 
The conversion between binary counts and physical radiance is defined by two linear scaling parameters in the image header (slope and offset) and the radiance for each spectral band is obtained as: \\Physical Units = offset + (slope $\times$ Level 1.5 Pixel Count) expressed in $mWm^{-2}sr^{-1}(cm^{-1})^{-1}$. It is worth noting that the slope and offset are fixed scaling factors that will normally not change and that are constant for all the analyzed 2010 dataset of MSG-2 (Meteosat 9) data~\cite{EUMETSATMSGLevel15}. 

Once we have the inputs in physical units, the conversion from radiance to top of atmosphere (TOA) reflectance for the SEVIRI reflective bands (VIS0.6, VIS0.8, NIR1.6, and HRV) is carried out. The set of solar irradiance values to be used to perform this conversion are provided for SEVIRI on-board MSG and the conversion to TOA reflectance (with no BRDF correction) is done as described in~\cite{EUMESATRad2Ref}. 
Analogously, the observed effective radiance in thermal bands is converted to equivalent brightness temperature in Kelvin as described in~\cite{EUMESATRad2BT}. 

\subsection{Database characterization}

\subsubsection{MSG Level 2 cloud mask}

The L2 cloud mask is used for the generation of the training and test sets, so it is interesting to have an estimation of the amount of cloudy pixels and wrong values present for each chip.
The pixel values of the L2 cloud mask provided for each chip are coded as: `0': space/no data, `50': water, `100': land, and `200': cloud. 
Additionally, from the MSG L2 cloud mask we also generate a land-cover (land-water) classification mask {\em per} landmark by computing the max-vote of the mask of all cloud-free landmarks. 
Next, we use the class `land' or `water' of each pixel to get a general land-cover mask by landmark. This land/water mask by landmark is also used to calculate a coastline mask by growing the coastline to the adjacent pixels with a $3\times3$ spatial neighborhood.

We have analyzed basic statistics and found two different problems with the L2 cloud mask: some landmark masks present space/no-data values and/or false positives over coastlines. These possibly wrong labels force us to implement an algorithm to detect such cases, and once they are detected, these misleading samples are removed from the training set.
Additionally, we detected two landmarks with different L2 cloud mask codification and both were discarded from the analysis (LM \#91 and \#98).


\subsubsection{MSG Level 1.5 data}

The MSG level 1.5 radiance data is analyzed in order to account for two main effects: the daily/seasonal cycle and the separability among classes in the spectral input space. 

The seasonal cycle is analyzed looking at the time evolution averaged over regions corresponding to each land cover class. The averaged radiance temporal profiles are computed for each image taking into account the land cover class of each acquisition (MSG L2 mask) and the static land-water mask of the landmark, which allows us to distinguish between clouds over ocean and land. Results show that radiances follow a clear daily cycle, especially noticeable for the visible range over land, while the thermal bands present a more complex and noisy time series (mostly due to undetected clouds in the MSG L2 mask).

The SEVIRI instrument also includes a High Resolution Visible (HRV) channel, which covers roughly half of the full disk image and is changing its coverage throughout the day. Statistical results shows that most of the landmarks present a huge amount of chips without usable HRV information. Therefore, the HRV band was finally discarded as input for classification.

\section{Proposed Scheme/Method}\label{sec:scheme}

\begin{figure}[t]
\centerline{\includegraphics[width=8.6cm]{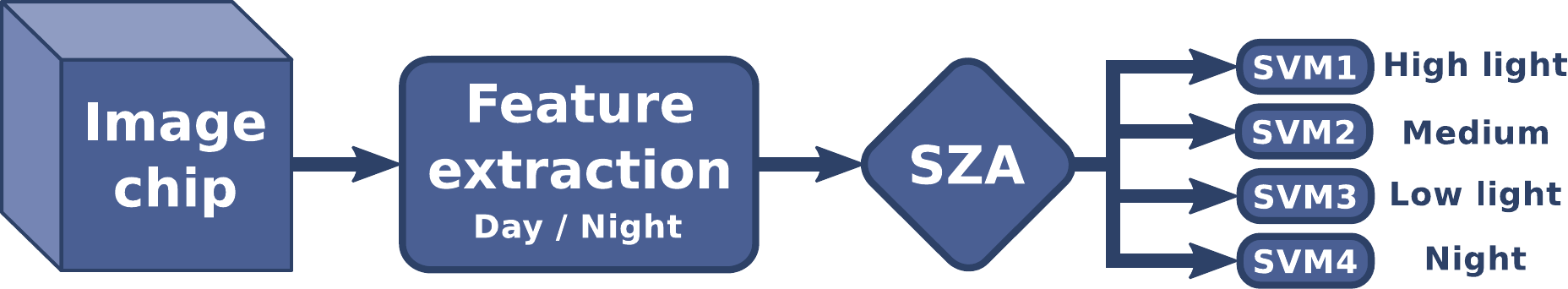}}
\vspace{-0.25cm}
\caption{Proposed scheme of a classification system for cloud detection over landmarks depending on the time-of-day.} 
\label{fig:scheme}
\end{figure}

Cloud detection over landmarks is tackled as a pixel-based classification problem following the requirements of providing a cloud mask prediction for each landmark chip. However, a global cloud detection classifier working for all considered test sites becomes computationally demanding, and usually only a limited number of samples (pixels) could be used for training such classifier. In order to alleviate this problem, we decided to follow a ``divide-and-conquer'' strategy and developed different landmark-specific classifiers, which are trained for specific sub-problems depending on the time-of-day. 
This allows us to reduce the complexity of the classification problem and to exploit the local characteristics of each particular landmark. 
\blue{We would like to remark that this approach meets the requirement of detecting clouds over landmarks. MSG landmark sites are selected attending to geographical features that must be easily recognized from satellite images, such as coastlines, islands, or inland waters, mainly in mid and low latitudes. 
This allows us to avoid the development of global cloud detection algorithms dealing with all critical cloud detection scenarios. For example, at mid latitudes the surface is rarely covered by snow and, in case of having a landmark presenting snow covers, the proposed landmark-specific classifier approach simplifies the cloud detection problem since it exploits the local characteristics of the landmark and adapts well to the particular problems: snow, ice, sand on coastlines, etc}. 
\blue{This way we obtain simple and accurate models at a landmark level}. 
The proposed scheme is shown in Fig.~\ref{fig:scheme} and we review the processing steps in the following sections.

\subsection{Feature Extraction}

Having a good set of features is of crucial importance because this information is directly fed to the classification algorithm. As explained in the previous section, TOA reflectance and brightness temperature are computed from the raw channels acquired by the satellite. 
\blue{Table~\ref{tab:feats} shows the selected features for day and night to build the classifiers. Note that visible channels and derived features from them are discarded at night.} 
\blue{The VIS0.6 and VIS0.8 visible channels help in the discrimination of clouds from the surface due to their higher intensity. The IR3.9, IR8.7, IR10.8 and IR12.0 channels help providing temperature of clouds, land and sea surfaces. The NIR1.6 channel is particularly useful to discriminate between clouds and snow, and the IR3.9 channel also helps to detect fog and very low clouds at night.}  
\blue{In addition, some extra informative features for cloud detection are also extracted, such as a band ratio to enhance clouds, a normalized difference spectral index specifically designed to improve the discrimination of clouds from snow, and the normalized difference vegetation index (NDVI), which helps discarding vegetation covers over land~\cite{Derrien05, Hocking11}.
Finally, contextual information and spatial homogeneity are taken into account by second order spatial statistics (mean and standard deviation) at two different scales (spatial windows of 3$\times$3 and 5$\times$5 pixels). Certainly, non-linear classifiers, such as SVMs, might learn complex feature mappings directly from the MSG-SEVIRI channels when enough labeled data is available. However, including additional features that enhance the separability between classes simplifies the training of the classifiers and usually improves results.}

\begin{table}[t]
\begin{center}
\blue{
\caption{Selected features for pixel-based classification.} \label{tab:feats}
\bgroup
\def\arraystretch{1.2}%
\begin{tabular}{llcc} \hline
Number & Features & Day & Night \\ \hline 
1 & R1 VIS 0.6 $\mu$m & \checkmark & $\times$ \\
2 & R2 VIS 0.8 $\mu$m & \checkmark & $\times$ \\
3 & R3 NIR 1.6 $\mu$m & \checkmark & $\times$ \\
4 & R4 IR 3.9 $\mu$m & \checkmark & \checkmark \\
5 & BT7 IR 8.7 $\mu$m & \checkmark & \checkmark \\
6 & BT9 IR 10.8 $\mu$m & \checkmark & \checkmark \\
7 & BT10 IR 12.0 $\mu$m & \checkmark & \checkmark \\
8 & Cloud Test:~$\frac{R2}{R1}$ & \checkmark & $\times$ \\[1ex]
9 & Snow Test:~$\frac{R1-R3}{R1+R3}$ & \checkmark & $\times$  \\[1ex]
10& NDVI:~$\frac{R2-R1}{R2-R1}$ & \checkmark & $\times$\\[1ex]
11& $mean_{3x3}(\text{R1})$  & \checkmark & $\times$ \\
12& $std_{3x3}(\text{R1})$  & \checkmark & $\times$ \\
13& $mean_{5x5}(\text{R1})$  & \checkmark & $\times$ \\
14& $std_{5x5}(\text{R1})$  & \checkmark & $\times$ \\
15& $mean_{3x3}(\text{BT9})$  & \checkmark & \checkmark \\
16& $std_{3x3}(\text{BT9})$  & \checkmark & \checkmark \\
\hline
\end{tabular}
\egroup
}
\end{center}
\end{table}

\subsection{Divide-and-conquer: splitting the day into four periods according to light conditions}

The problem of cloud detection over landmarks is tackled following a divide-and-conquer strategy.
Among the main advantages of this approach, we should mention: (i) local models are smaller in size, thus faster to train and use in the prediction phase; (ii) local models are accurate in their defined regions, usually better than global models that try to cover all possible situations; (iii) it is straightforward to obtain a parallel implementation; and (iv) local models are often simpler and easier to interpret. Therefore, we follow this strategy to divide the cloud detection problem in different sub-problems and train different classifiers with reduced datasets: specific classifiers per landmark and time-of-day (Fig.~\ref{fig:scheme}). \blue{This approach, besides exploiting the local characteristics of each particular landmark, allows us to deal with this large scale problem and to train our scheme using millions of MSG multispectral image chips.} 

\begin{figure*}[t]
 \begin{center}
  \begin{tabular}{ccc}
  LM0 Ad Dakhla ($lat=23.73$)  & LM83 Scotland ($lat=56.78$) & \small{SZA$_\textit{m}$} vs. Latitude \\
  \includegraphics[width=.35\textwidth]{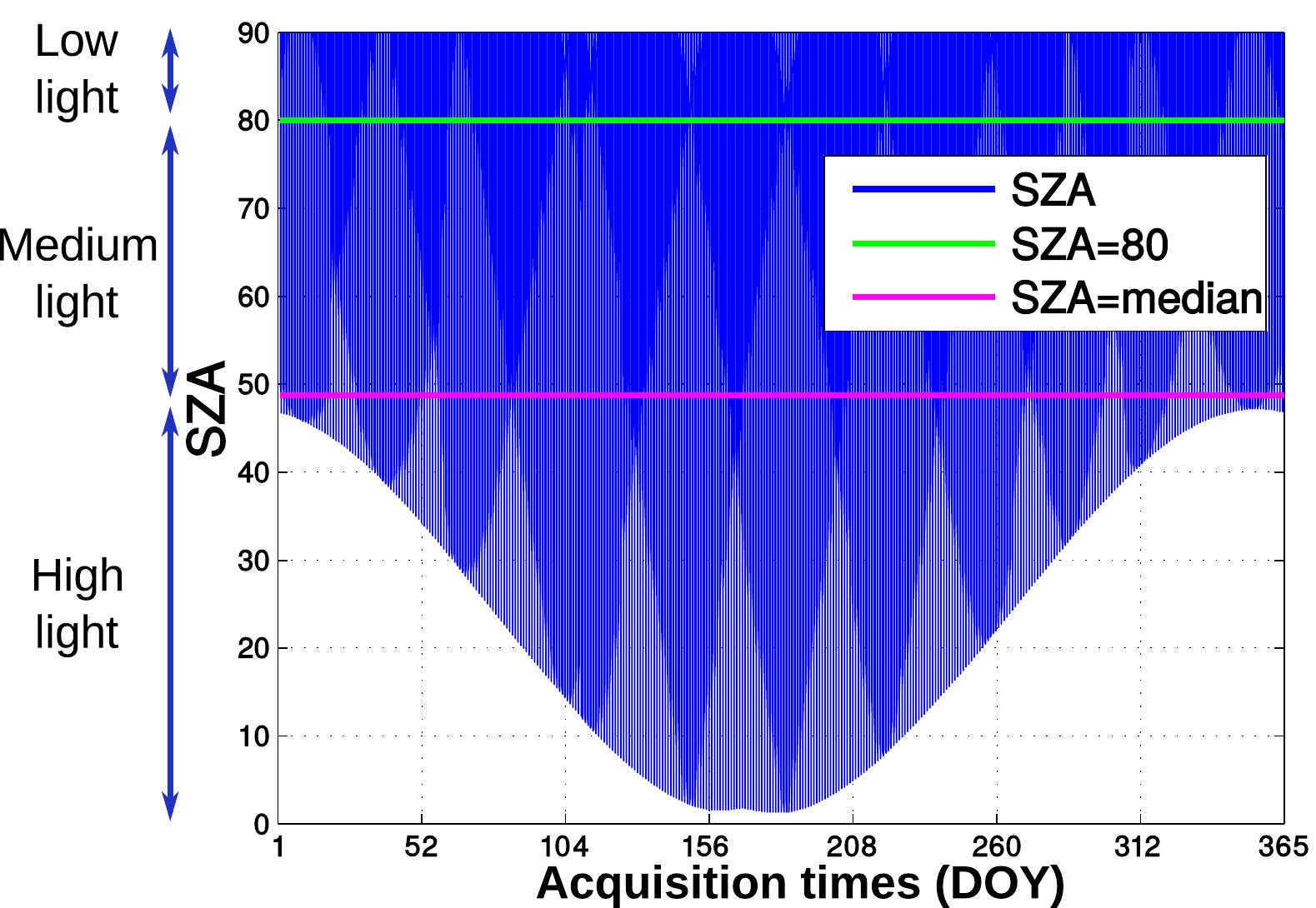} &
\includegraphics[width=.3\textwidth]{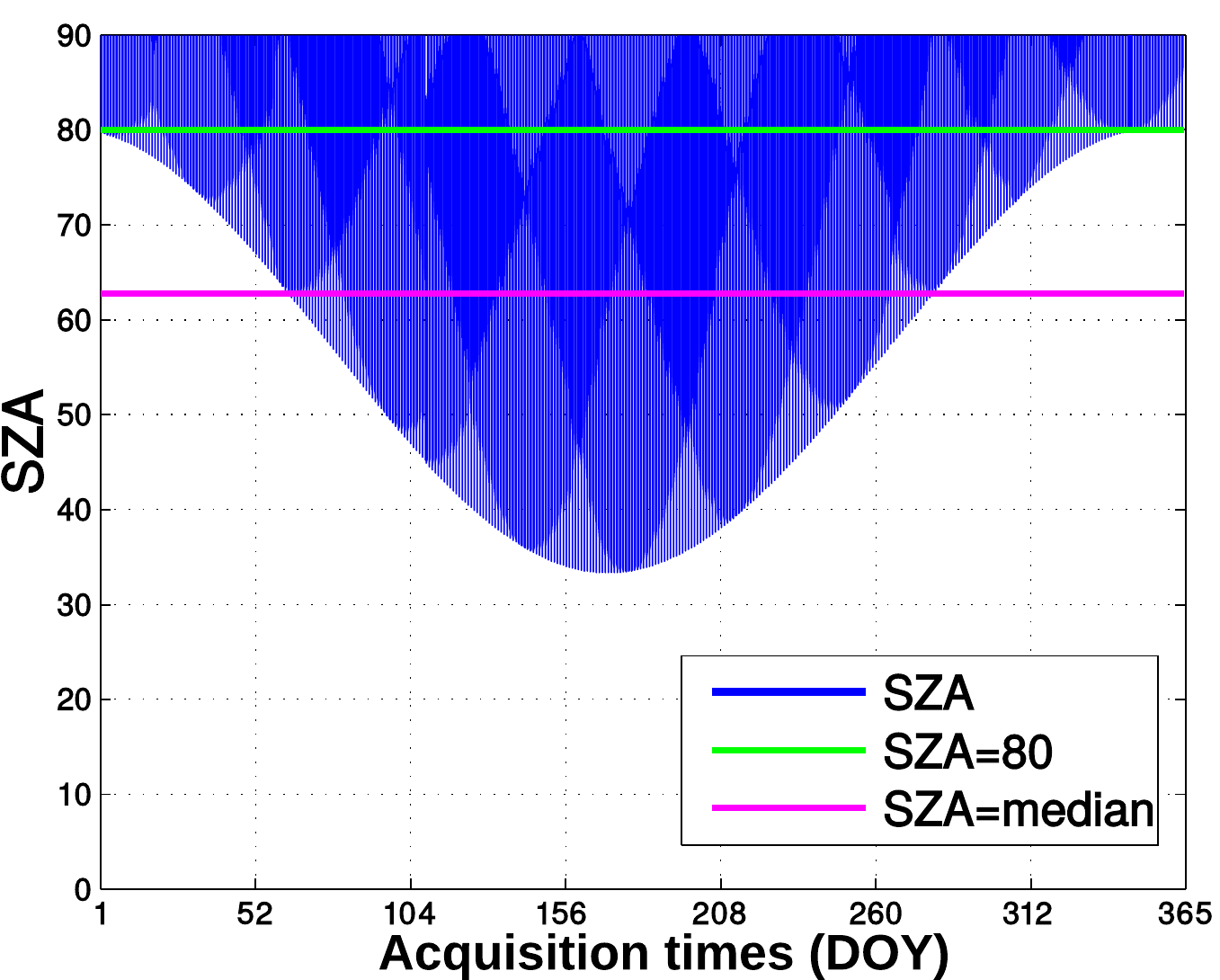} &
\includegraphics[width=.3\textwidth]{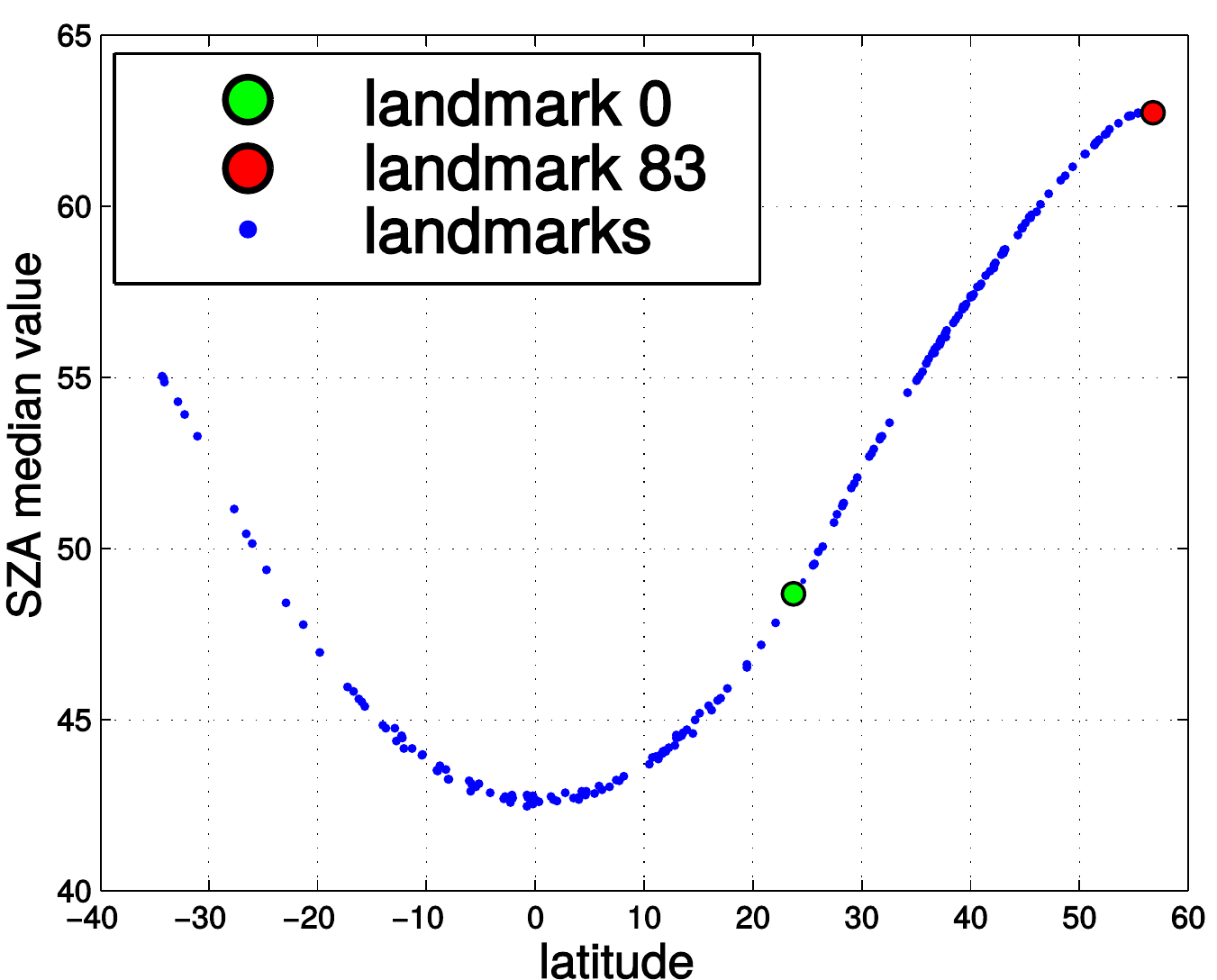} 
  \end{tabular}
\caption{Daytime SZA values (SZA$<$90\textdegree) for landmarks LM0 [left], LM83 [middle] including the different thresholds, and SZA$_\textit{m}$ threshold for all the landmarks vs. their latitude [right]. The landmarks LM0 and LM83 are marked with big dots. \label{fig:SZAranges}}
\end{center} 
\end{figure*}

\blue{The division of the problem to solve into different day periods is based on the study of the different illumination conditions over the landmarks.} \blue{Previous studies~\cite{Derrien05, Hocking11} showed that when SZA is taken into account in the cloud detection process, results are typically improved. However, a trade-off between the number of sub-problems and number of available training samples per sub-problem exists. The proposed approach provides a good compromise between complexity and accuracy by splitting the day in four ranges (sub-problems) according to the following solar zenith angle values}:
\begin{itemize}
 \item high-light conditions (midday sun): SZA$<$SZA$_{m}$
 \item medium-light conditions: SZA$_{m}<$SZA$<80$\textdegree
 \item low-light conditions (sunrise/twilight): 80\textdegree$<$SZA$<$90\textdegree
 \item night: SZA$>$90\textdegree
\end{itemize}
The SZA curve presents a similar trend for all landmarks, but the minimum SZA value (maximum-light condition) for each landmark depends on its latitude (see Fig.~\ref{fig:SZAranges}). \blue{The low-light threshold is fixed to 80\textdegree\ following~\cite{Derrien05} in order to deal with low radiance values in the visible channels which can mislead the classifiers.} The remaining SZA angles from 0\textdegree\ to 80\textdegree\ depend on a threshold, SZA$_{m}$, that is different for each landmark. \blue{This threshold is selected to split all acquired chips in 2010 for this landmark during daytime ($0$\textdegree$<$SZA$<80$\textdegree) in two sets of equal size (50\%), i.e. SZA$_{m}$ is the median value of all SZA bellow $80$\textdegree. It allows us to divide the problem into high and medium light conditions while taking into account the Sun illumination dependency on the landmark latitude. Fig.~\ref{fig:SZAranges} shows clearly the dependence of this threshold on latitude for different landmarks.} 

\subsection{Machine learning classifier} 

During the last decades, machine learning (and the related fields of pattern recognition and statistics) has deeply advanced in developing automatic techniques for data classification. The field of supervised classification is a very active one and many algorithms are currently available, from standard neural networks~\cite{Haykin94}, nearest neighbors~\cite{Altman92}, random forest~\cite{Breiman01,Hastie09} and kernel methods like Support Vector Machines (SVMs)~\cite{Scholkopf02,CampsValls09}. SVMs for example have found a wide application and acceptance in remote sensing data/image classification in general~\cite{CampsValls05,CampsValls09shaker,CampsValls09wiley,CampsValls14} and for cloud detection in particular~\cite{GomezChovaTGARS07,GomezChovaGRSL08,GomezChovaTGARS10,GomezChovaIGARSS13kcda,GomezChovaIGARSS13cci,GomezChovaJARS17}. 
In this particular application, we select the SVM classifier for the cloud detection over landmarks which usually provides a very good results in terms of accuracy and robustness.

The SVM is a non-parametric kernel method that separates the classes fitting an optimal linear hyperplane in a higher dimensional representation (feature) space. To do this, the method maximizes the margin (separation between samples of different classes, which is related to the norm of the classification hyperplane) while minimizing the classification error. This introduces a hyperparameter to be tuned, the so-called regularization, or cost, parameter $C$. Since this classification is done in a projected space, the mapping function needs to be defined. The advantage of SVMs over other nonlinear, yet parametric, classifiers is that the mapping function is defined implicitly. In practice, samples are never mapped explicitly to the feature space. It can be shown that only the similarity between samples in the mapped space is needed to construct the classifier, and this can be actually computed through a similarity function (called kernel function) that takes two samples and returns a scalar. Such kernel function needs to be parametrized, and hence the associated hyperparameters to be tuned. In summary, to obtain an SVM classifier one needs to optimize two parameters: (1) $C$ in order to adjust the level of regularization (prevent overfitting), and (2) kernel function parameters (mapping space dimensionality). SVMs are typically fast to train and apply in moderate sized problems, but slow with many labeled examples (usually more than 10,000 examples).

\subsection{Setup for classification}

The proposed classification scheme is developed with a limited number of training samples for the selected classifiers. 
Hence, sample selection is a critical issue that directly affects the performance of the trained classifiers. We adopted different strategies to alleviate this issue, and also to account for the land-cover types in each landmark. 

We select samples that cover all months/dates with a balanced number of cloud-free and cloudy samples over land, water, and coastline pixels using the L2 cloud mask and the calculated land-cover map for each landmark.
For all the analyzed sub-problems, we split the labeled dataset into two disjoint sets: the so-called training and testing sets. For tuning the SVM parameters only the training set is used, and the test set is only used to report the classification performance in such unseen data by the classifier.
Training the classifiers was done through standard $v$-fold cross-validation, i.e. the training set is split in $v$ folds and $v$ different classifiers are obtained for several combinations of parameters. The best combination of parameters in terms of average accuracy over all folds is selected and used to generate the final classifier for each sub-problem. 

The validation scheme assesses model performance on data never used in the training procedure (testing set). The essence of statistical learning is to derive algorithms that can generalize well to unseen situations (data). The performance is thus evaluated in the test sets using statistical classification scores: confusion matrices, overall accuracy, and the estimated Cohen's kappa statistic.

\section{Experimental Results}\label{sec:results}

\begin{table*}[t]
\begin{center}
\caption{Kappa statistics and Overall Accuracy [$\kappa$ (OA\%)] for the selected landmarks. Best results are highlighted in bold.}\label{tab:final} 
\begin{tabular}{|c|c|cccc|c|} \hline \hline

\multirow{2}{*}{Name} & \multirow{2}{*}{\#LM} & High light & Medium light & Low light & Night & \multirow{2}{*}{\bf Global}\\ 
& & {SZA $\leq$ SZA$_{m}$} & {SZA$_{m}$ $\leq$ SZA $\leq$ 80} & {80 $\leq$ SZA $\leq$ 90} & {SZA $\geq$ 90} &  \\ \hline
Ad Dakhla \blue{(Morocco)} & 0 & 0.62 (83.24) & 0.71 (87.03) & 0.63 (83.51) & 0.65 (85.78) & 0.66 (85.35) \\
Aqaba2 \blue{(Saudi Arabia)} & 14 & 0.50 (82.74) & 0.56 (83.34) & 0.57 (84.95) & 0.65 (90.06) & 0.59 (86.65) \\
Azores5 \blue{(Portugal)} & 17 & 0.72 (87.61) & 0.64 (86.20) & {0.56} ({80.98}) & 0.56 (79.86) & 0.61 (82.94) \\
Chad2 \blue{(Chad)} & 48 & 0.76 (87.94) & 0.74 (87.09) & 0.64 (81.97) & 0.58 (78.75) & 0.65 (82.83) \\
Danger \blue{(South Africa)} & 63 & 0.82 (90.89) & 0.81 (90.36) & 0.68 (84.22) & 0.63 (81.57) & 0.71 (85.64) \\
Grampian \blue{(Scotland)} & 83 & 0.70 (89.99) & 0.69 (88.90) & 0.57 (81.92) & 0.48 (78.32) & 0.57 (82.95) \\
Libreville \blue{(Gabon)} & 107 & 0.69 (87.93) & 0.73 (89.52) & 0.69 (87.34) & 0.68 (88.25) & 0.69 (88.40) \\
Messina \blue{(Sicilia)} & 120 & 0.80 (90.09) & 0.80 (89.92) & 0.73 (86.47) & 0.71 (85.73) & 0.75 (87.61) \\
Nasser2 \blue{(Egypt)} & 131 & 0.57 (89.16) & 0.59 (88.33) & 0.63 (90.08) & 0.71 (94.17) & 0.64 ({\bf 91.52}) \\
Rhodes \blue{(Greece)} & 154 & 0.80 (91.54) & 0.77 (88.73) & 0.72 (86.44) & 0.72 (86.50) & 0.75 (88.05) \\
Tenerife \blue{(Spain)} & 177 & 0.77 (88.46) & 0.71 (85.41) & 0.63 (81.30) & 0.67 (83.18) & 0.69 (84.68) \\
Val\`encia \blue{(Spain)} & 190 & 0.83 (91.59) & 0.84 (92.18) & 0.76 (87.88) & 0.73 (86.72) & {\bf 0.78} (89.01) \\
\hline \hline
\end{tabular}
\end{center}
\end{table*}

As previously mentioned, different landmark-specific classifiers are developed in order to provide a cloud mask per image chip.
Therefore, this section presents the cloud detection results over landmarks of the proposed pool of dedicated SVMs trained for particular light conditions depending on the time-of-the-day (SZA). 
In order to set the final experimental setup, some critical parameters have been empirically studied. A preliminary experimentation has been designed taking into account a subset consisting of 12 representative landmarks. 
\blue{Among them we include the most cloudy test site (Grampian, in Scotland), deserts (Ad Dakhla, in Morocco; Aqaba2, in Saudi Arabia; and Nasser lake, in Egypt), islands (Azores; Rhodes, in Greece; and Tenerife island, in Spain), and the Nasser lake that is the less cloudy test site.} 
We assess cloud detection performance in the selected landmarks in terms of classification accuracy, dependence on the SZA associated to the acquired chip, day of the year, and seasonal variations of the obtained accuracy, as well as dependence on the land cover (predictions over land, water, and coastlines) both numerically and visually.
\blue{For the interested reader, we additionally provide illustrative videos with the results for a set of representative landmarks in a dedicated web page (\href{http://isp.uv.es/code/landmarks.html}{http://isp.uv.es/code/landmarks.html})\footnote{\red{Source code is available under request.}}.}

\subsection{Experimental setup}

Data normalization or feature scaling is one pre-processing step to scale (or standardize) the range of the features so that each one contributes proportionately to the final decision function implemented by the classifier.
Commonly, the values of each feature in the data are scaled between minimum and maximum values, or standardized to have zero-mean and unit-variance. For the cloud detection problem, the features that fed the classifiers are scaled in the 0-1 range. Note that this normalization step should be applied just before the training or test of each particular classifier. Hence, it is applied to the features extracted after the conversion of input radiance to TOA reflectance and brightness temperature, and after the `divide-and-conquer' strategy to train independent classifiers per landmark and SZA range.

\blue{Then, we split the labeled dataset for each sub-problem according to the SZA into the trained and testing sets with fixed sizes of $10,000$ and $100,000$ pixels, respectively. 
Training was done through the standard $v$-fold cross-validation with $v=10$. In this setting, each model is obtained as follows. The training set is split into $v=10$ subsets, where 9 are used for training the model parameters and the other for validation. This process is repeated 10 times. The combination of parameters that obtained the best results during the 10-folds is selected to train the final model with the full training set. Finally, classification results are computed on the (unseen) test set of $100,000$ pixels}.

\subsection{Classification results} 

Table~\ref{tab:final} shows the main results for the selected landmarks. We highlight in blue and red the best and worst results, respectively. 
The main conclusions at this point are summarized as follows: (1) In general terms, good and consistent results are obtained across the different landmarks. (2) Uneven results are obtained when comparing the overall accuracy (OA) and the kappa statistic. This may lead to the idea that some systematic bias of the classifiers seems to overpredict one class (cloudy) versus the other (cloud-free). This observation is further studied in the following sections.  (3) Finally, as expected, it is also observed that lower results are obtained for night and twilight times (SZA$>$80). This is a consequence of developing classifiers with a lower number of features (only thermal channels can be used at night) with a poorer discrimination capability. It is well-known the fact that classification at night constitutes a more challenging problem because the lack of the optical channels, but also because atmospheric-land energy exchanges and drop-outs in temperature.

\begin{figure*}[]
\begin{center}
\footnotesize
\setlength{\tabcolsep}{1pt}
\renewcommand{\arraystretch}{1}
\begin{tabular}{cccc}
\blue{LM0: Ad Dakhla (Morocco)} & \blue{LM14: Aqaba2 (Saudi Arabia)}  & \blue{LM 17: Azores5 (Portugal)} & \blue{LM48: Chad2 Lake (Chad)} \\
\includegraphics[width=.25\textwidth]{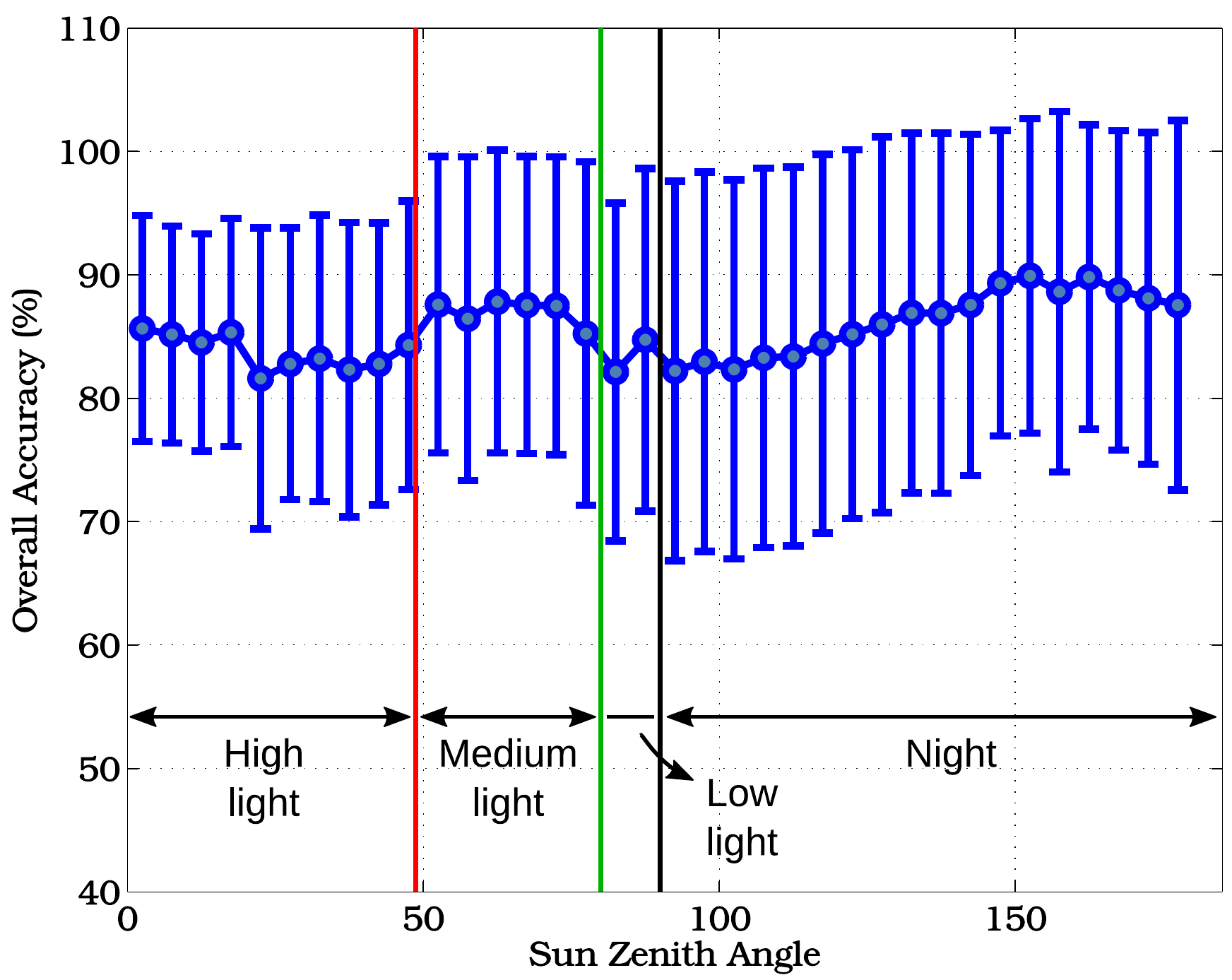} &
\includegraphics[width=.25\textwidth]{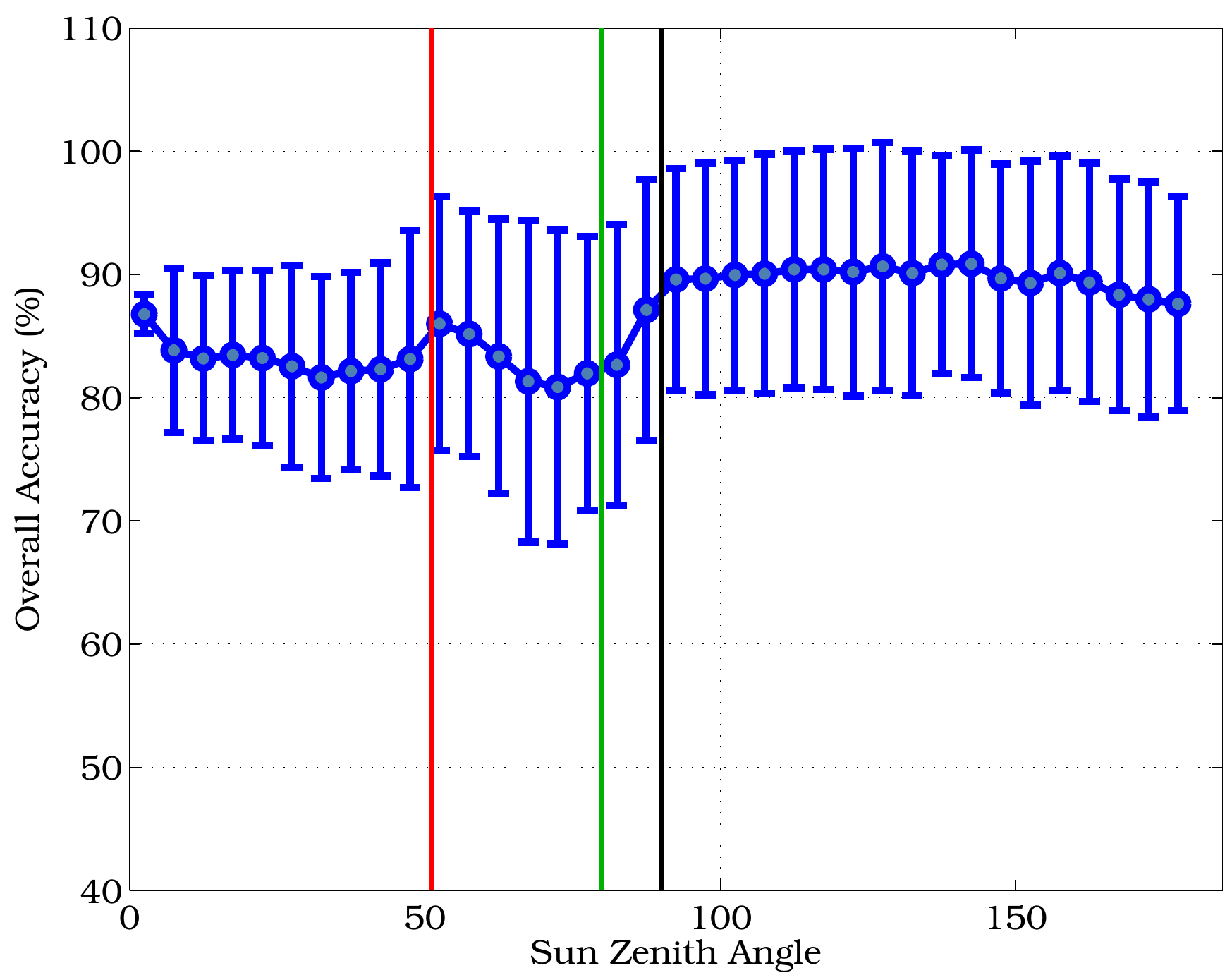} &
\includegraphics[width=.25\textwidth]{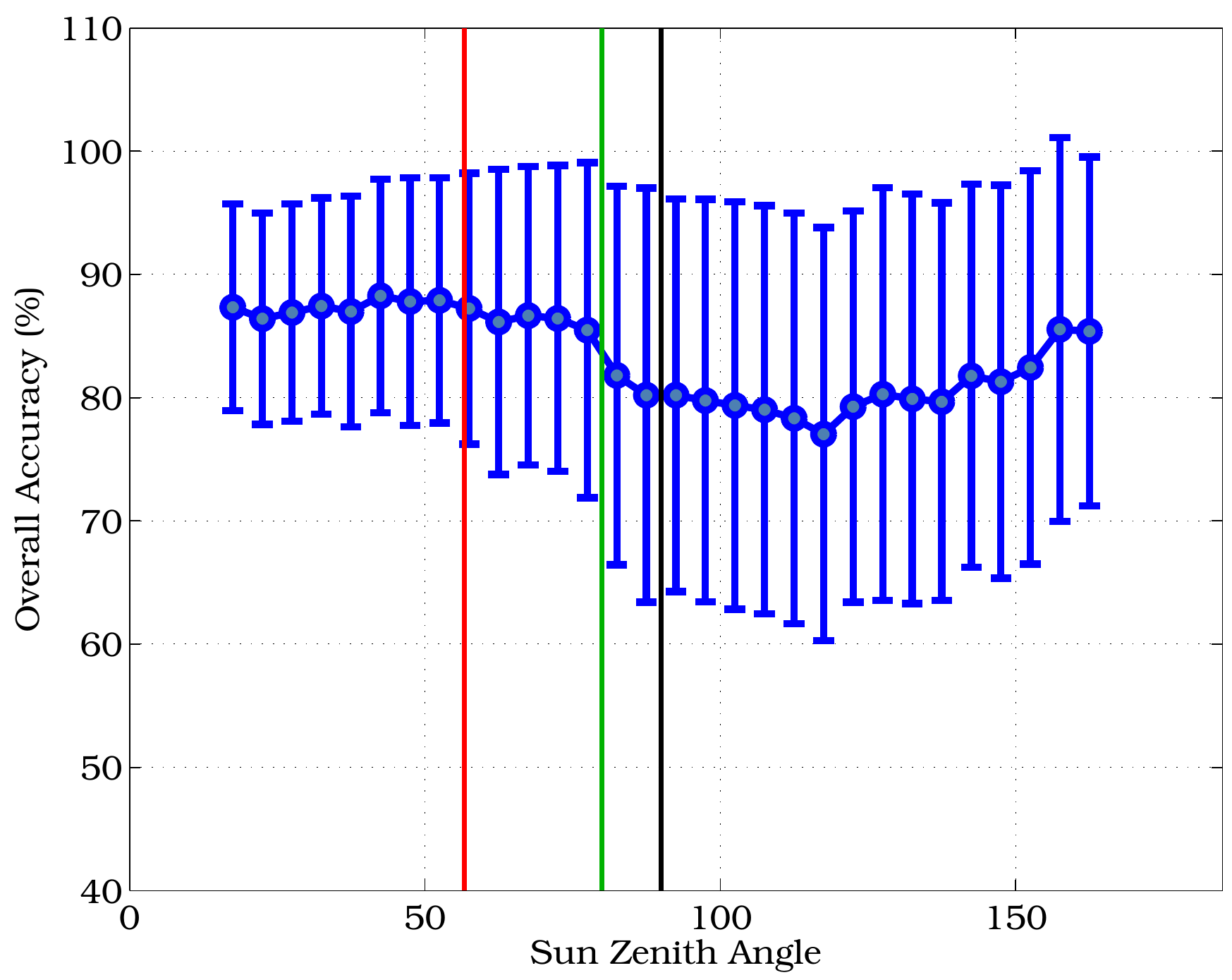} &
\includegraphics[width=.25\textwidth]{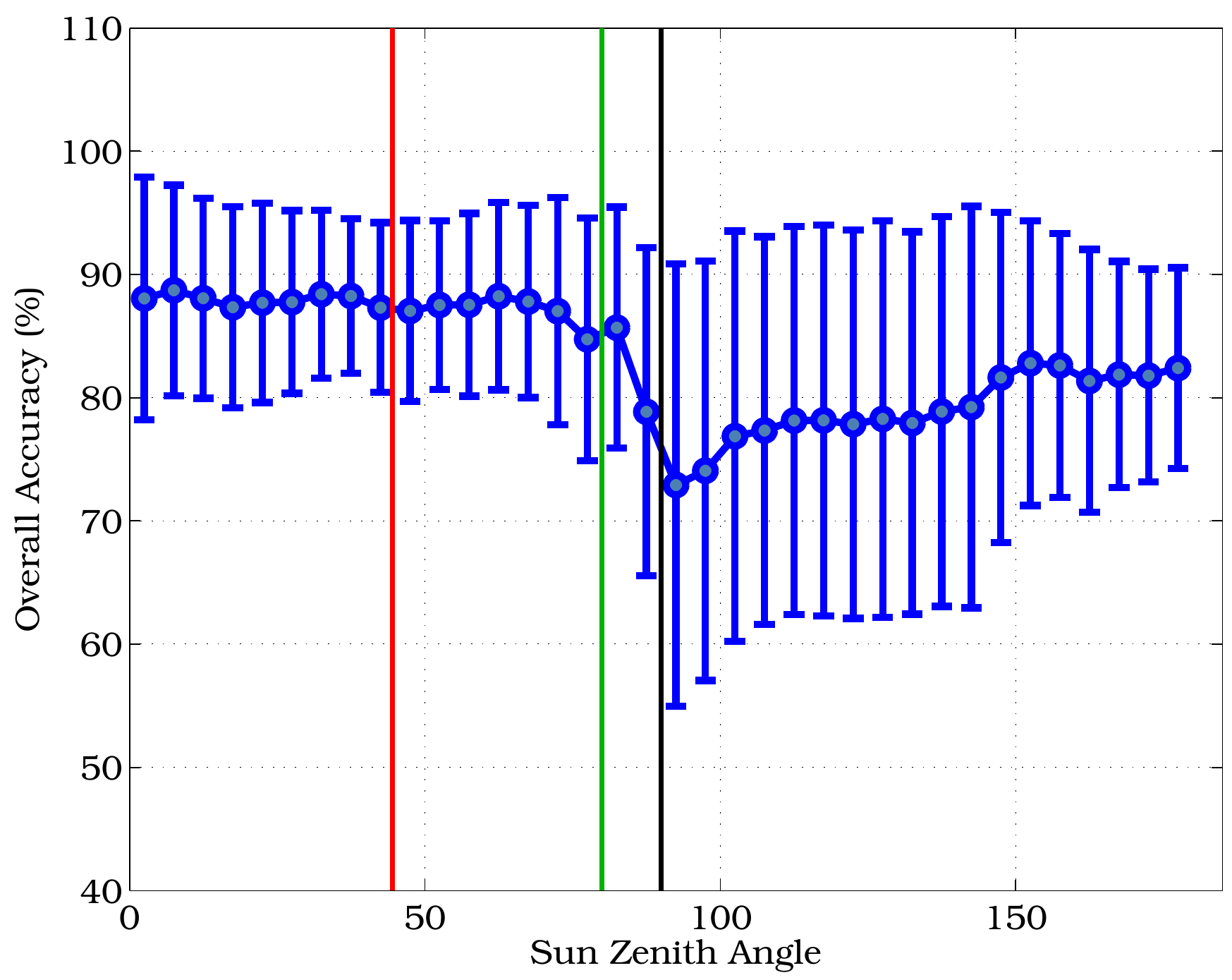}\\
\blue{LM63: Danger (South Africa)} & \blue{LM83: Grampian (Scotland)} & \blue{LM107: Libreville (Gabon)} & \blue{LM120: Messina (Sicilia)} \\
\includegraphics[width=.25\textwidth]{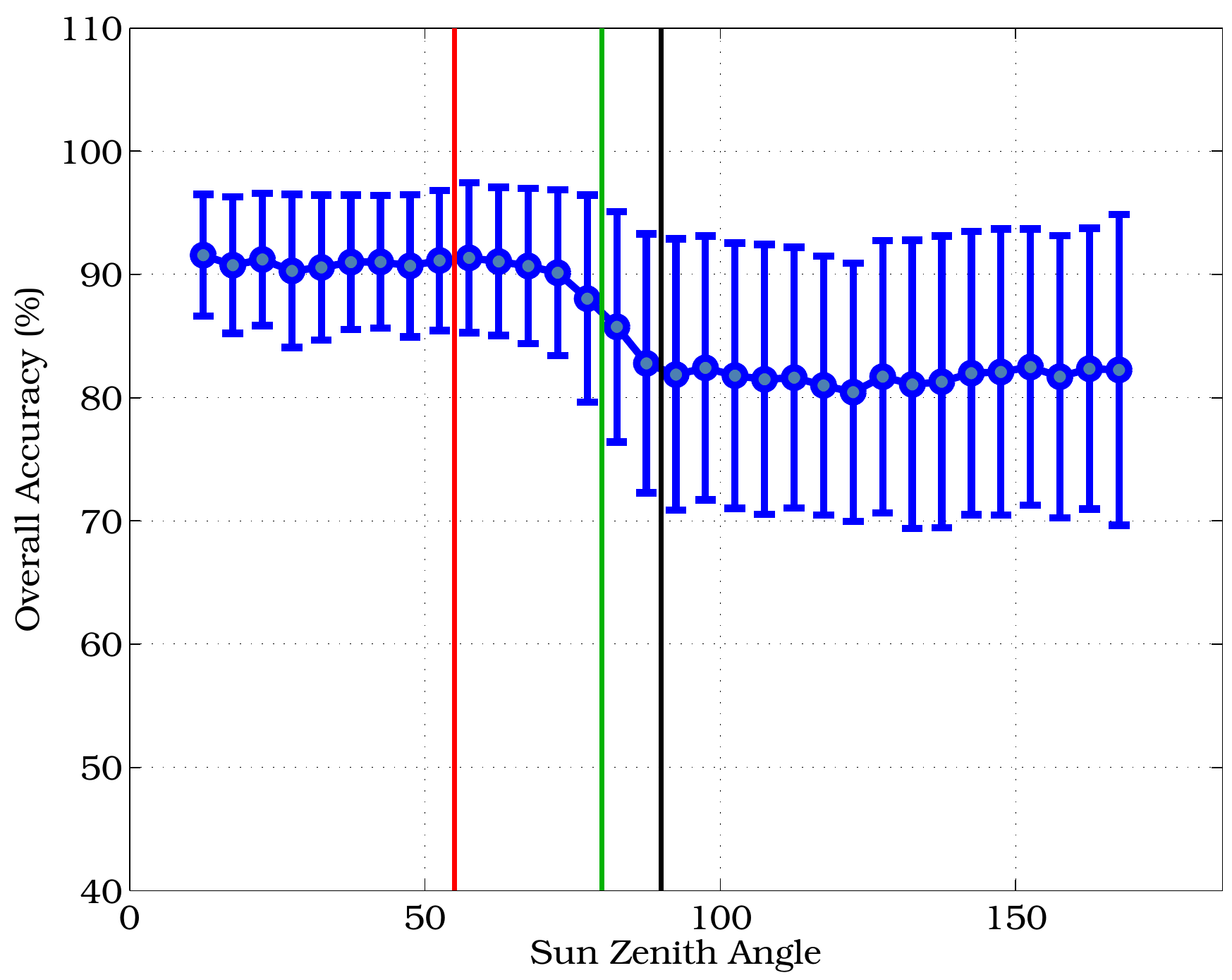} &
\includegraphics[width=.25\textwidth]{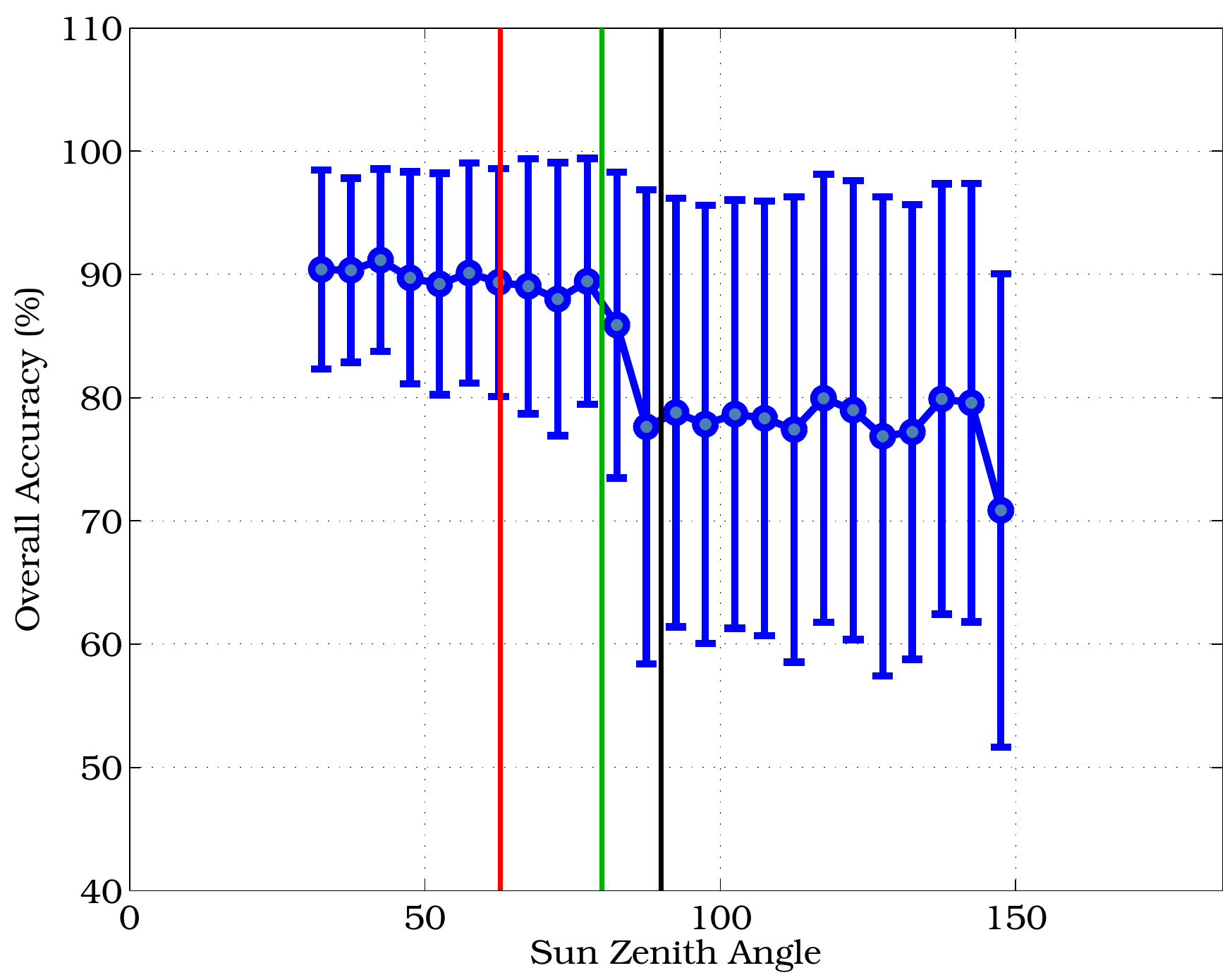} &
\includegraphics[width=.25\textwidth]{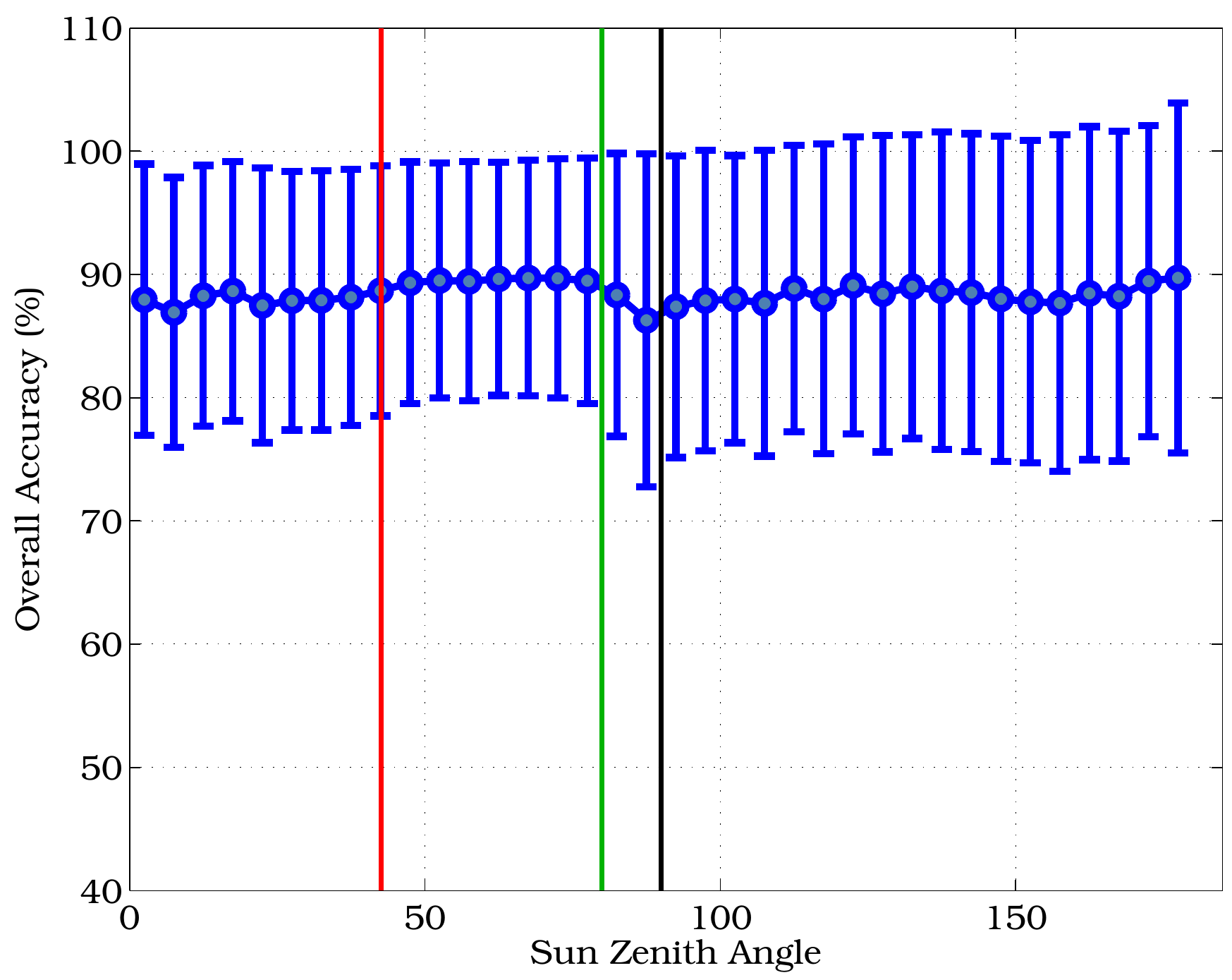} &
\includegraphics[width=.25\textwidth]{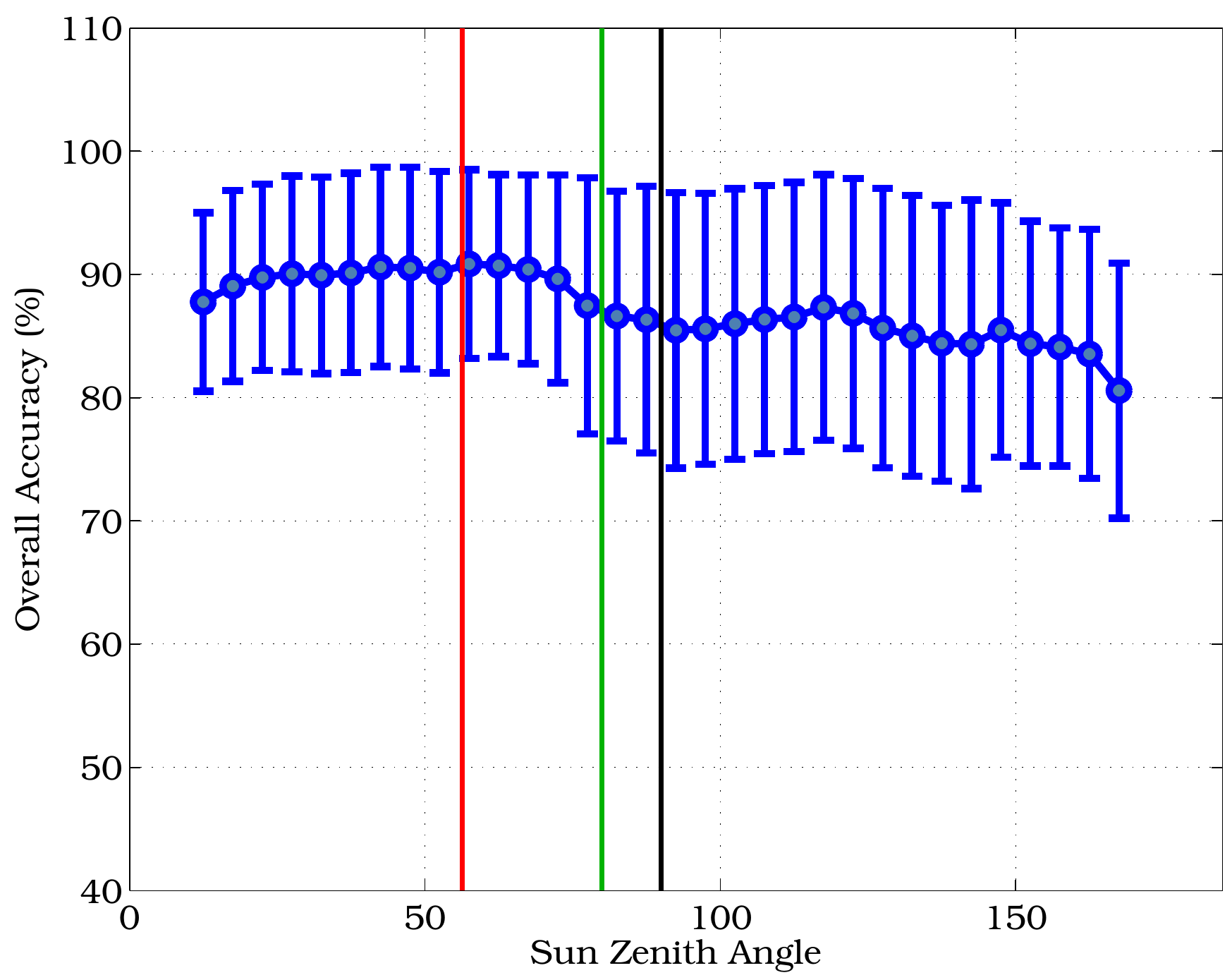}\\
\blue{LM131: Nasser2 Lake (Egypt)} & \blue{LM154: Rhodes (Greece)} & \blue{LM177: Tenerife (Spain)} & \blue{LM190: Val\`encia (Spain)}  \\
\includegraphics[width=.25\textwidth]{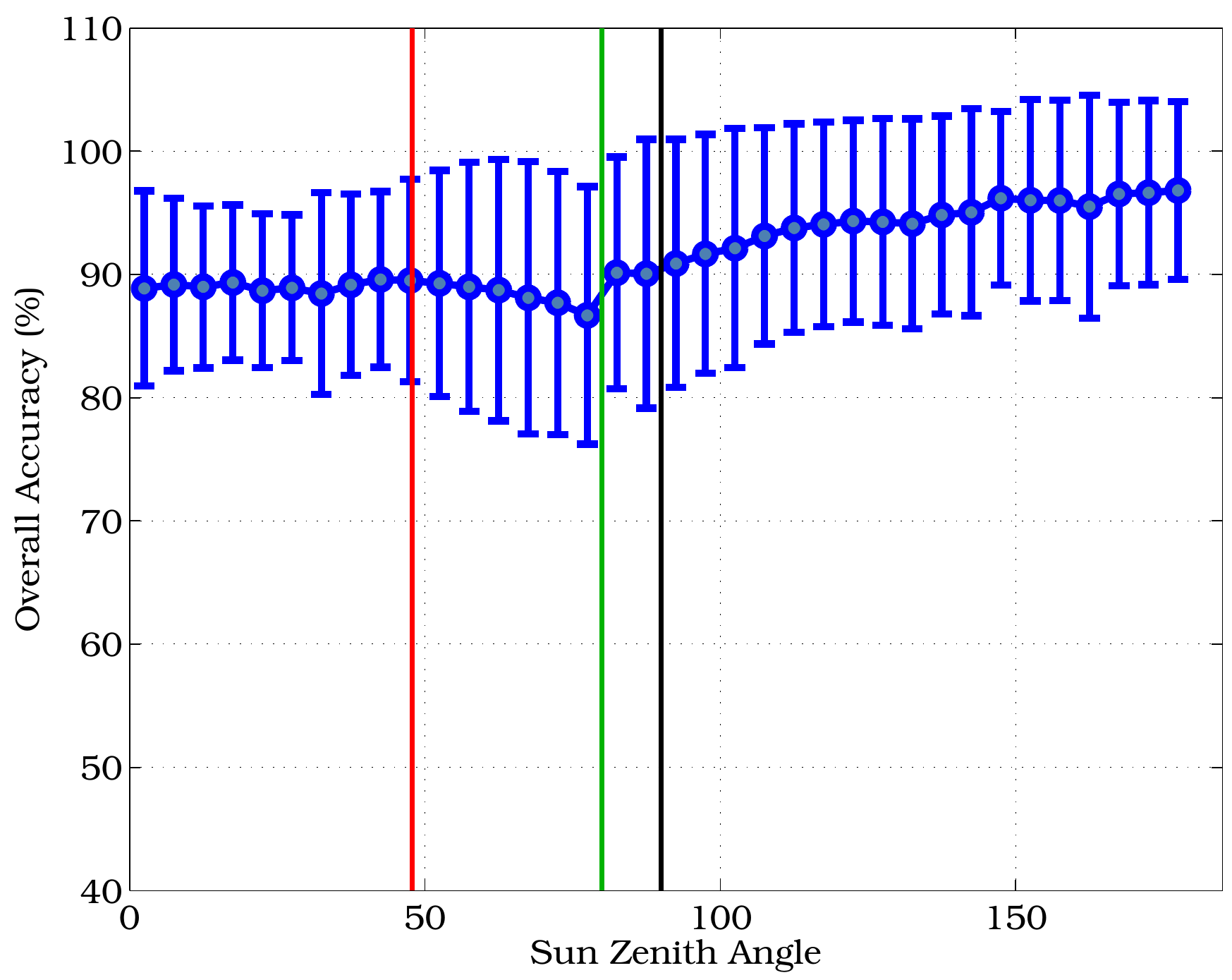} &
\includegraphics[width=.25\textwidth]{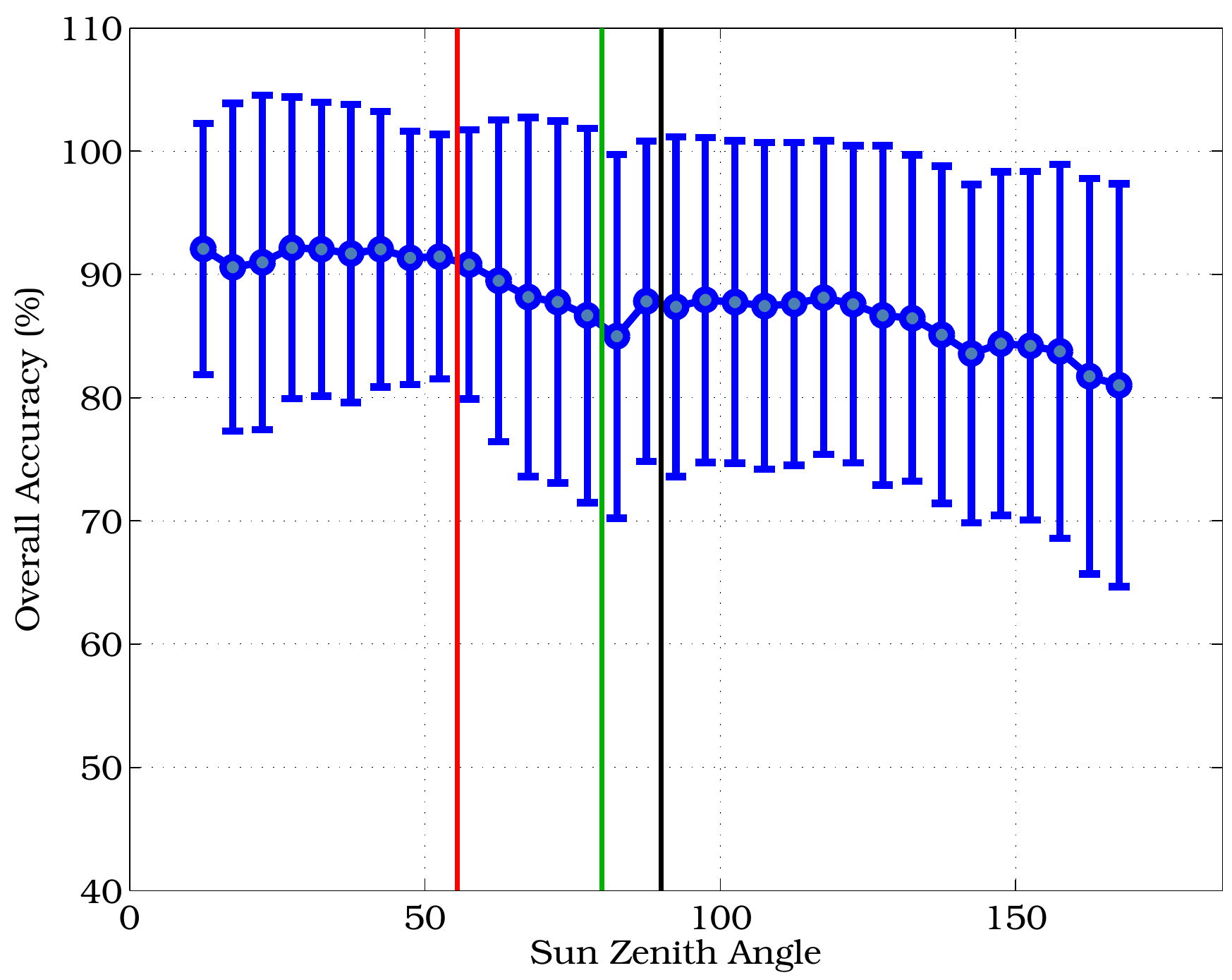} &
\includegraphics[width=.25\textwidth]{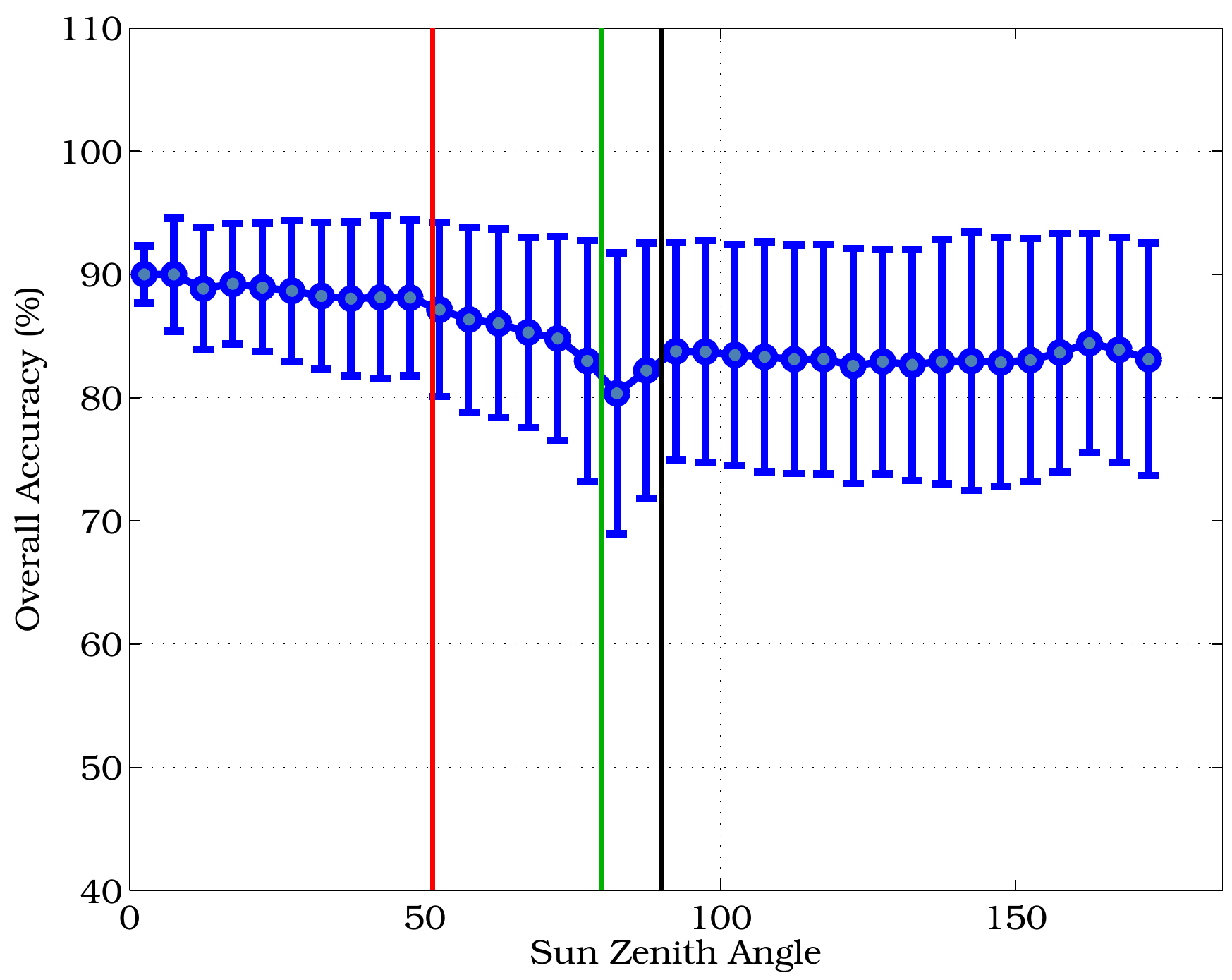} &
\includegraphics[width=.25\textwidth]{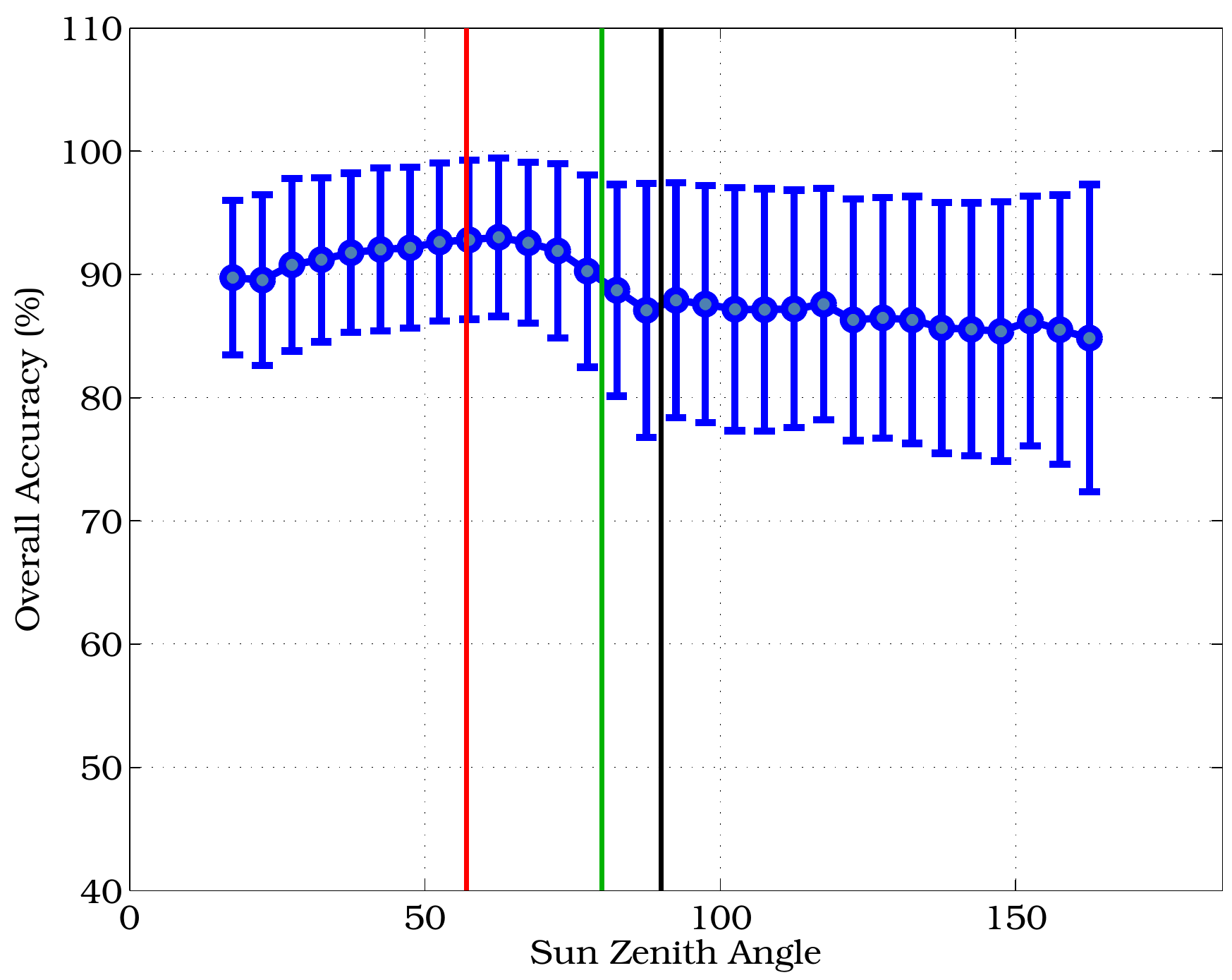}\\
\end{tabular}
\end{center}
\caption{Overall accuracy considering the different illumination conditions according to different values of sun zenith angle (SZA). The vertical lines divide the daytime in 4 parts, and are landmark-specific, which correspond to the four trained SVM classifiers. The green and black lines correspond to SZA=80 and 90, respectively. The red line is the median SZA value (from $0$\textdegree$<$SZA$<80$\textdegree) of all the chips by landmark.\label{fig:illu_analysis1}}
\end{figure*}

\subsection{Results as a function of the illumination conditions}

In this section we show the results obtained as a function of the illumination conditions. Results are shown in Fig.~\ref{fig:illu_analysis1}. \blue{We can see that the obtained average accuracy is higher than 80\% in almost all situations and landmarks. We also observe a lower detection accuracy for higher SZA values, i.e. with low light conditions, similarly to the results obtained in Table~\ref{tab:final} and, as expected, for the twilight and night cases. However, in landmarks LM0 (Morocco), LM14 (Saudi Arabia) and LM131 (Egypt) the accuracy is lower for high lighting conditions. These landmarks are located in desert areas with very high radiance values over sand in the coastline, and the L2 mask used as ground truth in this work systematically classified these pixels as clouds. A further analysis of the spatial patterns and the L2 mask consistency is done in the following sections.}

Additionally, we observe that in general the accuracy drift is well captured by the SZA thresholds, indicating that the strategy for selecting the threshold, which is dependent of the particular landmark, is adequate. \blue{It is also noted that in some cases the standard deviation of the detection accuracy increases at higher SZA for some landmarks, such as LM48 (Chad lake), LM63 (South Africa), and LM83 (Scotland). These landmarks present a lot of vegetation and clouds. Note that LM83 is the most cloudy landmark in the database. As night,  classifiers present usually a lower performance that the day classifiers to detect clouds, this higher probability of having clouds explains the lower overall accuracy and higher standard deviation. The relation of these results with the type of land cover and the landmark spatial patterns is further analyzed in the following sections.
}

\begin{figure*}[]
\begin{center}
\footnotesize
\setlength{\tabcolsep}{1pt}
\renewcommand{\arraystretch}{1}
\begin{tabular}{cccc}
\blue{LM0: Ad Dakhla (Morocco)} & \blue{LM14: Aqaba2 (Saudi Arabia)}  & \blue{LM 17: Azores5 (Portugal)} & \blue{LM48: Chad2 Lake (Chad)} \\
\includegraphics[width=.25\textwidth]{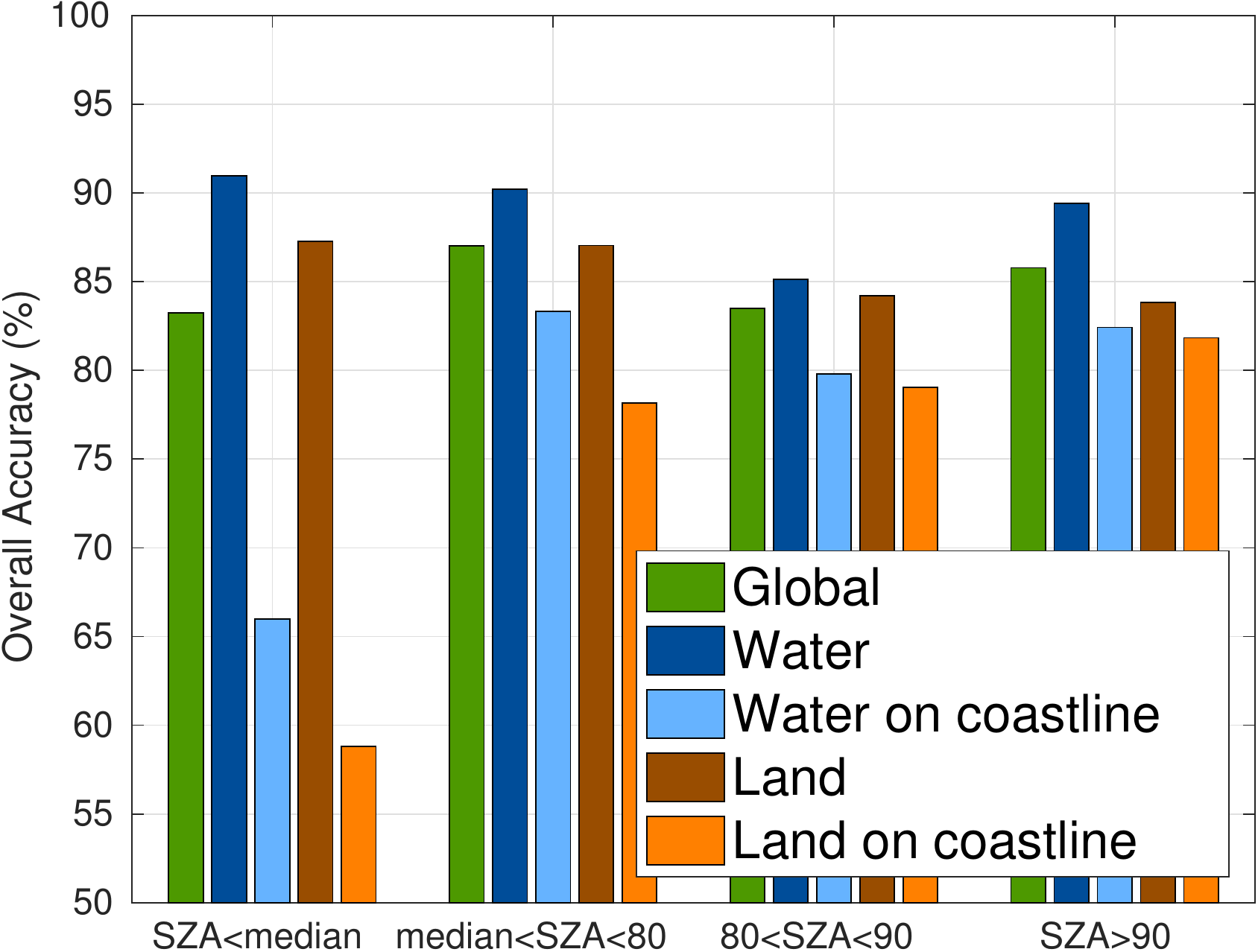} &
\includegraphics[width=.25\textwidth]{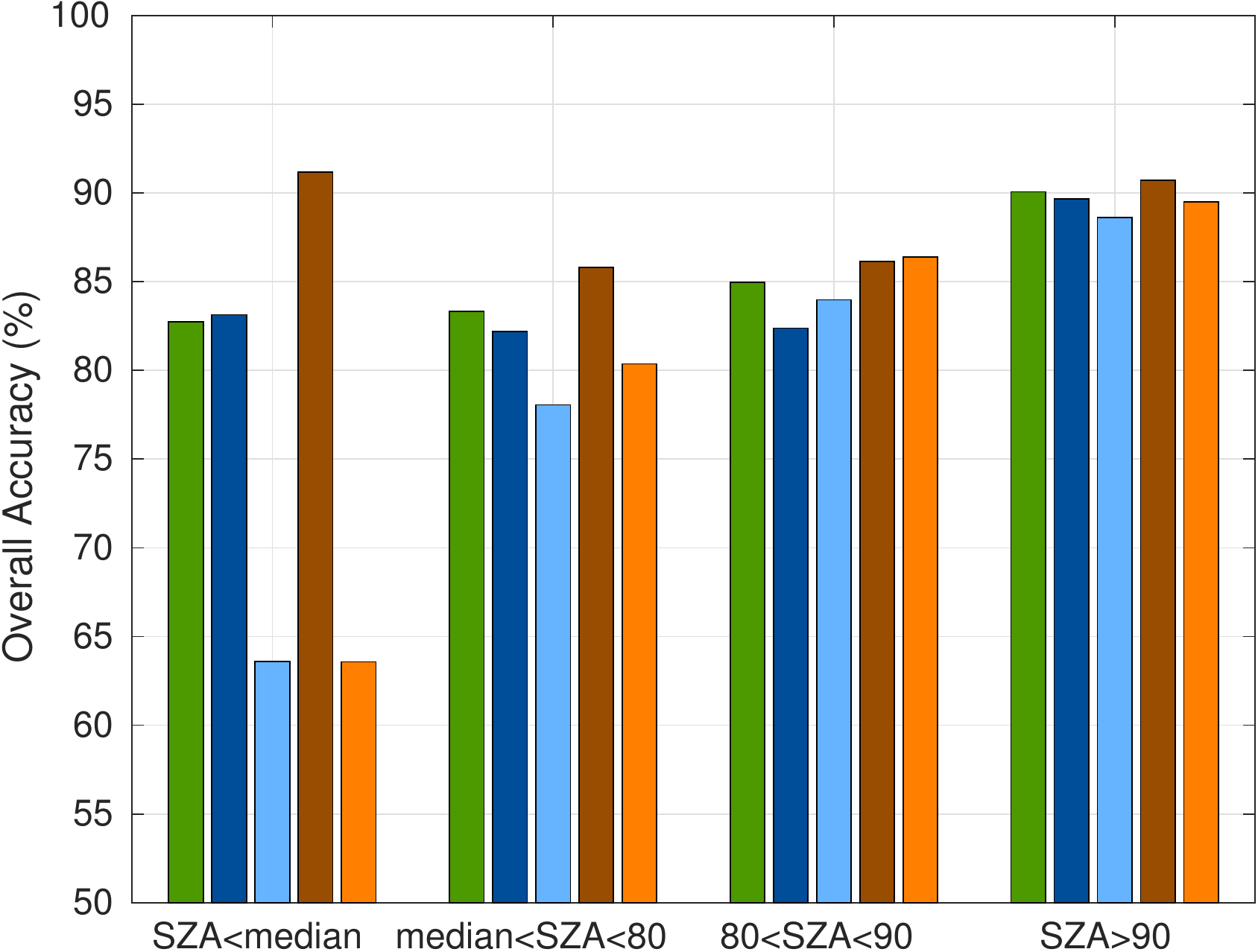} &
\includegraphics[width=.25\textwidth]{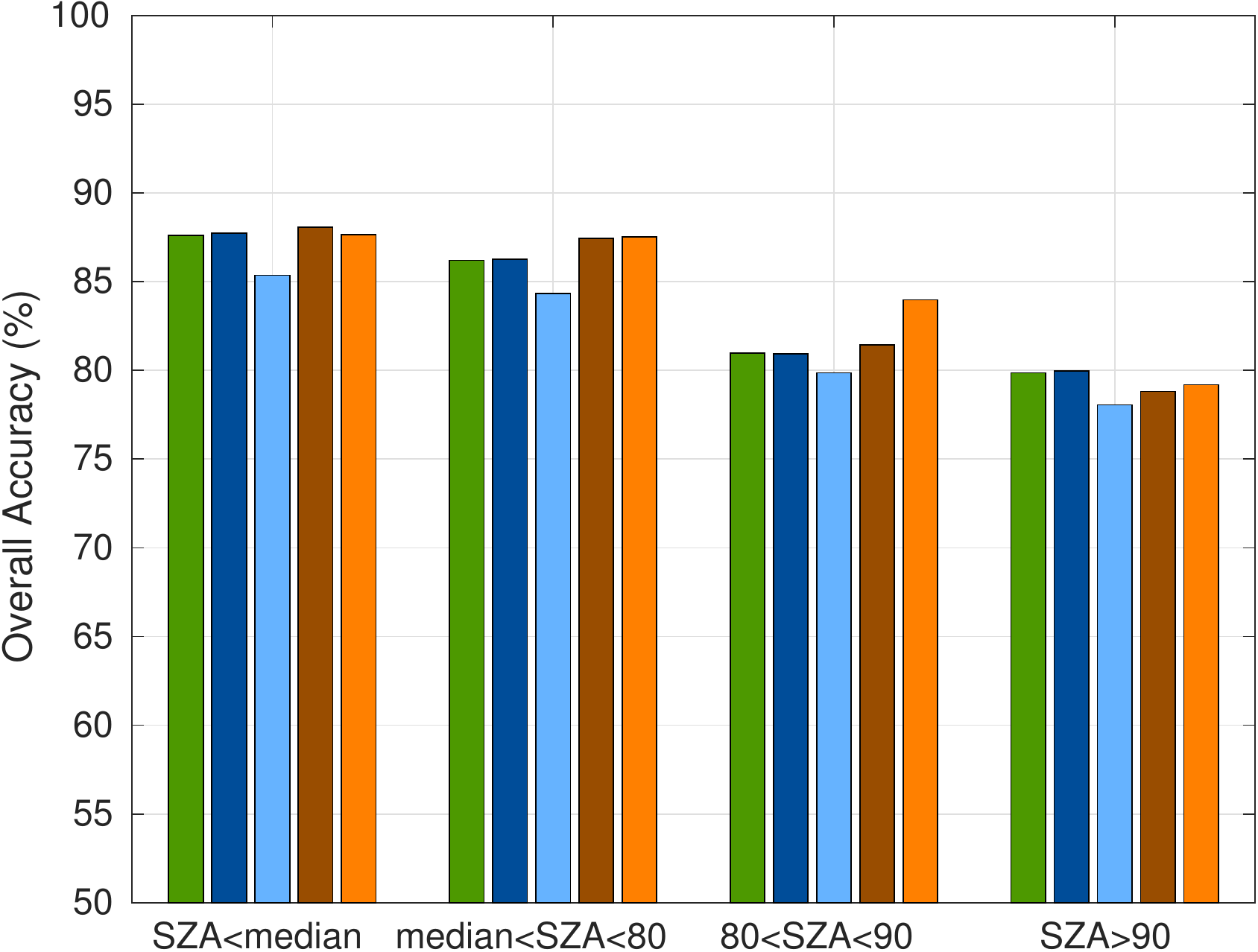} &
\includegraphics[width=.25\textwidth]{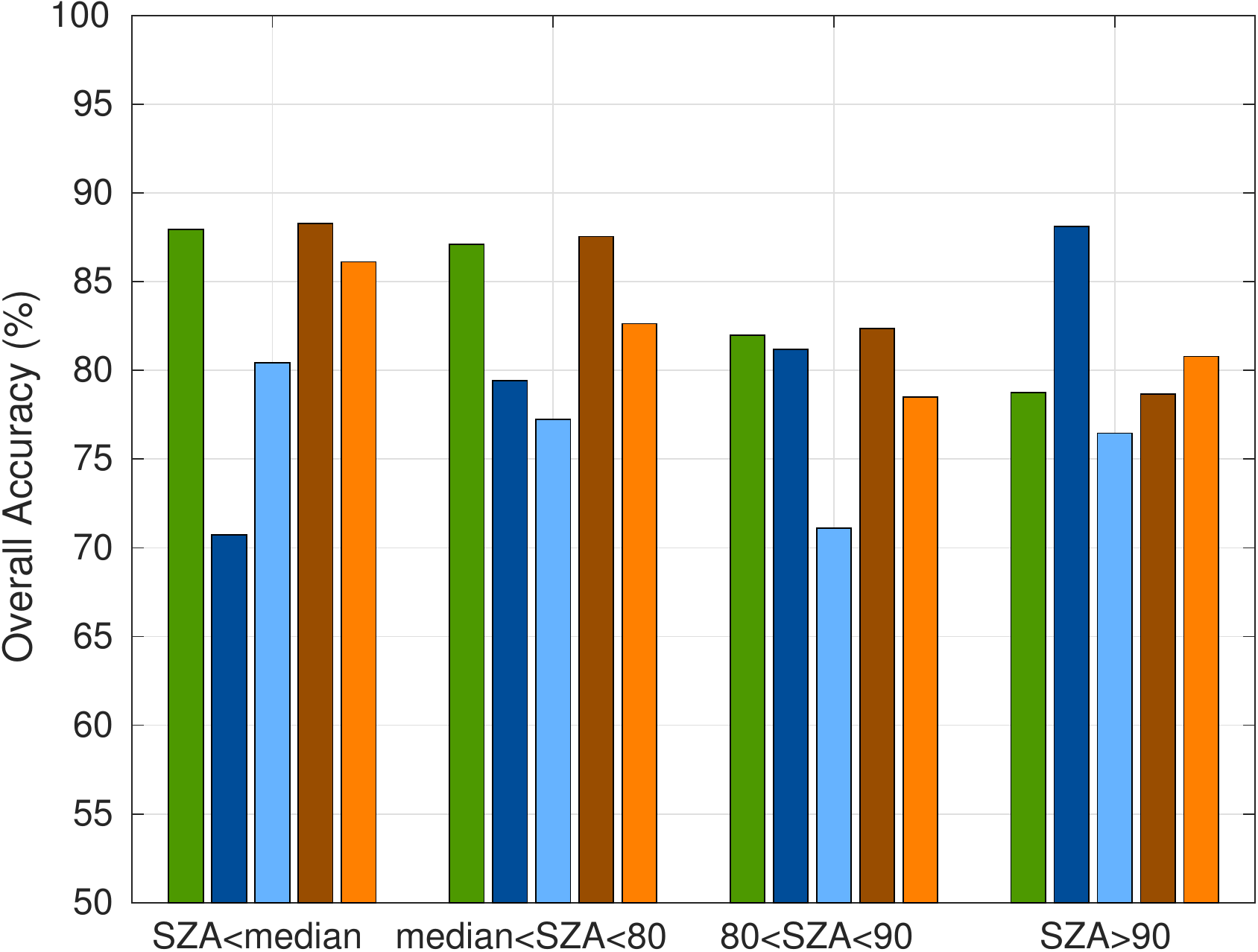}\\
\blue{LM63: Danger (South Africa)} & \blue{LM83: Grampian (Scotland)} & \blue{LM107: Libreville (Gabon)} & \blue{LM120: Messina (Sicilia)} \\
\includegraphics[width=.25\textwidth]{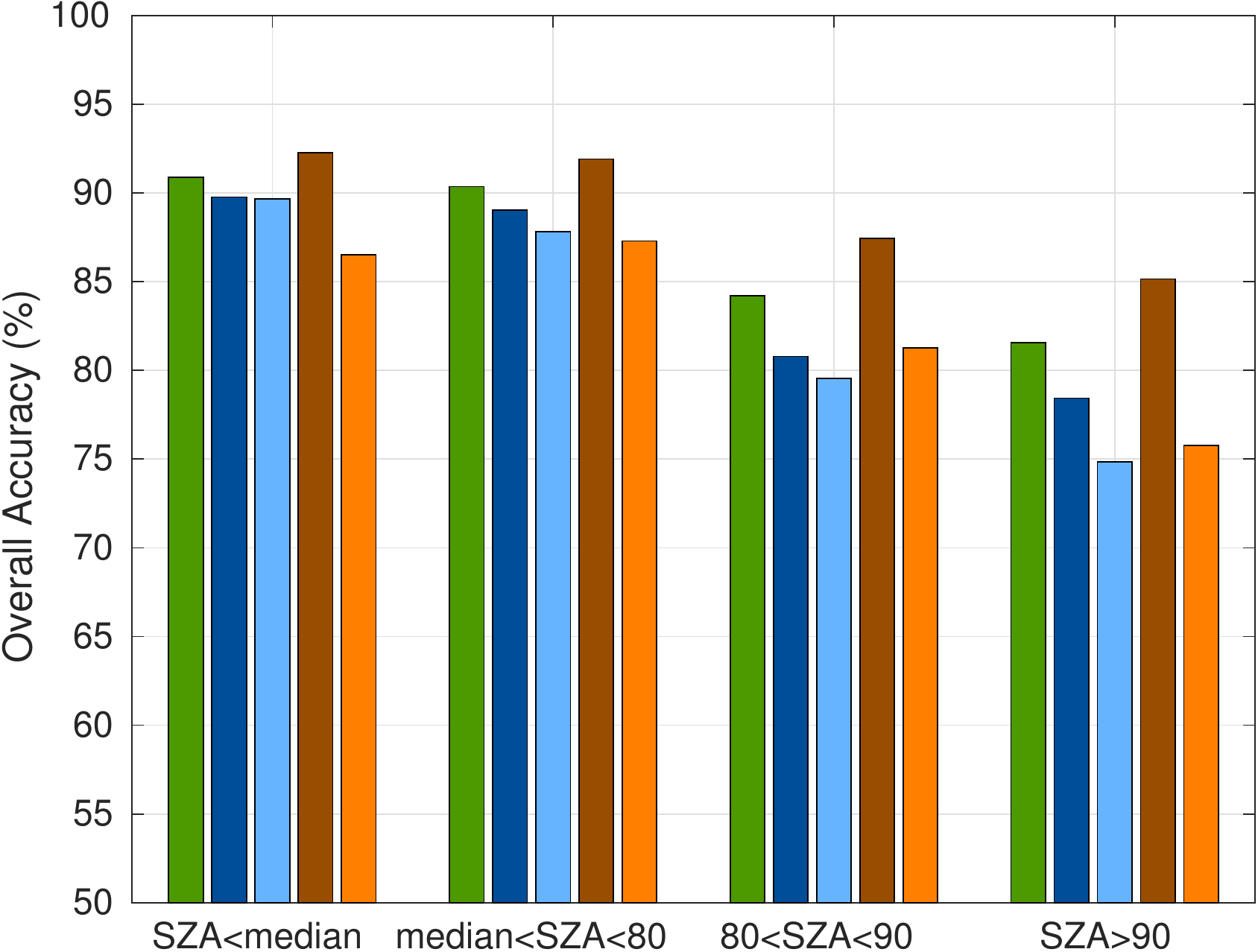} &
\includegraphics[width=.25\textwidth]{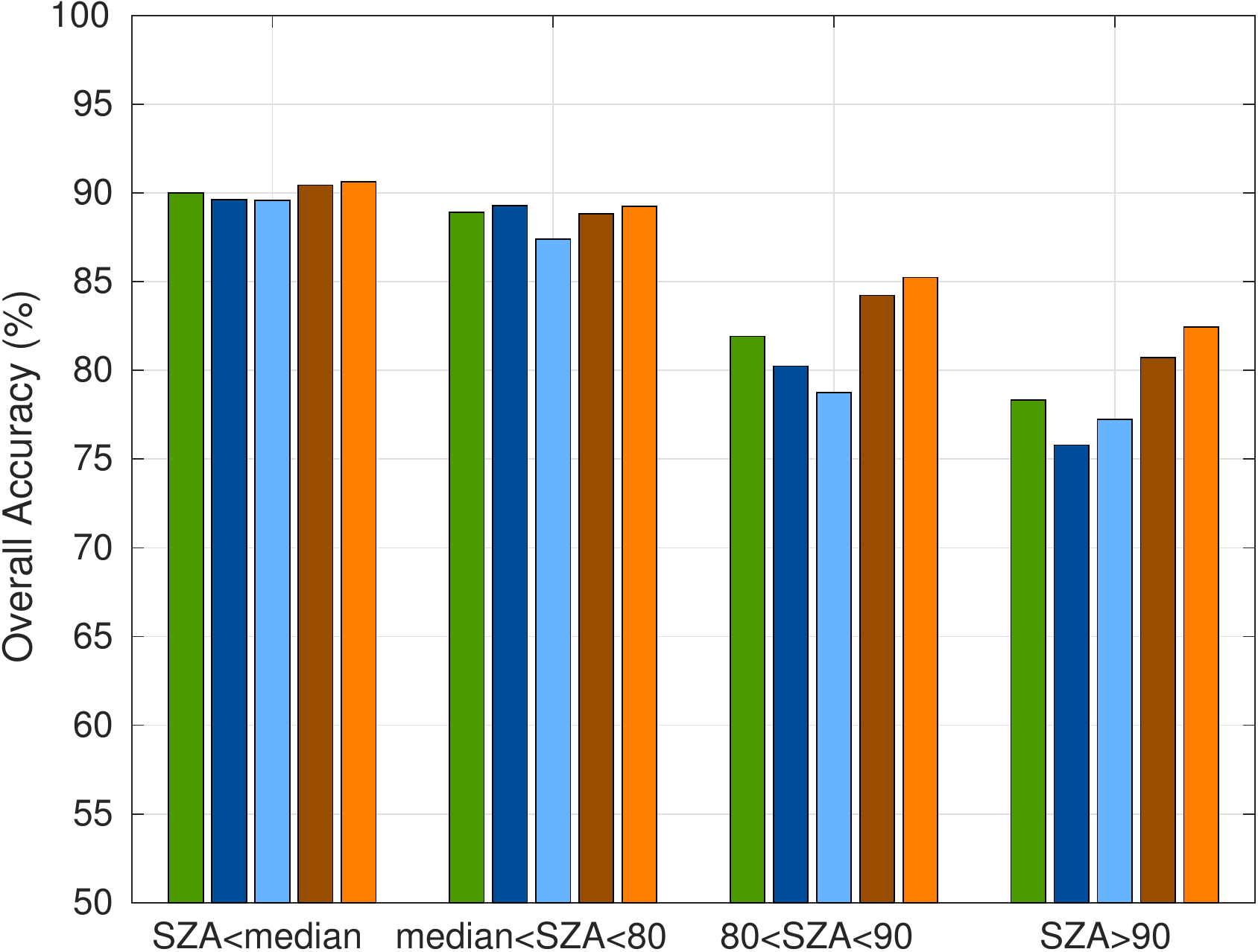} &
\includegraphics[width=.25\textwidth]{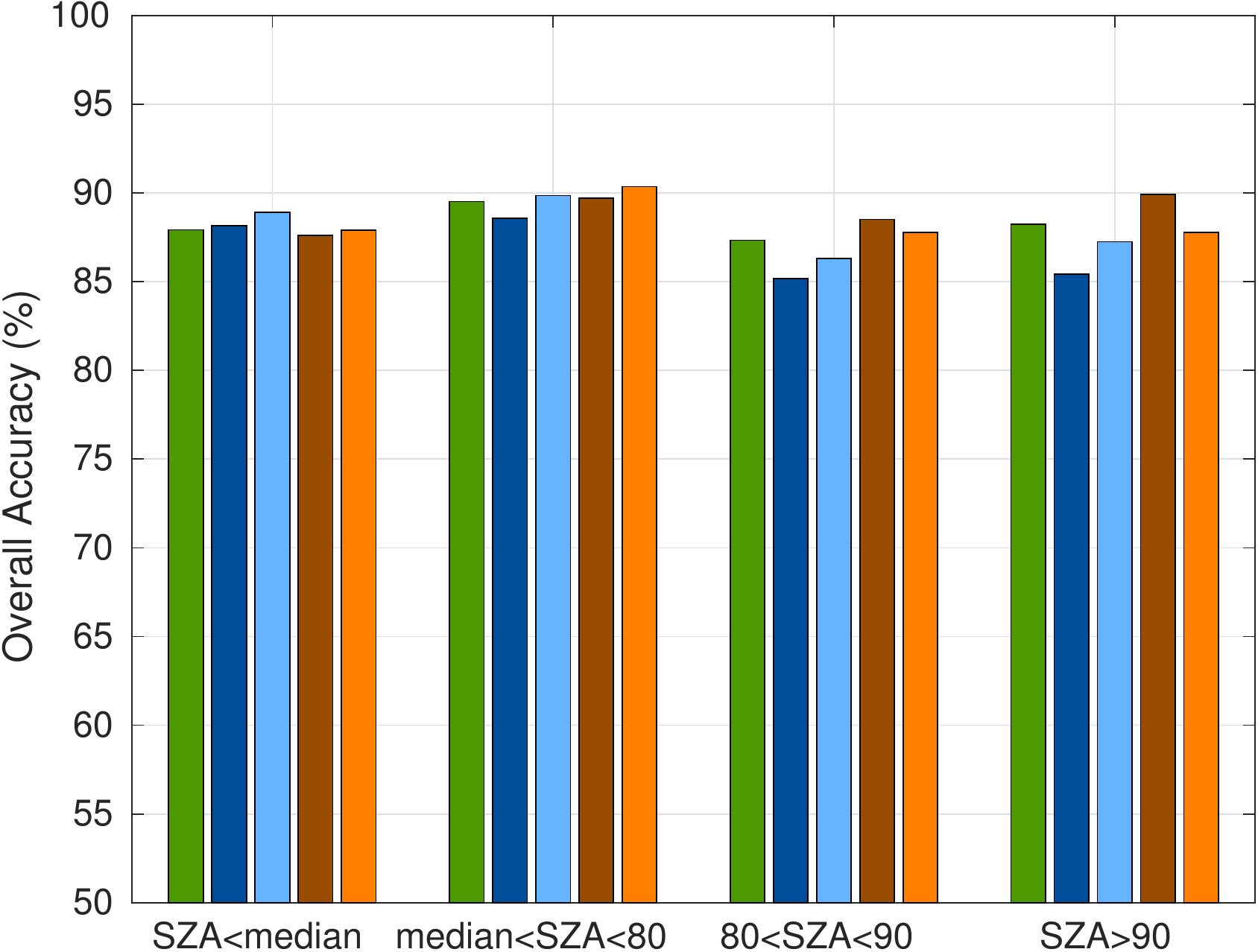} &
\includegraphics[width=.25\textwidth]{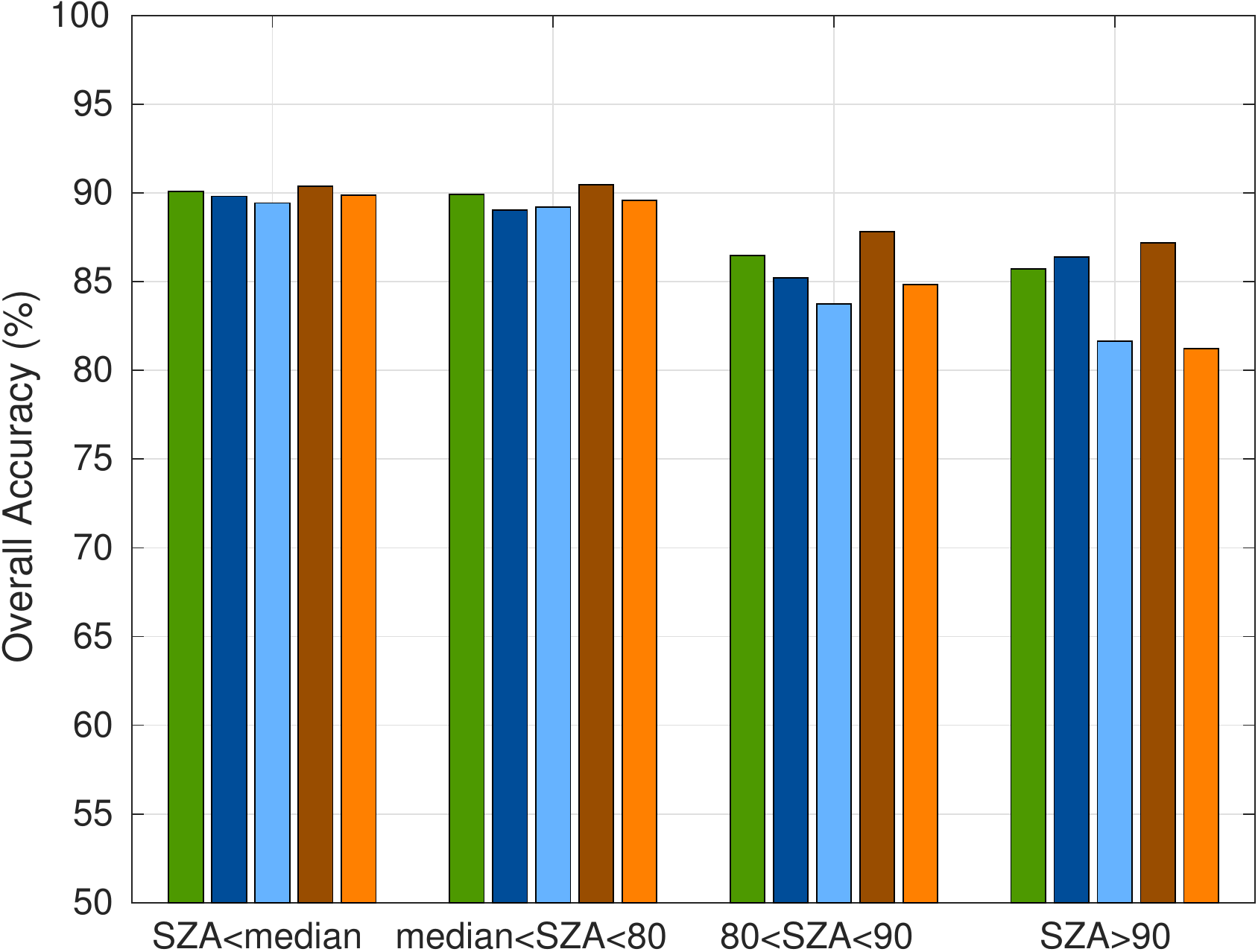}\\
\blue{LM131: Nasser2 Lake (Egypt)} & \blue{LM154: Rhodes (Greece)} & \blue{LM177: Tenerife (Spain)} & \blue{LM190: Val\`encia (Spain)}  \\
\includegraphics[width=.25\textwidth]{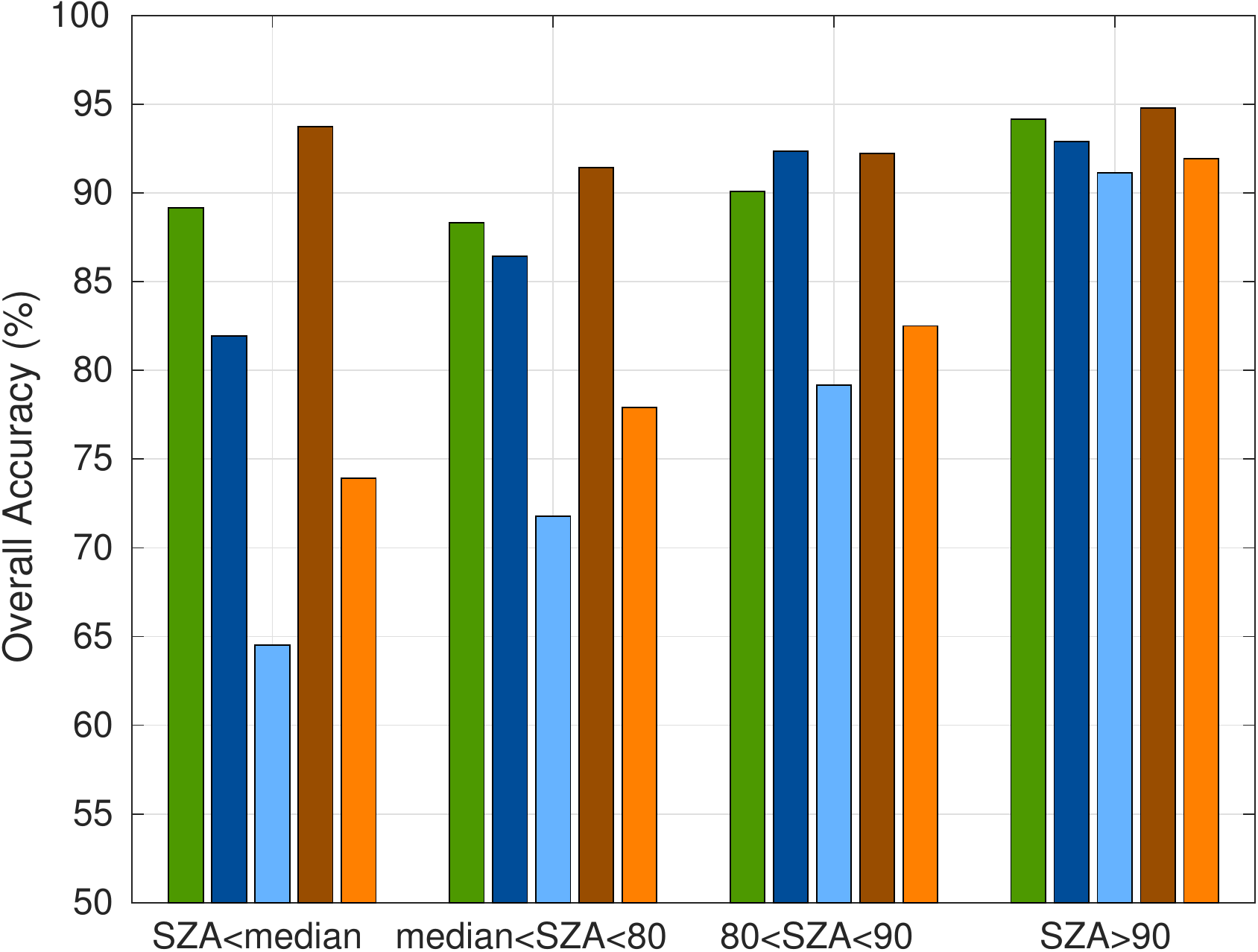} &
\includegraphics[width=.25\textwidth]{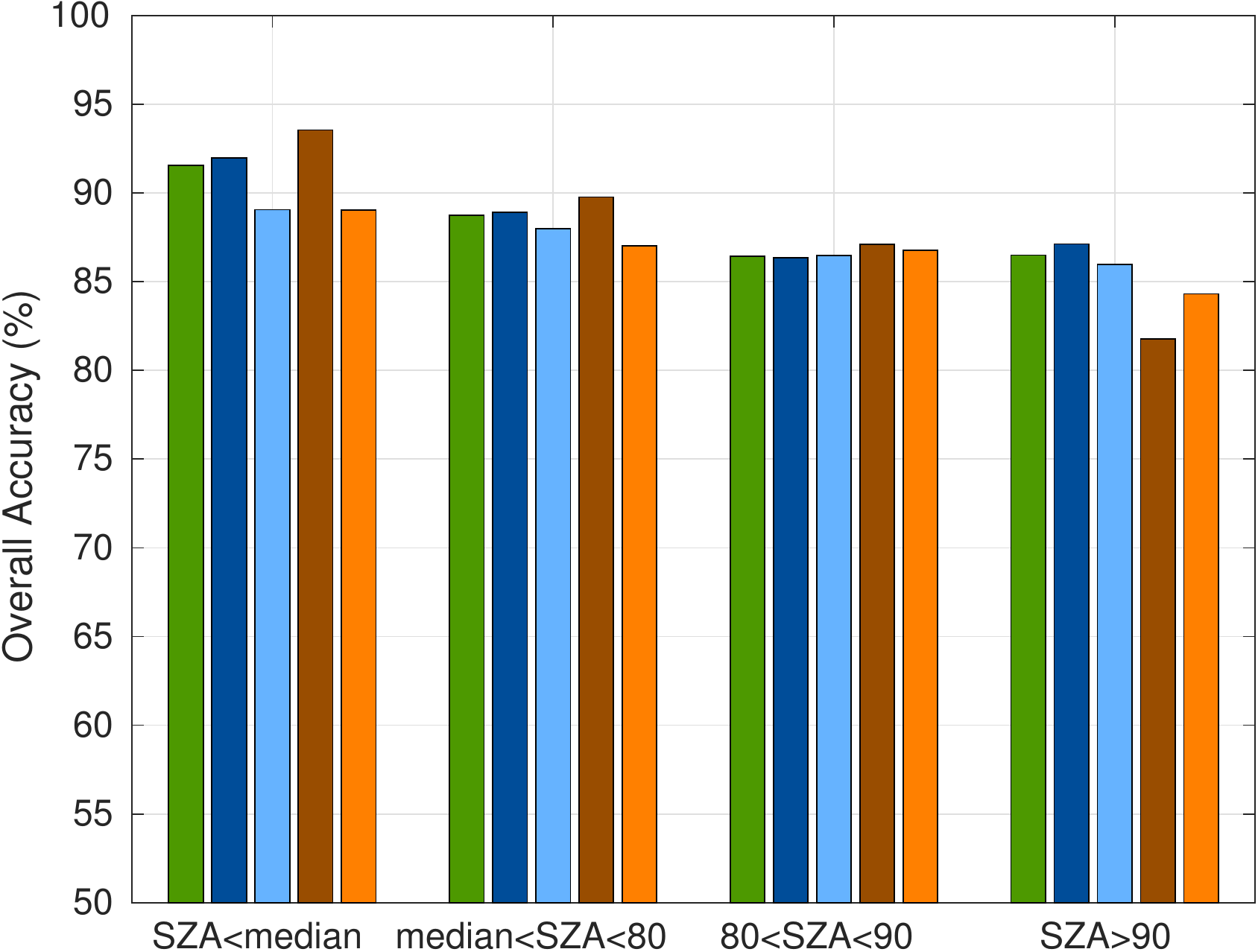} &
\includegraphics[width=.25\textwidth]{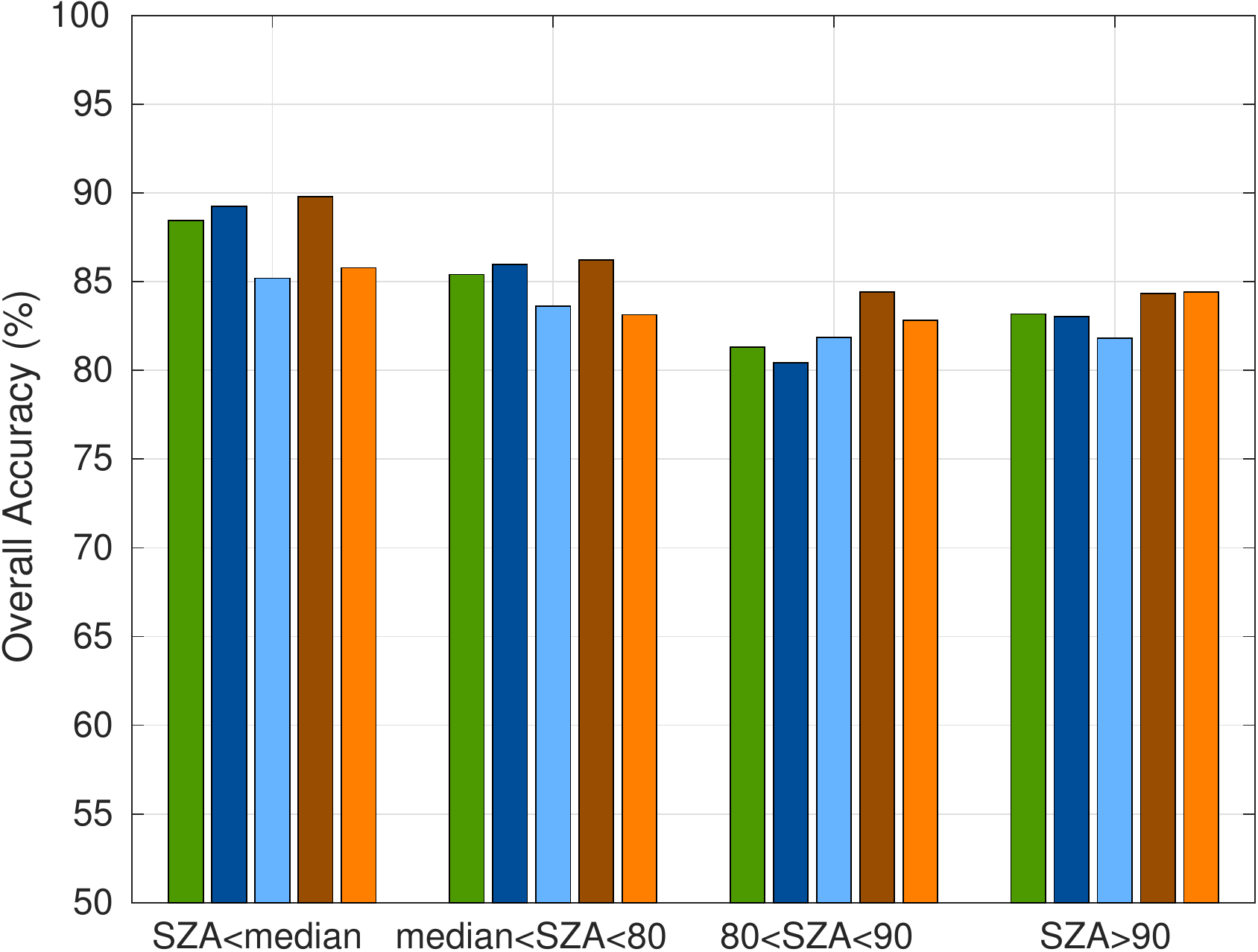} &
\includegraphics[width=.25\textwidth]{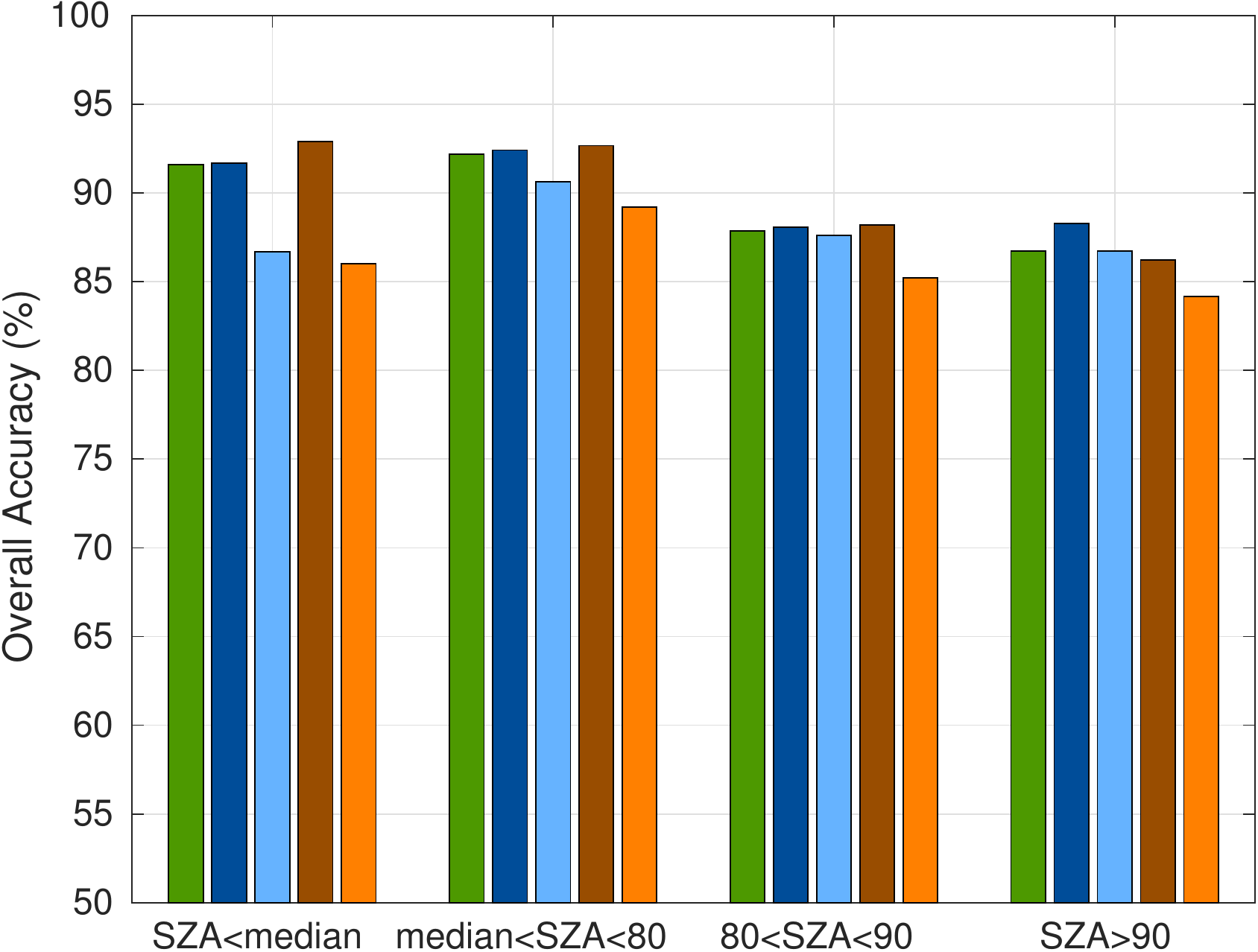}\\
\end{tabular}
\end{center}
\caption{Overall accuracy considering the different land covers (water, land) and coastline over water/land. \label{fig:LC_OA}}
\end{figure*}

\subsection{Results as a function of the land cover}

In this section, we show the results obtained as a function of the land cover in each particular landmark. 
For the analysis, we have included also the coastline pixels and they are analyzed separately. 
Results are shown in Fig.~\ref{fig:LC_OA}. The four bar groups on the x-axis correspond to the four trained classifiers with different illumination conditions according to the sun zenith angle (SZA). Each bar group provides the global agreement between the predicted and the level-2 mask for all the chips of this landmark. Then, we analyze the OA only considering land and water areas without including the coastline (3 pixels width). Finally, OA for coastline pixels over water and land is also calculated.

The main conclusions at this point are summarized as follows. In general, lower accuracy is obtained at night and twilight, as expected. This is observed for almost all landmarks considered in this study.
\blue{For some landmarks (LMs 0, 14, 131) we observed a poor performance at coastline pixels with high illumination conditions, which is mainly due to errors in the L2 mask that labels bright pixels in desert areas over the coastline as cloudy pixels systematically. The proposed cloud mask properly label these pixels as cloud-free and thus disagree with the L2 mask, decreasing the OA.} %
Finally, similar accuracy is generally achieved over land and over water, and hence there is not a bias in the classification accuracy with regard to the specific land cover heterogeneity of the landmark.

\subsection{Analysis of the averaged accuracy in the spatial domain}

\begin{figure}[t!]
\scriptsize
\begin{center}
\setlength{\tabcolsep}{1.5pt}
\renewcommand{\arraystretch}{0.5}
\begin{tabular}{ccccccc}

\bf{LM} & \bf{Land cover} & \bf{\tiny{SZA $\leq$ SZA$_{m}$}} & \bf{\tiny{SZA$_{m}$ $\leq$ SZA $\leq$ 80}} & \bf{\tiny{80 $\leq$ SZA $\leq$ 90}} & \bf{\tiny{SZA $\geq$ 90}} \\

\bf{0} & \includegraphics[width=.085\textwidth]{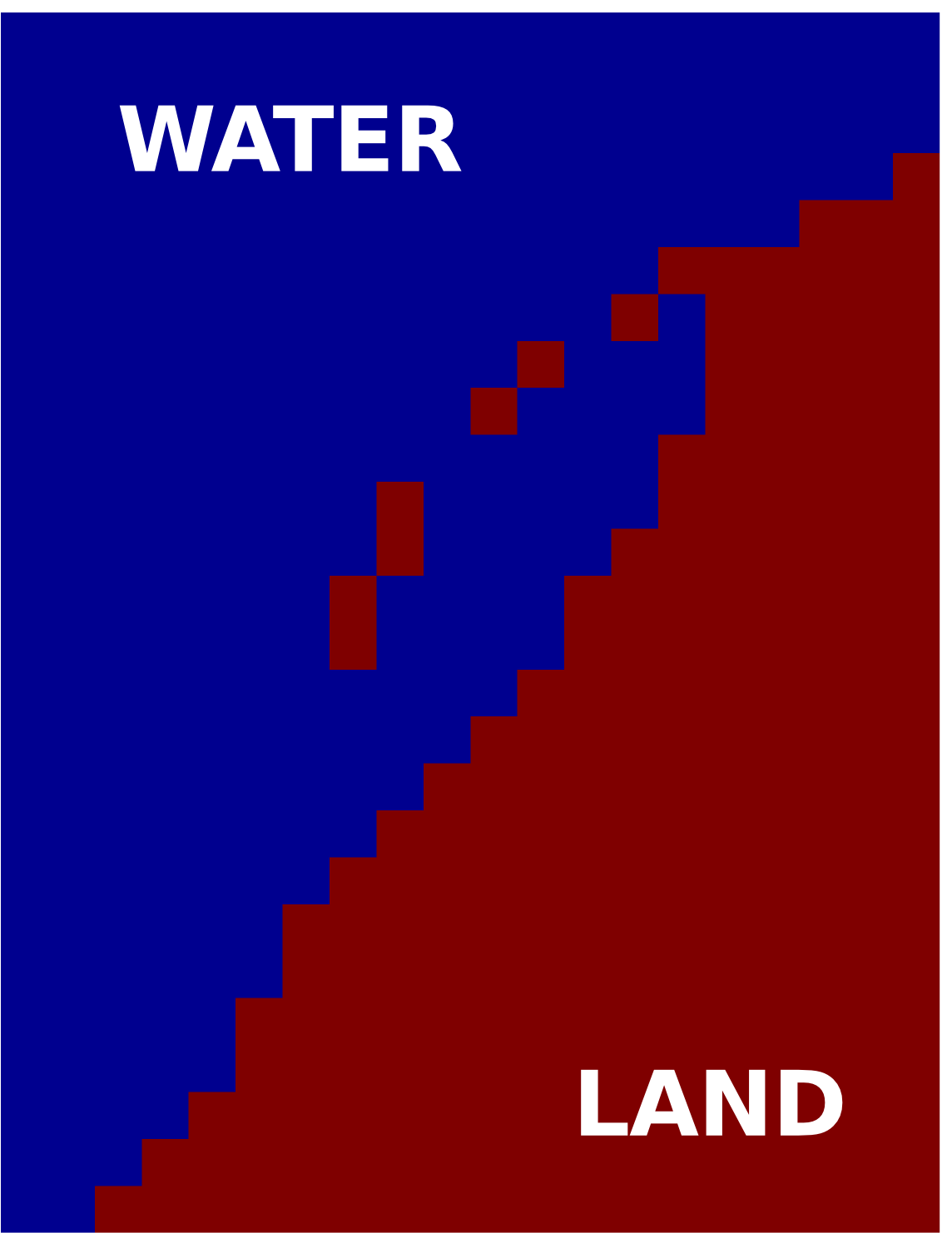} &
\includegraphics[width=.085\textwidth]{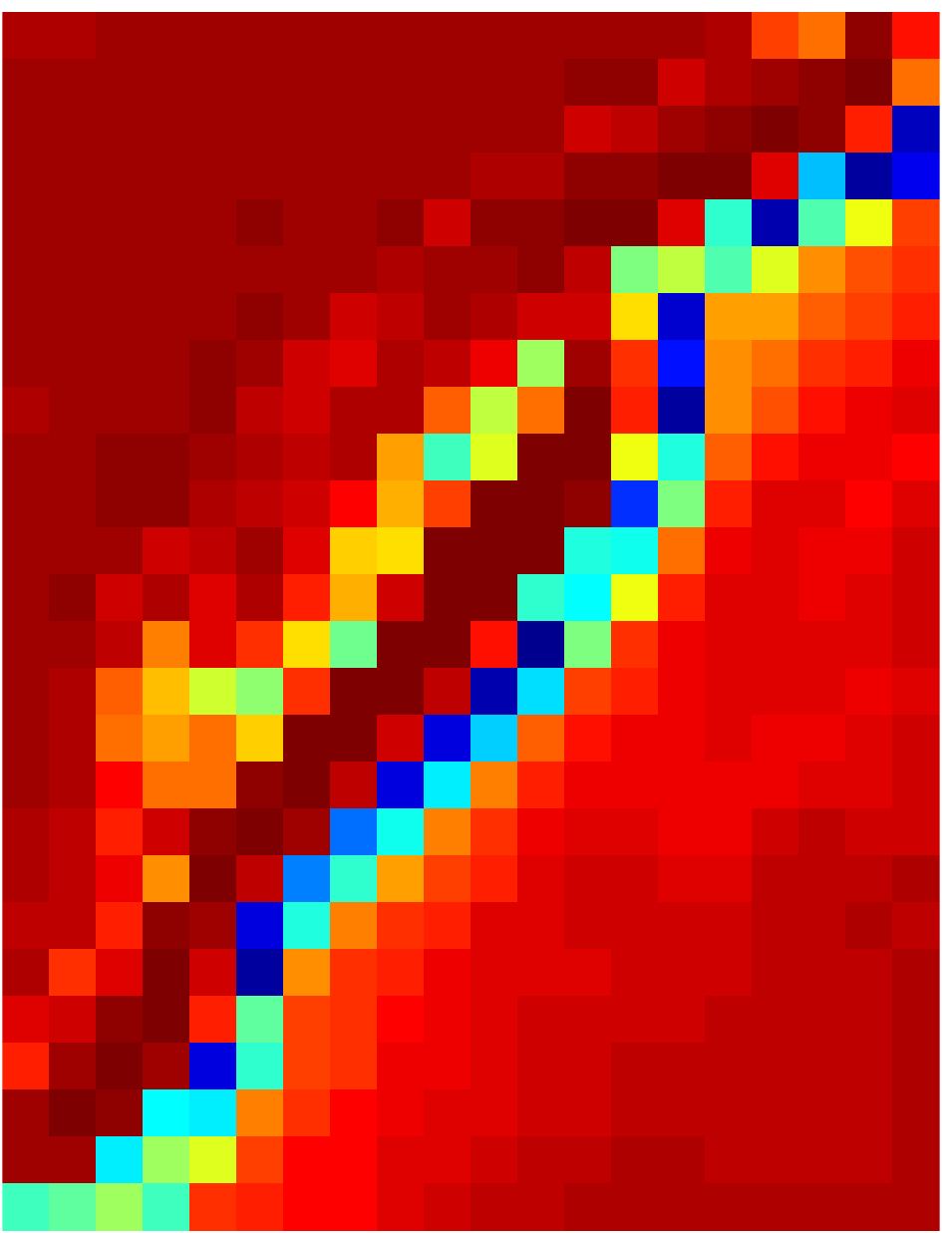} &
\includegraphics[width=.085\textwidth]{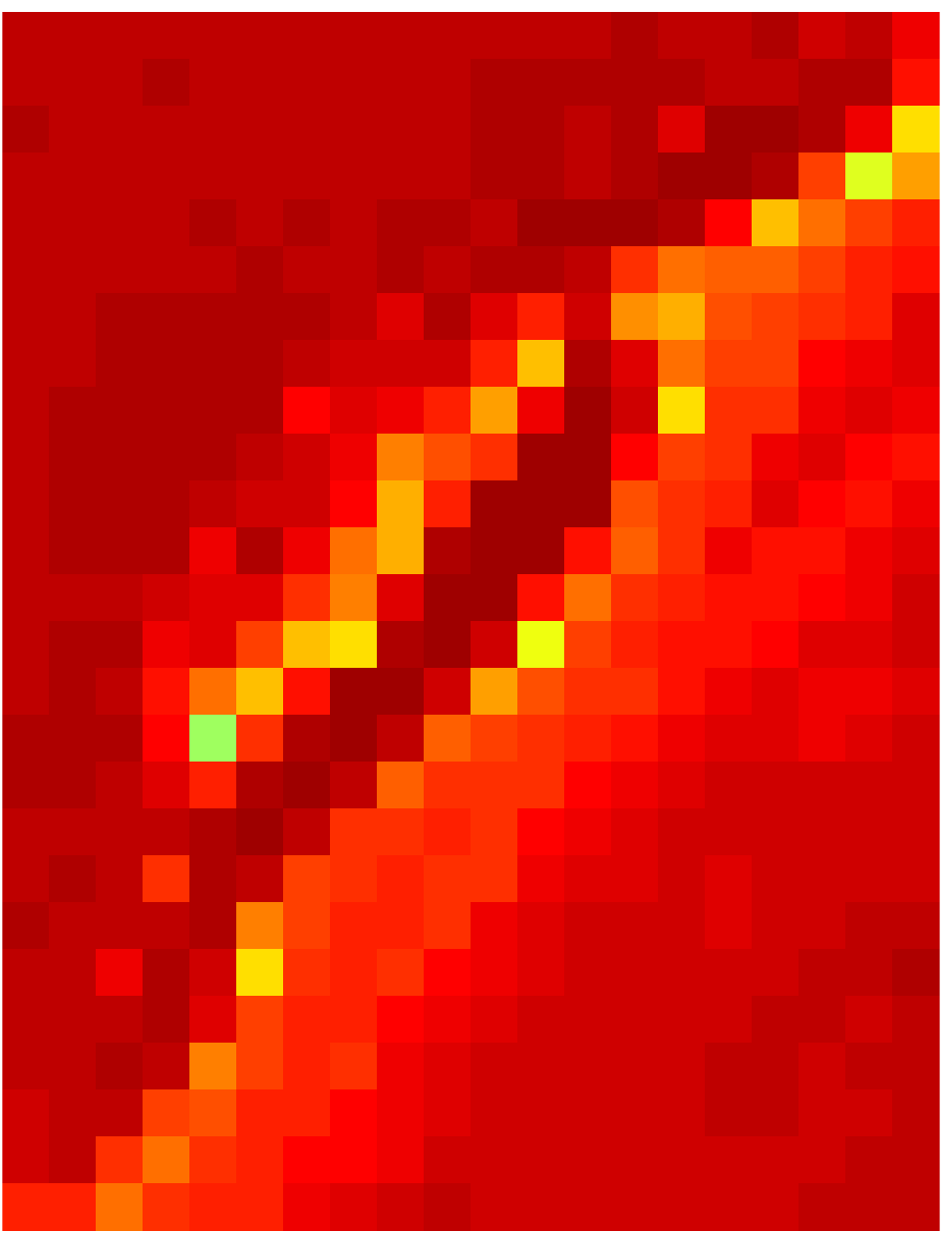} &
\includegraphics[width=.085\textwidth]{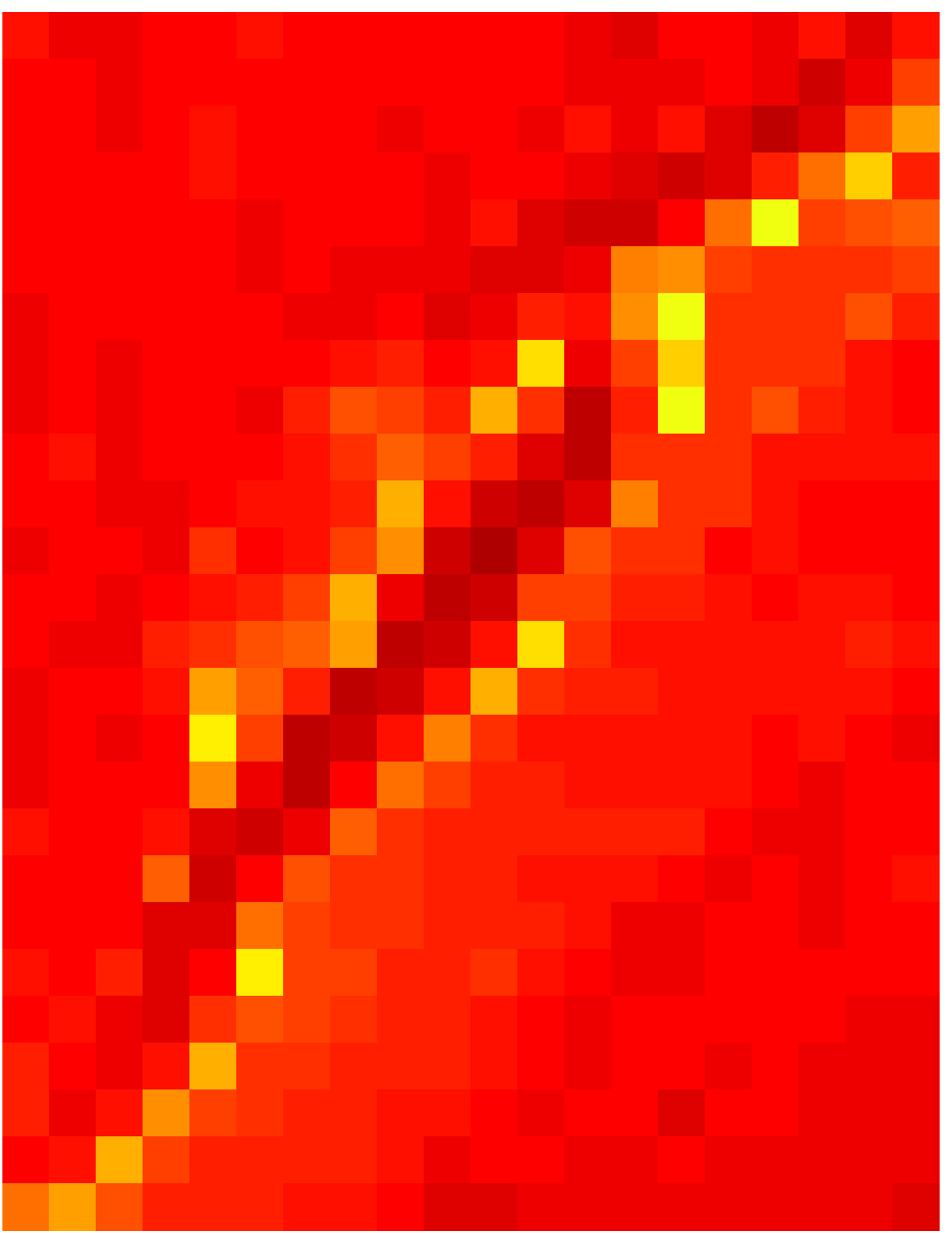} &
\includegraphics[width=.085\textwidth]{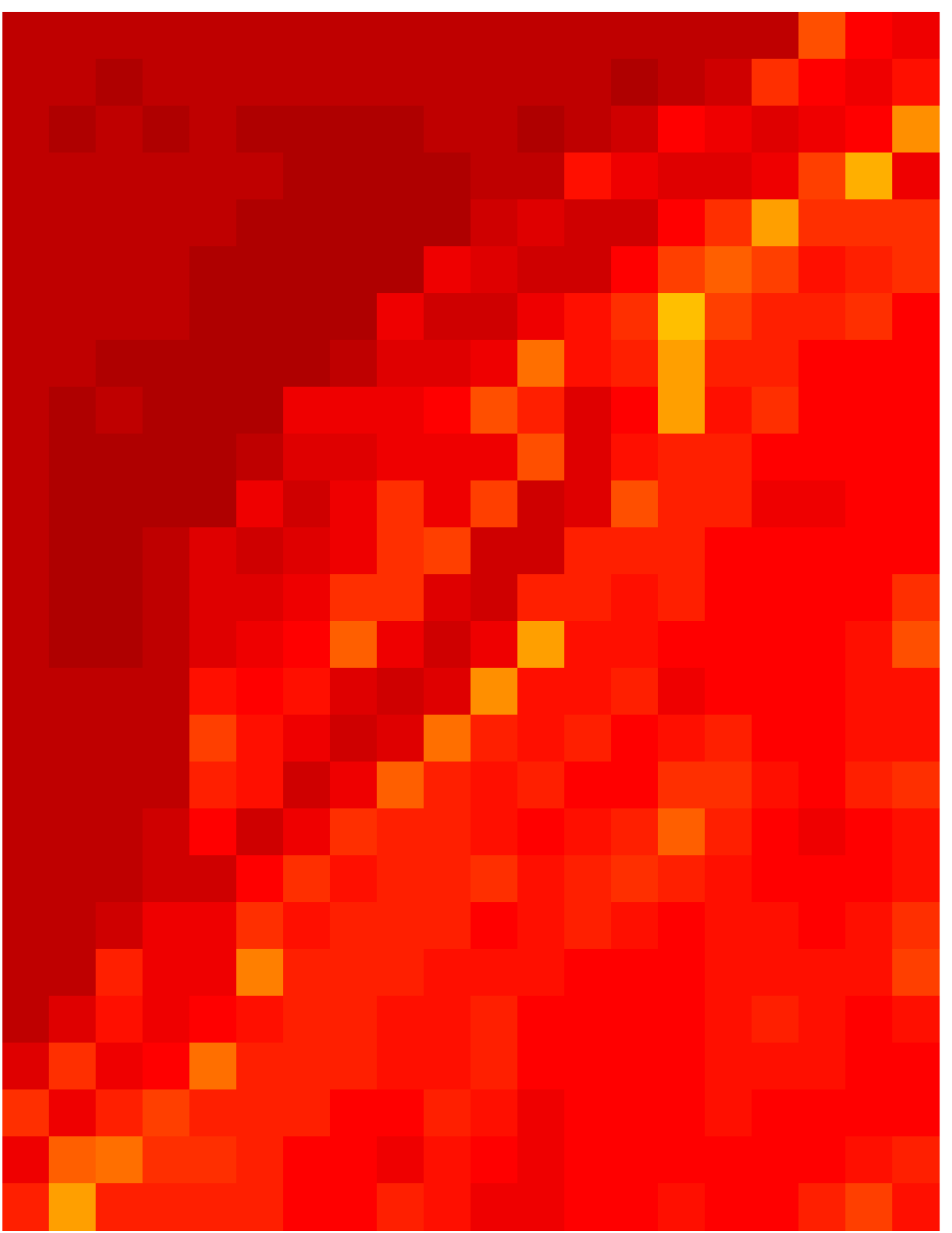} 
& \includegraphics[trim=400 5 0 0, clip, width=.022\textwidth]{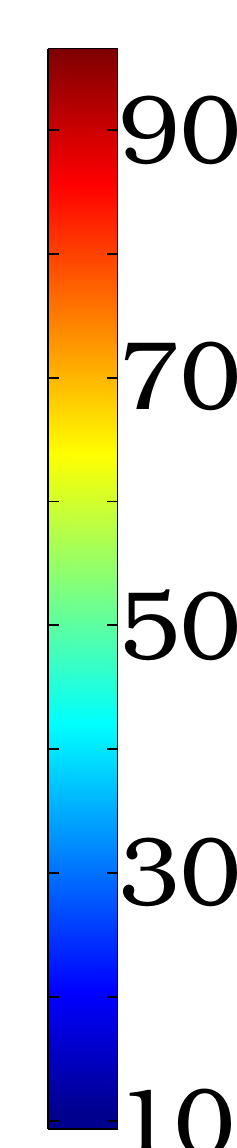} 
\\

\bf{14} & 
\includegraphics[width=.085\textwidth]{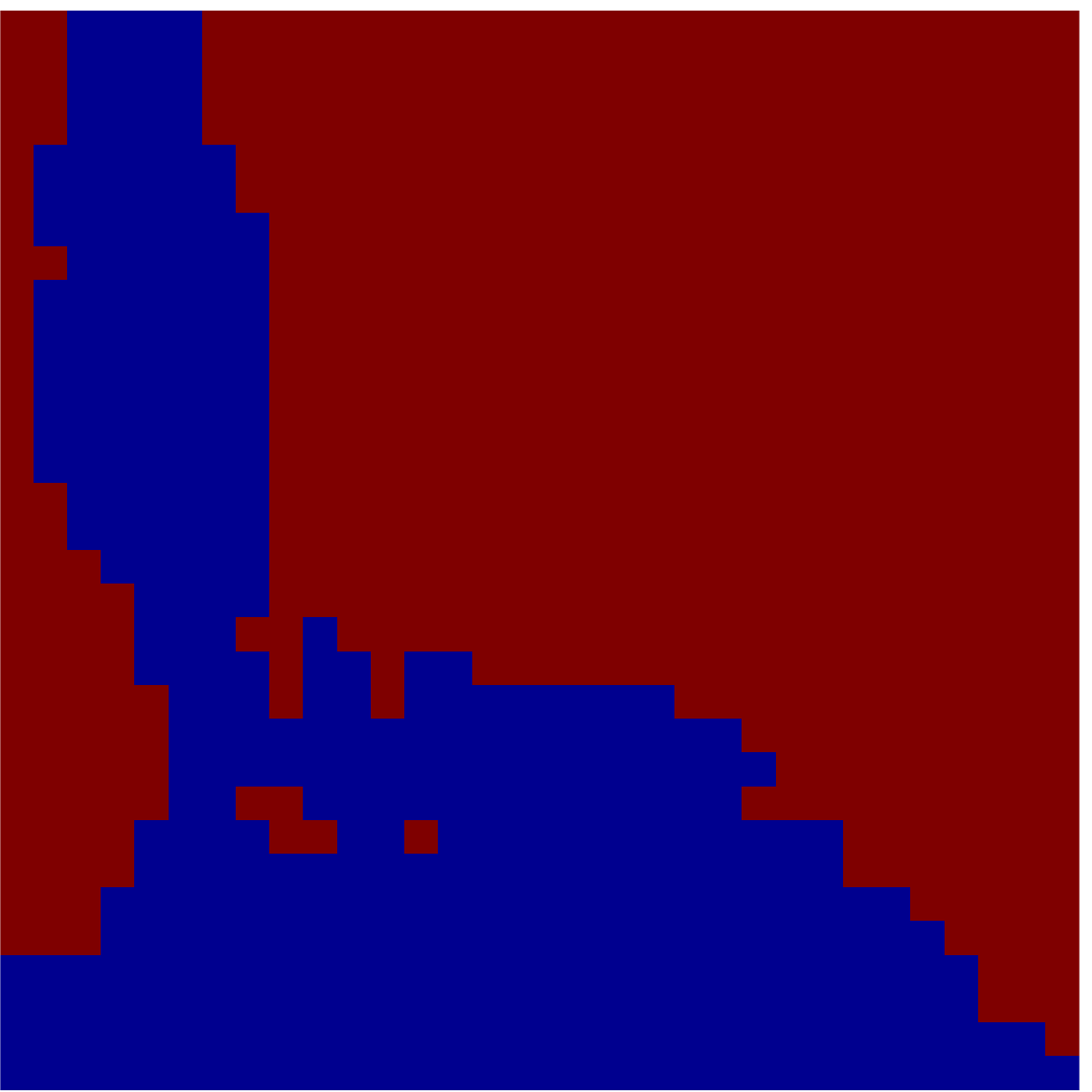} &
\includegraphics[width=.085\textwidth]{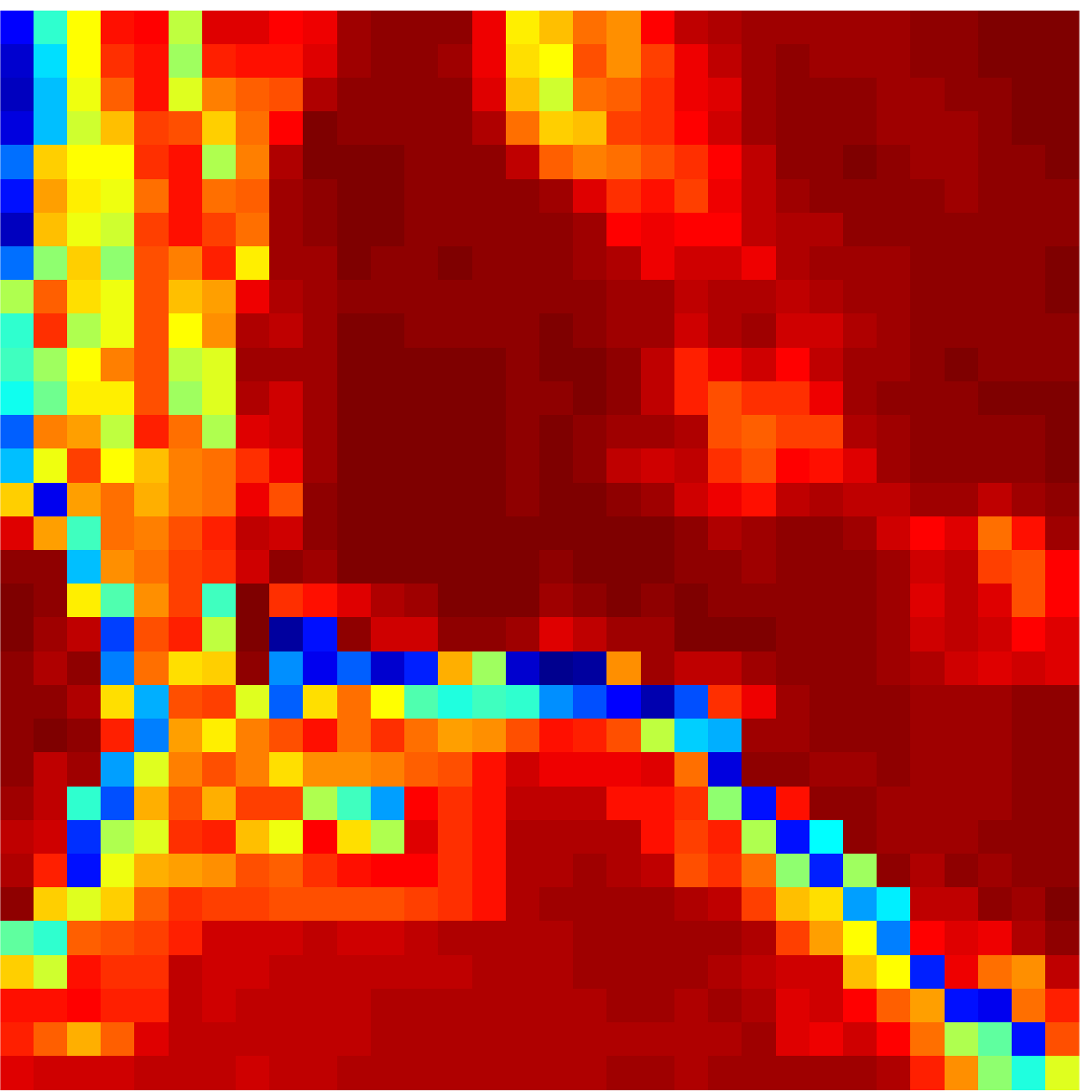} &
\includegraphics[width=.085\textwidth]{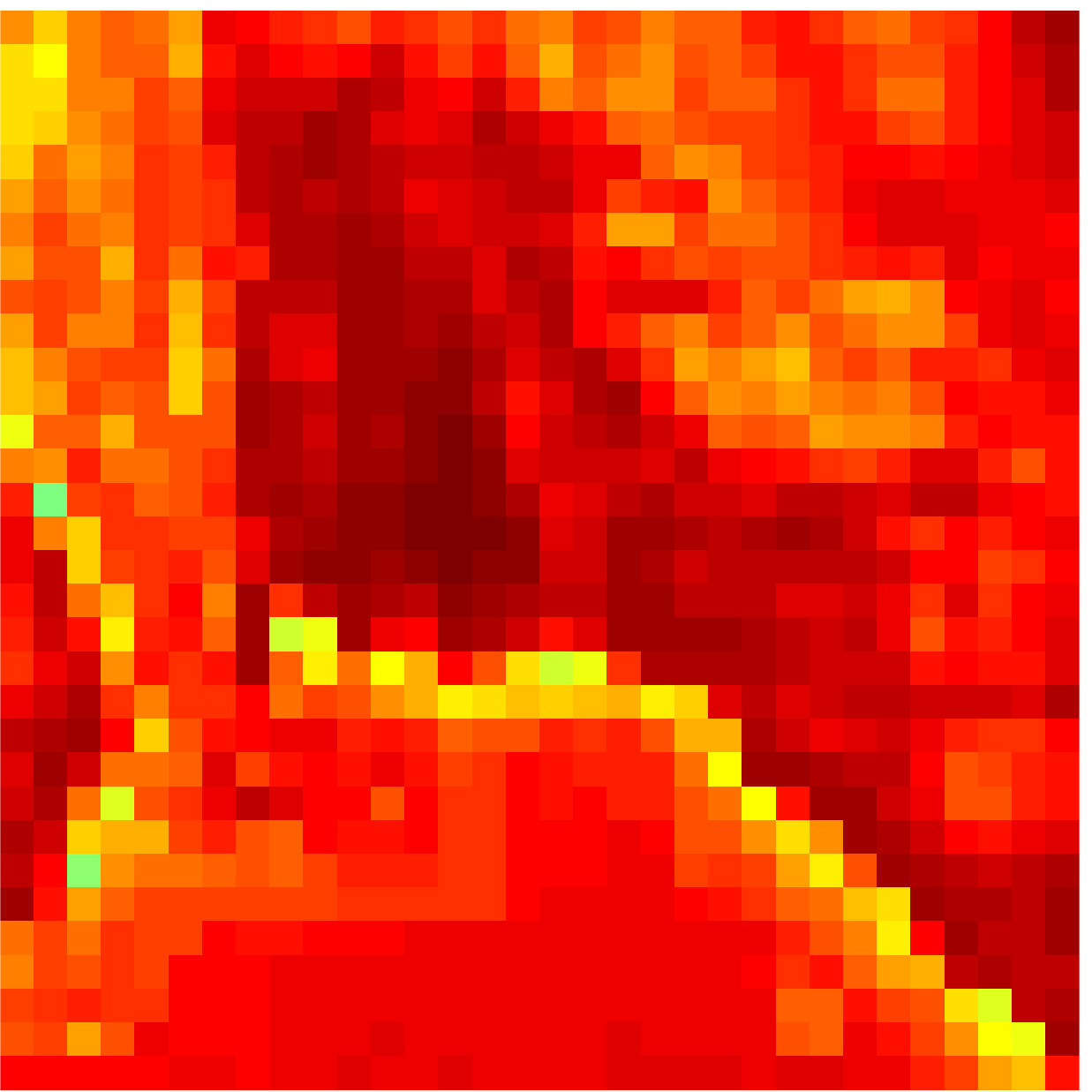} &
\includegraphics[width=.085\textwidth]{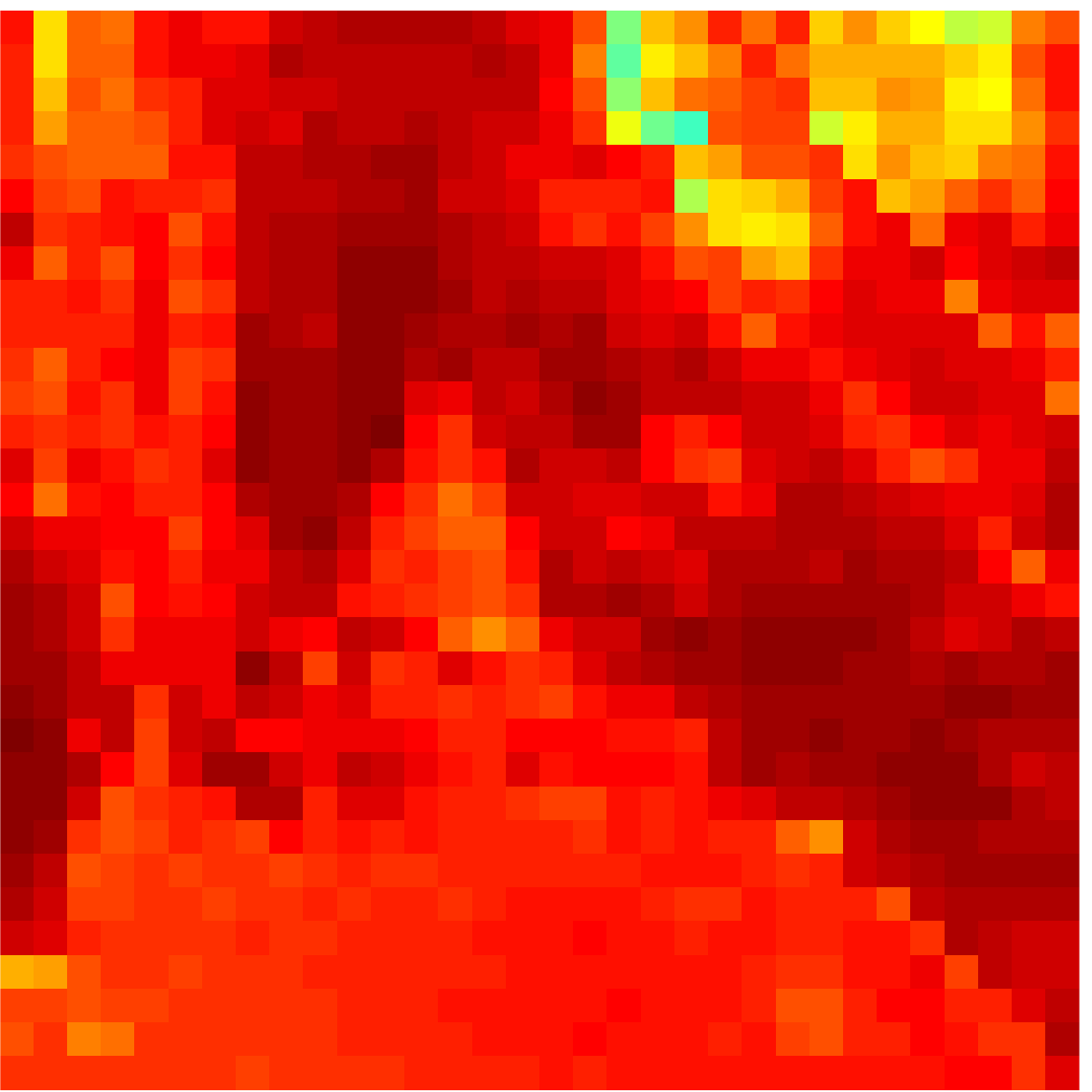} &
\includegraphics[width=.085\textwidth]{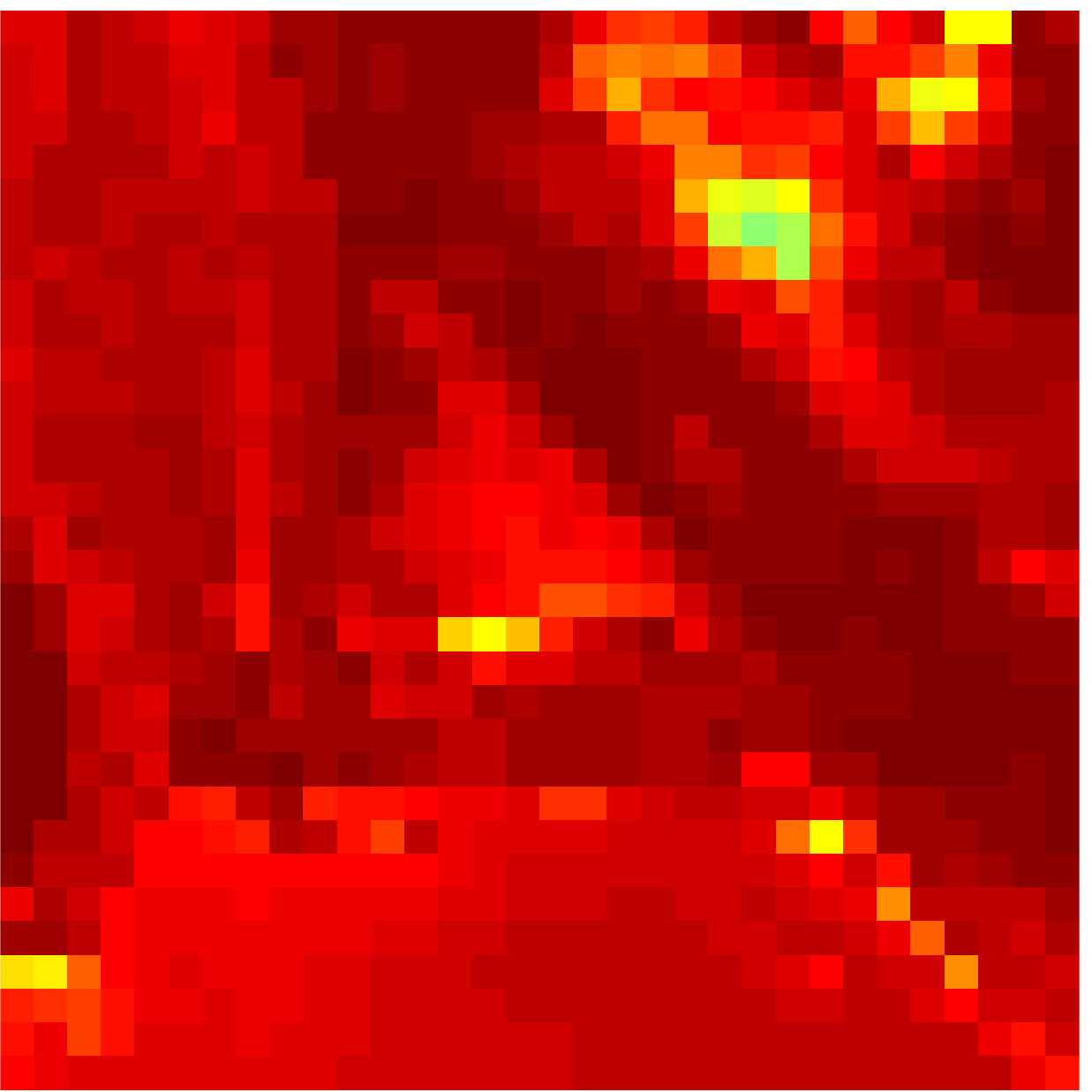} 
& \includegraphics[trim=400 5 0 0, clip, width=.02\textwidth]{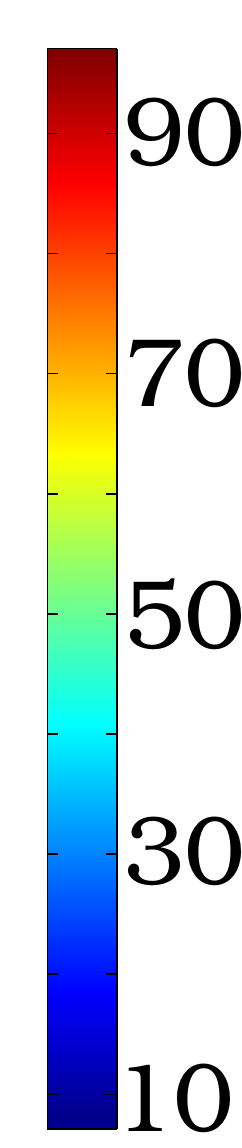} 
\\

\bf{17} & 
\includegraphics[width=.085\textwidth]{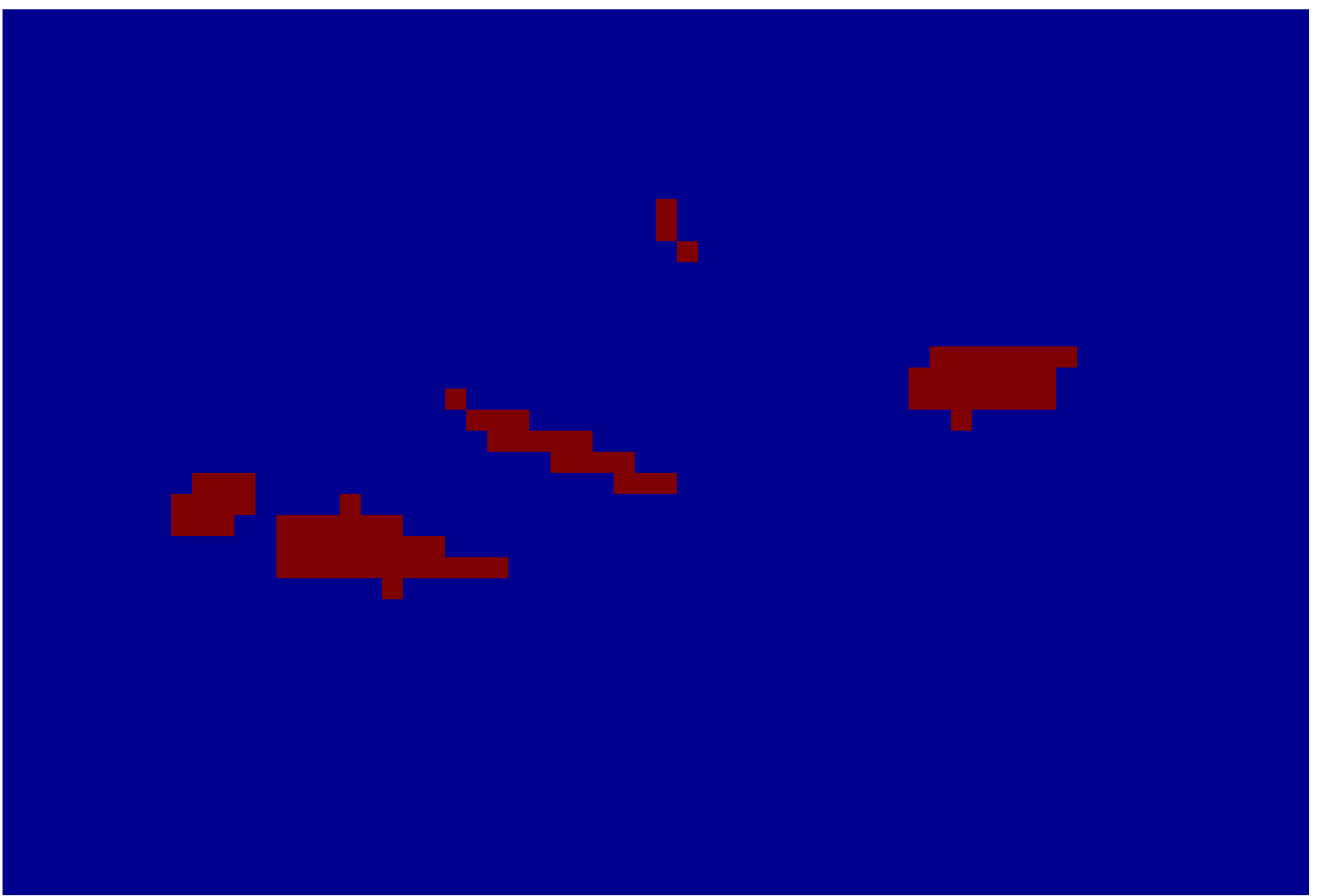} &
\includegraphics[width=.085\textwidth]{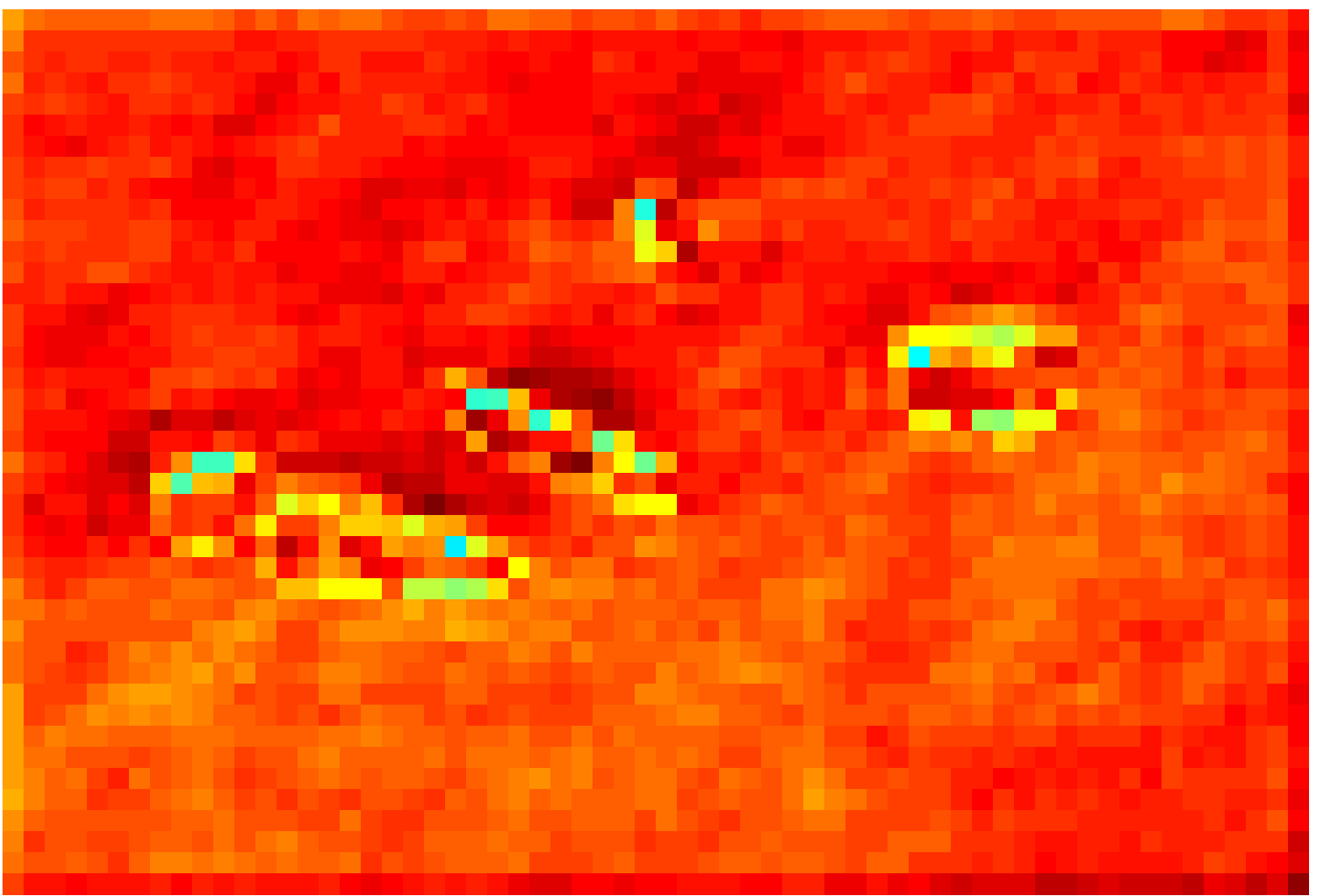} &
\includegraphics[width=.085\textwidth]{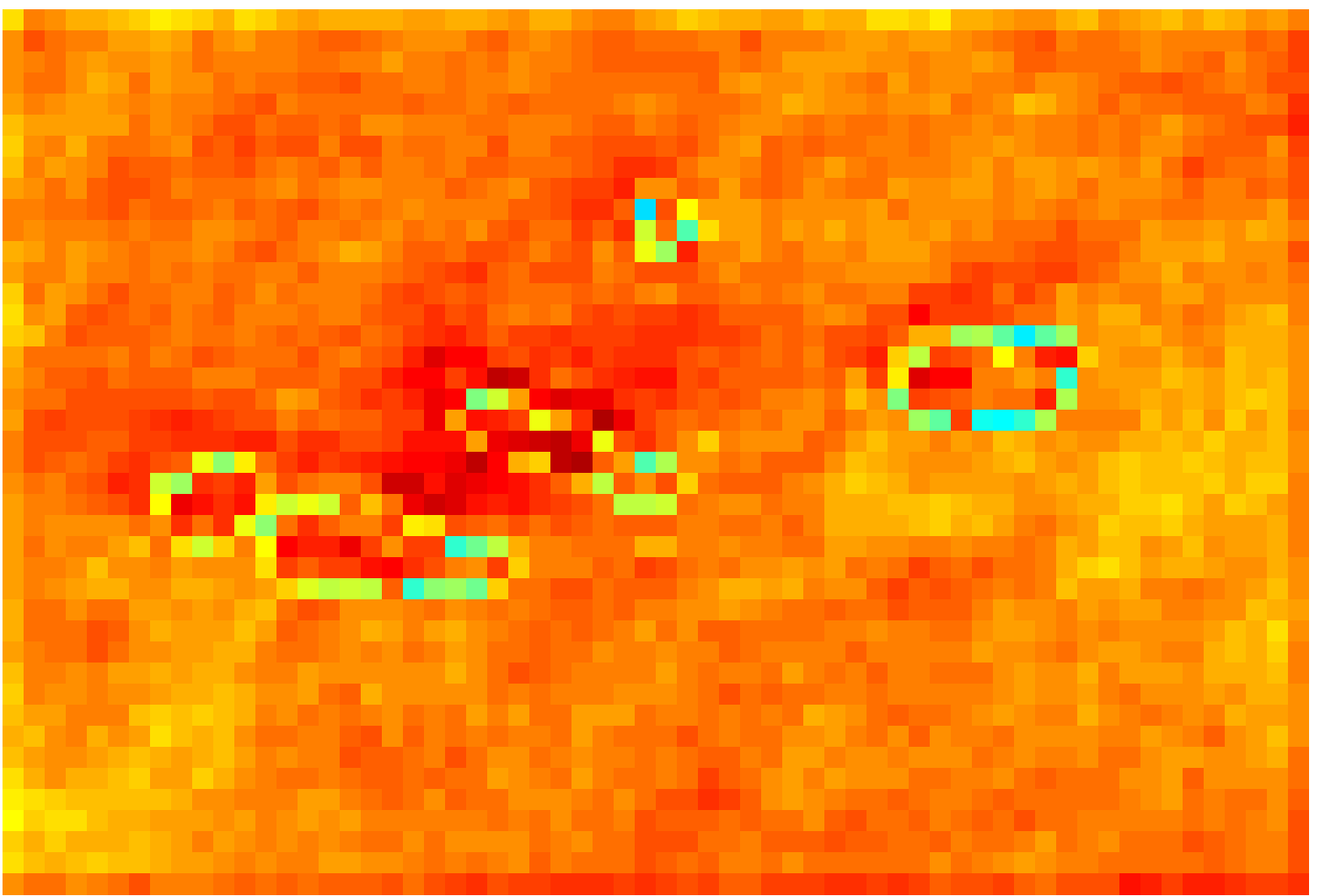} &
\includegraphics[width=.085\textwidth]{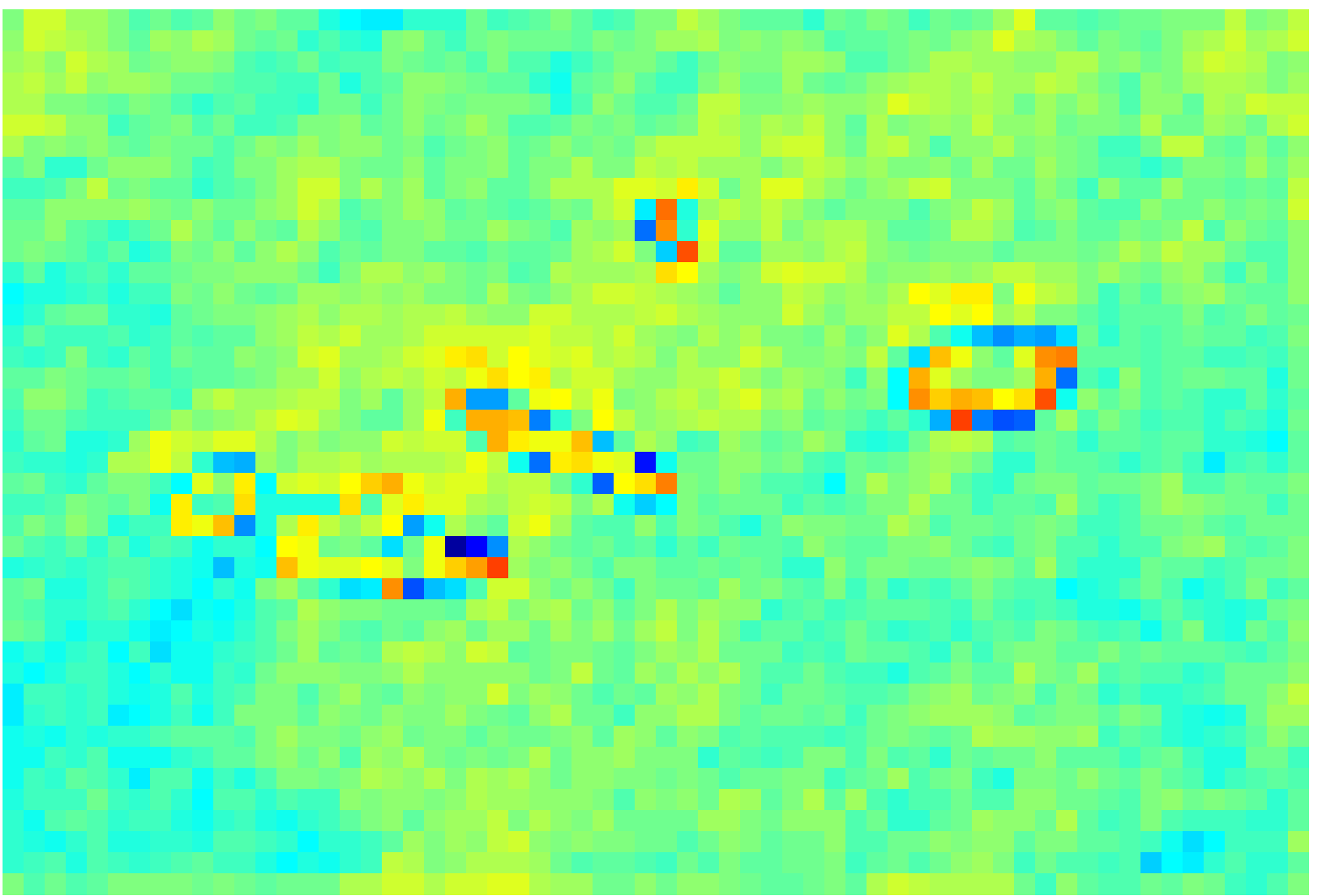} &
\includegraphics[width=.085\textwidth]{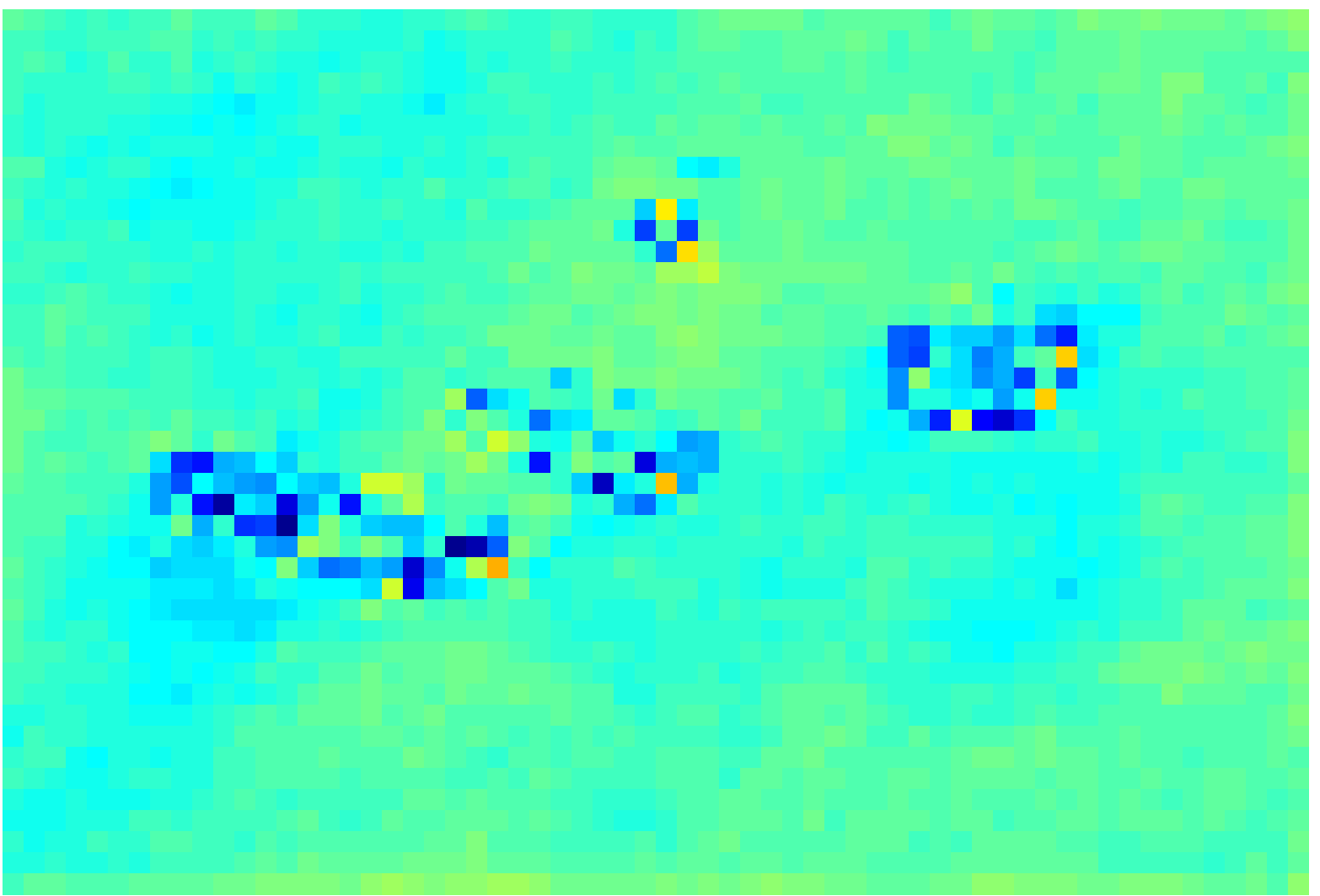} 
& \includegraphics[trim=400 5 0 0, clip, width=.015\textwidth]{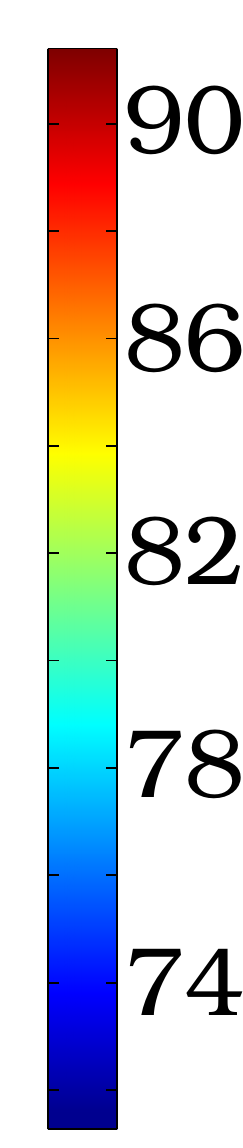} 
\\

\bf{48} & 
\includegraphics[width=.085\textwidth]{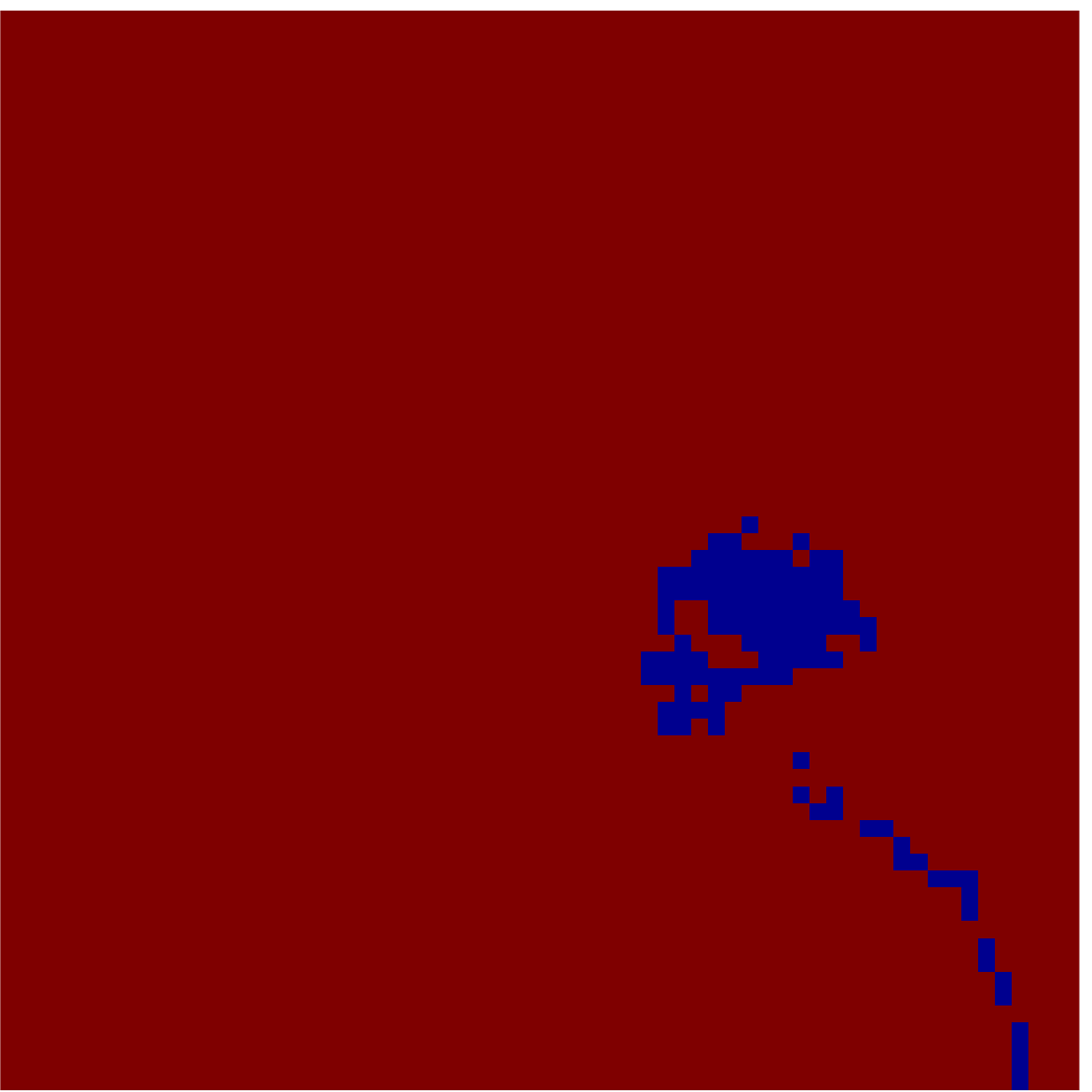} &
\includegraphics[width=.085\textwidth]{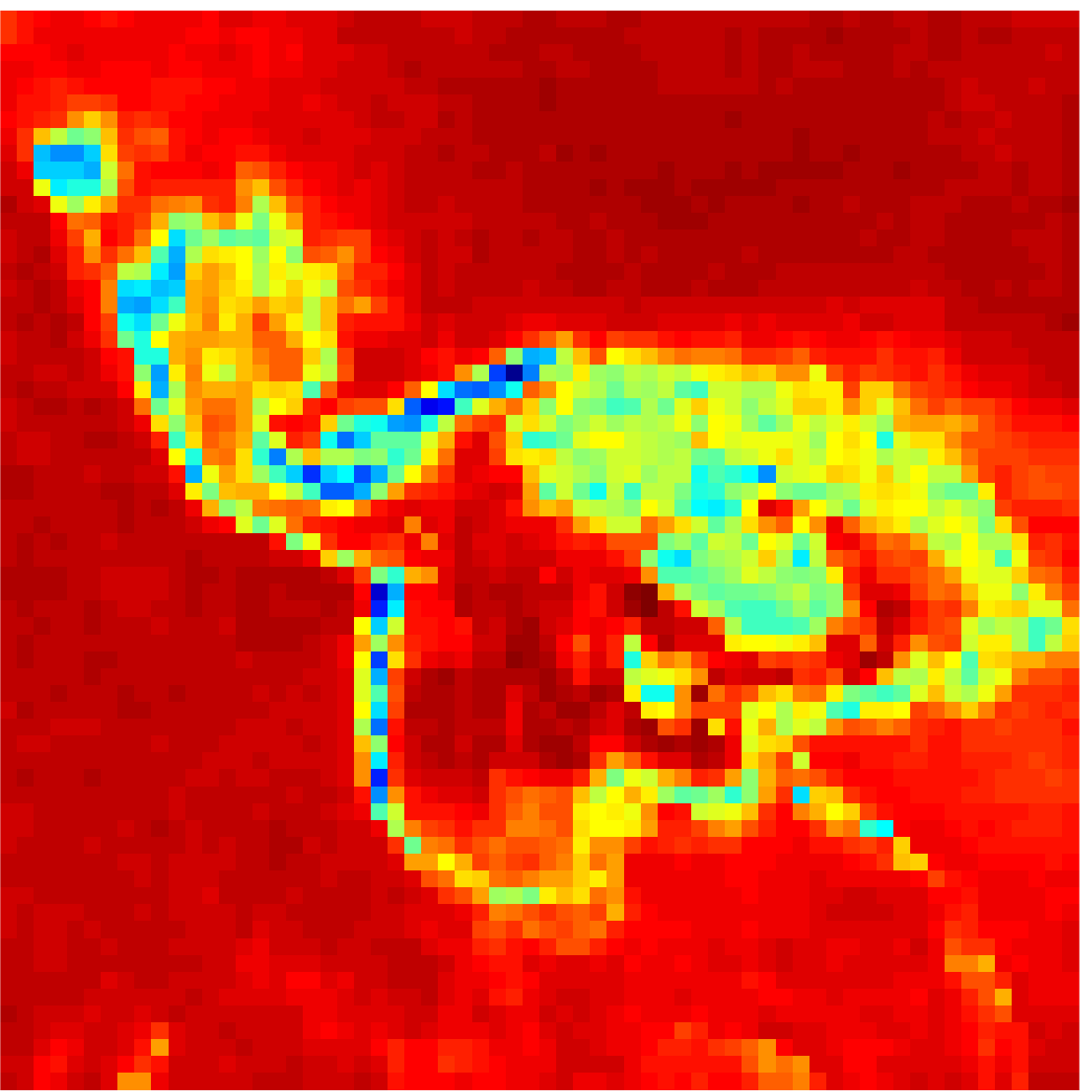} &
\includegraphics[width=.085\textwidth]{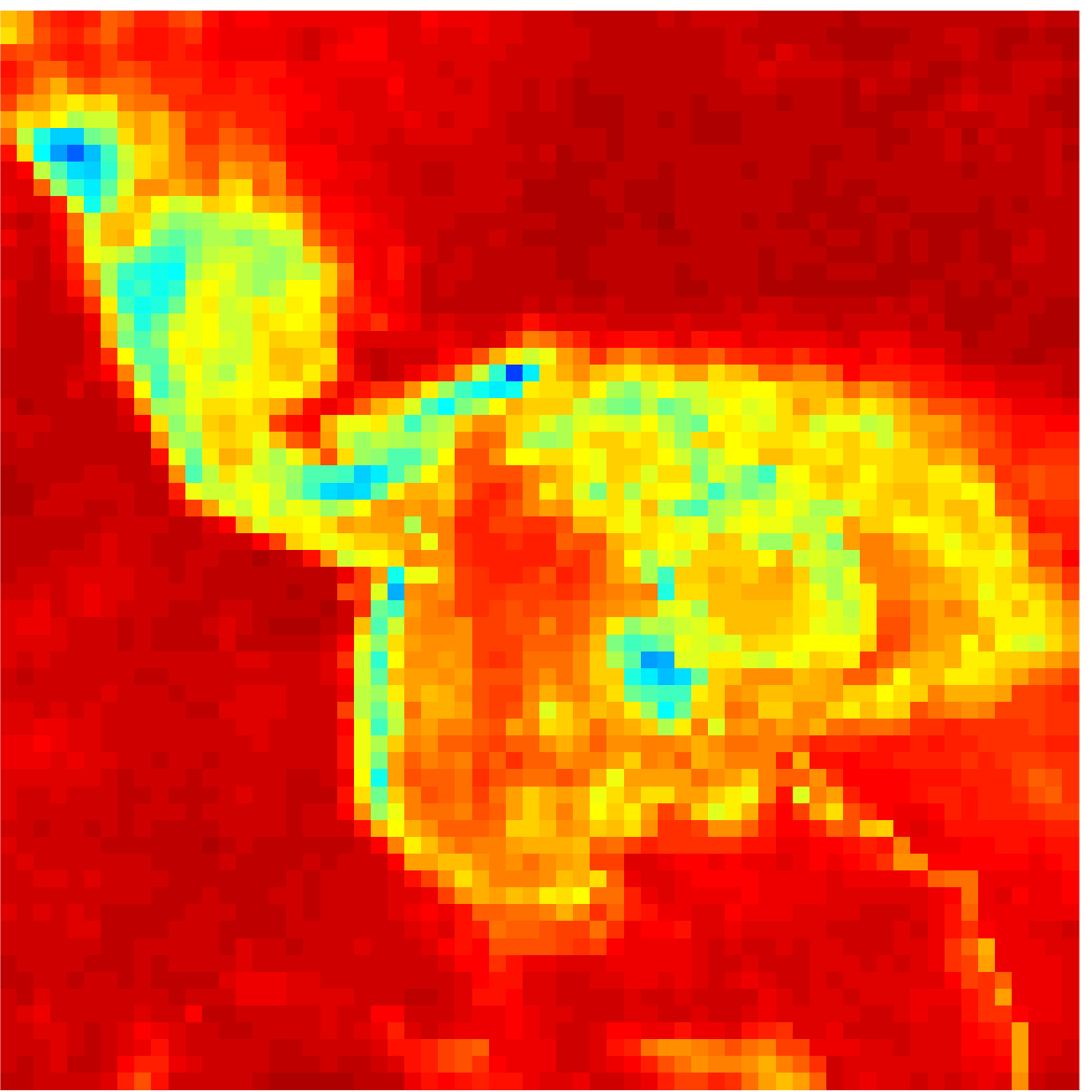} &
\includegraphics[width=.085\textwidth]{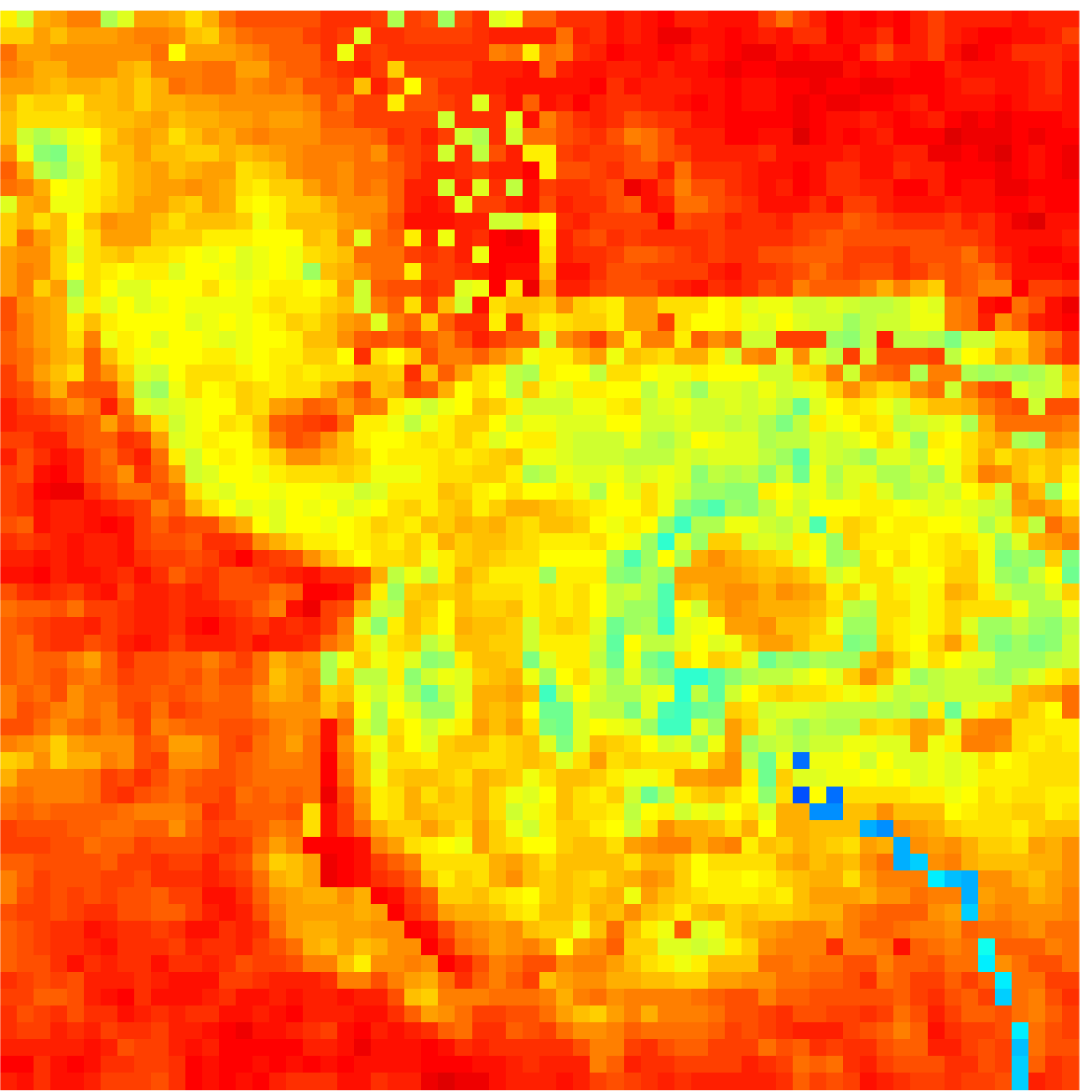} &
\includegraphics[width=.085\textwidth]{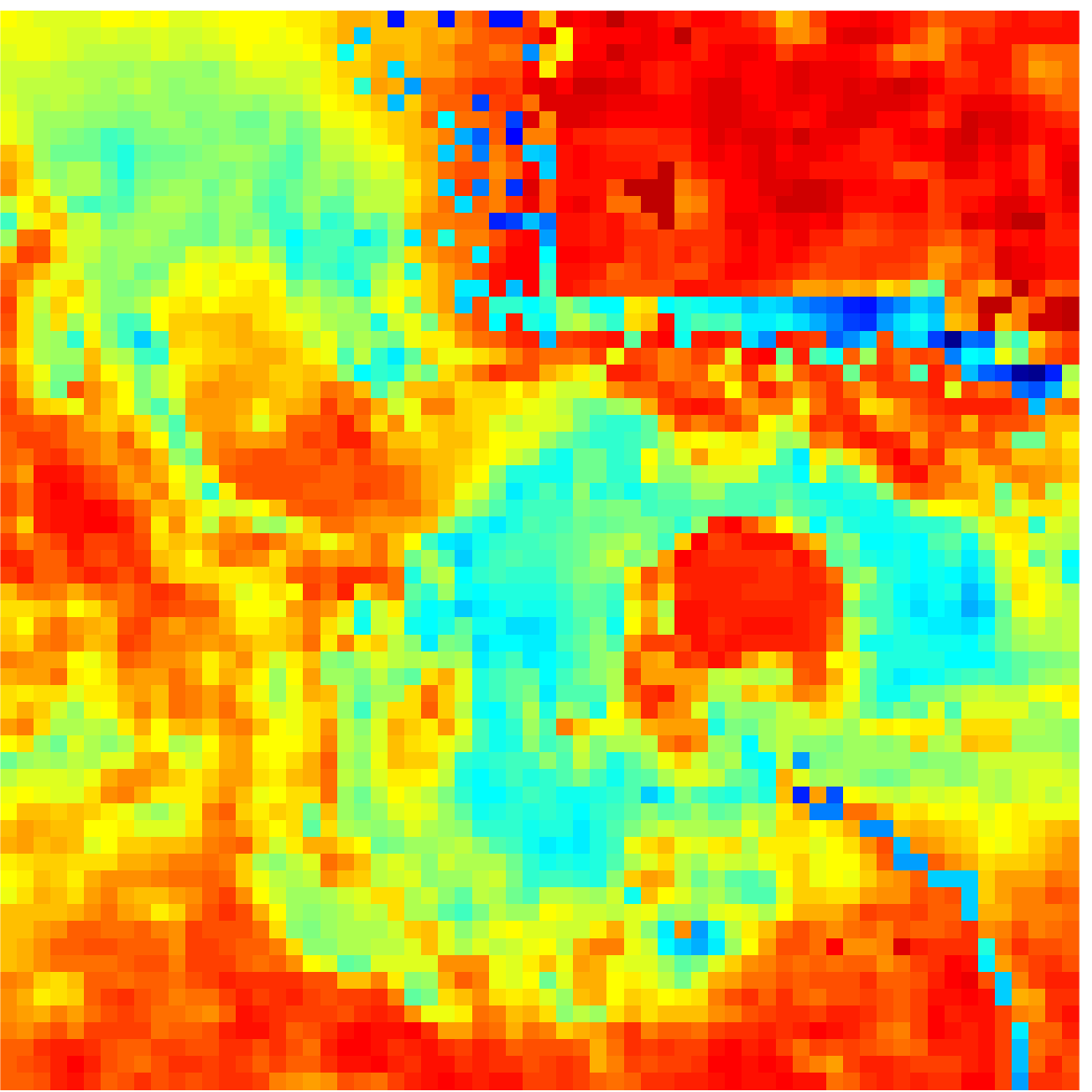} 
& \includegraphics[trim=400 5 0 0, clip, width=.02\textwidth]{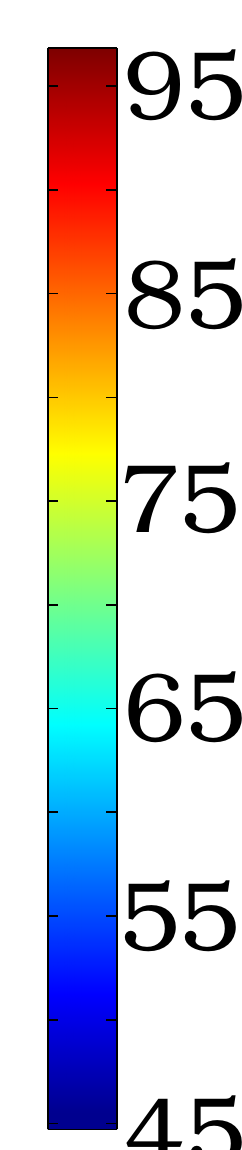} 
\\

\bf{63} & 
\includegraphics[width=.085\textwidth]{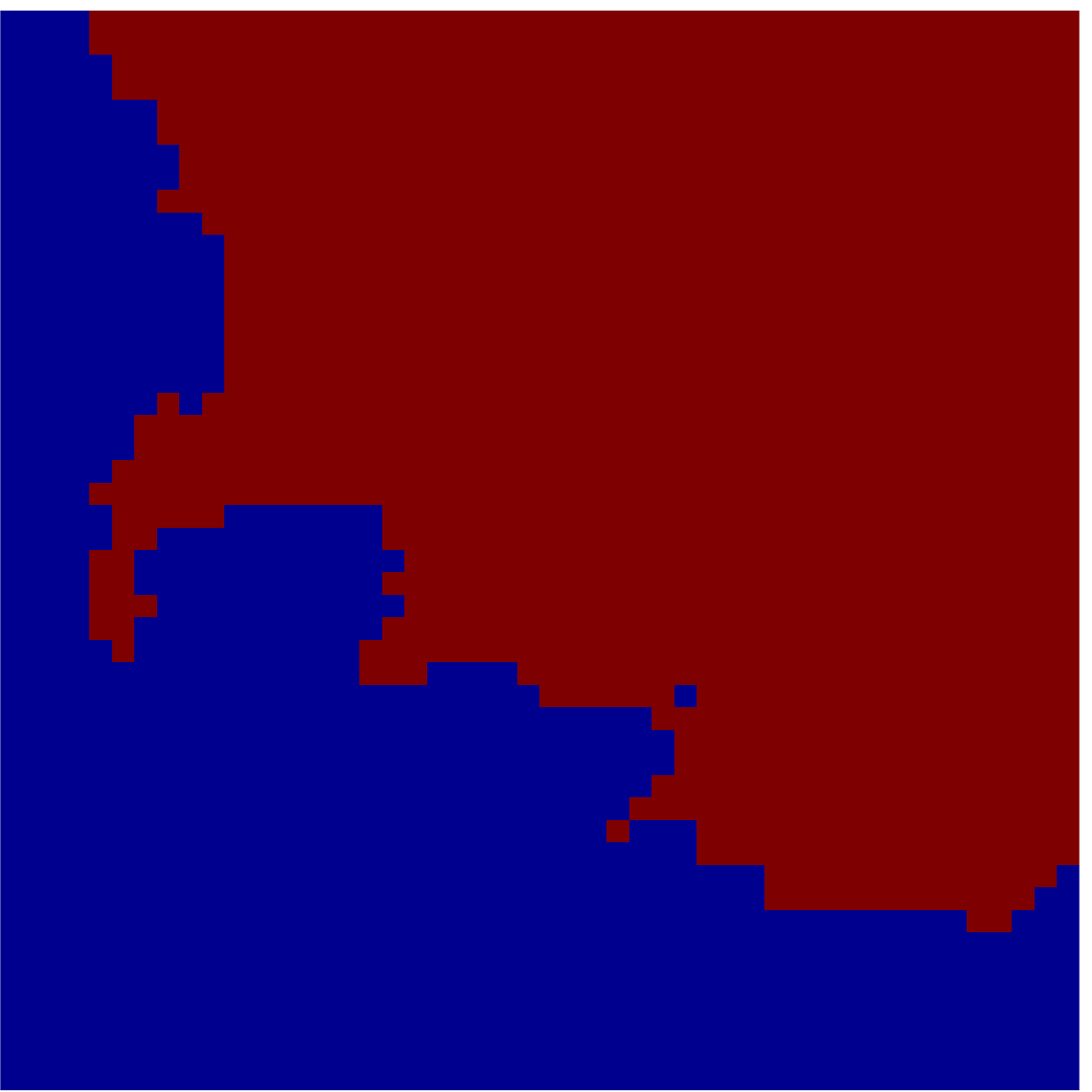} &
\includegraphics[width=.085\textwidth]{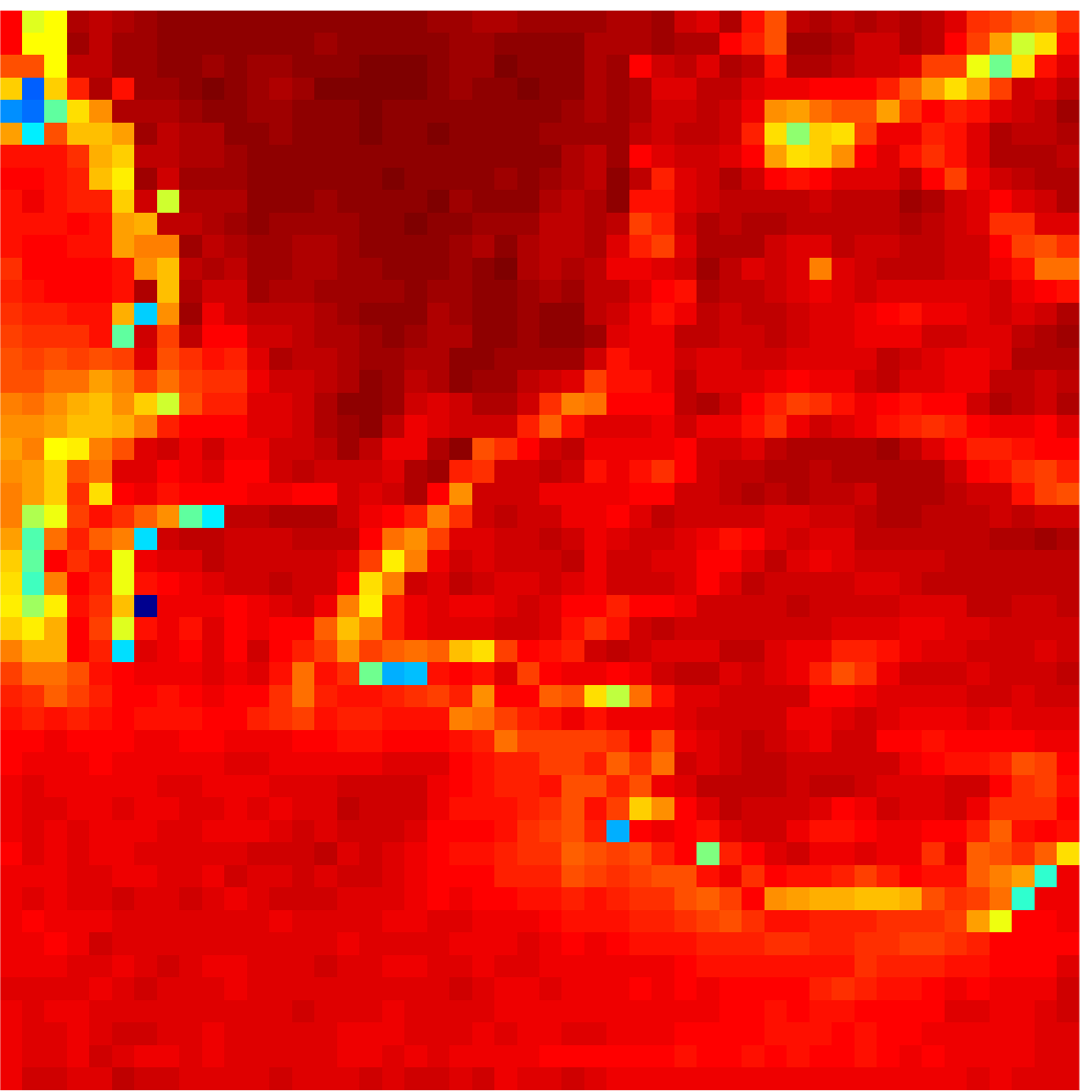} &
\includegraphics[width=.085\textwidth]{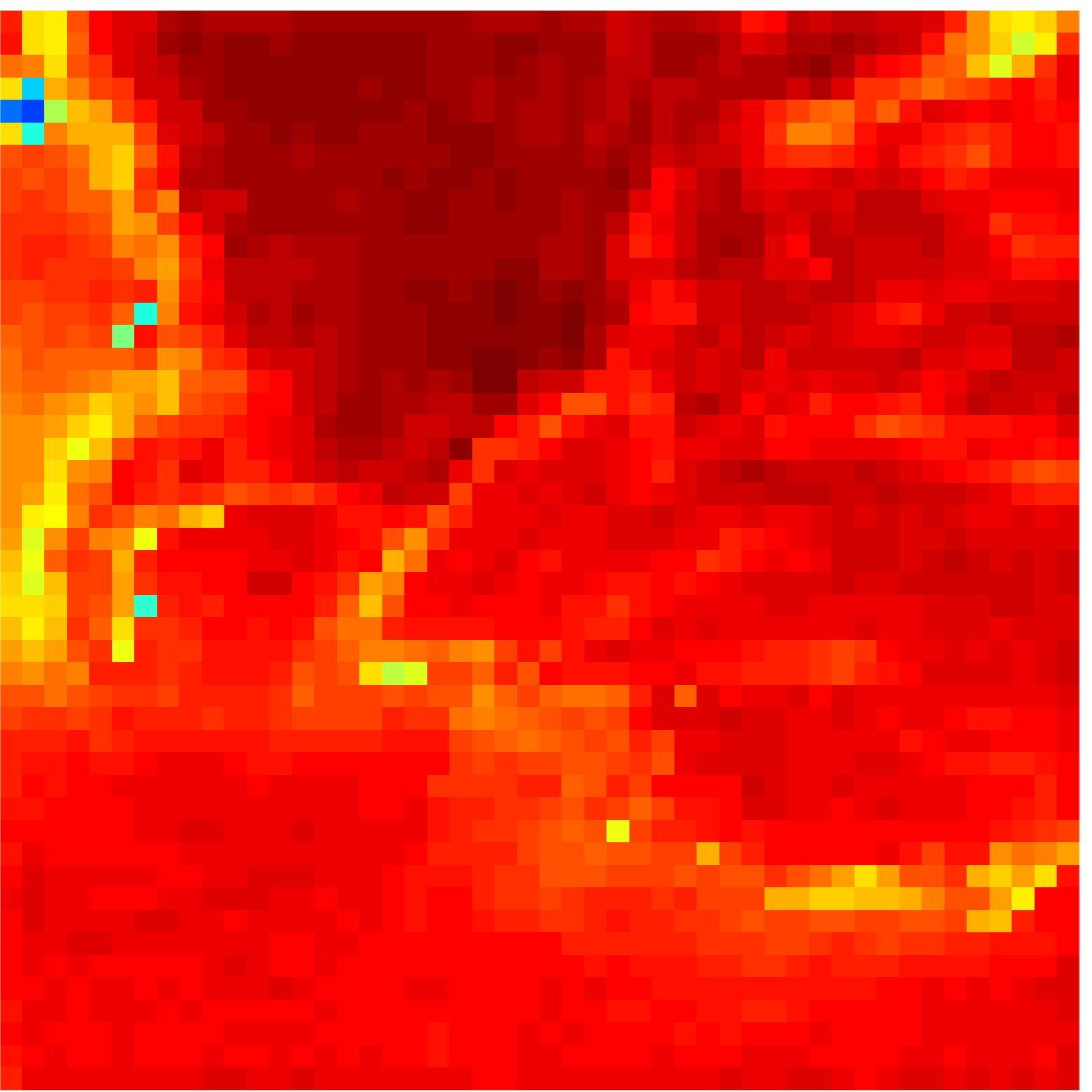} &
\includegraphics[width=.085\textwidth]{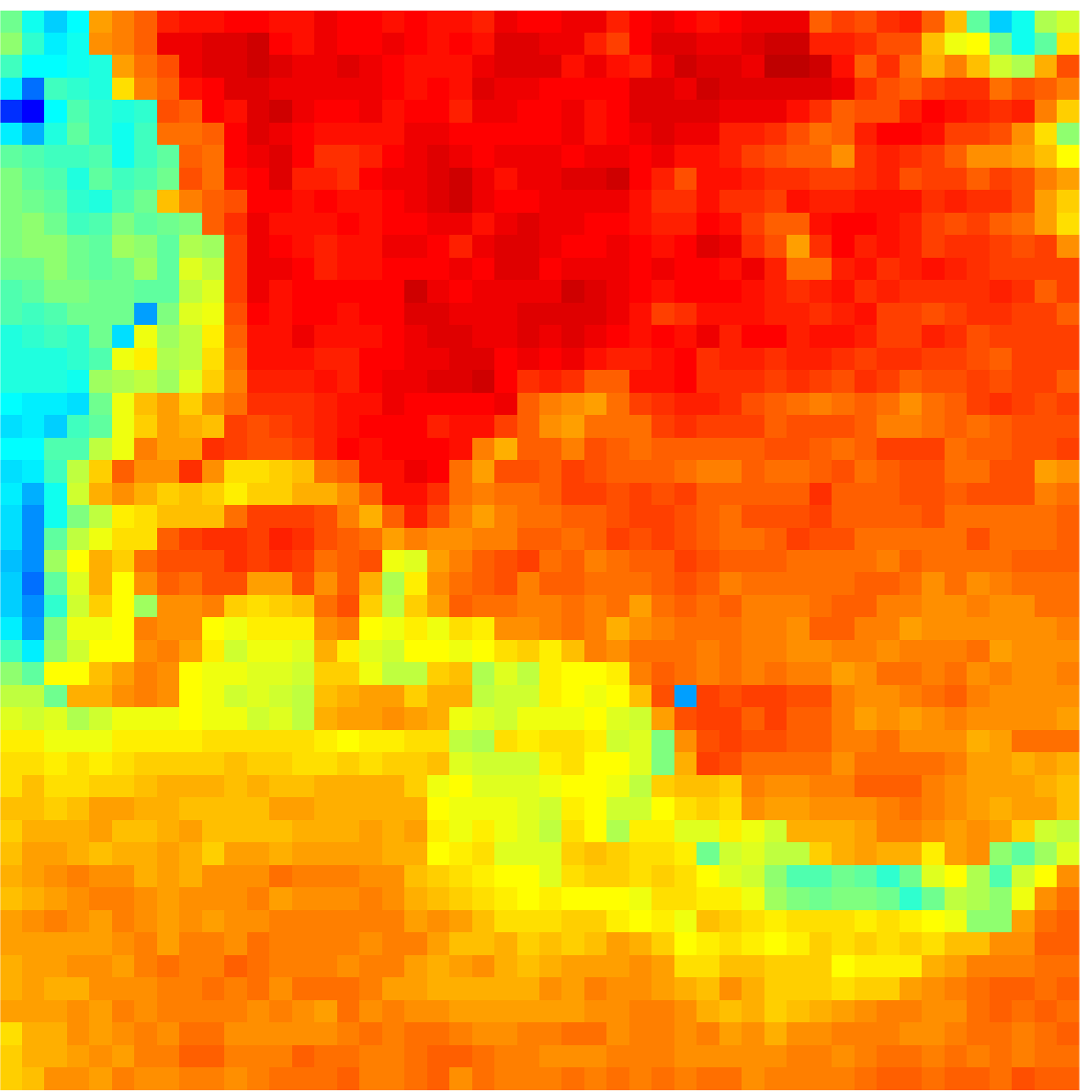} &
\includegraphics[width=.085\textwidth]{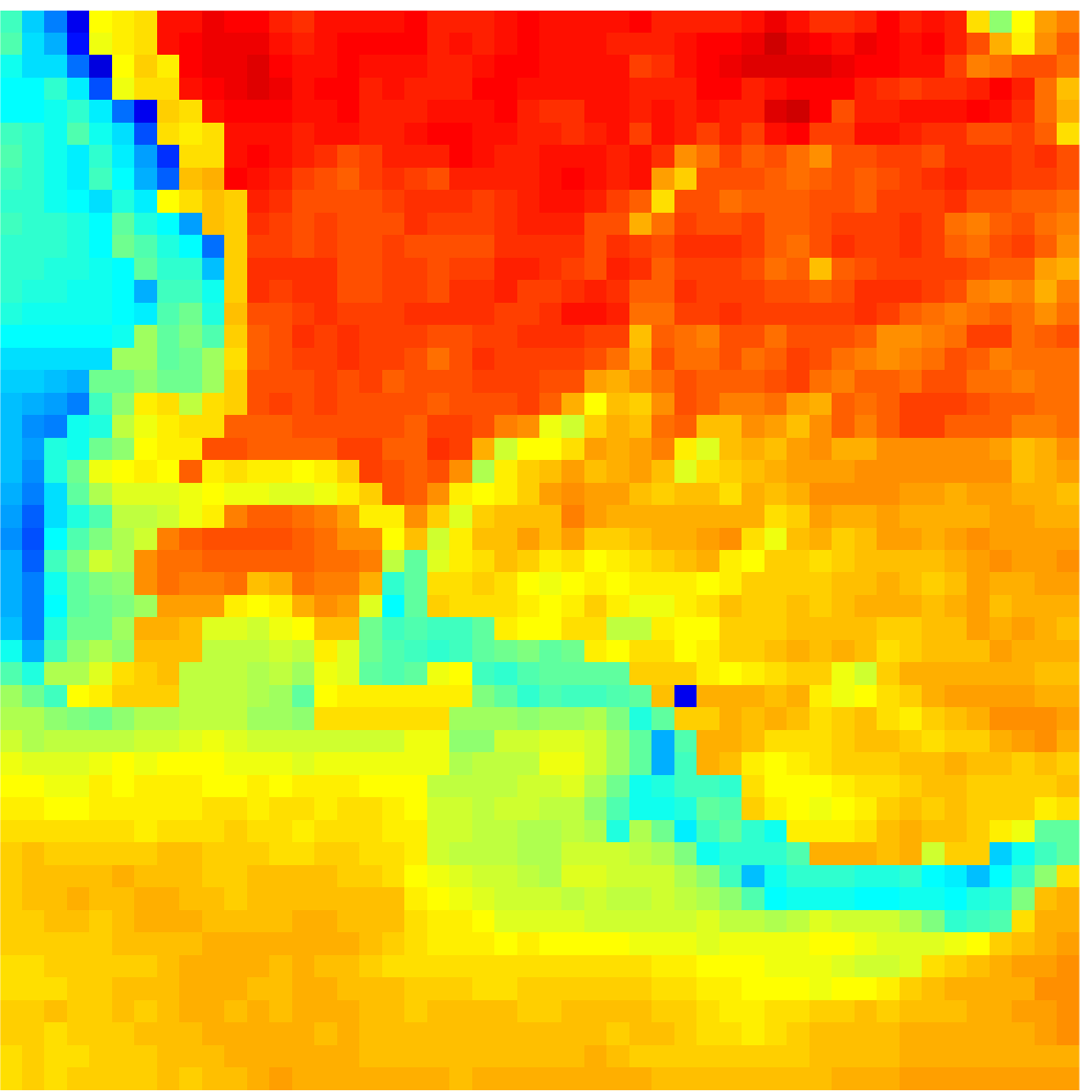} 
& \includegraphics[trim=400 5 0 0, clip, width=.02\textwidth]{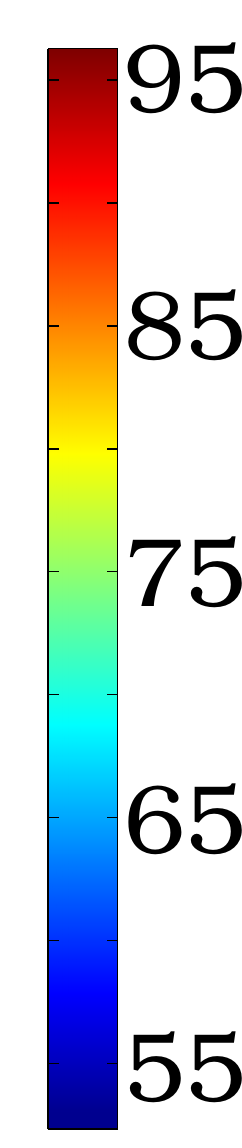} 
\\

\bf{83} & 
\includegraphics[width=.085\textwidth]{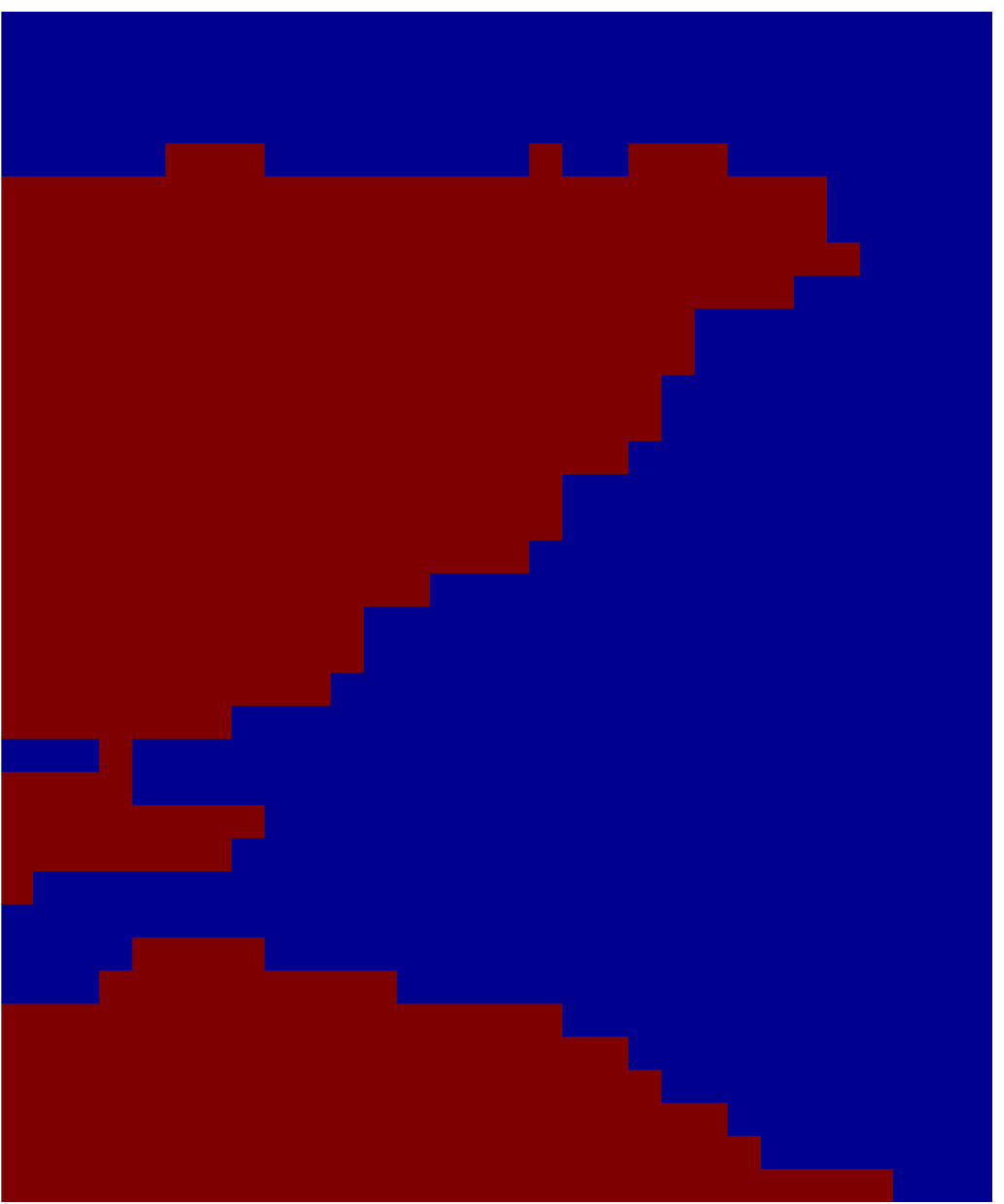} &
\includegraphics[width=.085\textwidth]{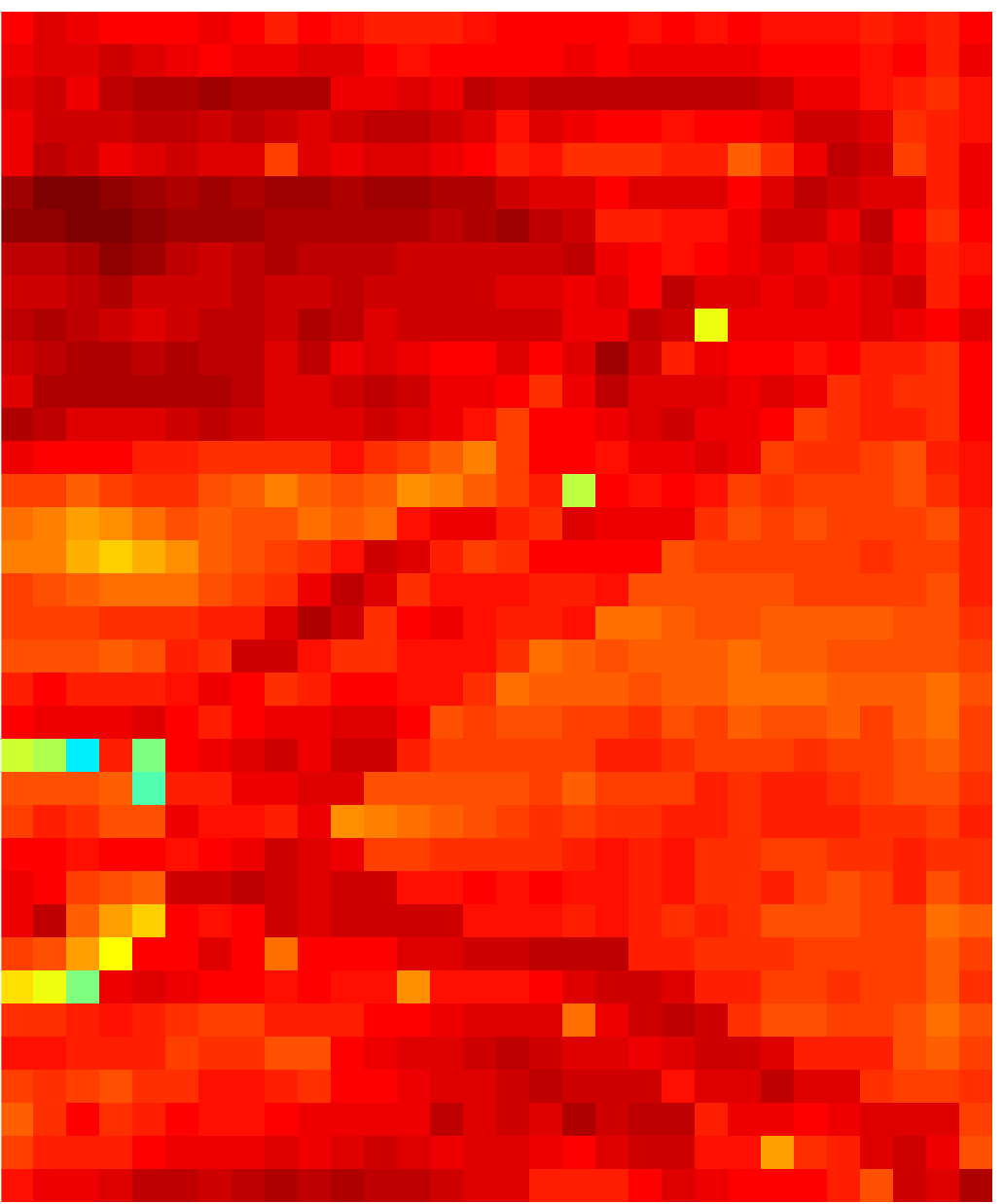} &
\includegraphics[width=.085\textwidth]{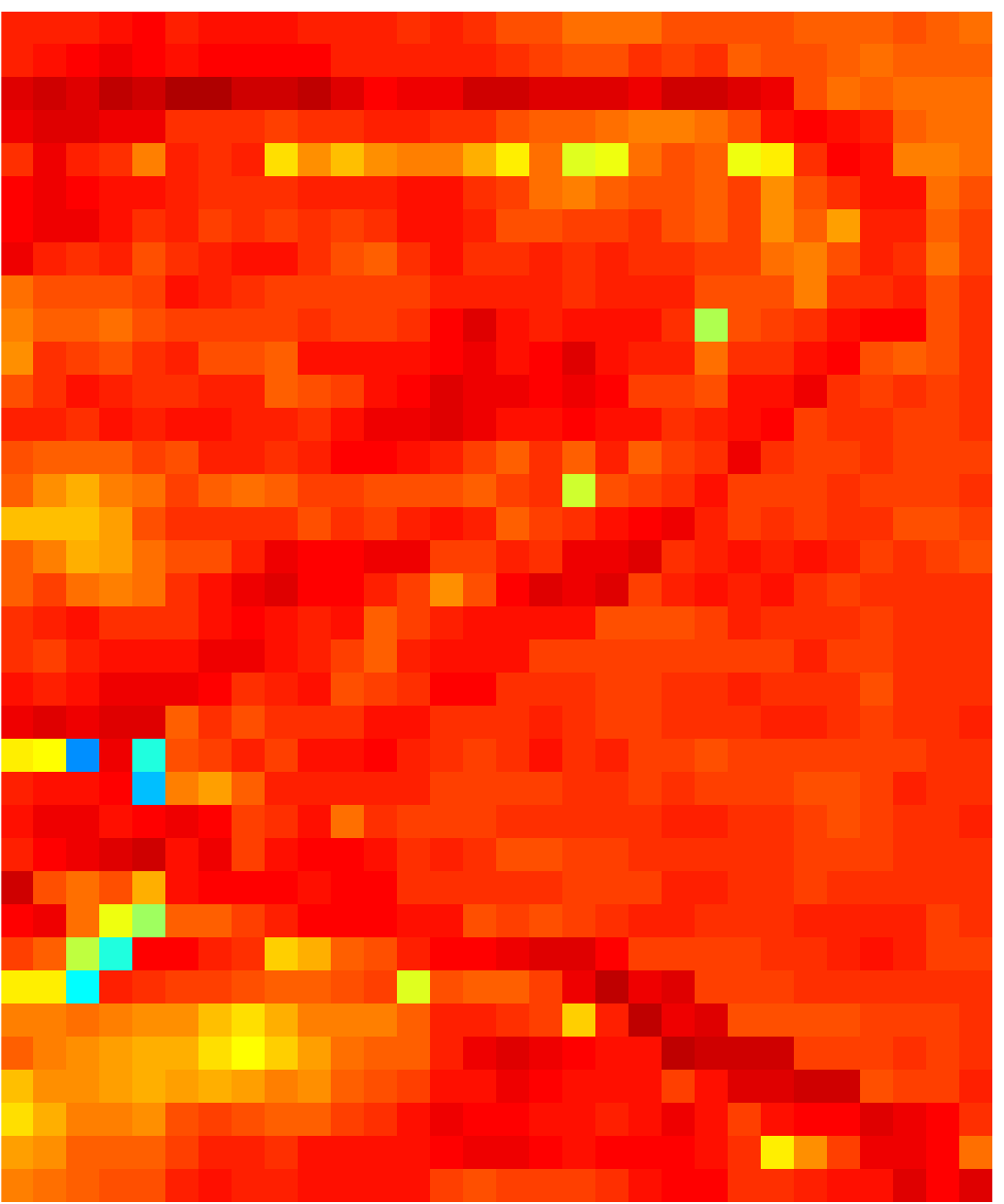} &
\includegraphics[width=.085\textwidth]{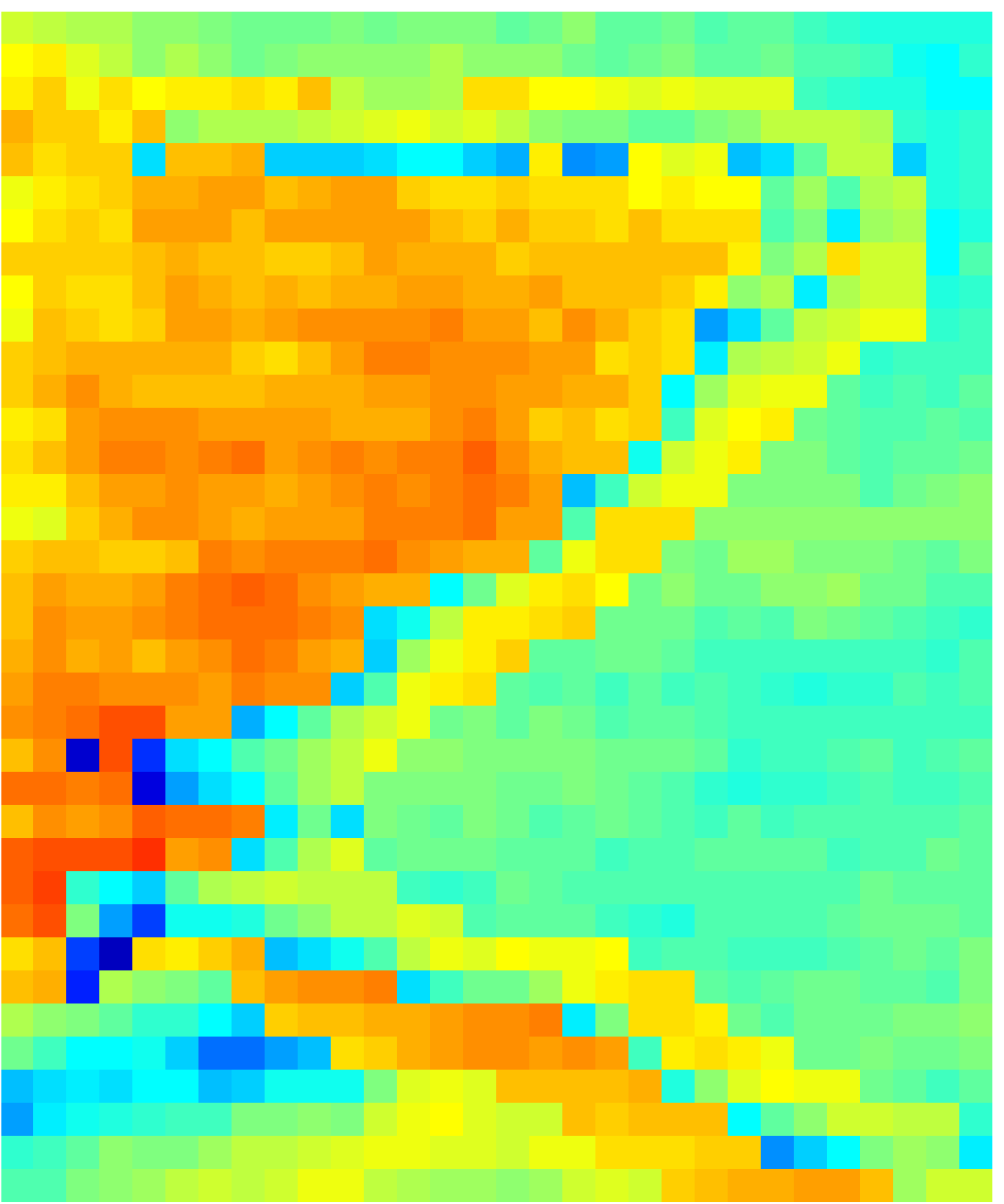} &
\includegraphics[width=.085\textwidth]{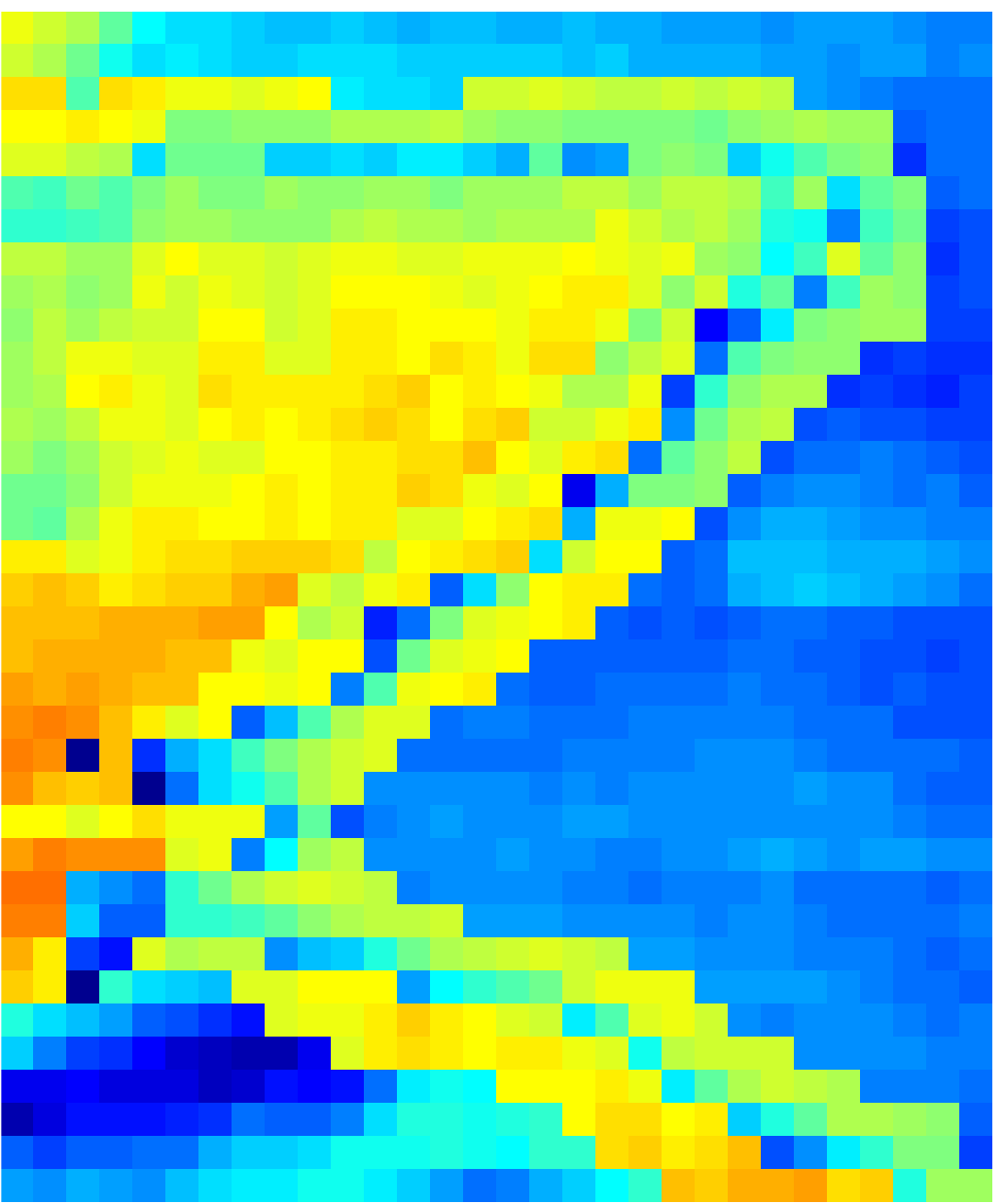} 
& \includegraphics[trim=400 5 0 0, clip, width=.02\textwidth]{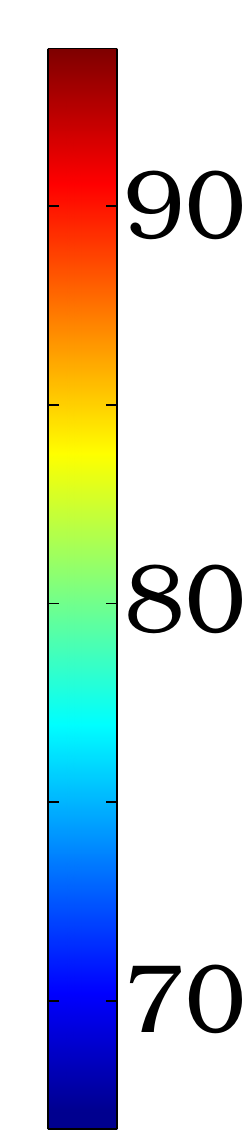} 
\\

\bf{107} & 
\includegraphics[width=.085\textwidth]{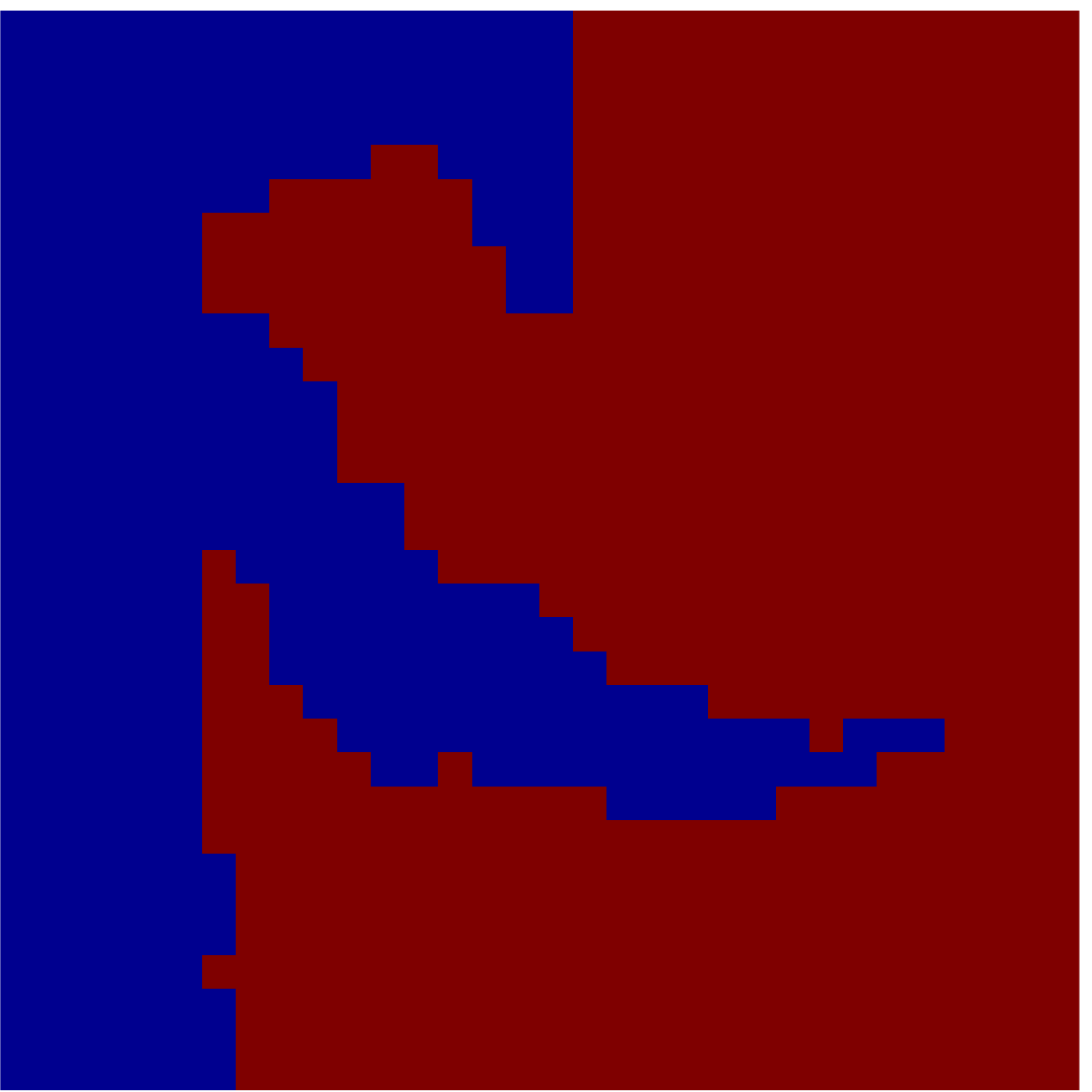} &
\includegraphics[width=.085\textwidth]{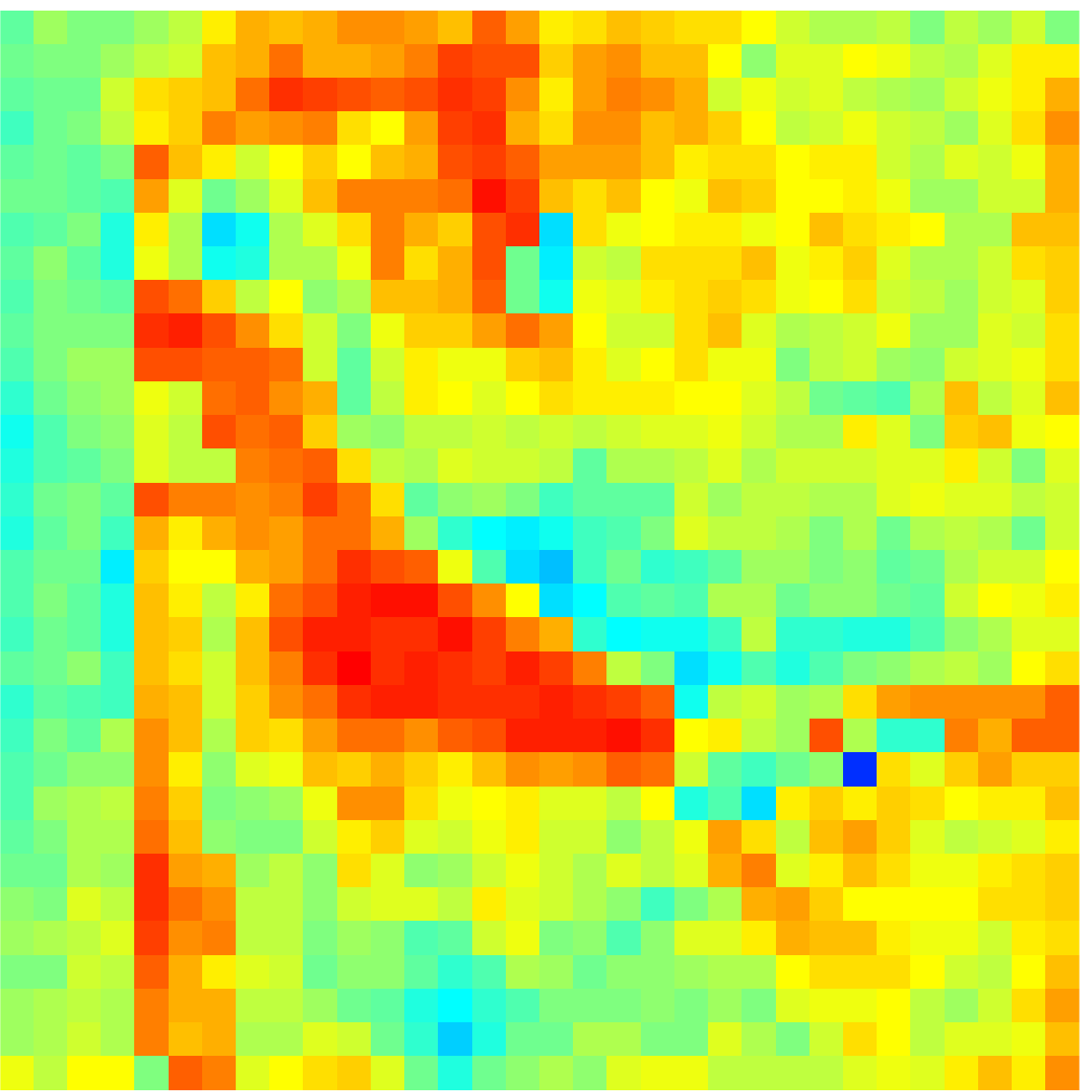} &
\includegraphics[width=.085\textwidth]{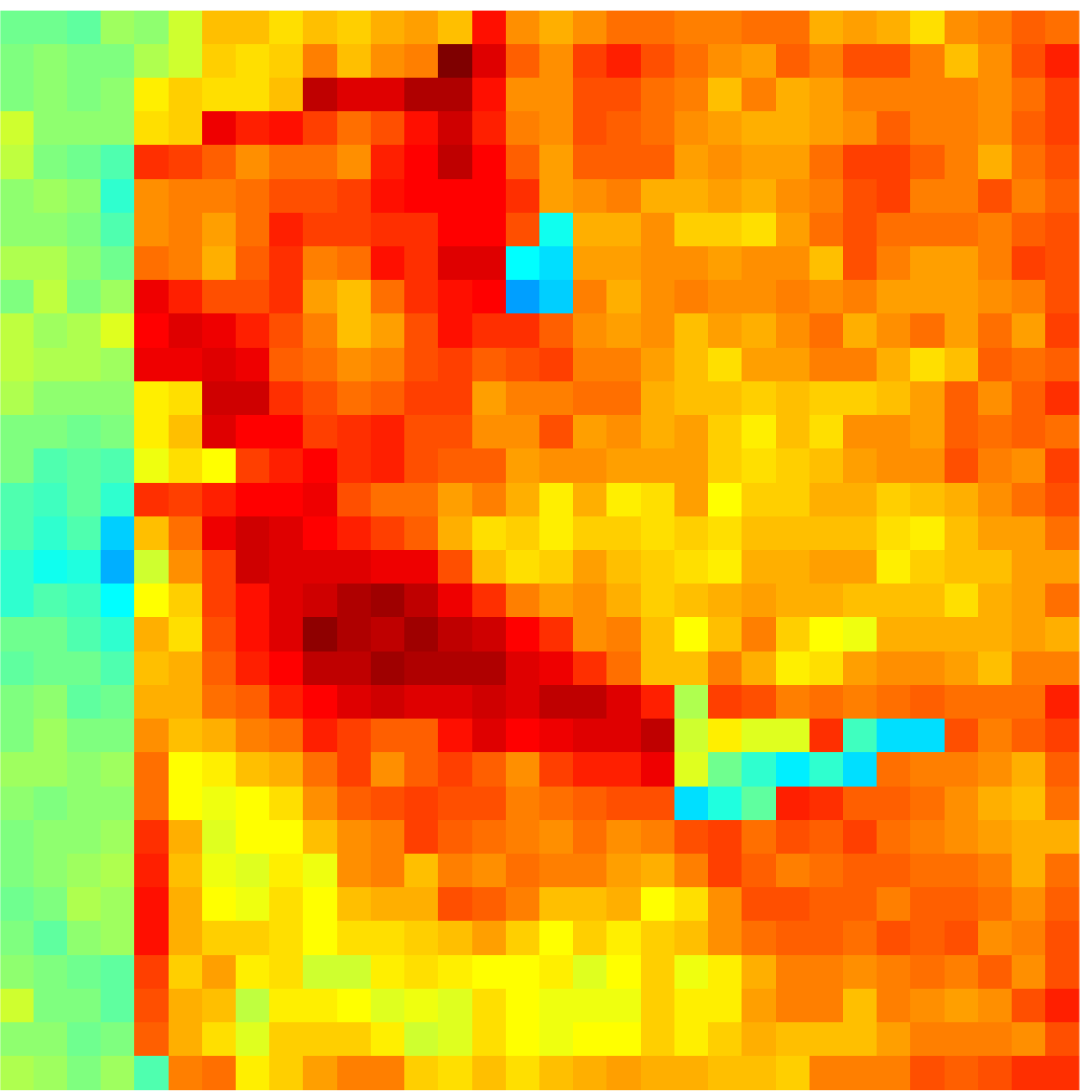} &
\includegraphics[width=.085\textwidth]{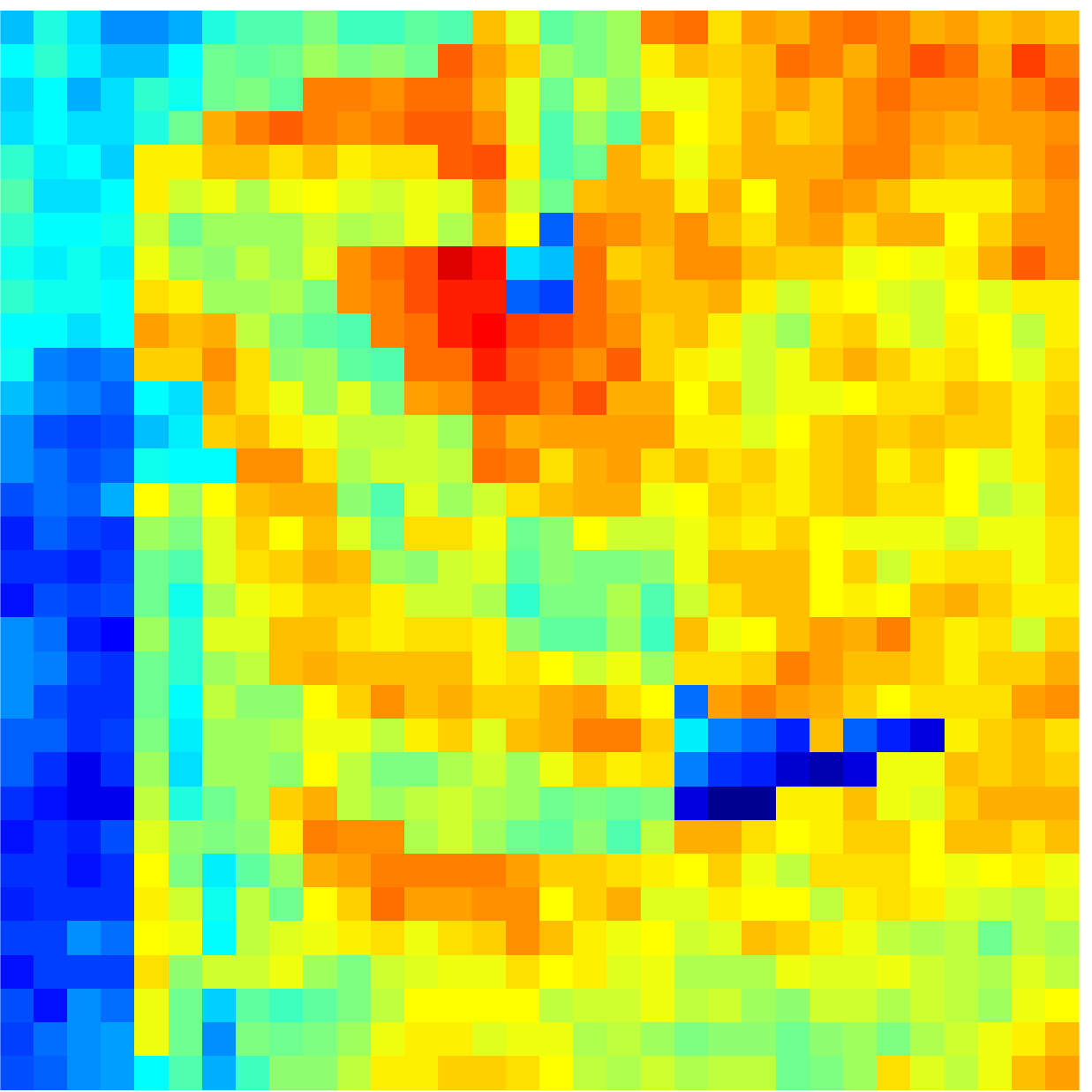} &
\includegraphics[width=.085\textwidth]{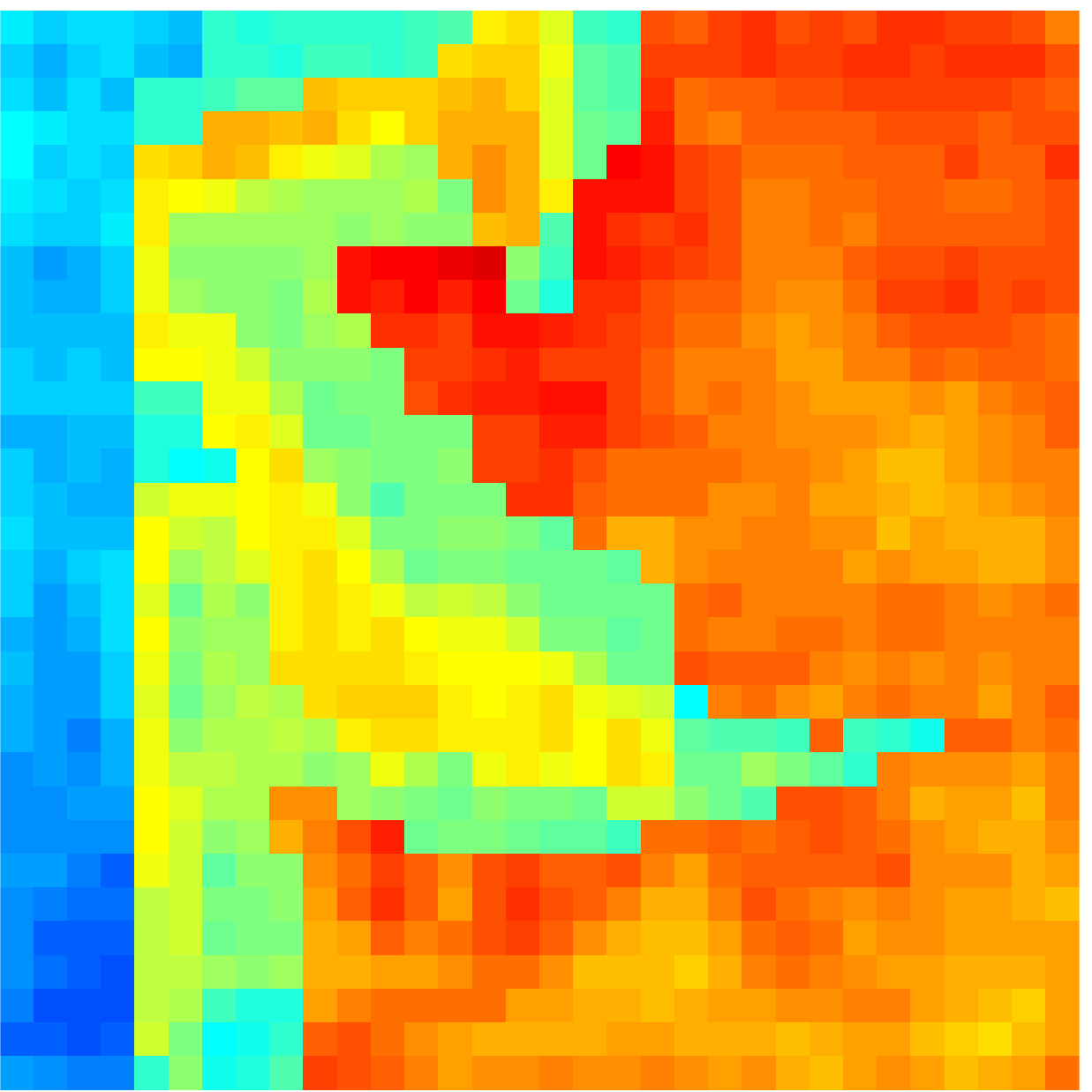} &
\includegraphics[trim=400 5 0 0, clip, width=.02\textwidth]{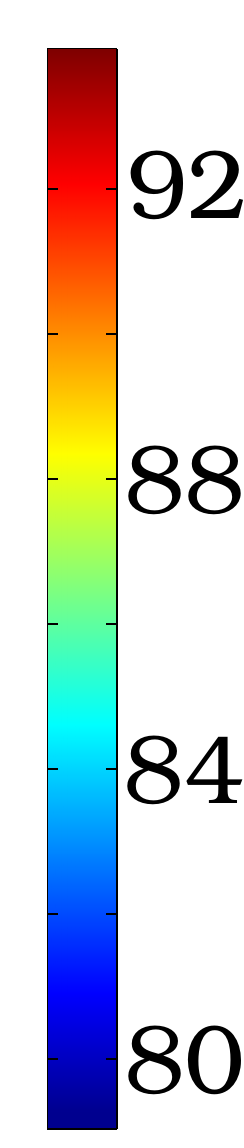} \\

\bf{120} & 
\includegraphics[width=.085\textwidth]{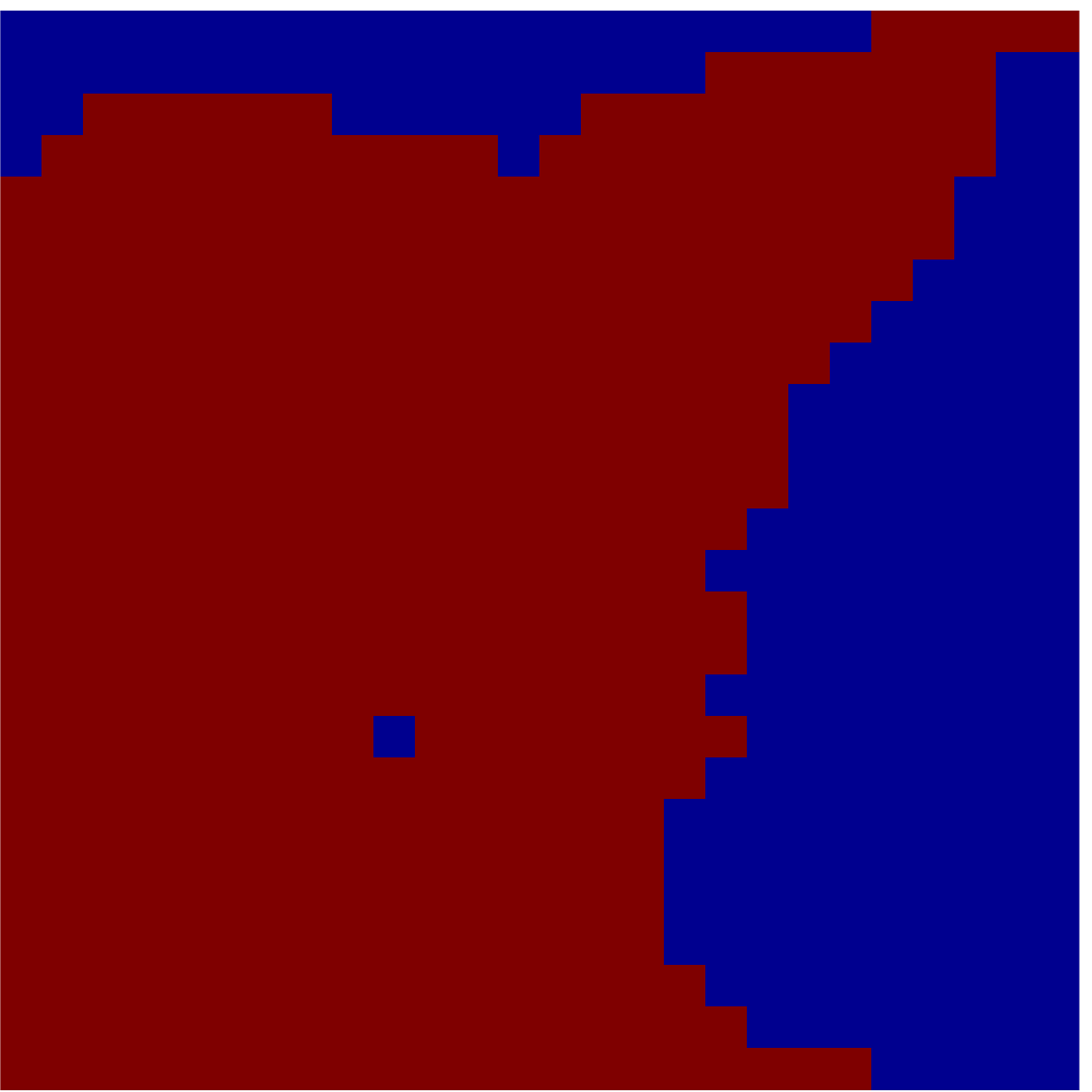} &
\includegraphics[width=.085\textwidth]{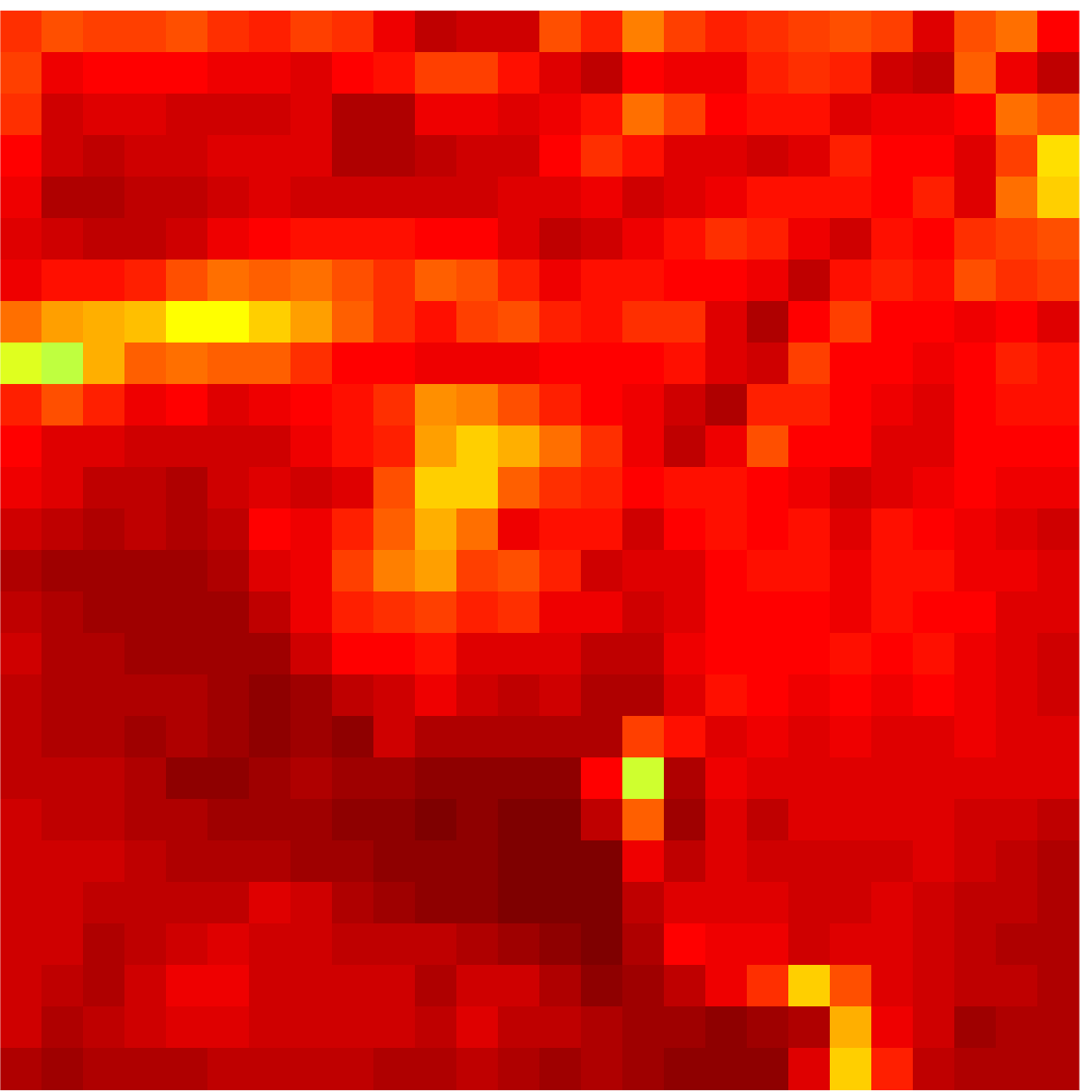} &
\includegraphics[width=.085\textwidth]{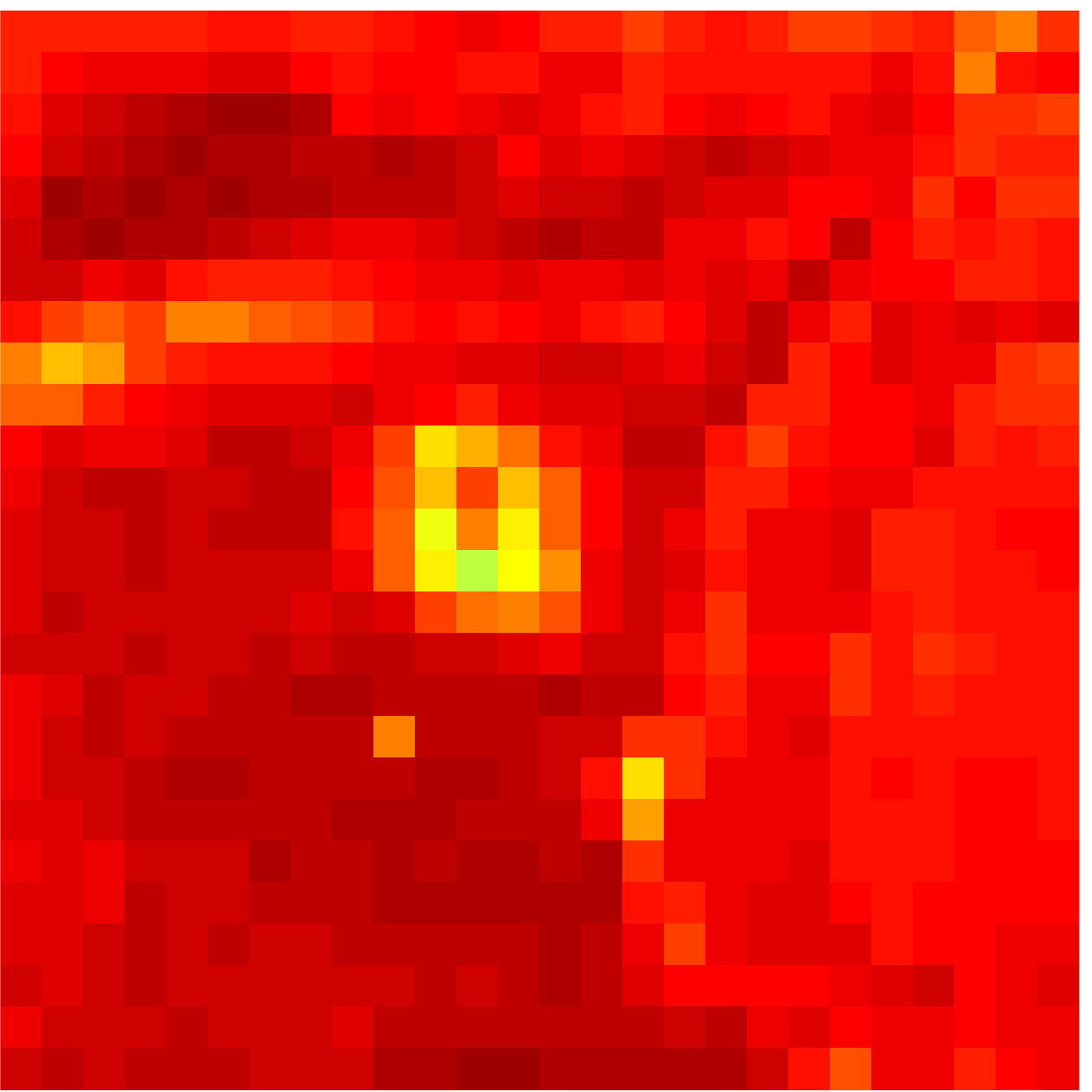} &
\includegraphics[width=.085\textwidth]{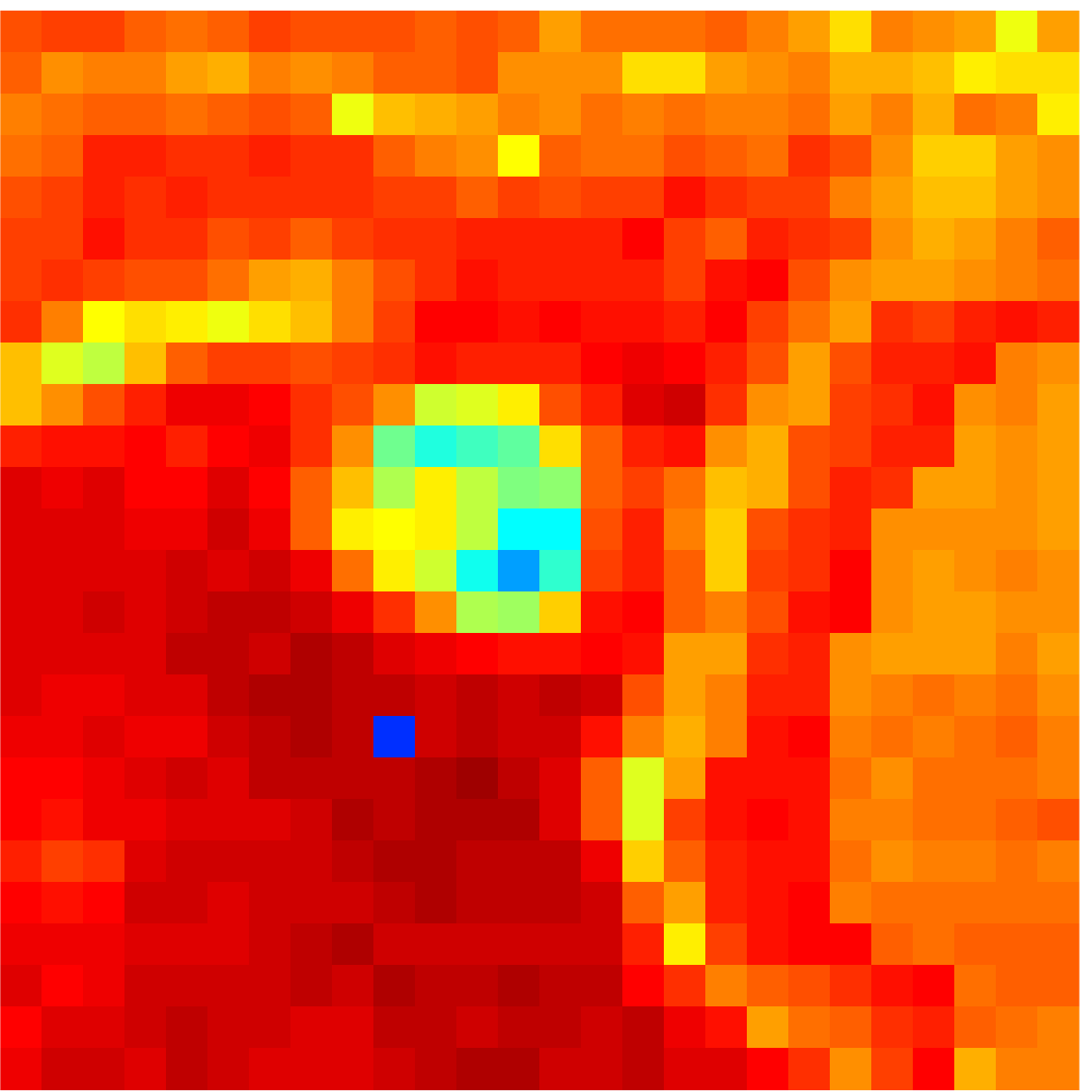} &
\includegraphics[width=.085\textwidth]{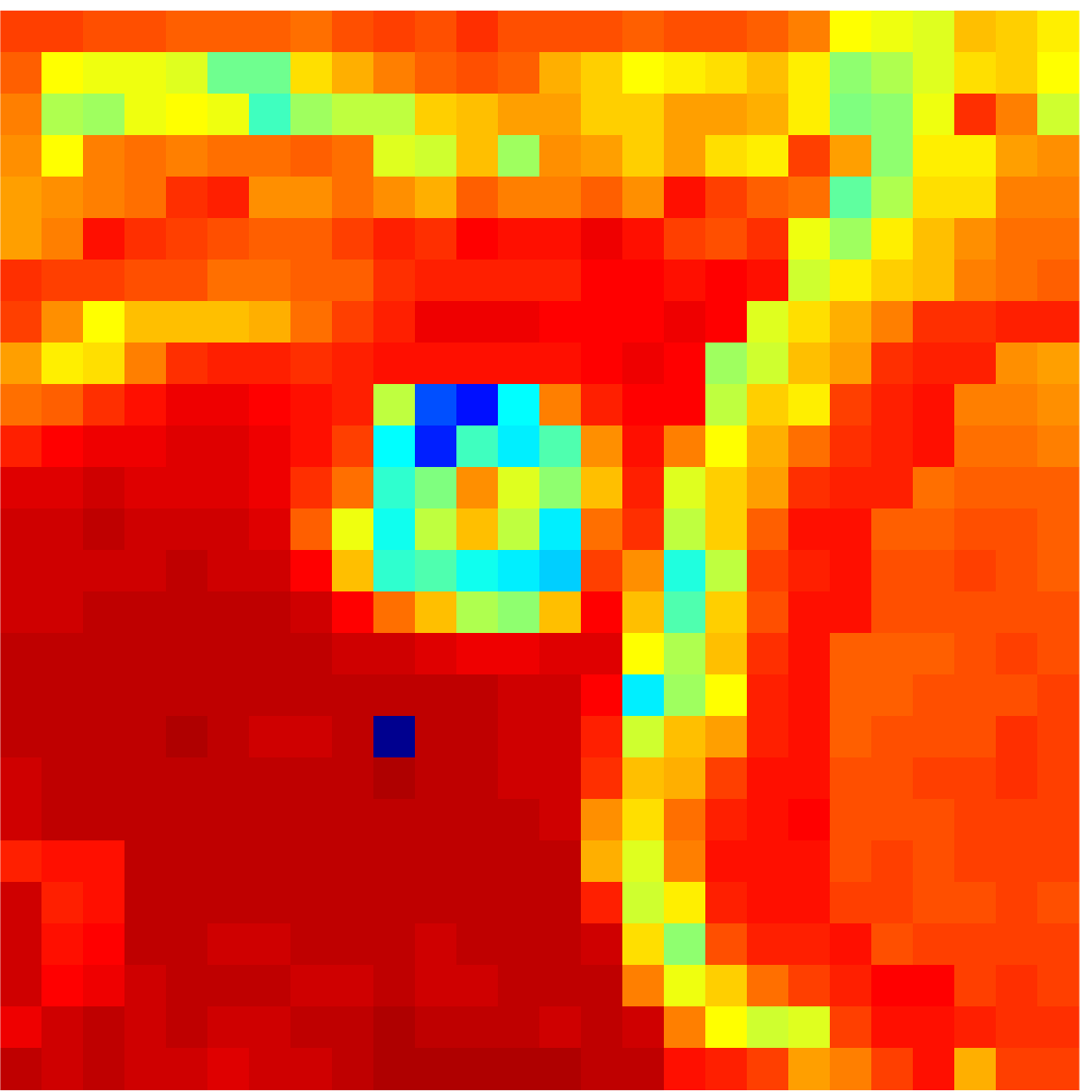} &
\includegraphics[trim=400 5 0 0, clip, width=.02\textwidth]{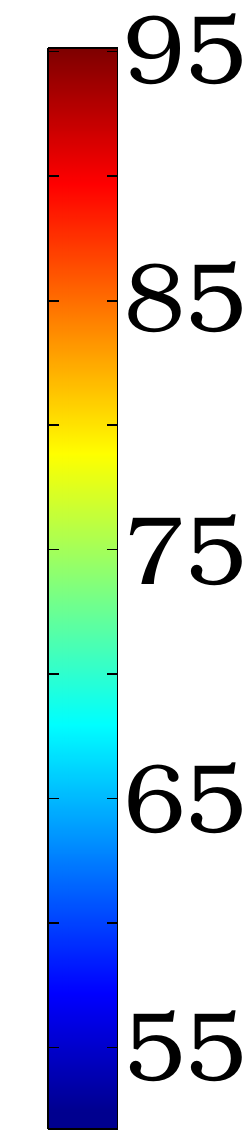} \\

\bf{131} & 
\includegraphics[width=.085\textwidth]{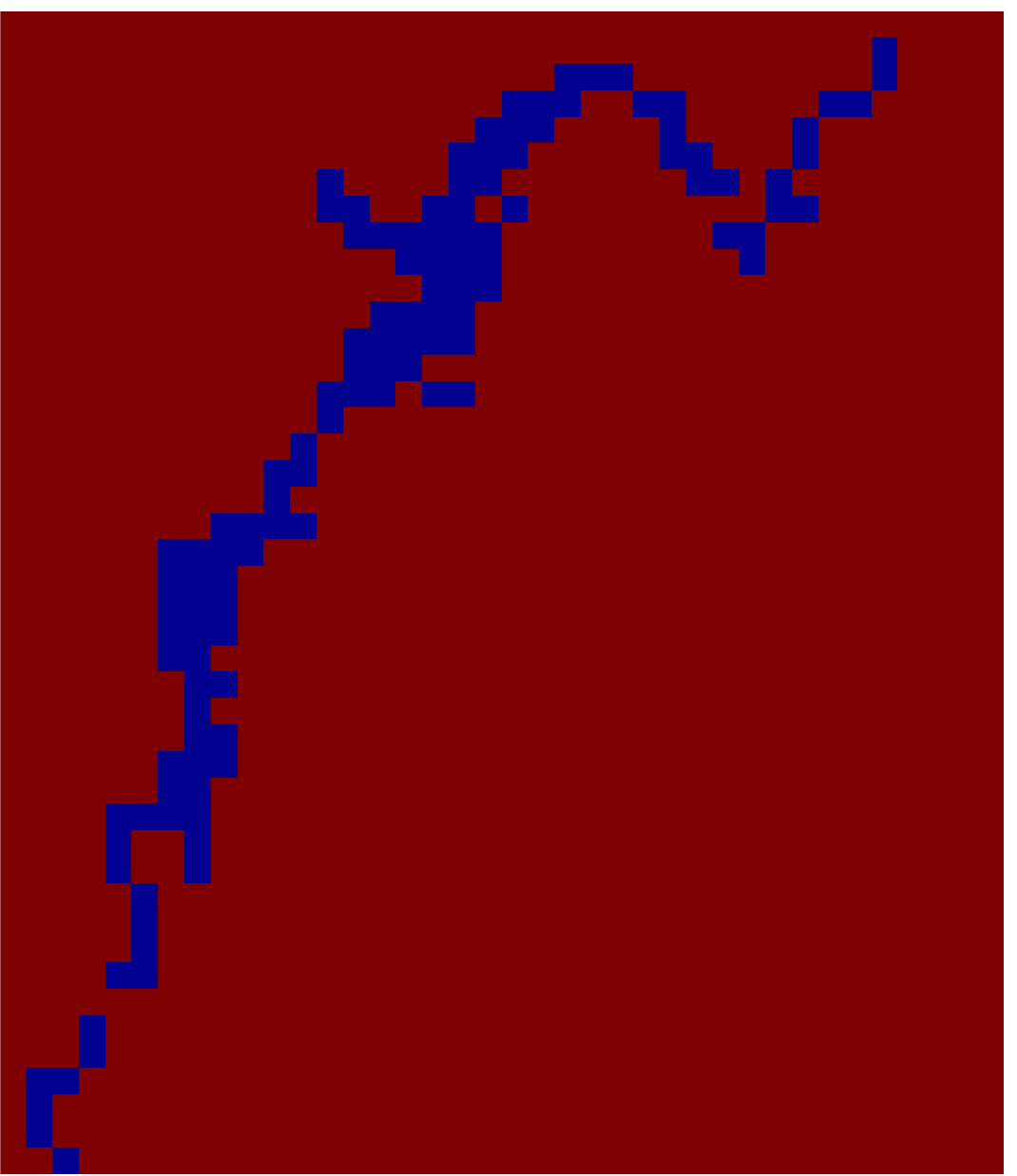} &
\includegraphics[width=.085\textwidth]{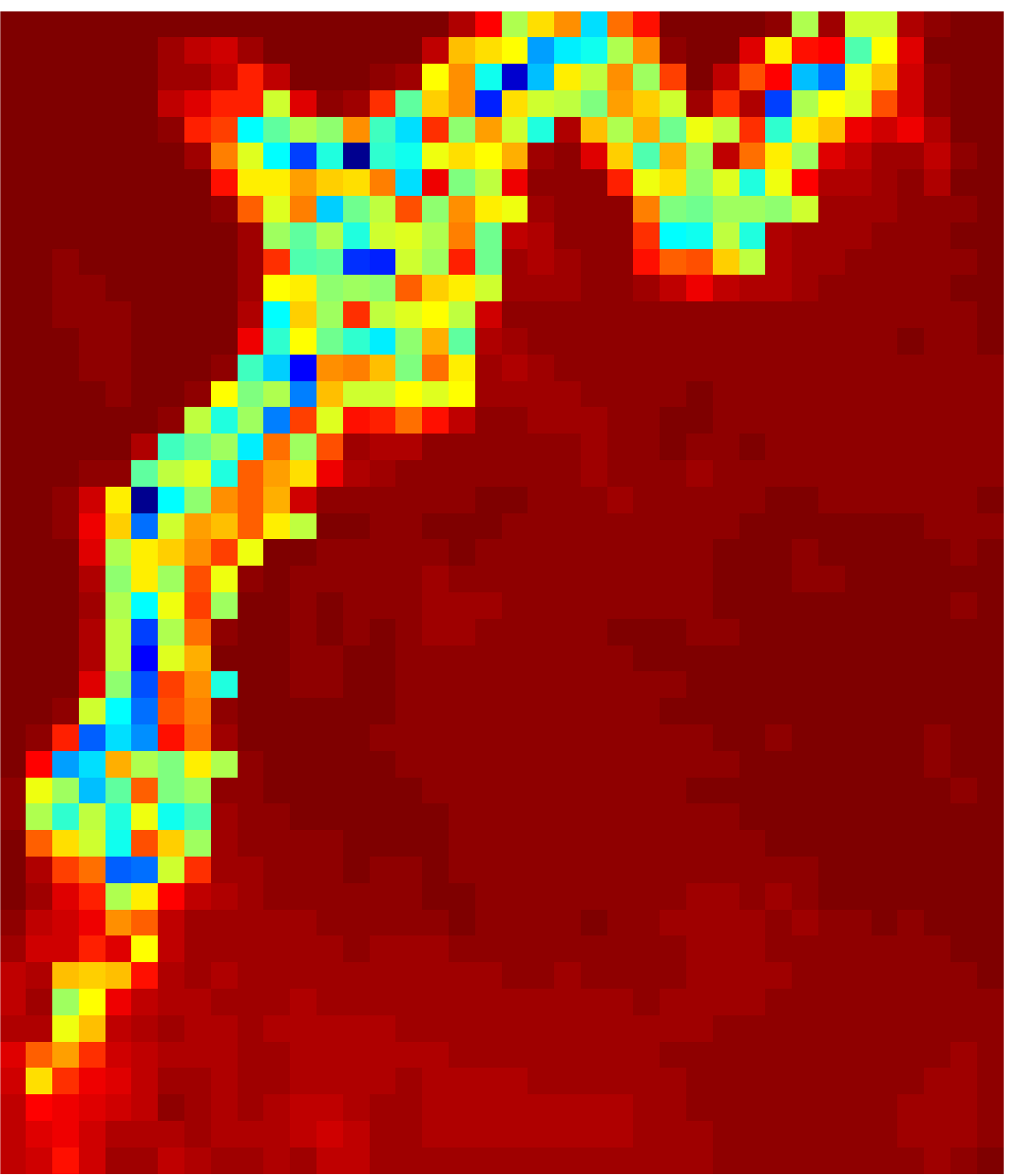} &
\includegraphics[width=.085\textwidth]{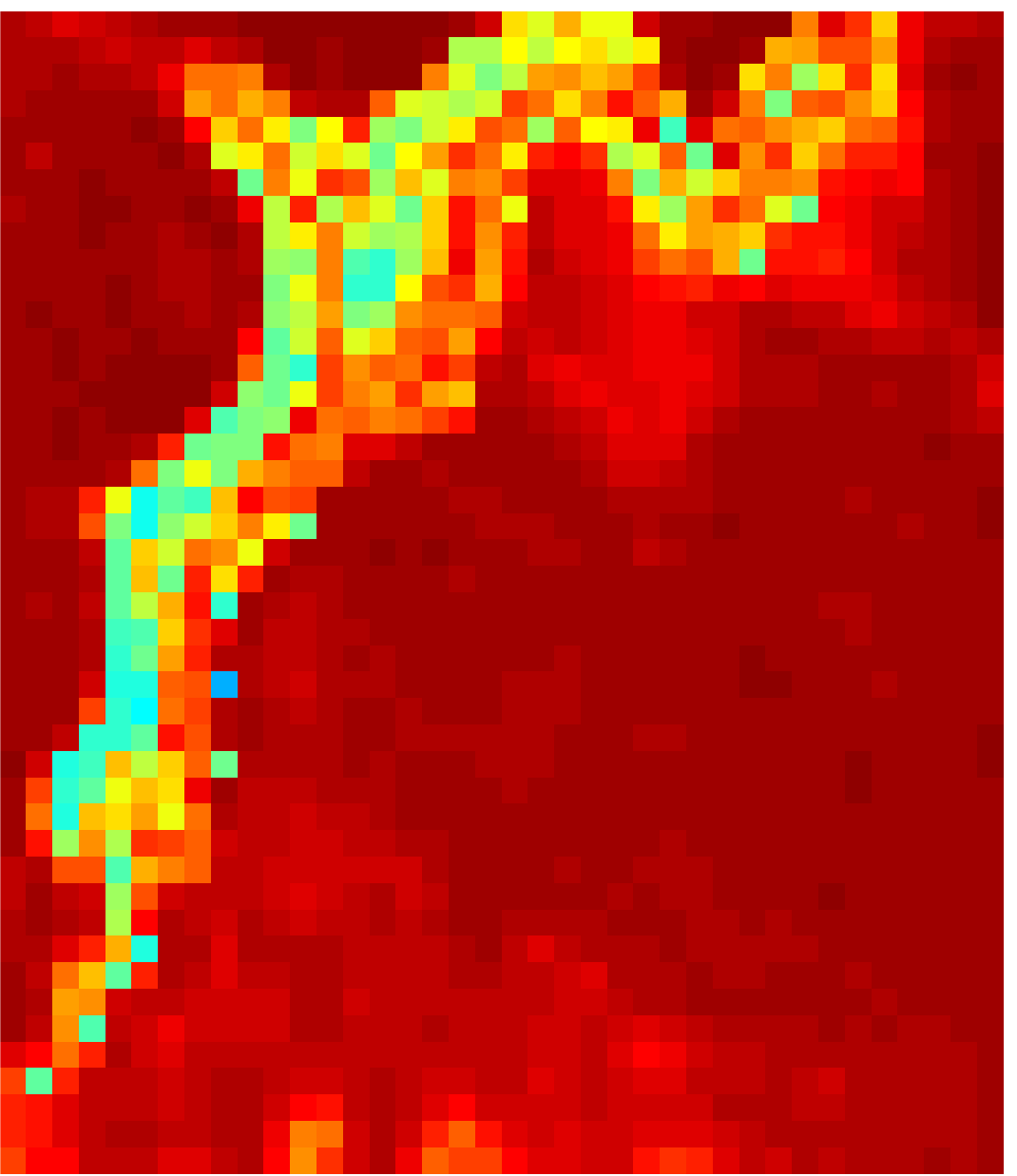} &
\includegraphics[width=.085\textwidth]{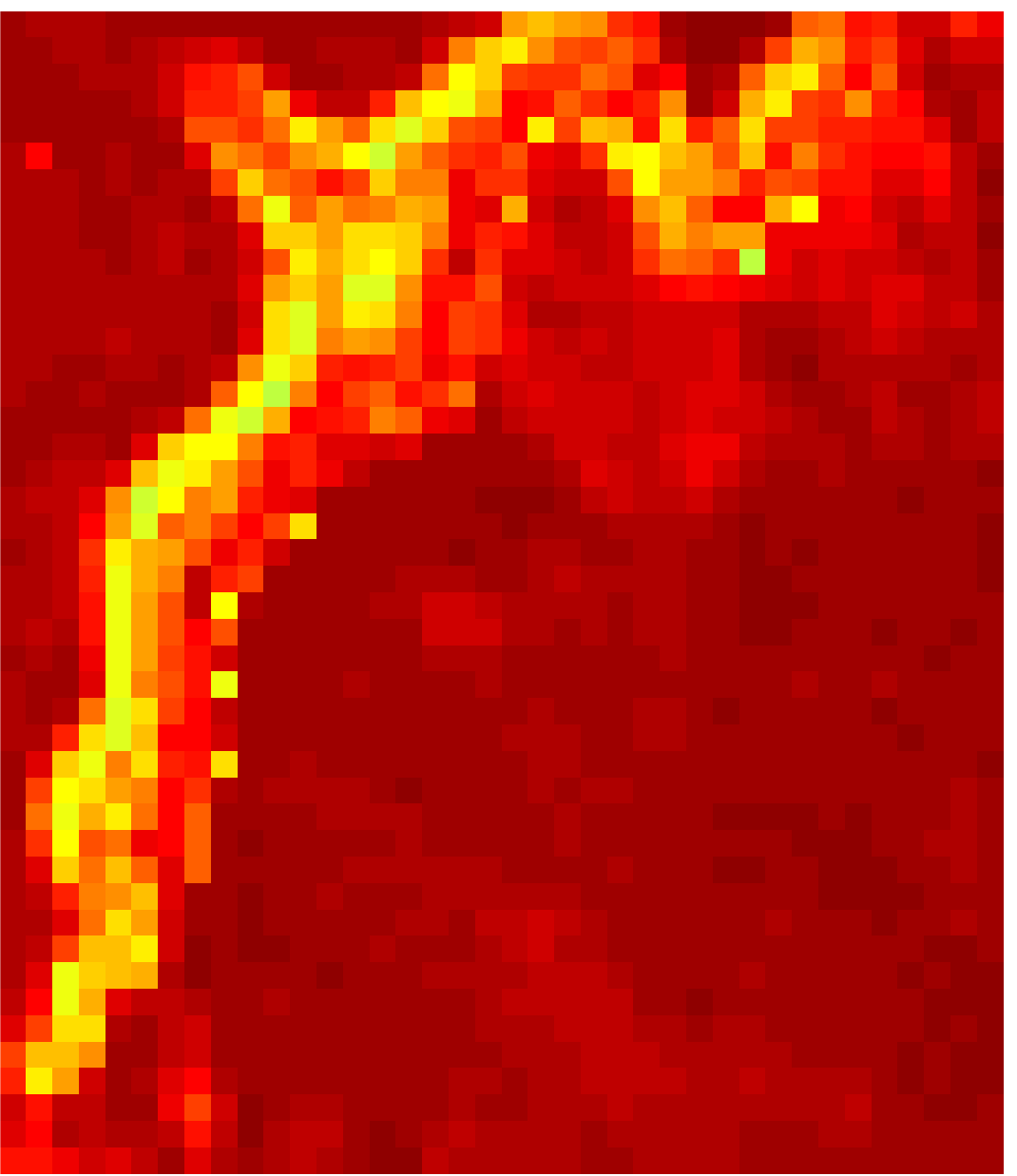} &
\includegraphics[width=.085\textwidth]{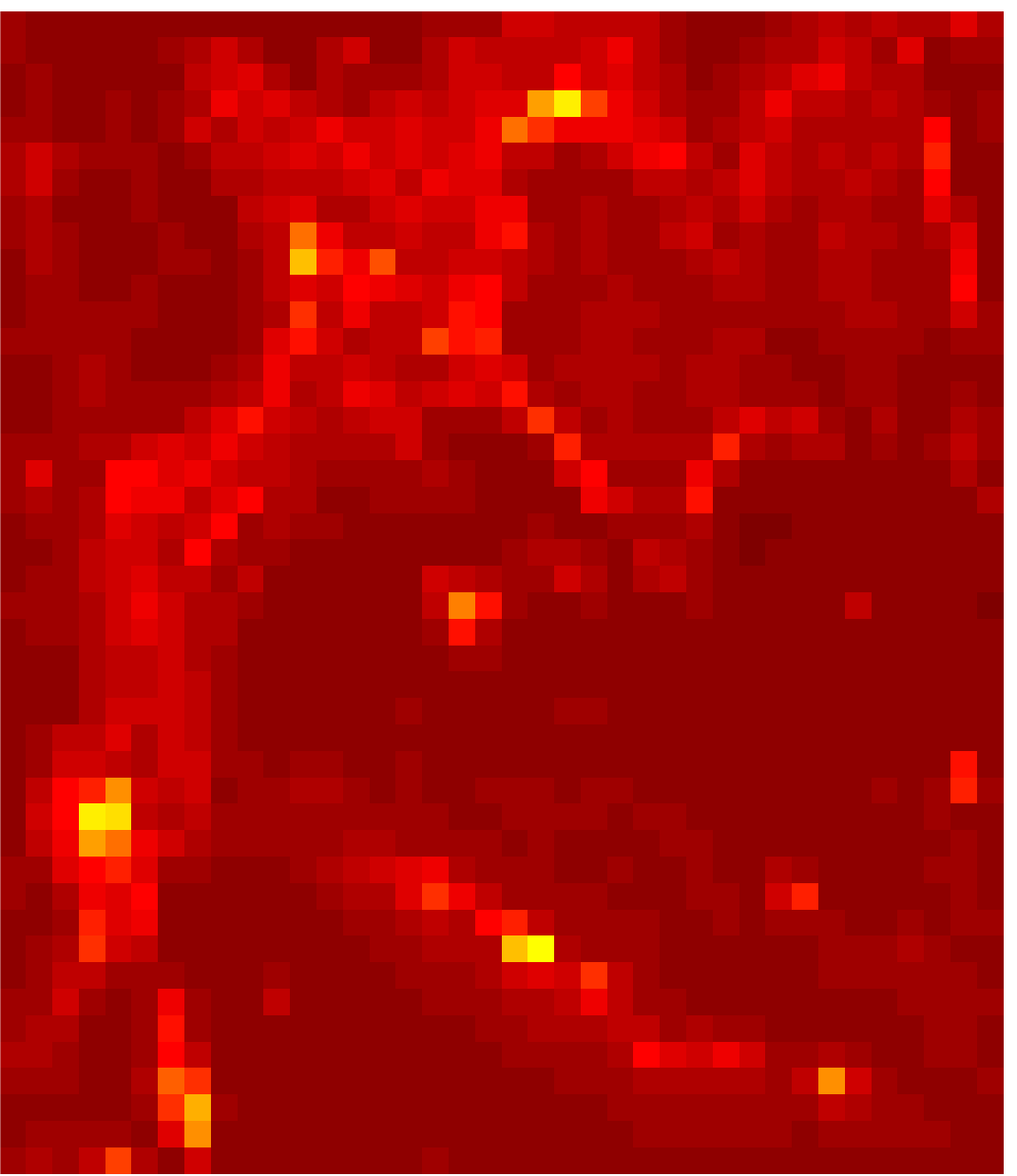} &
\includegraphics[trim=400 5 0 0, clip, width=.02\textwidth]{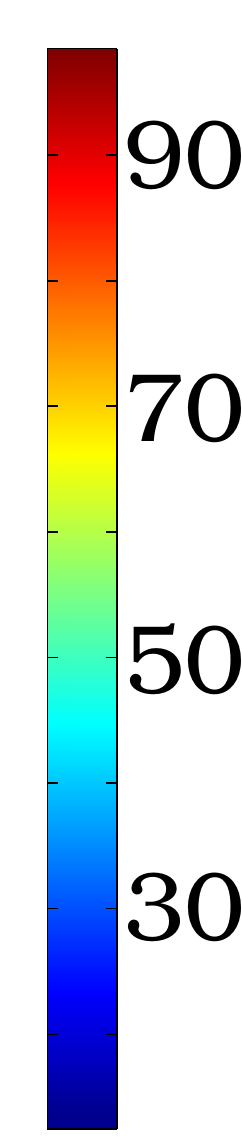} \\

\bf{154} & 
\includegraphics[width=.085\textwidth]{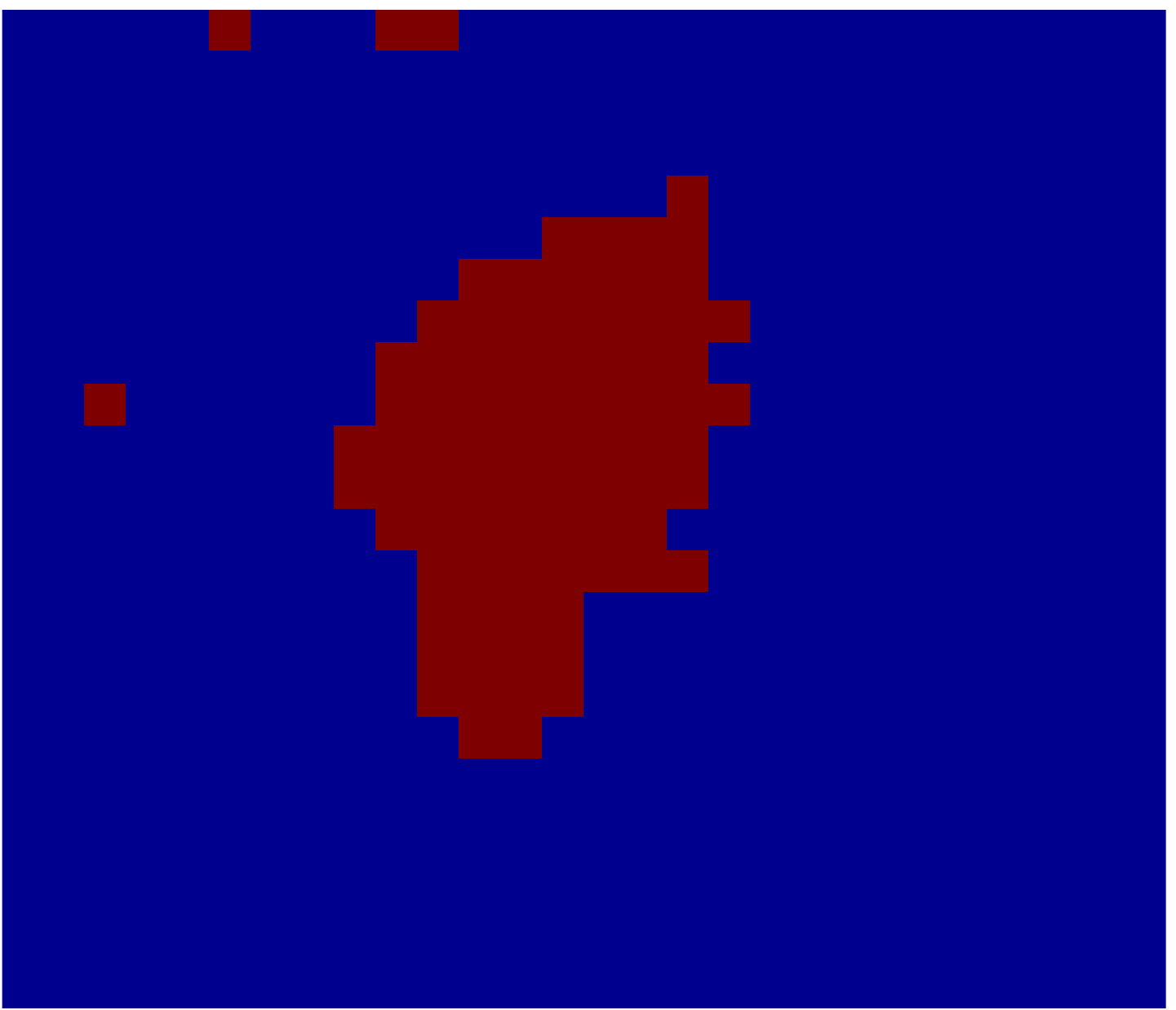} &
\includegraphics[width=.085\textwidth]{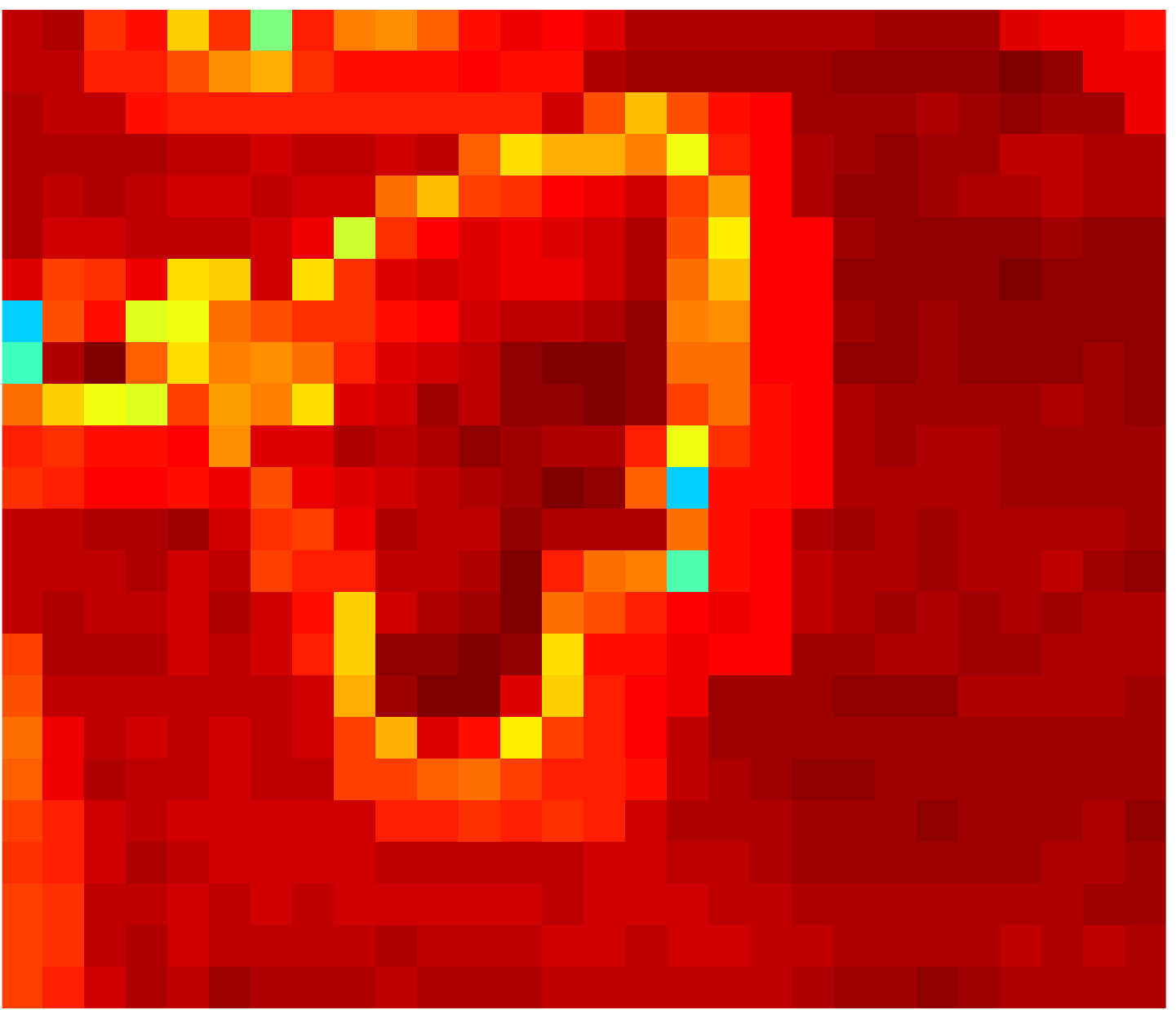} &
\includegraphics[width=.085\textwidth]{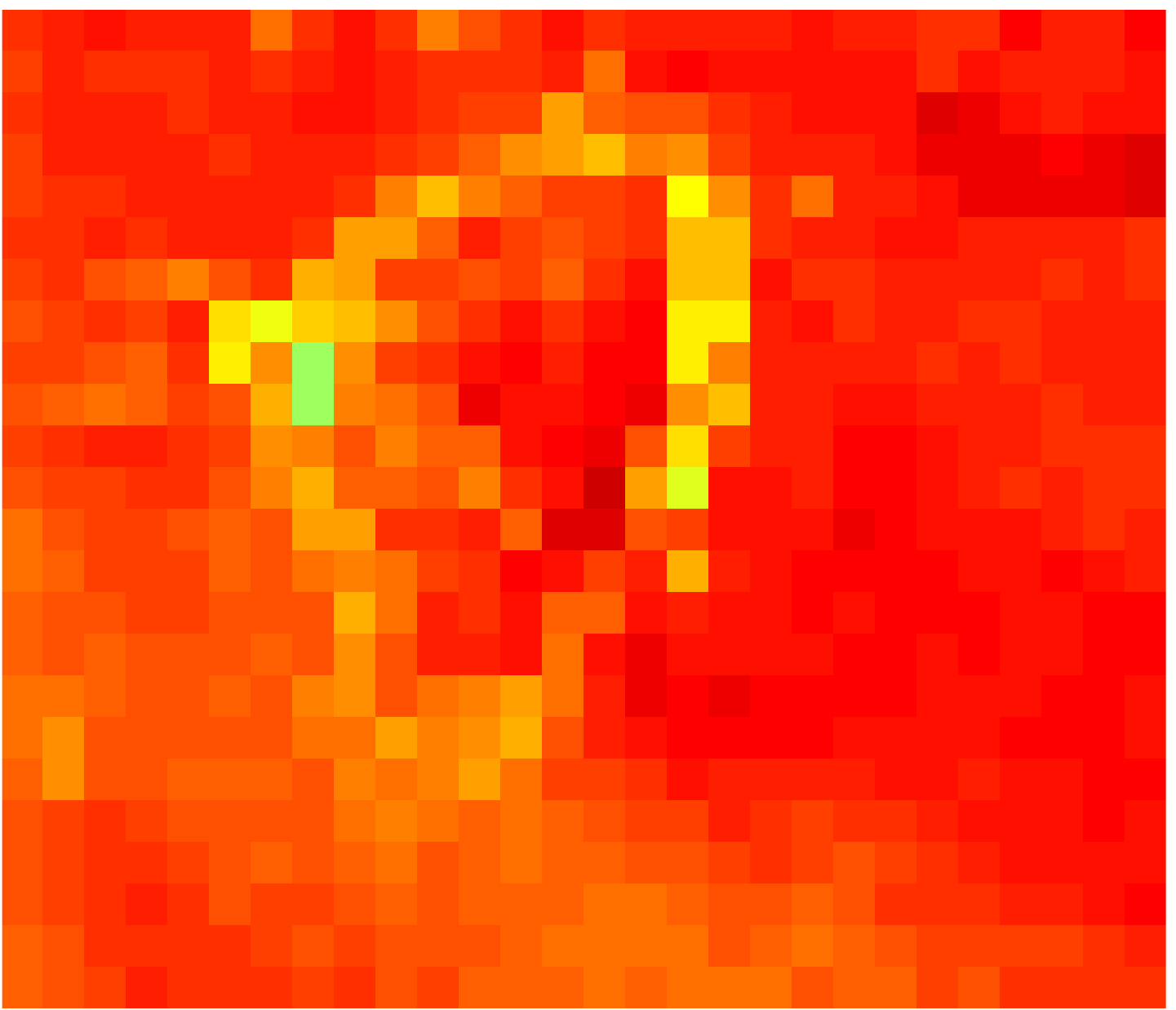} &
\includegraphics[width=.085\textwidth]{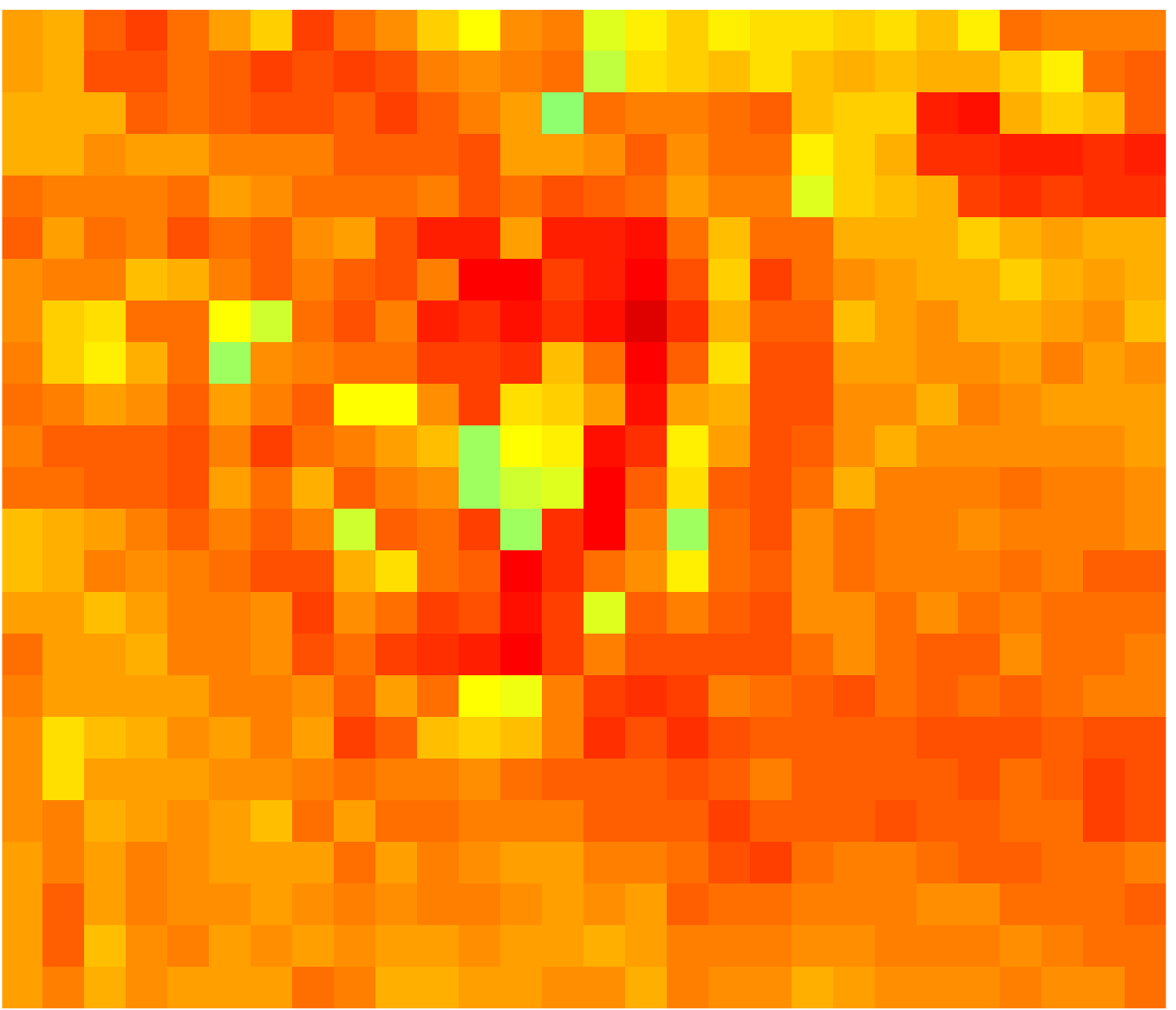} &
\includegraphics[width=.085\textwidth]{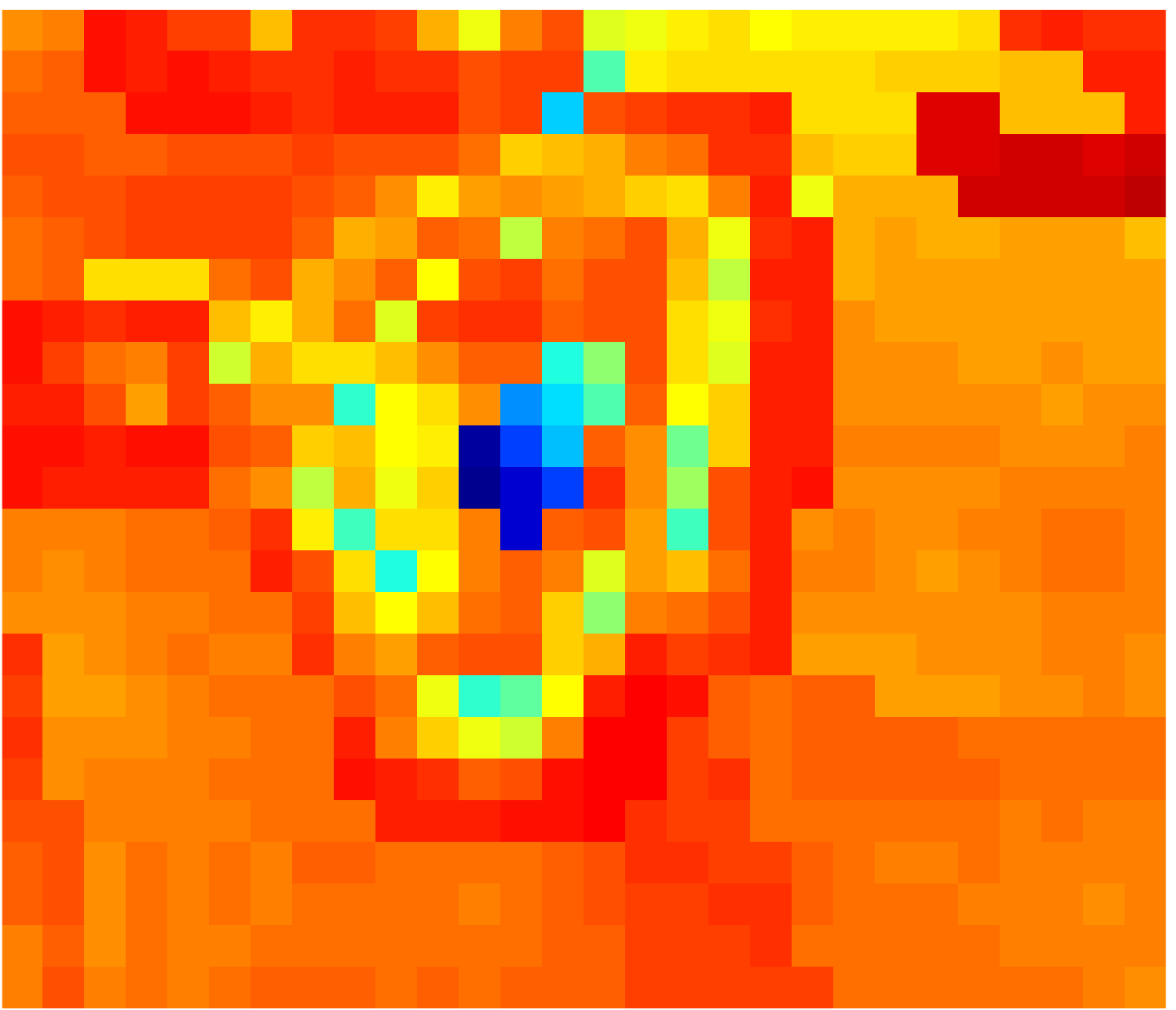} &
\includegraphics[trim=400 5 0 0, clip, width=.015\textwidth]{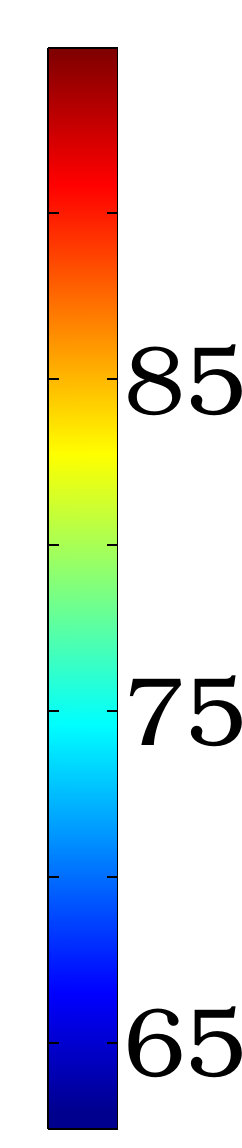} \\

\bf{177} & 
\includegraphics[width=.085\textwidth]{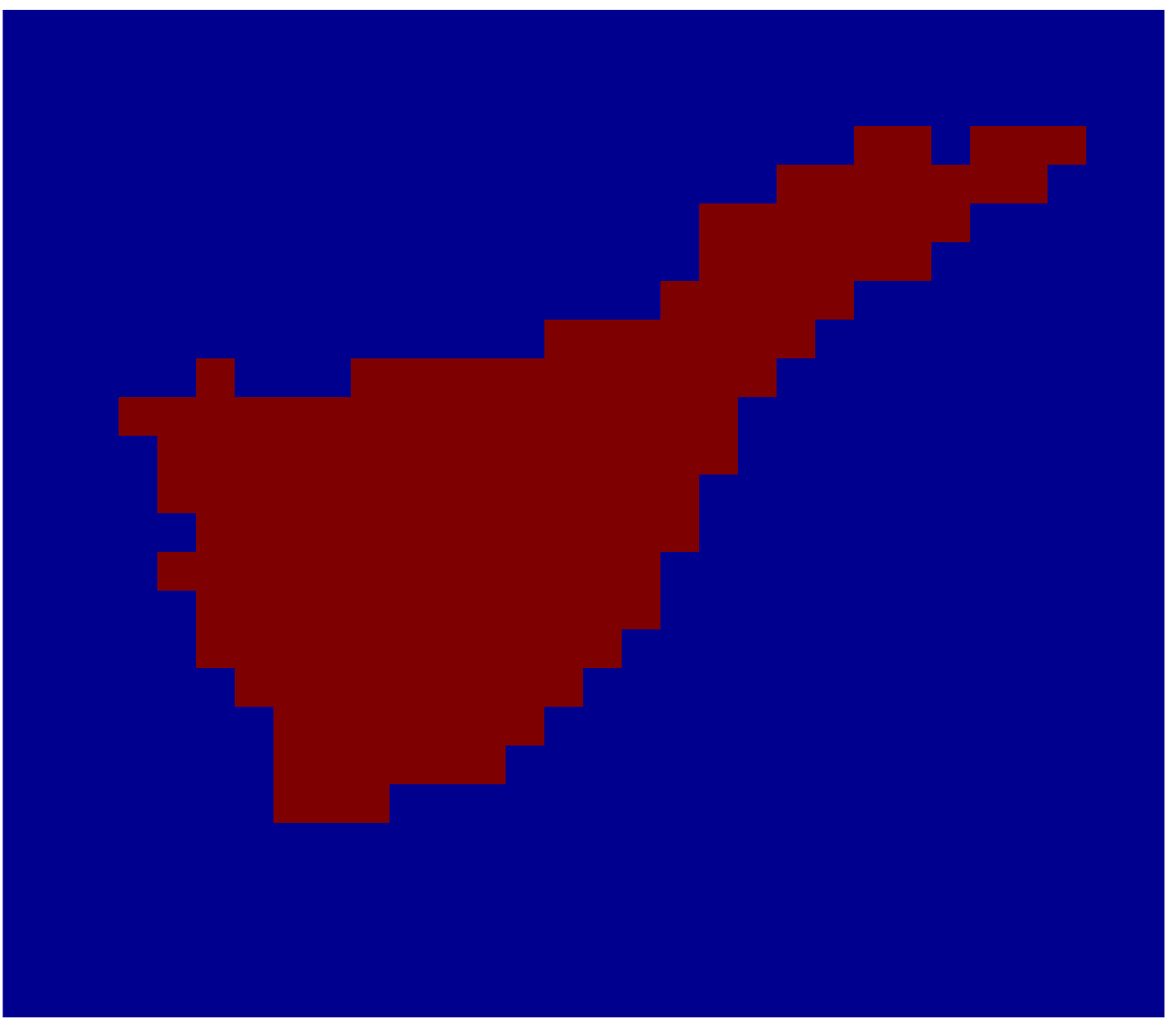} &
\includegraphics[width=.085\textwidth]{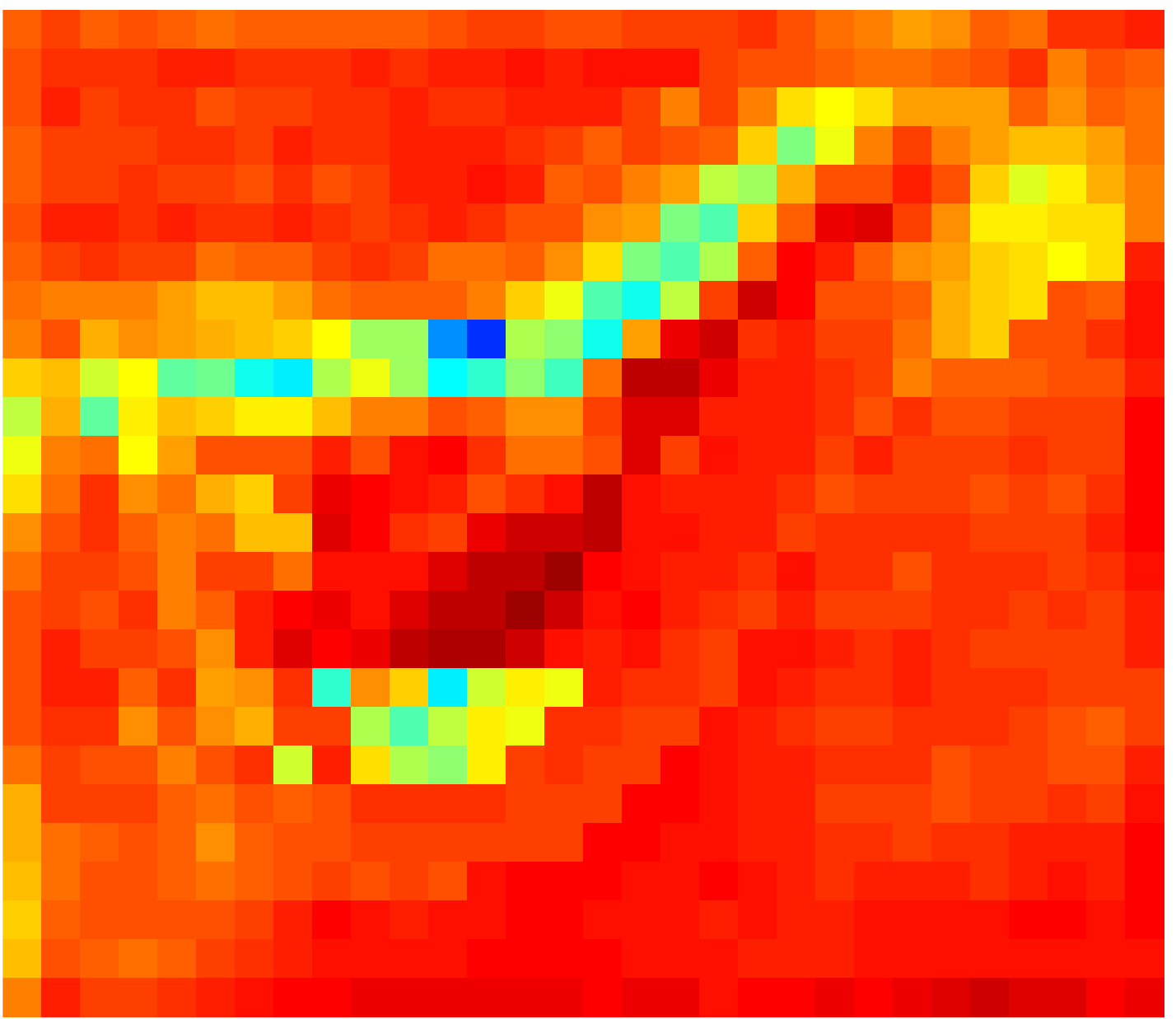} &
\includegraphics[width=.085\textwidth]{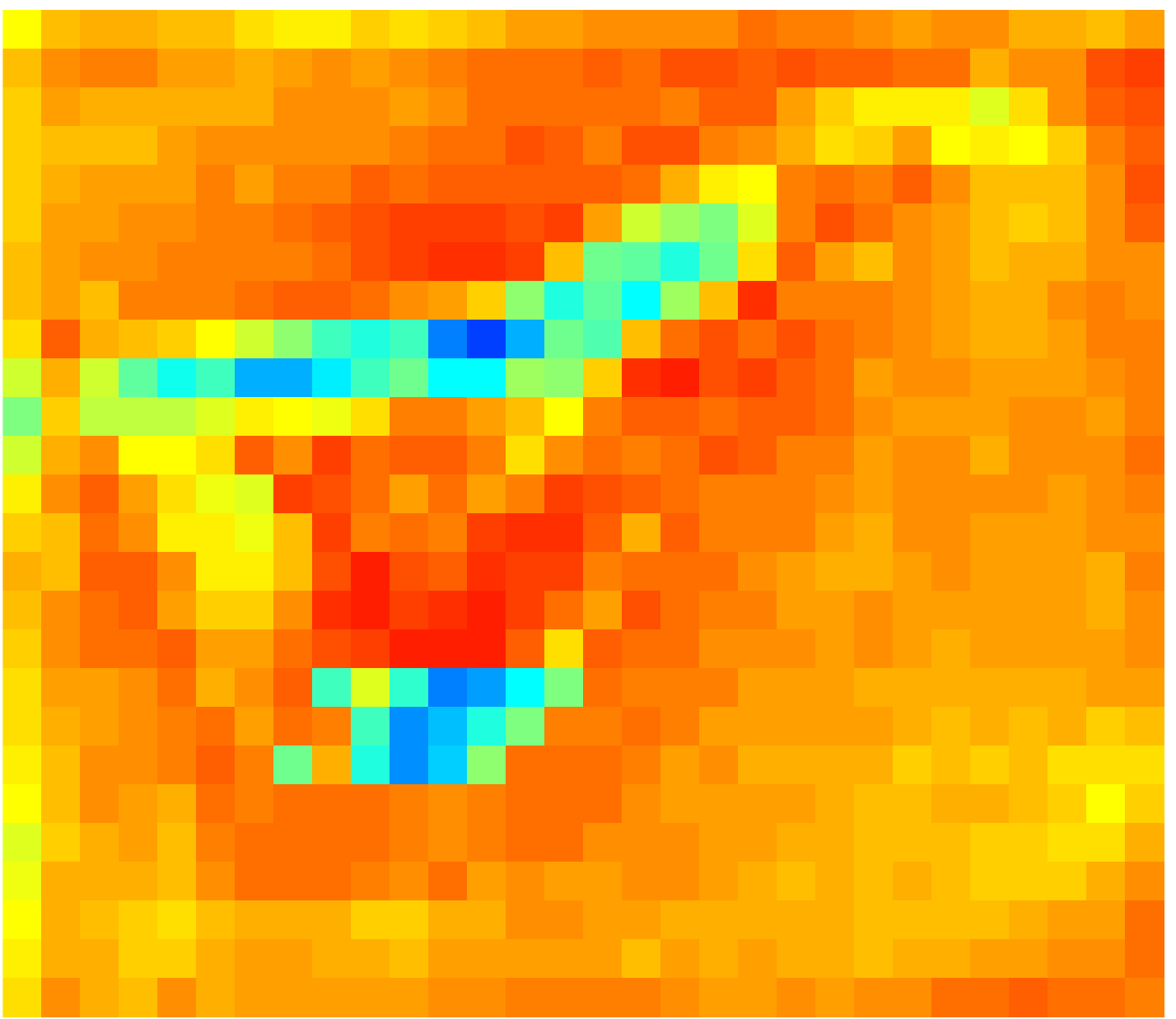} &
\includegraphics[width=.085\textwidth]{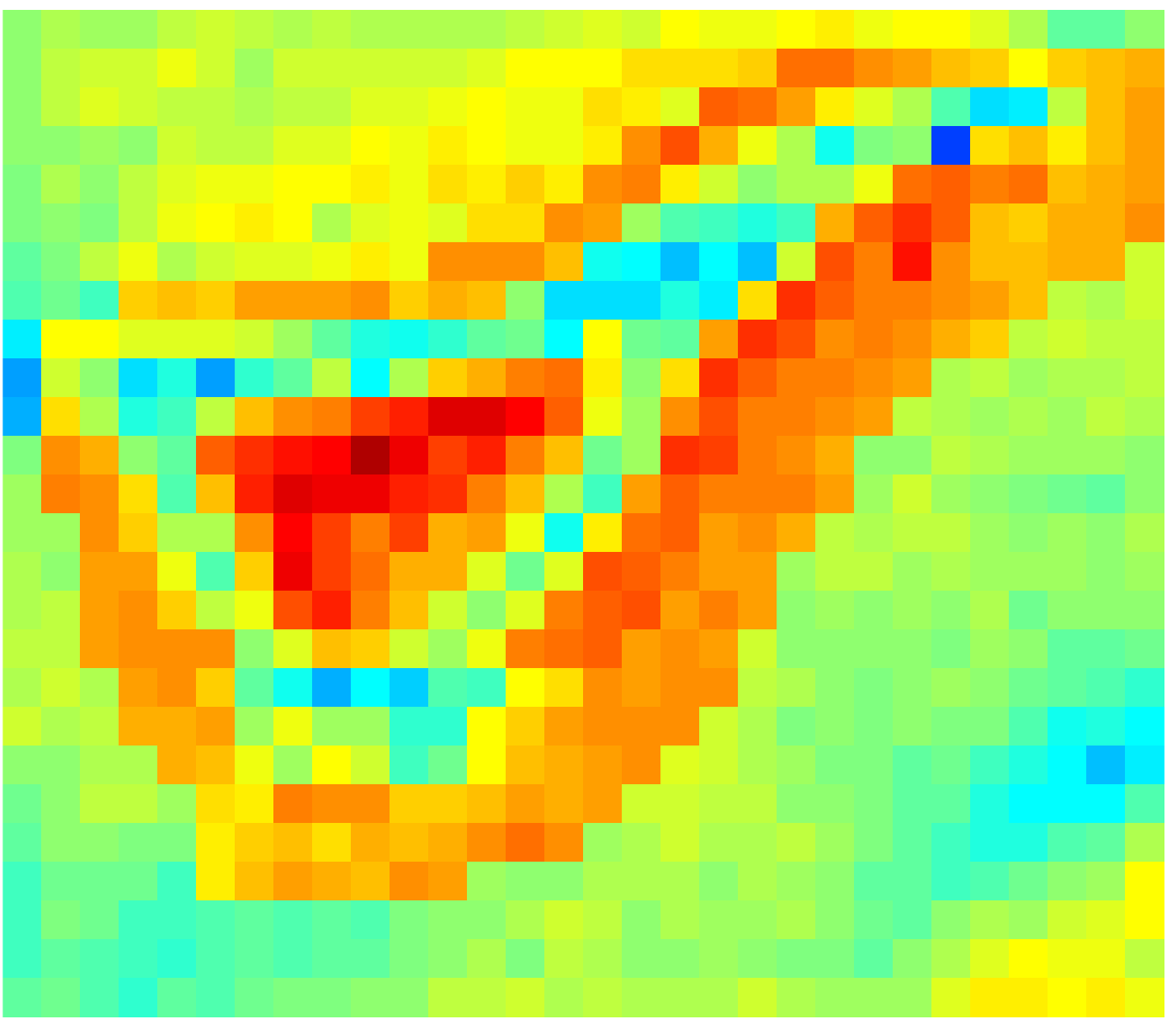} &
\includegraphics[width=.085\textwidth]{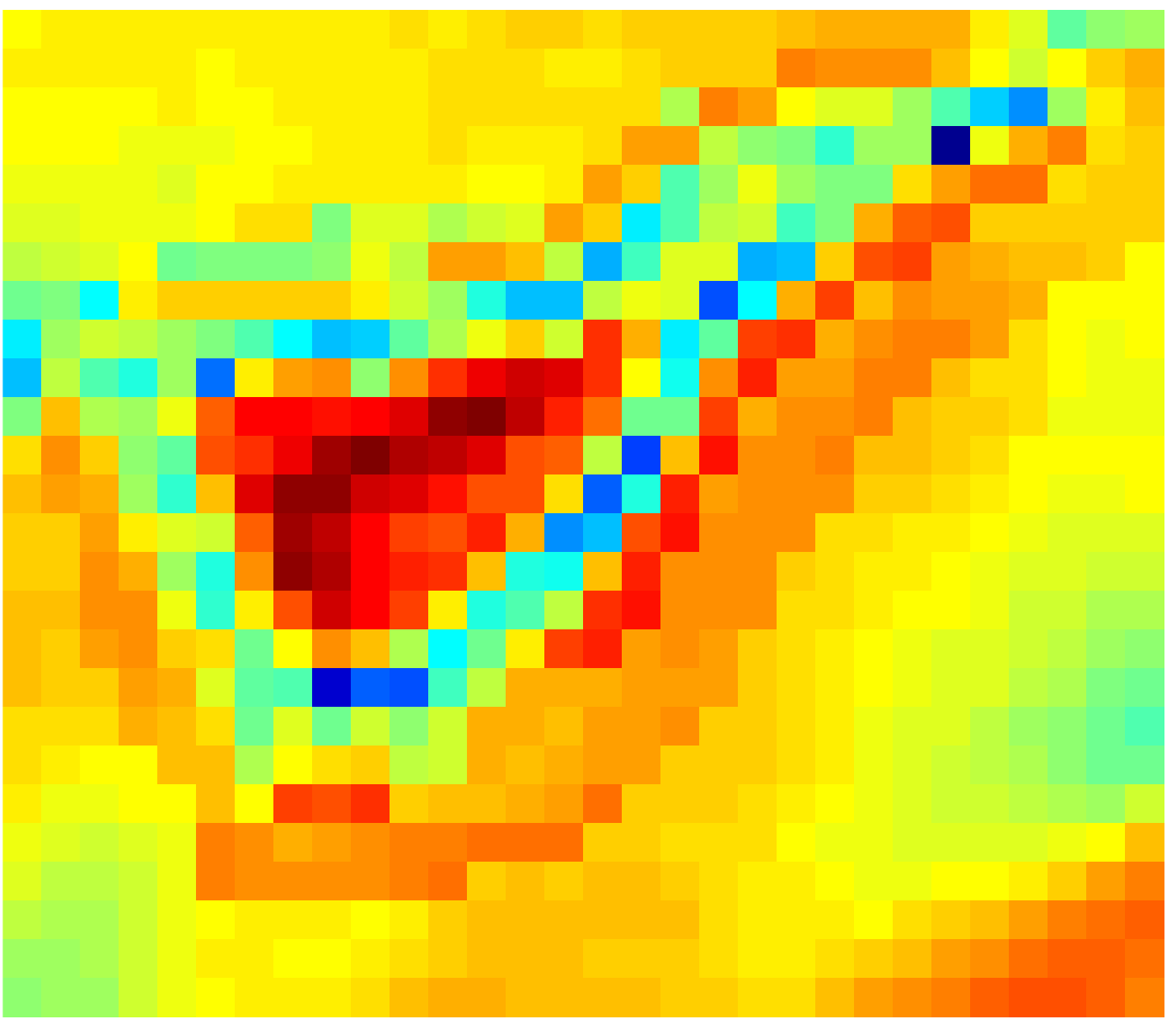} &
\includegraphics[trim=400 5 0 0, clip, width=.015\textwidth]{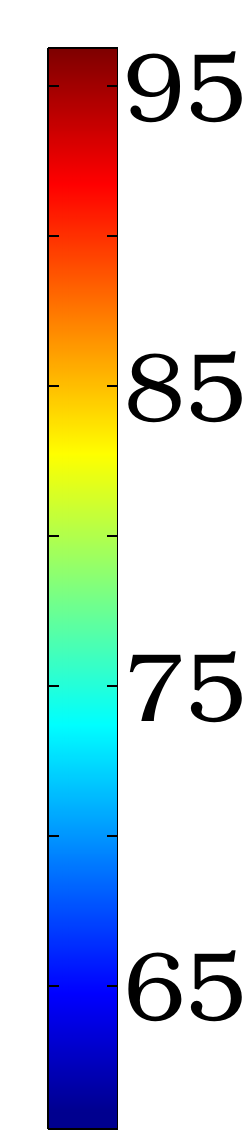} \\


\bf{190} & 
\includegraphics[width=.085\textwidth]{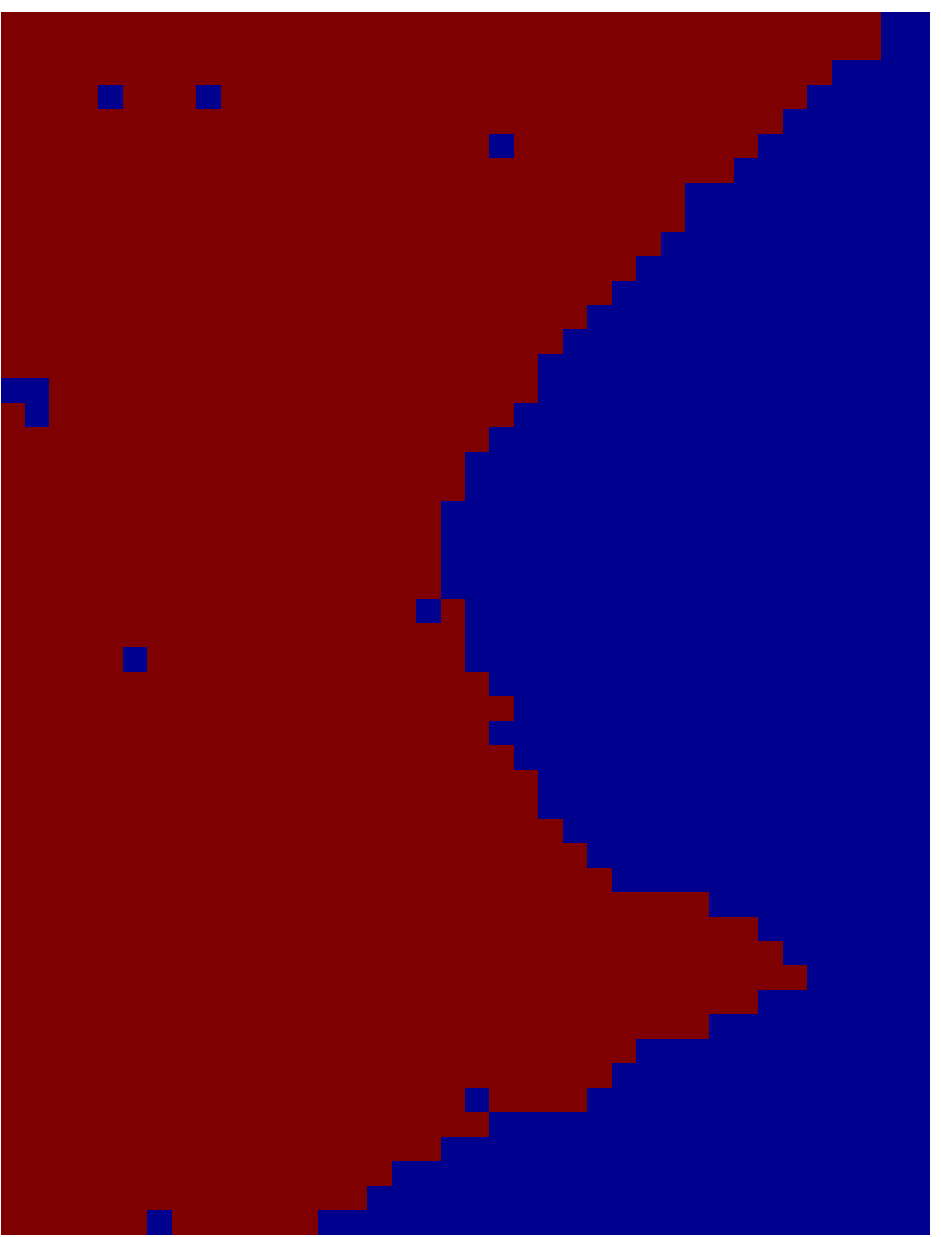} &
\includegraphics[width=.085\textwidth]{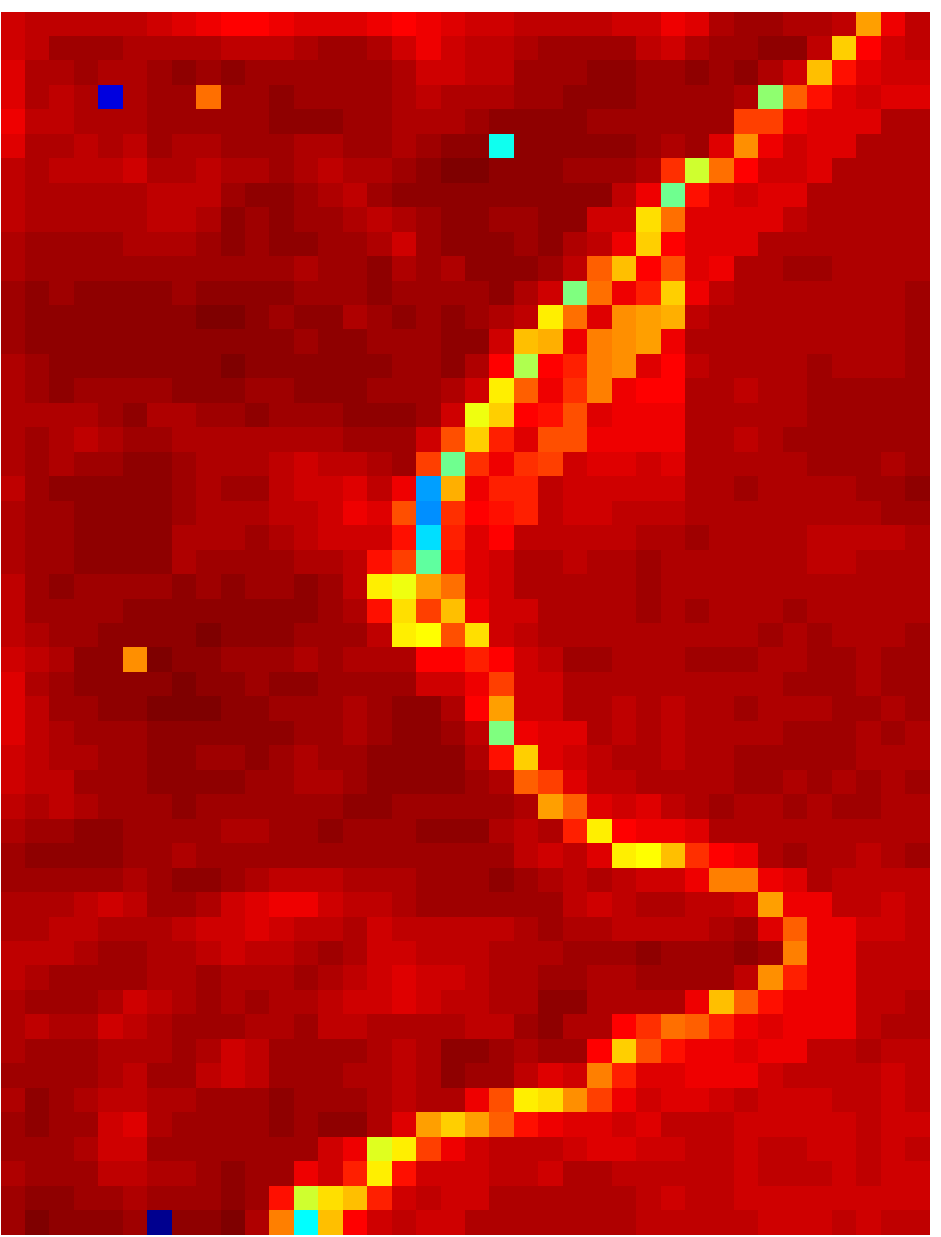} &
\includegraphics[width=.085\textwidth]{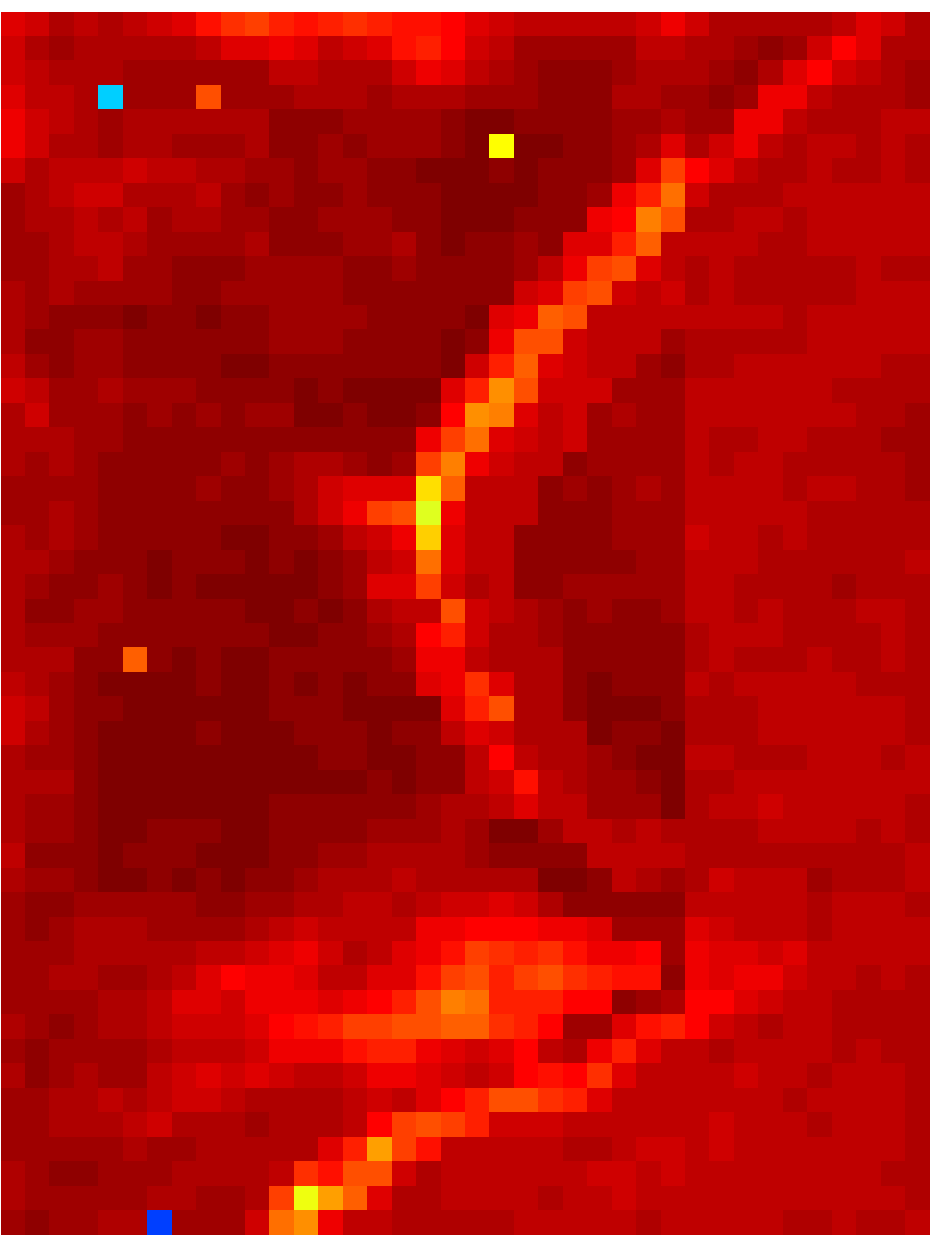} &
\includegraphics[width=.085\textwidth]{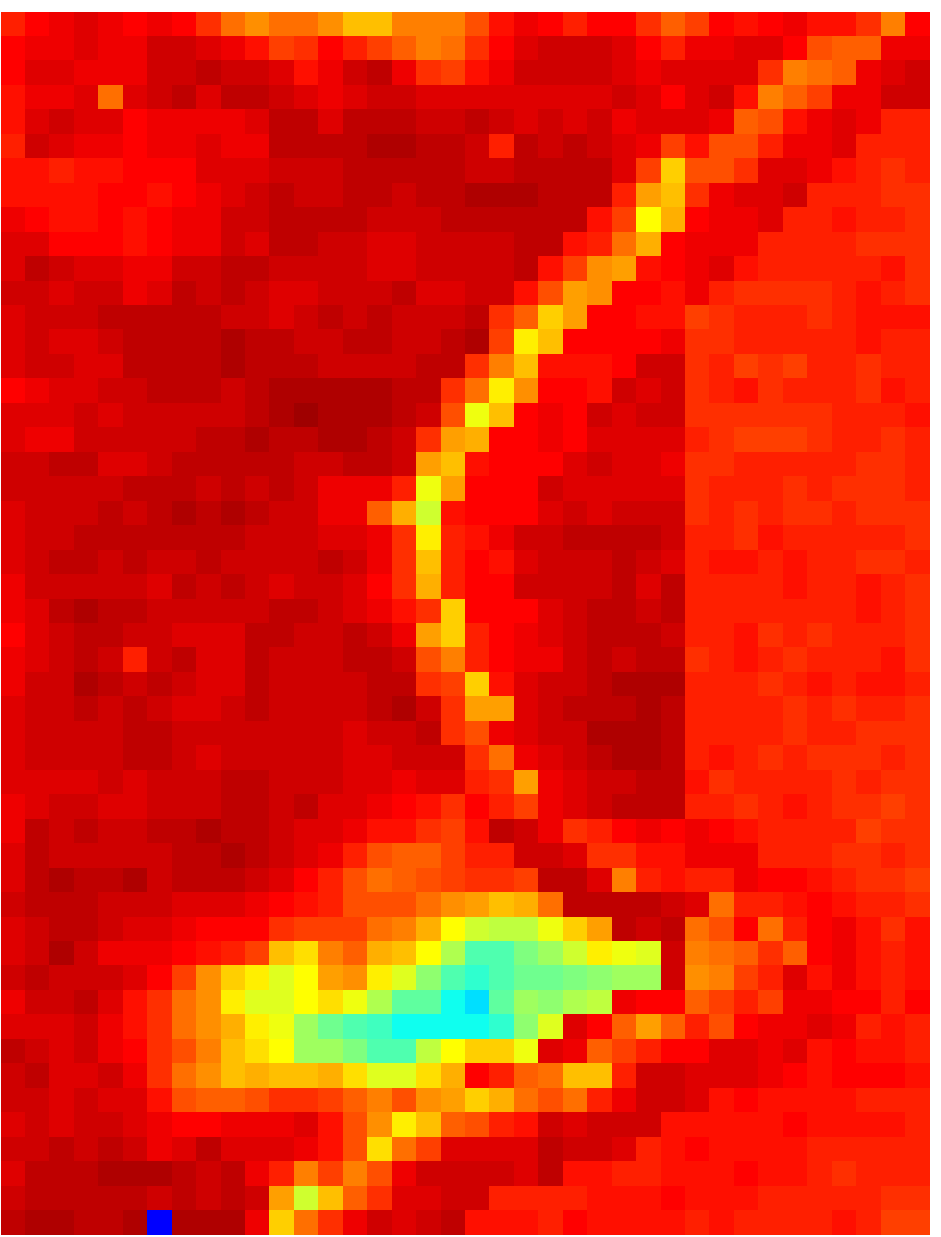} &
\includegraphics[width=.085\textwidth]{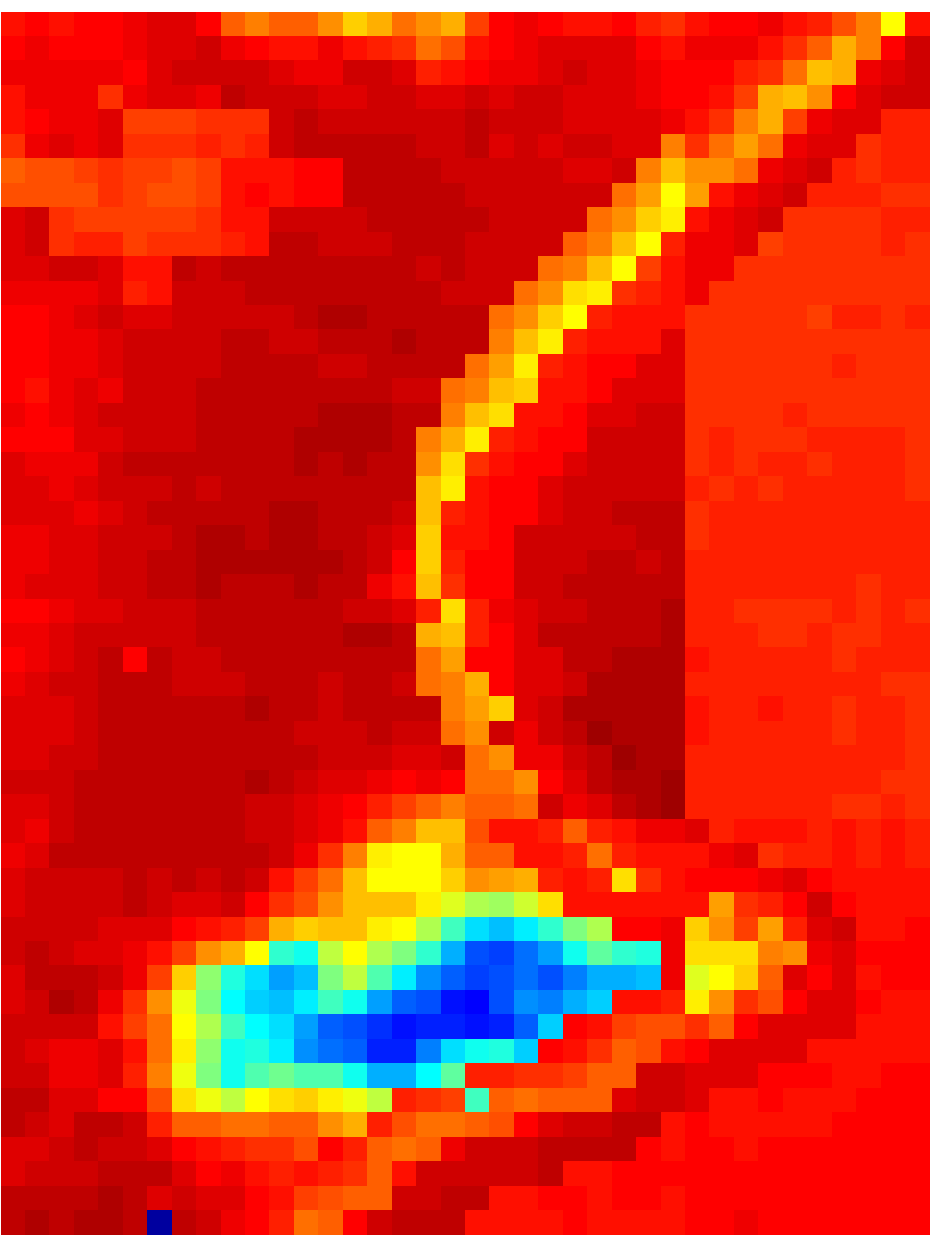} &
\includegraphics[trim=400 5 0 0, clip, width=.022\textwidth]{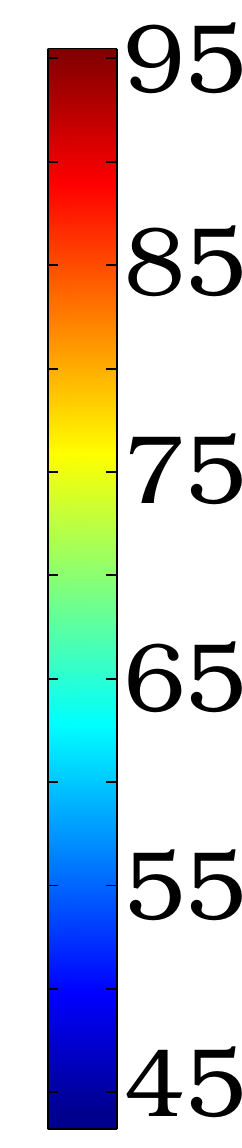} \\

\end{tabular}
\end{center}
\vspace{-0.2cm}
\caption{Averaged accuracy maps over the whole time series for different landmarks (rows) and different SZA ranges (columns).\label{fig:spatial_analysis2}}
\end{figure}

In this section we further analyze the obtained results, but in this case in the spatial domain by paying attention to the average classification accuracy per pixel throughout the year 2010. 
Results are shown in Fig.~\ref{fig:spatial_analysis2} for the 12 considered landmarks. \blue{Clear spatial patterns of the classification error can be observed in almost all landmarks. The OA patterns match the coastline in most cases, which agrees with previous results regarding the illumination conditions (Fig.~\ref{fig:illu_analysis1}) and land cover (Fig.~\ref{fig:LC_OA}) at high light intensity conditions near the coastlines. For the LMs 0, 14 and 131, these poor results in coastline areas are probably due to the misclassification of bright sand.}
\blue{Additionally, we should stress that these spatial patterns on the OA maps are noticeable for some landmarks also at night times and during twilight, especially over coastlines (e.g. LMs 83, 120, and 190) and islands with high mountains (e.g. LMs 120, 154, and 177). For this latter case, in LM120 (Etna Volcano, 3.350~m, Sicily) and LM154 (Mount Attavyros,  1.216~m, Rhodes) we observe spots on the middle of the islands, especially at night. In the case of Tenerife (LM177), we observe similar patterns but in all the center of the island (from 1000~m altitude) and not only at the Teide Volcano (3.718~m).}
\blue{These errors in the L2 mask are further analyzed in next section, since it poses the crucial question about the trustworthiness of the L2 cloud mask over the coastlines and islands. In particular, we present some examples showing such eventual errors in the L2 mask (used as ground truth) and hence also in the classifier predictions.}

\subsection{Visual analysis of L2 mask consistency and errors over coastlines} 

\begin{figure}[t!]
\scriptsize
\begin{center}
\setlength{\tabcolsep}{1.5pt}
\begin{tabular}{lcccc}
{\bf LM} & {\bf Land cover}  & {\bf RGB/Band 9}  & {\bf L2 mask}  & {\bf Prediction}  \\
\bf{0} &
\includegraphics[width=1.50cm]{./images/accuracy/0_LC_mod.png} &
\includegraphics[width=1.50cm]{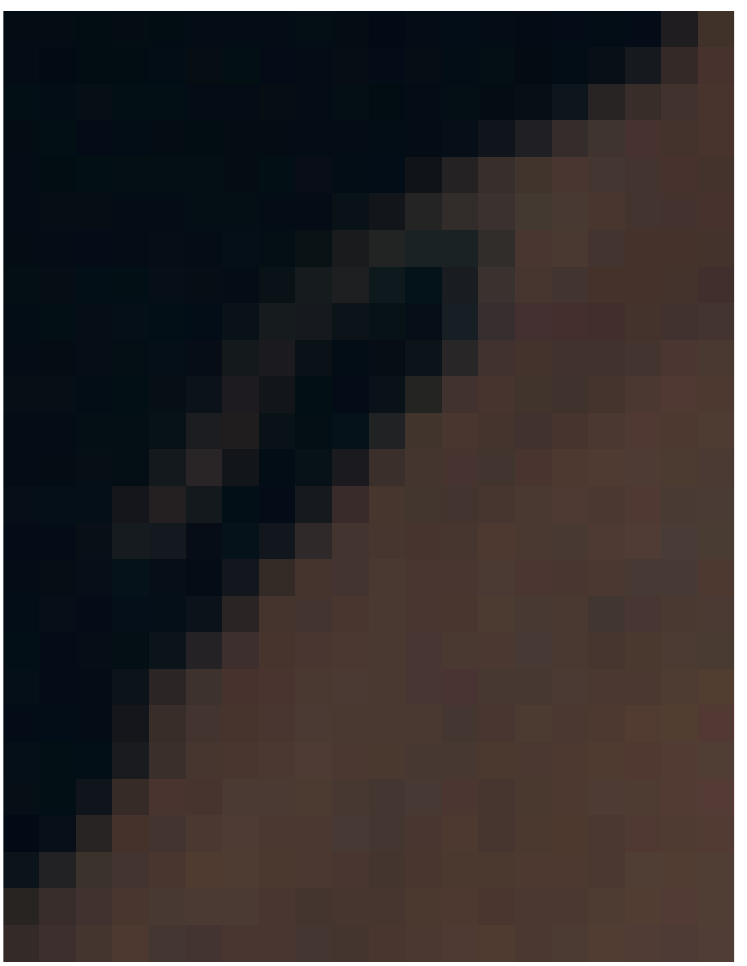} &
\includegraphics[width=1.50cm]{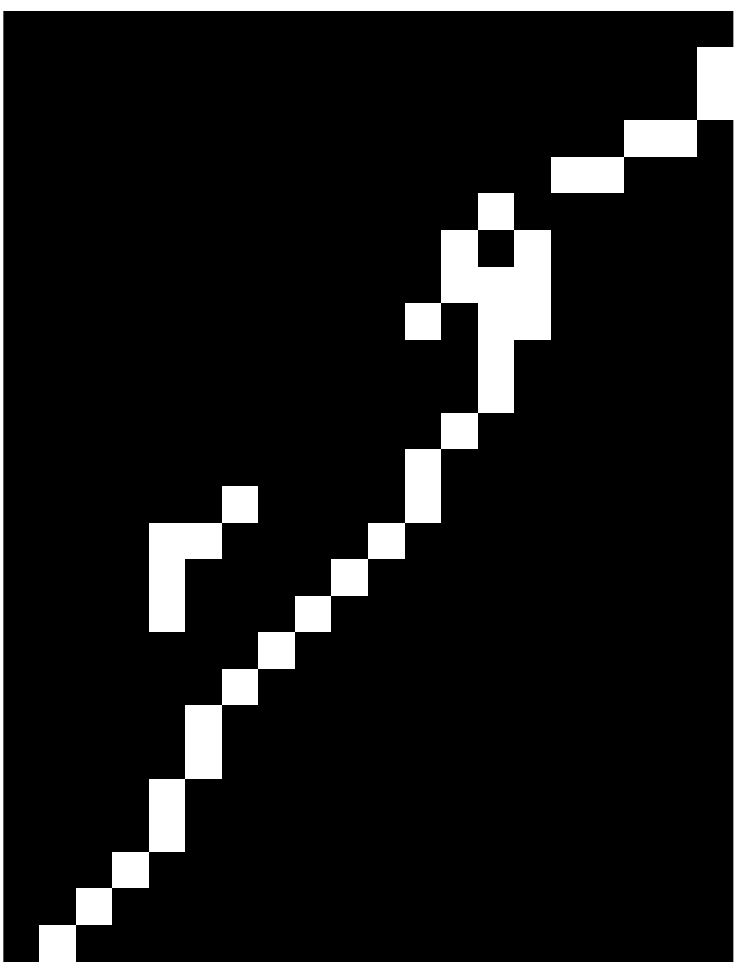} &
\includegraphics[width=1.50cm]{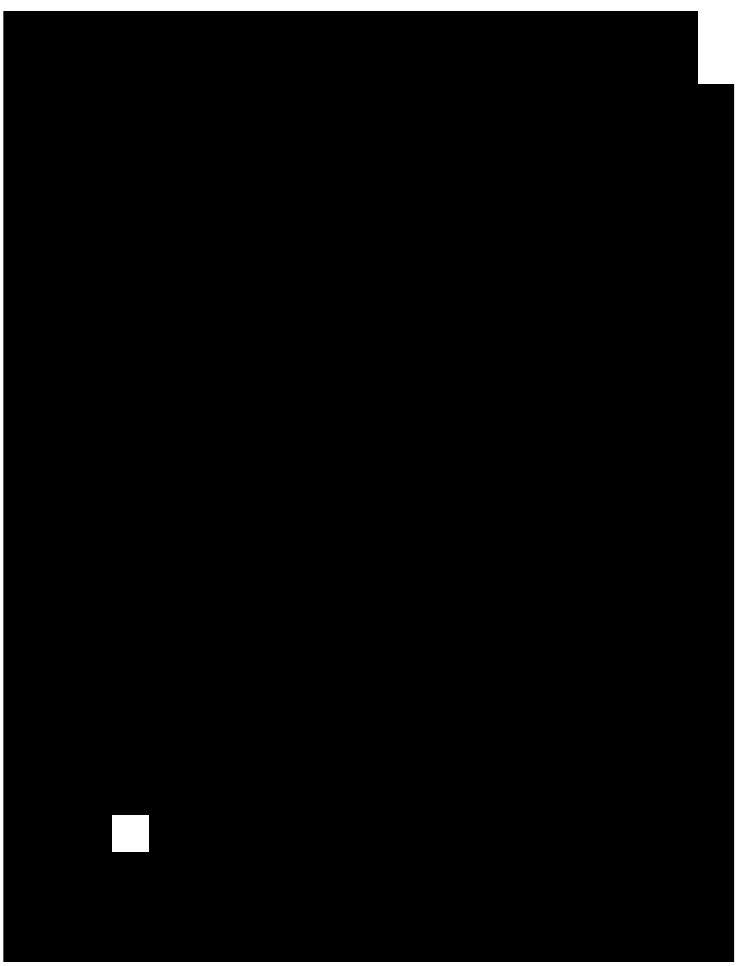}
\\
\bf{14} &
\includegraphics[width=1.50cm]{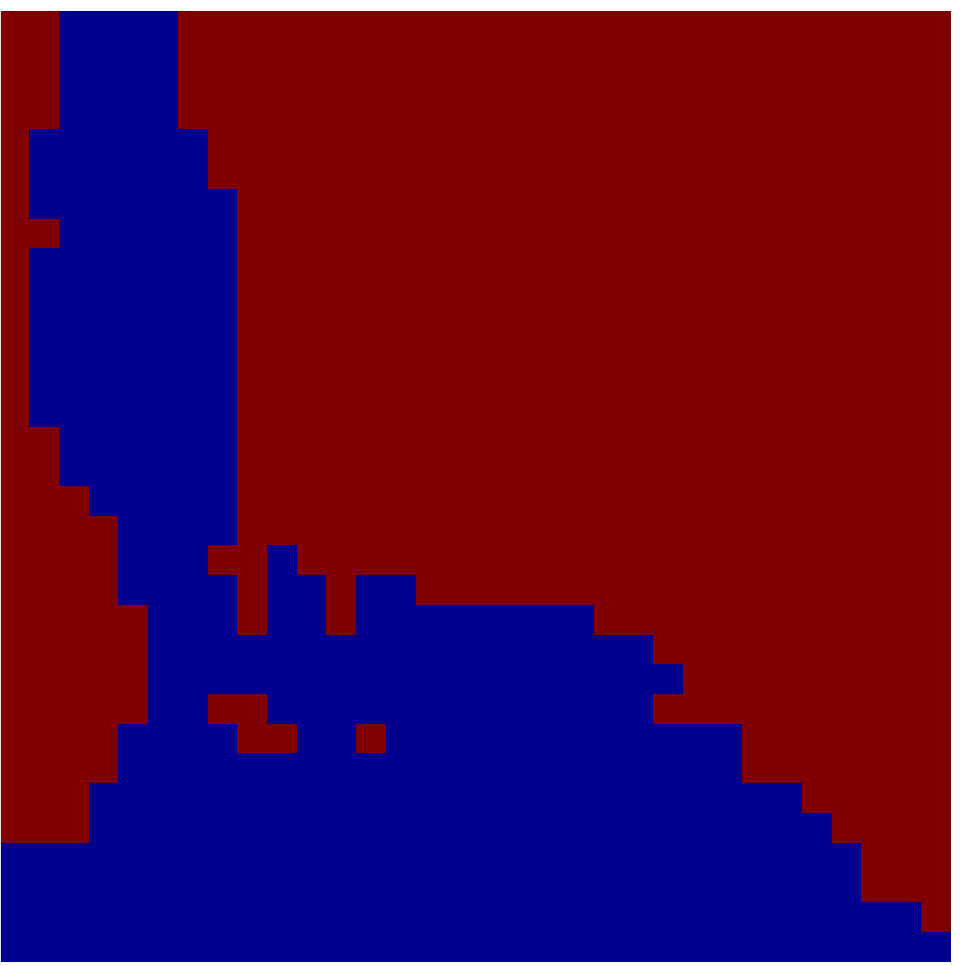} &
\includegraphics[width=1.50cm]{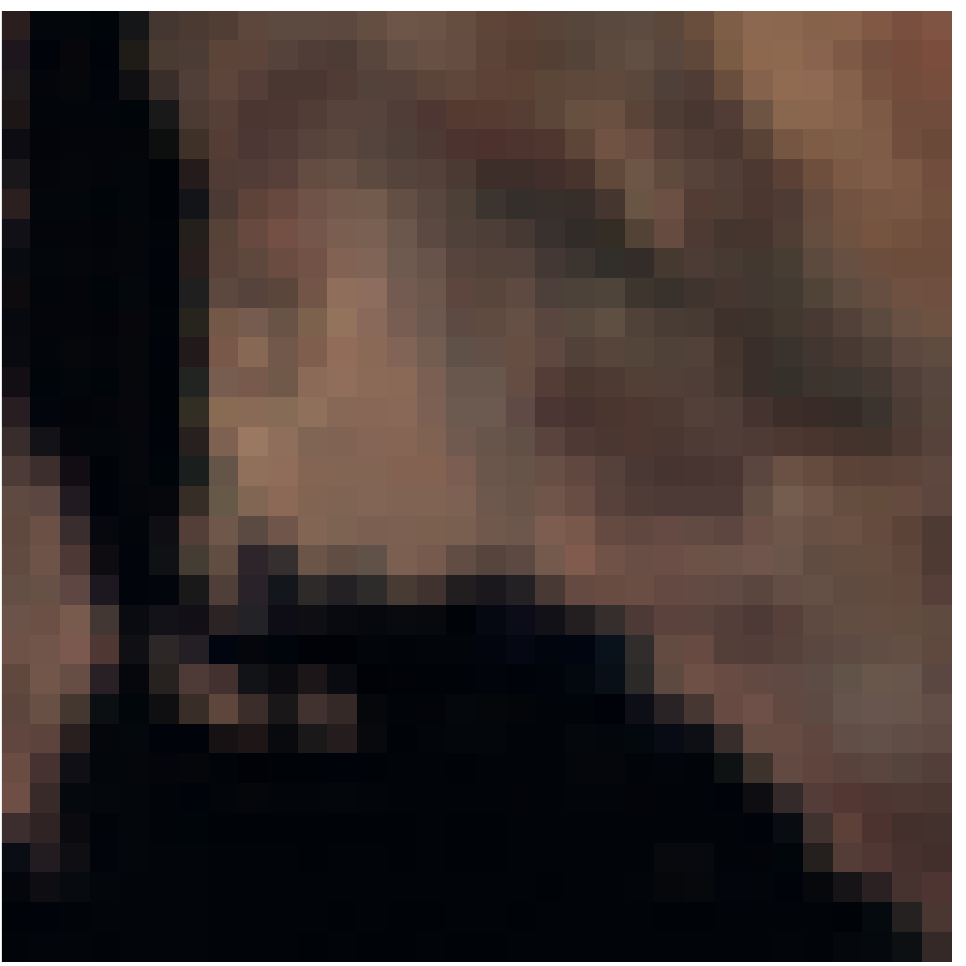} &
\includegraphics[width=1.50cm]{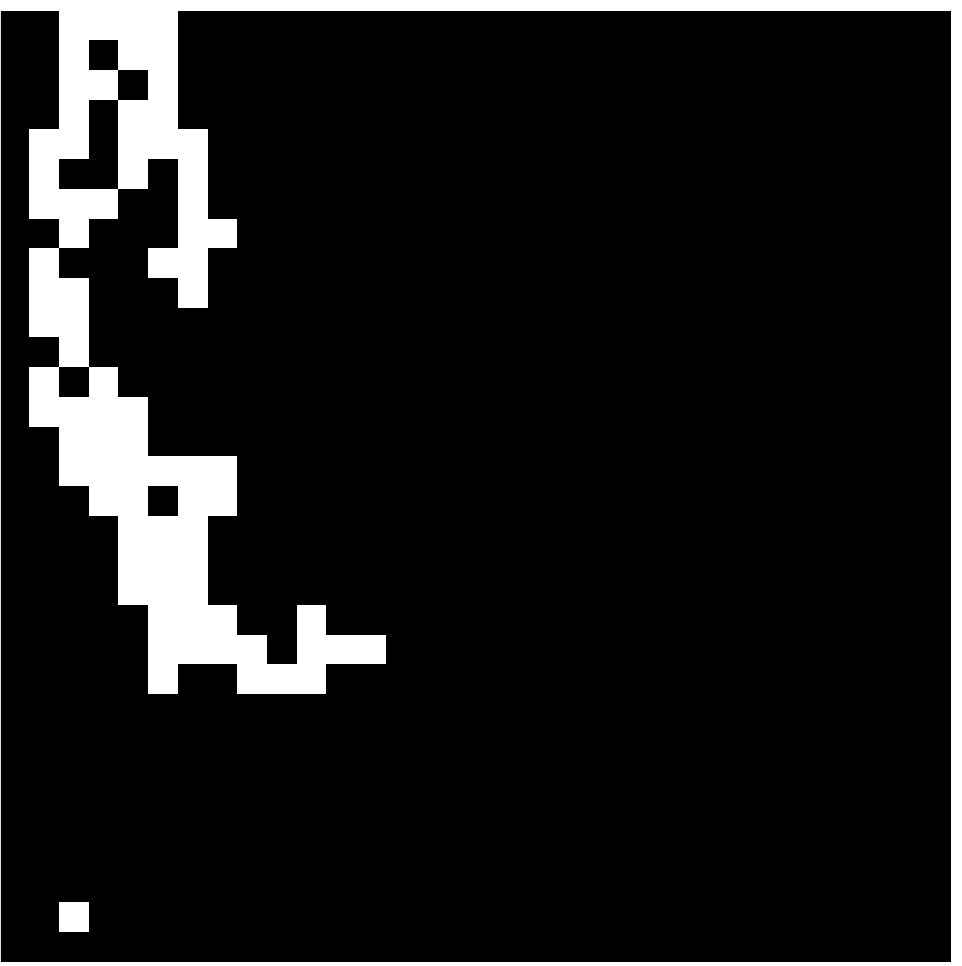} & 
\includegraphics[width=1.50cm]{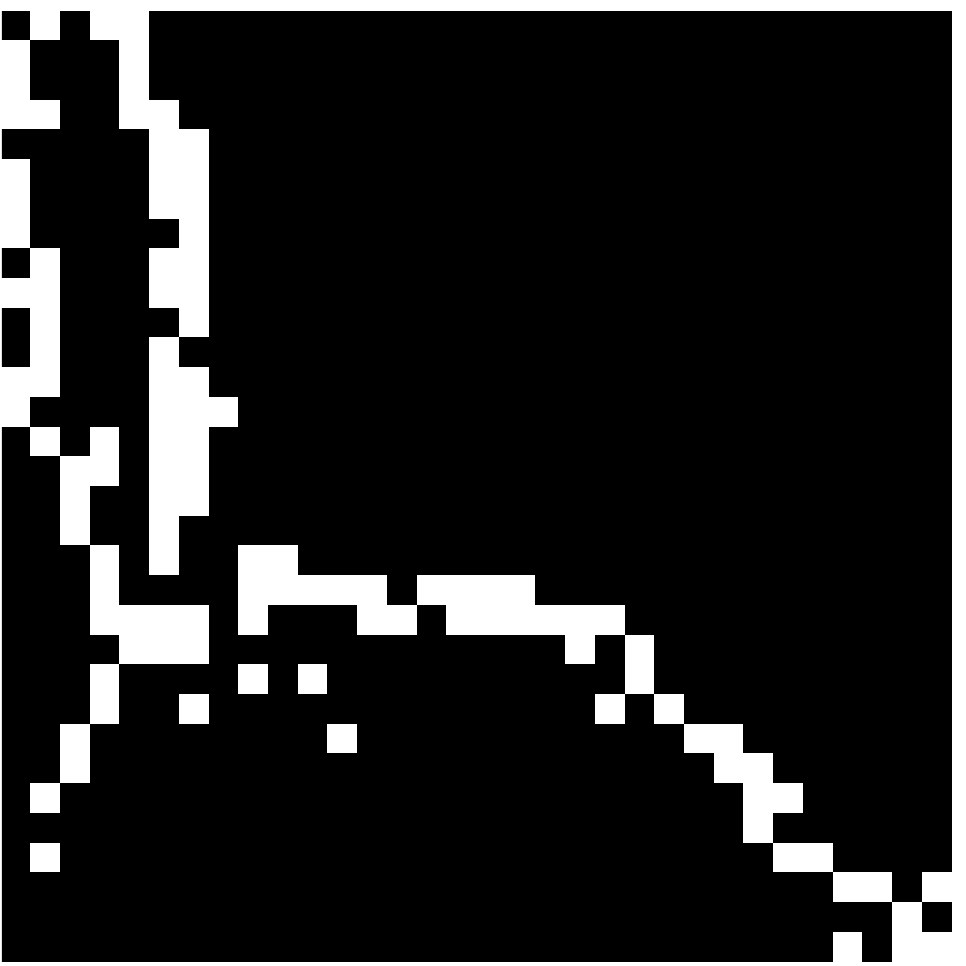}
\vspace{0.45cm} 
\\
\bf{17} &
\includegraphics[width=1.50cm]{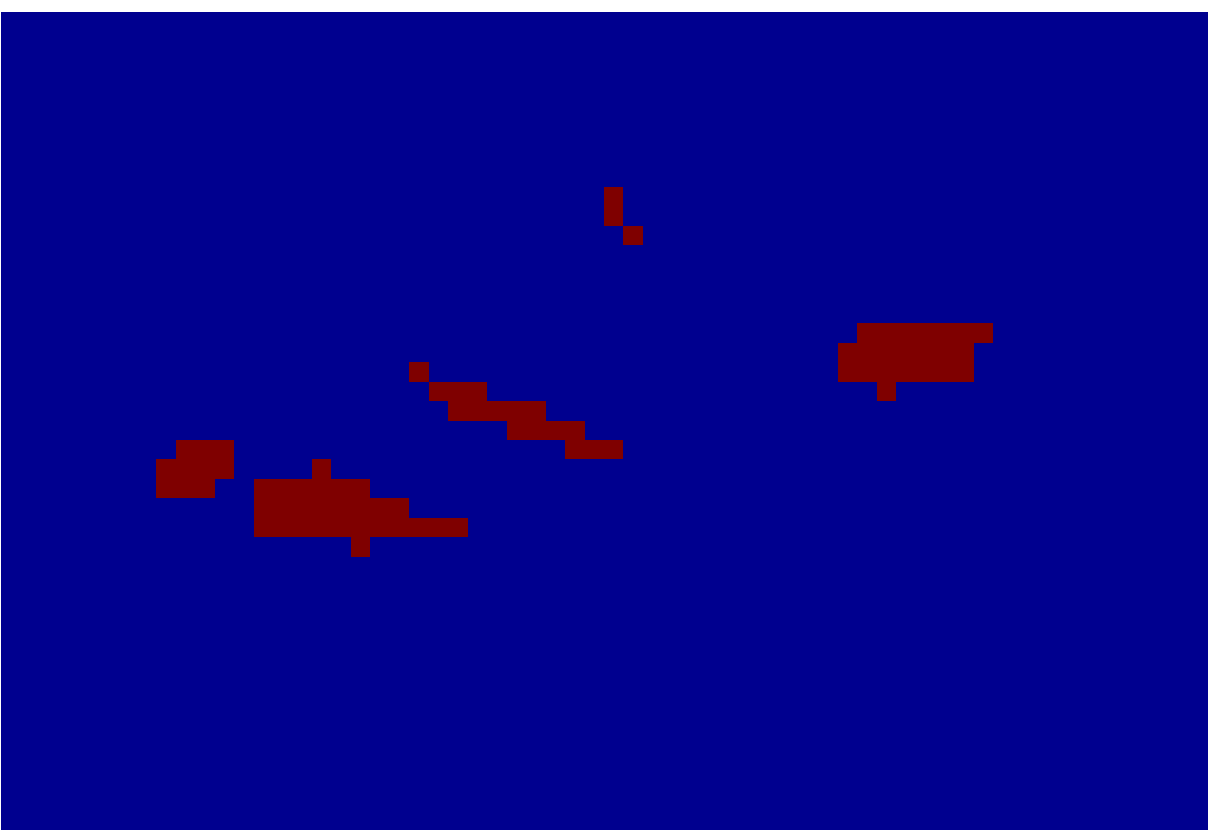} &
\includegraphics[trim=0 0 0 1.2cm, clip, width=1.50cm]{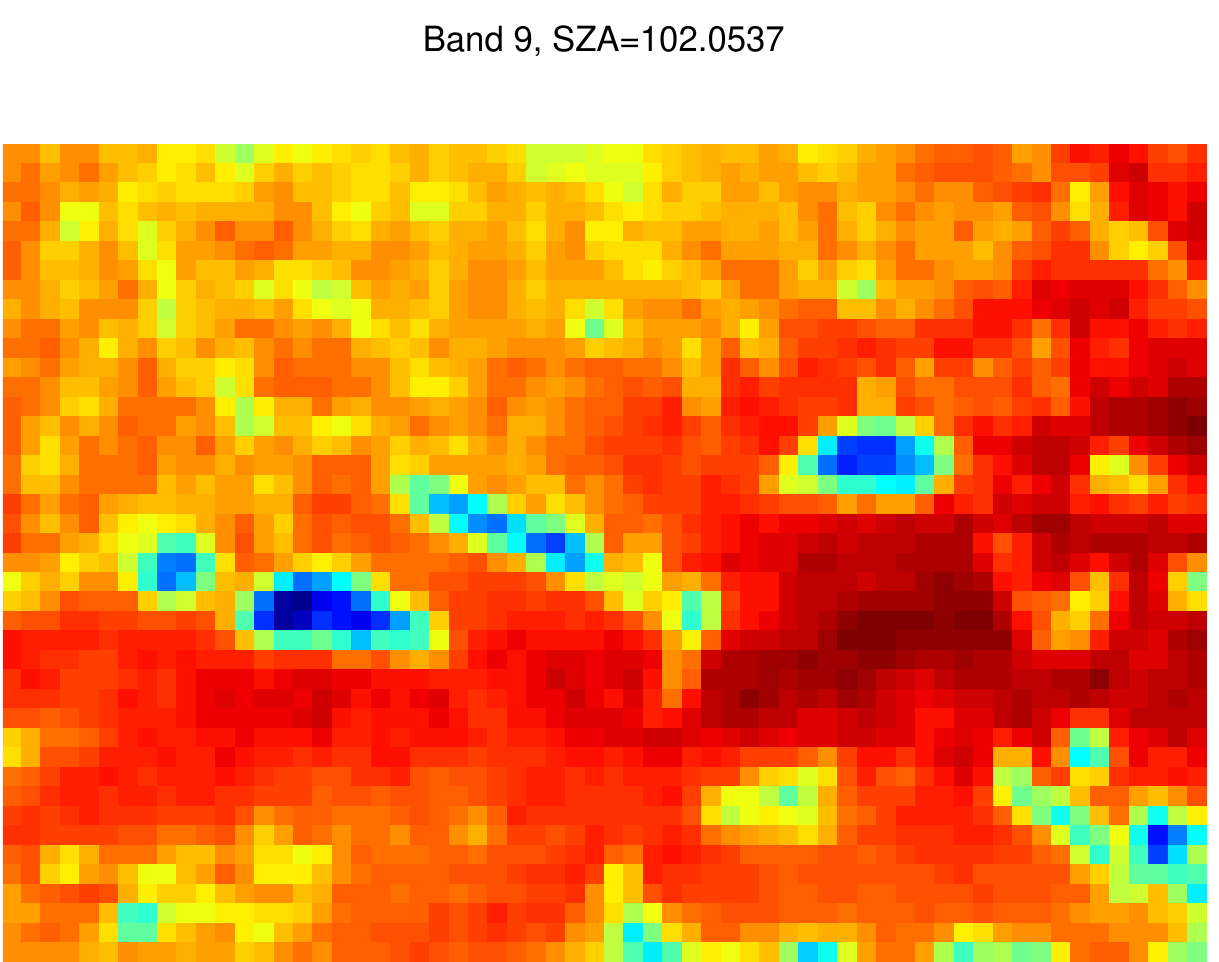} &
\includegraphics[width=1.50cm]{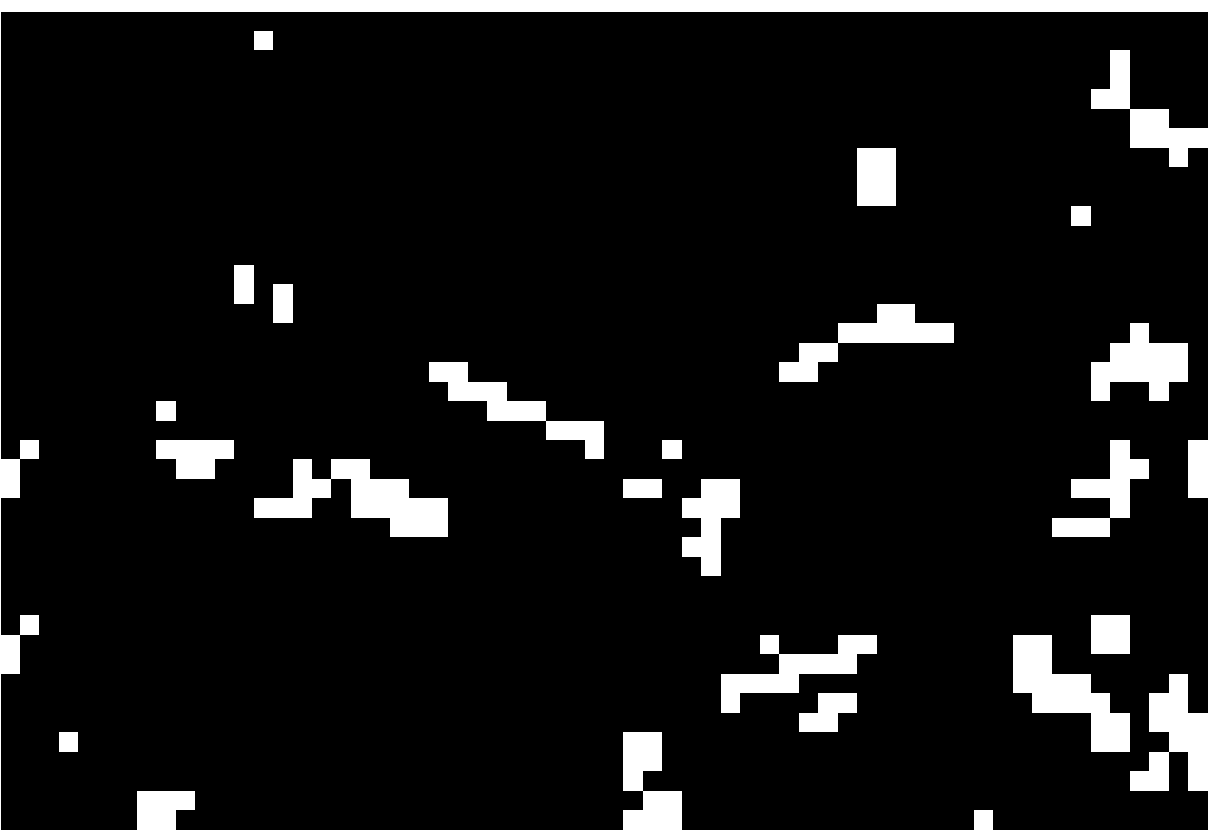} & 
\includegraphics[width=1.50cm]{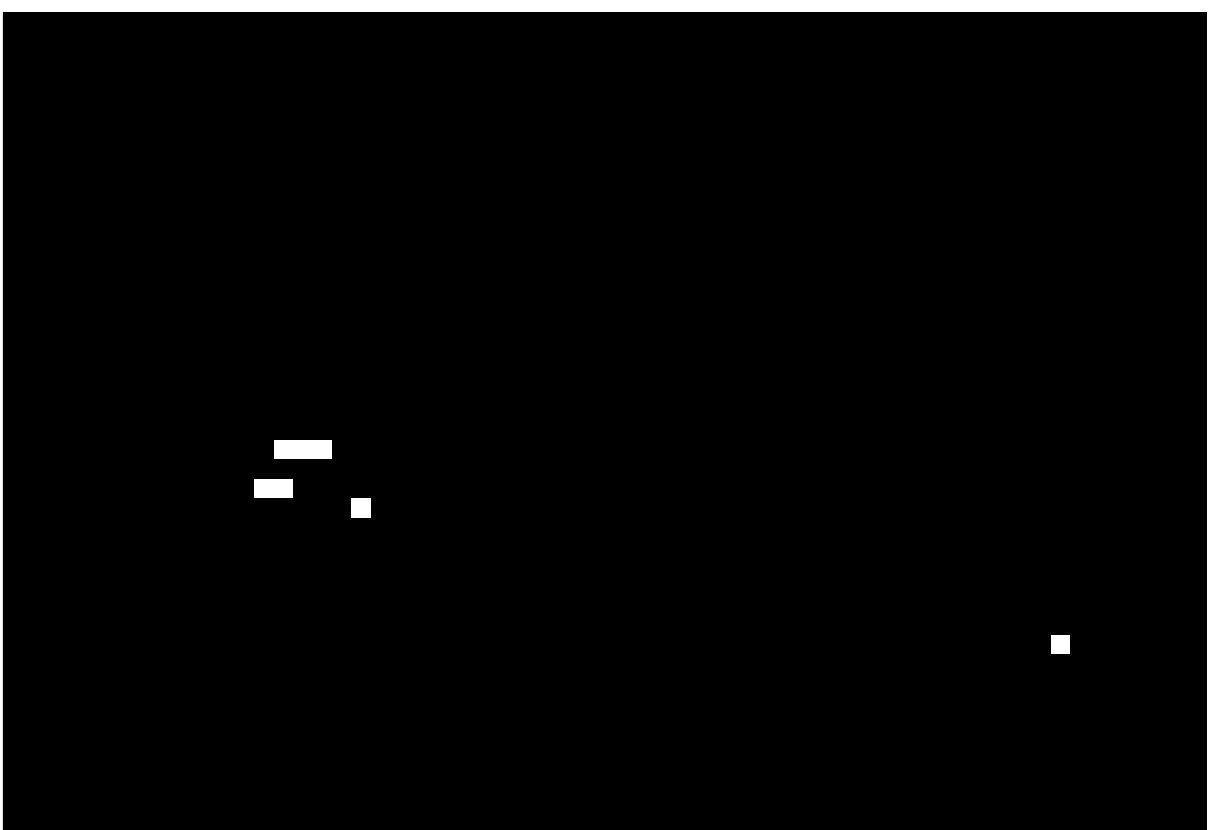}
\\
\bf{48} &
\includegraphics[width=1.50cm]{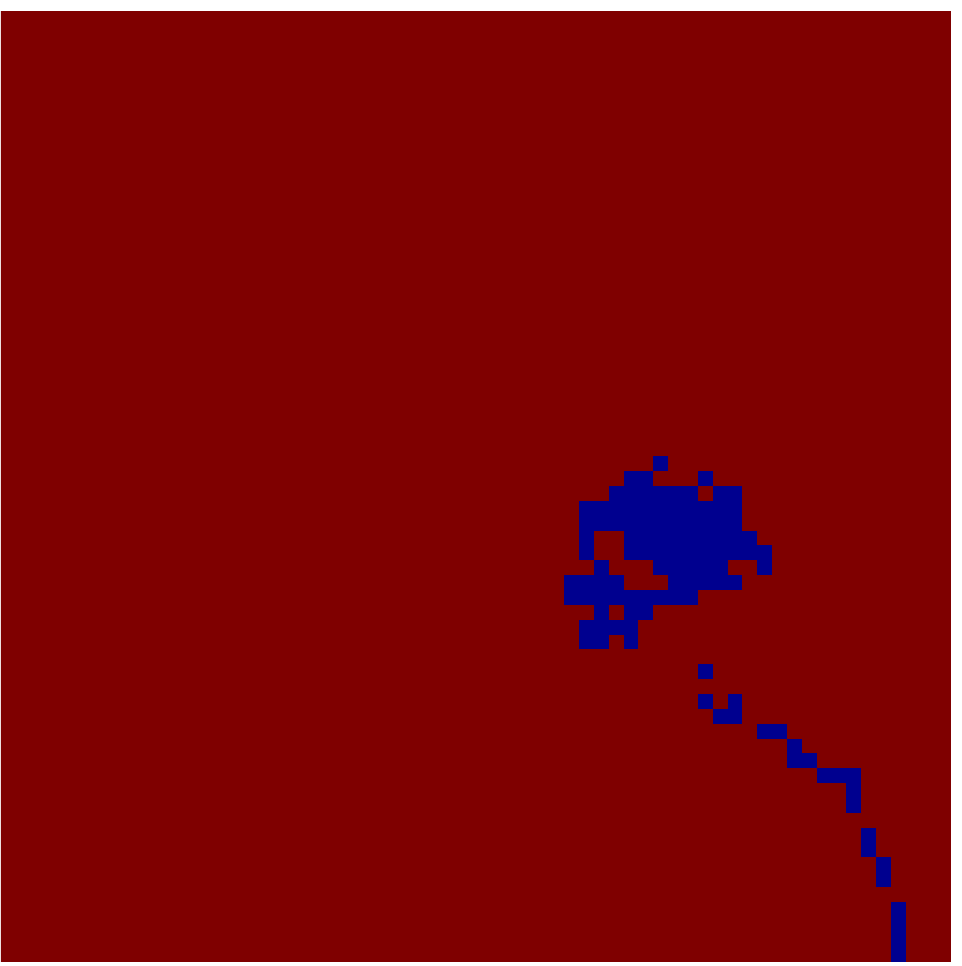} &
\includegraphics[width=1.50cm]{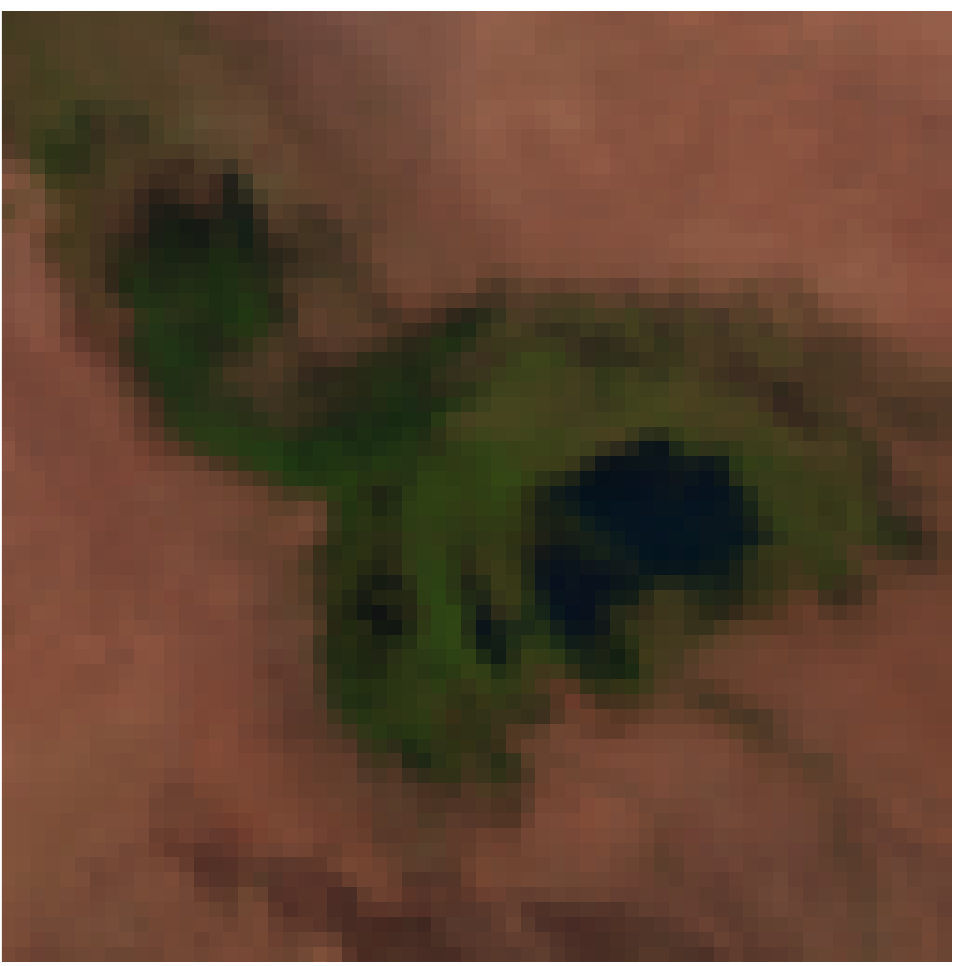} &
\includegraphics[width=1.50cm]{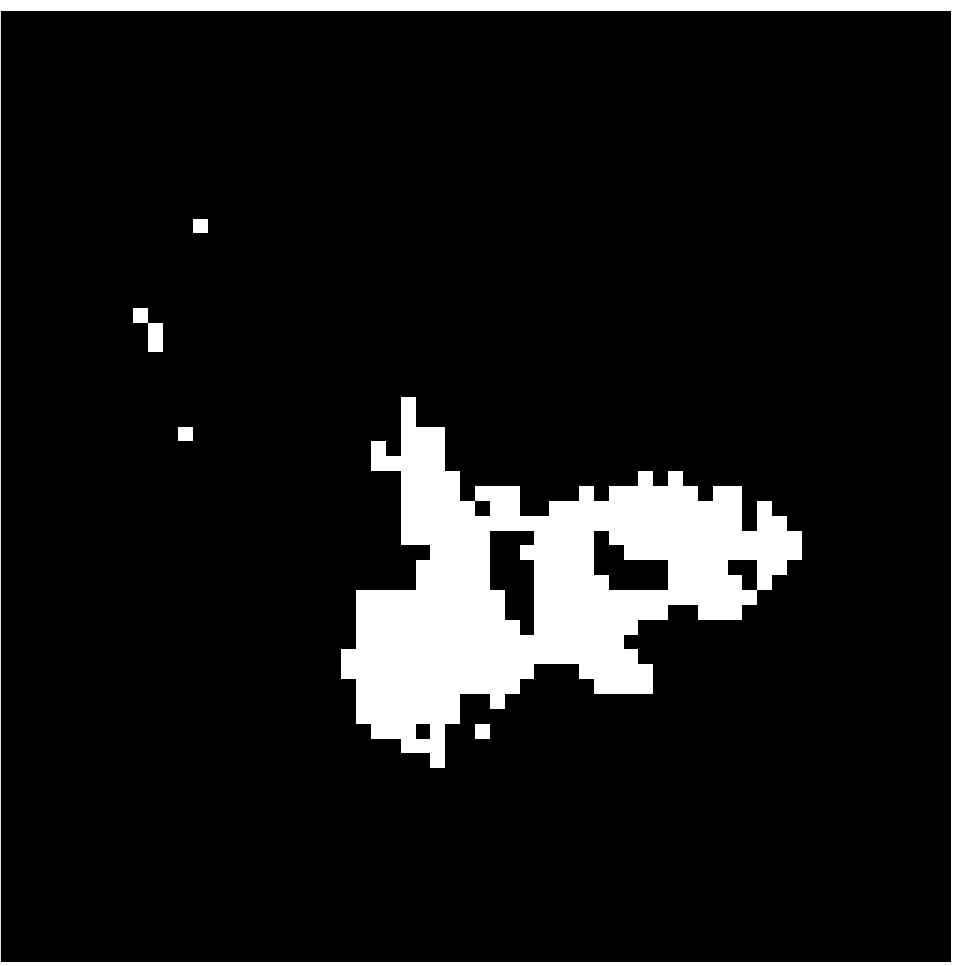} & 
\includegraphics[width=1.50cm]{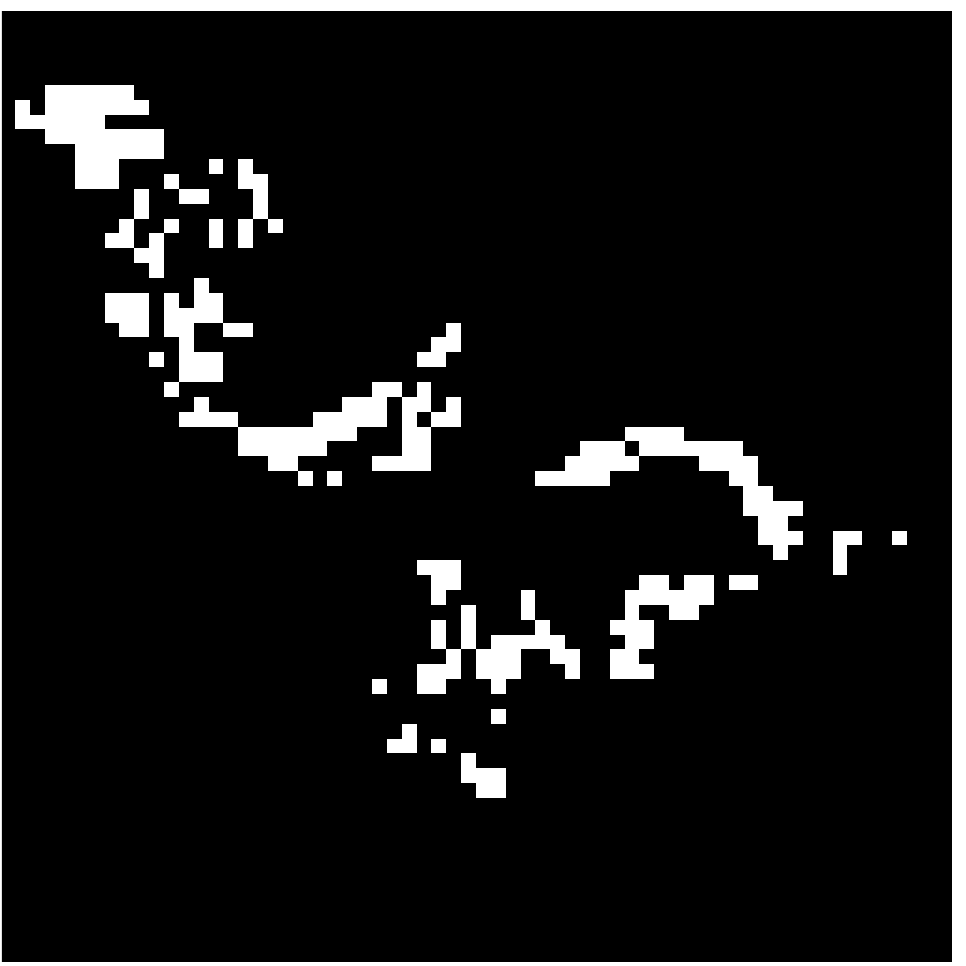}
\\
\bf{63} &
\includegraphics[width=1.50cm]{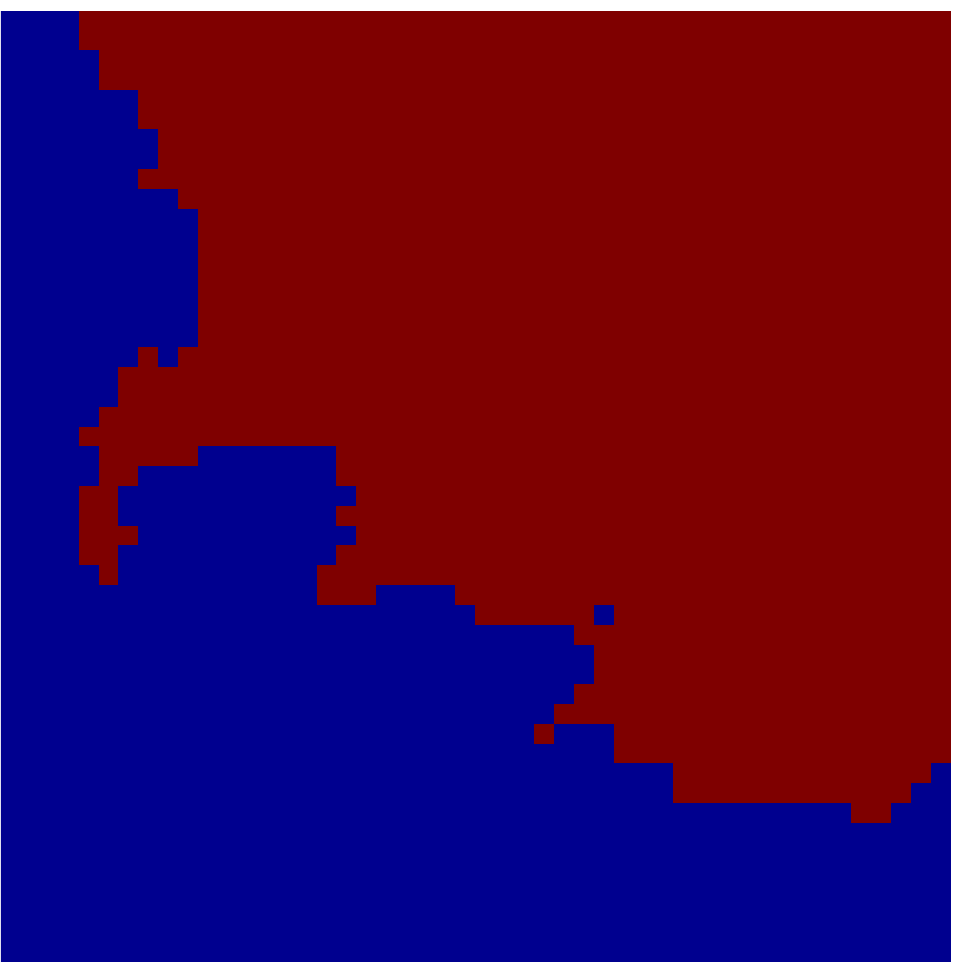} &
\includegraphics[trim=0 0 0 0.8cm, clip, width=1.50cm]{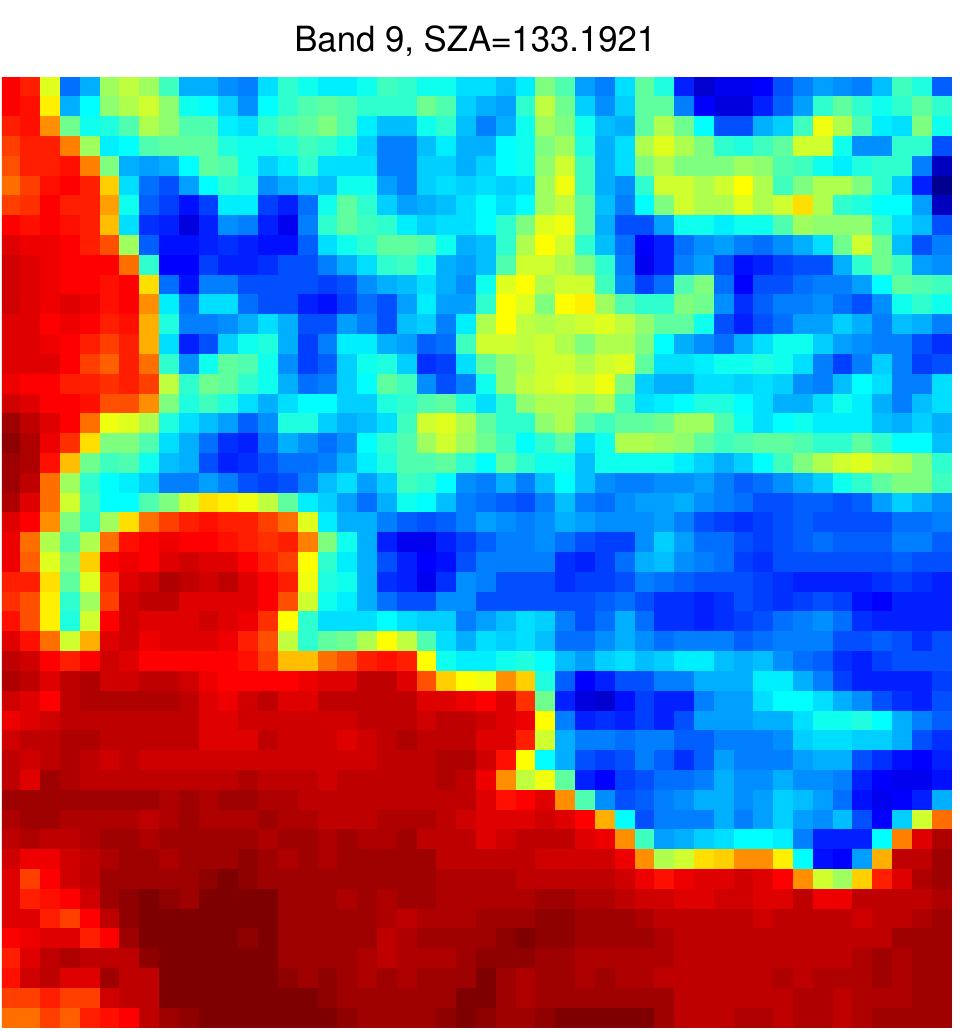} &
\includegraphics[width=1.50cm]{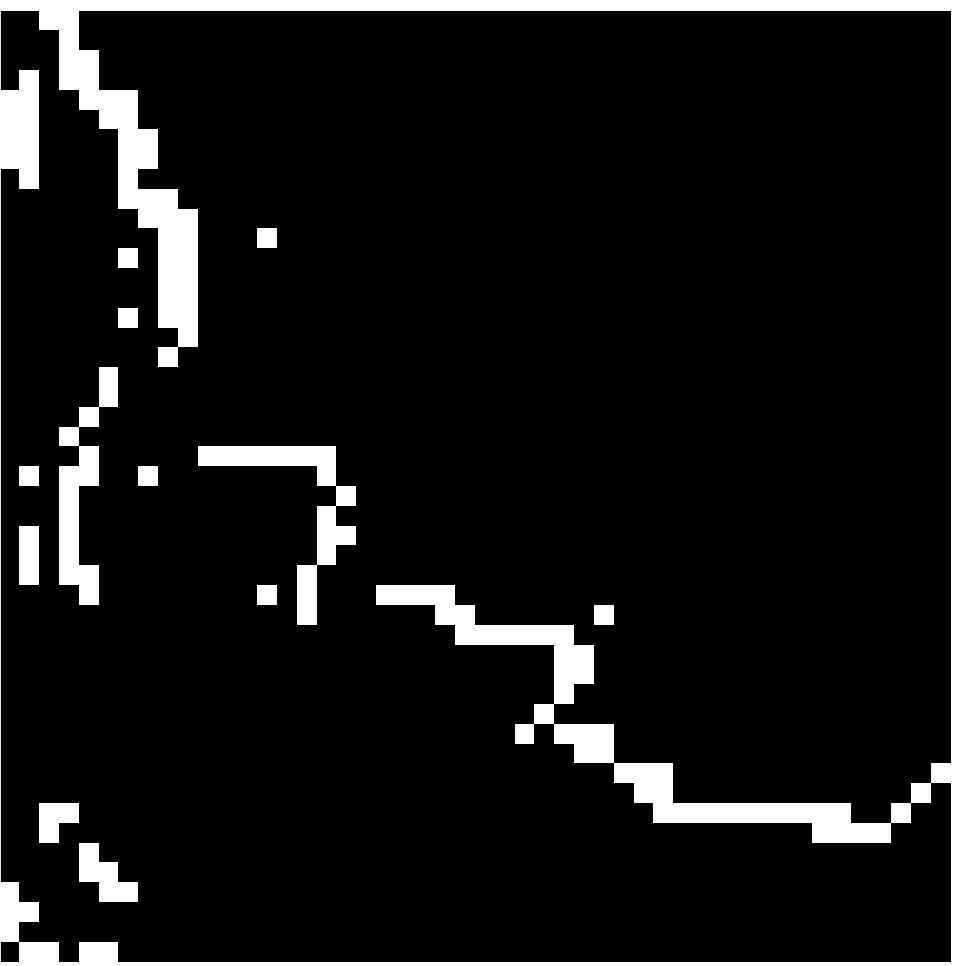} & 
\includegraphics[width=1.50cm]{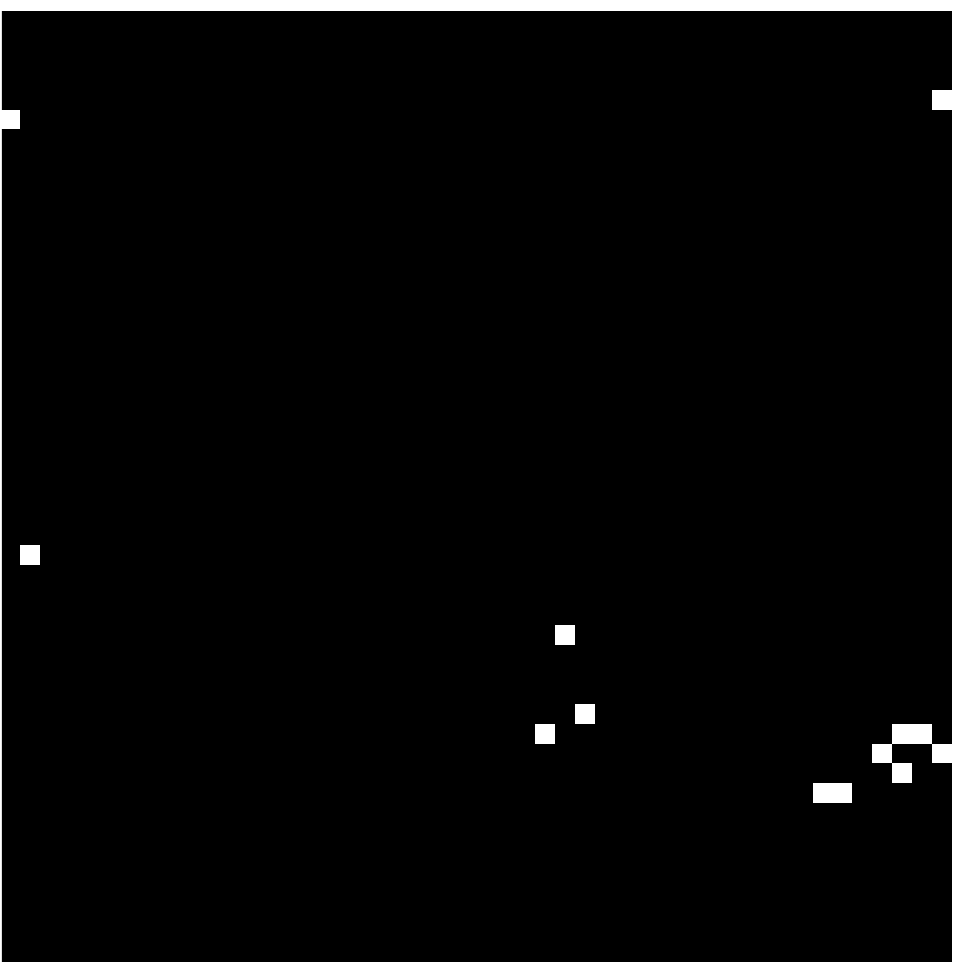}
\\
\bf{83} &
\includegraphics[width=1.50cm]{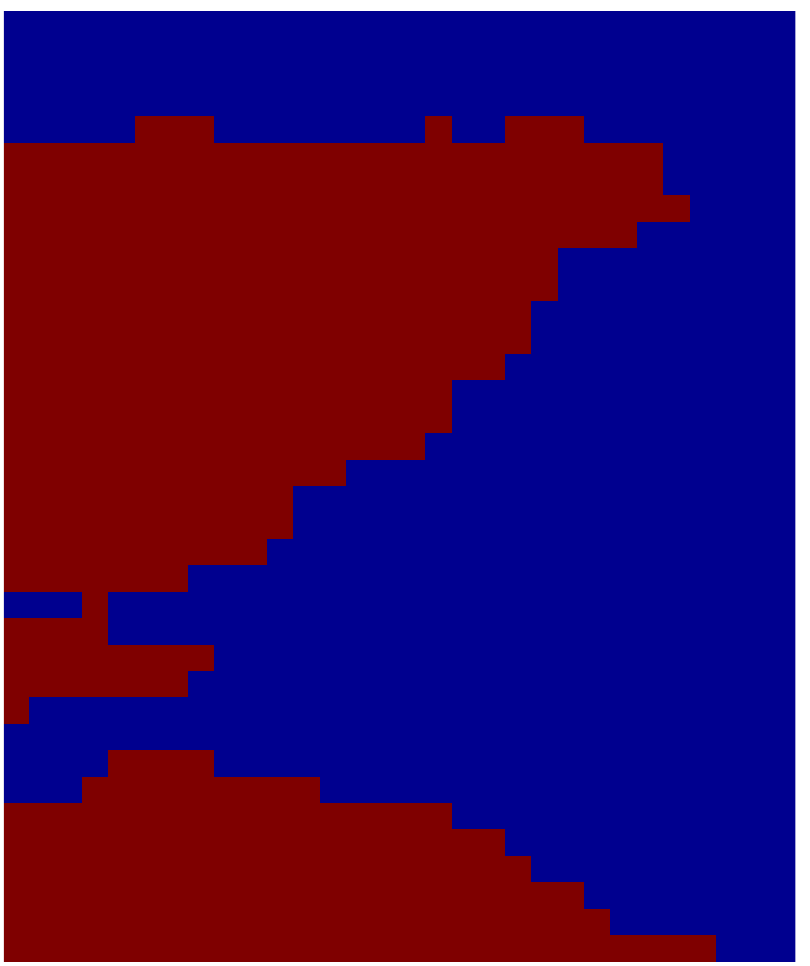} &
\includegraphics[width=1.50cm]{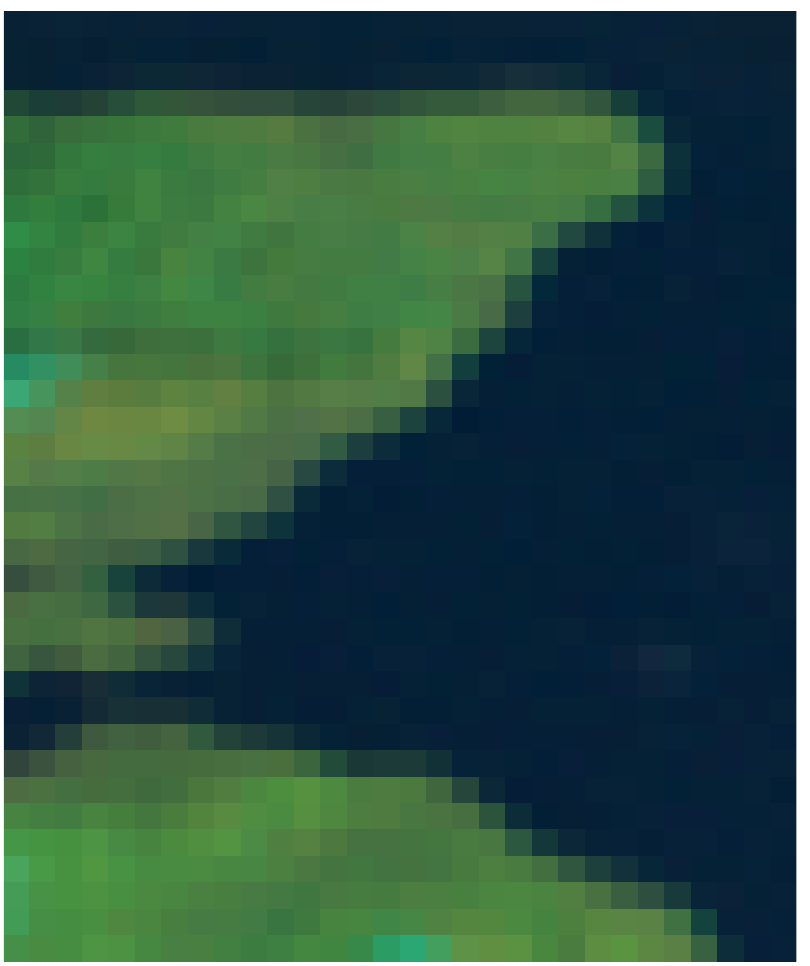} &
\includegraphics[width=1.50cm]{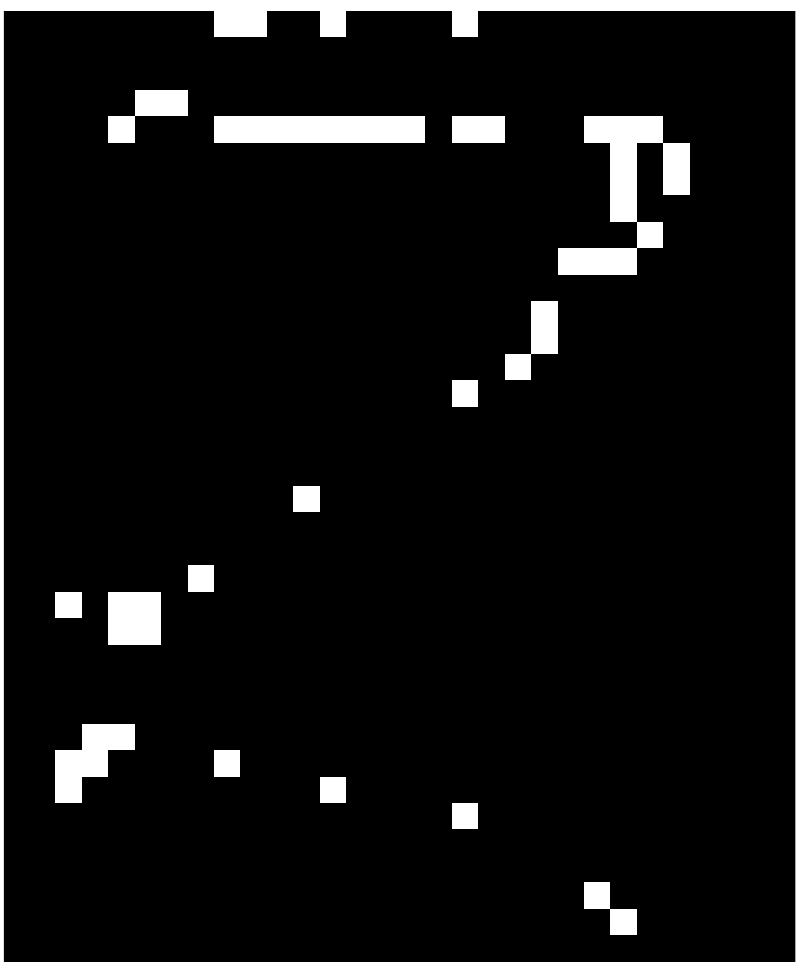} & 
\includegraphics[width=1.50cm]{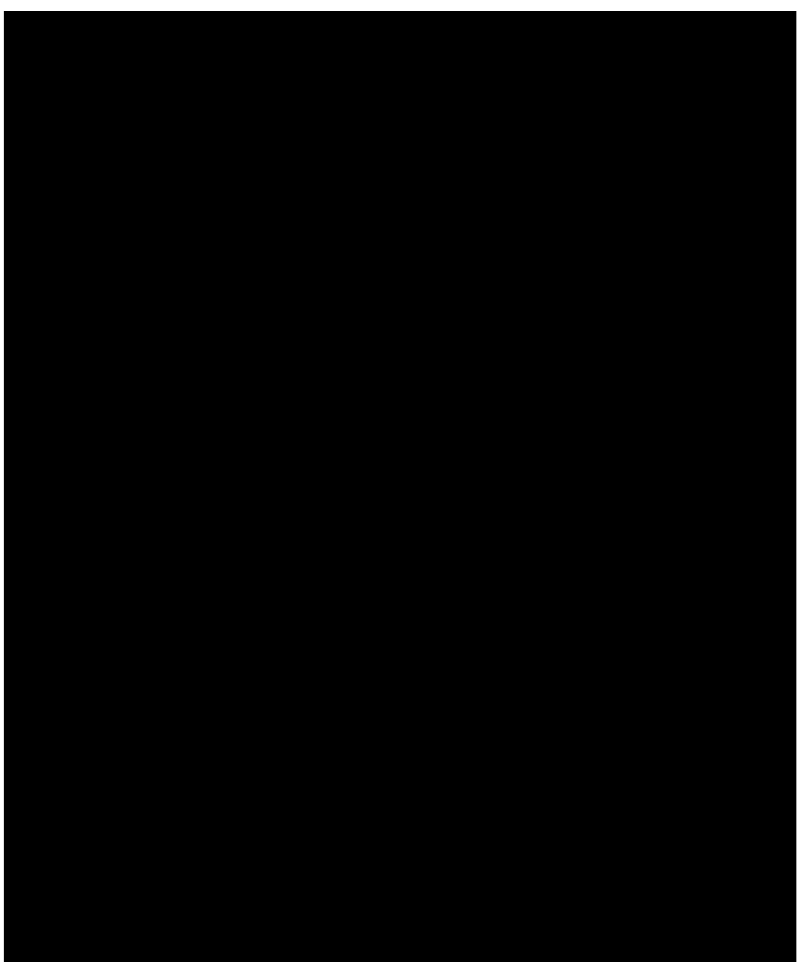}\\
\bf{107} &
\includegraphics[width=1.50cm]{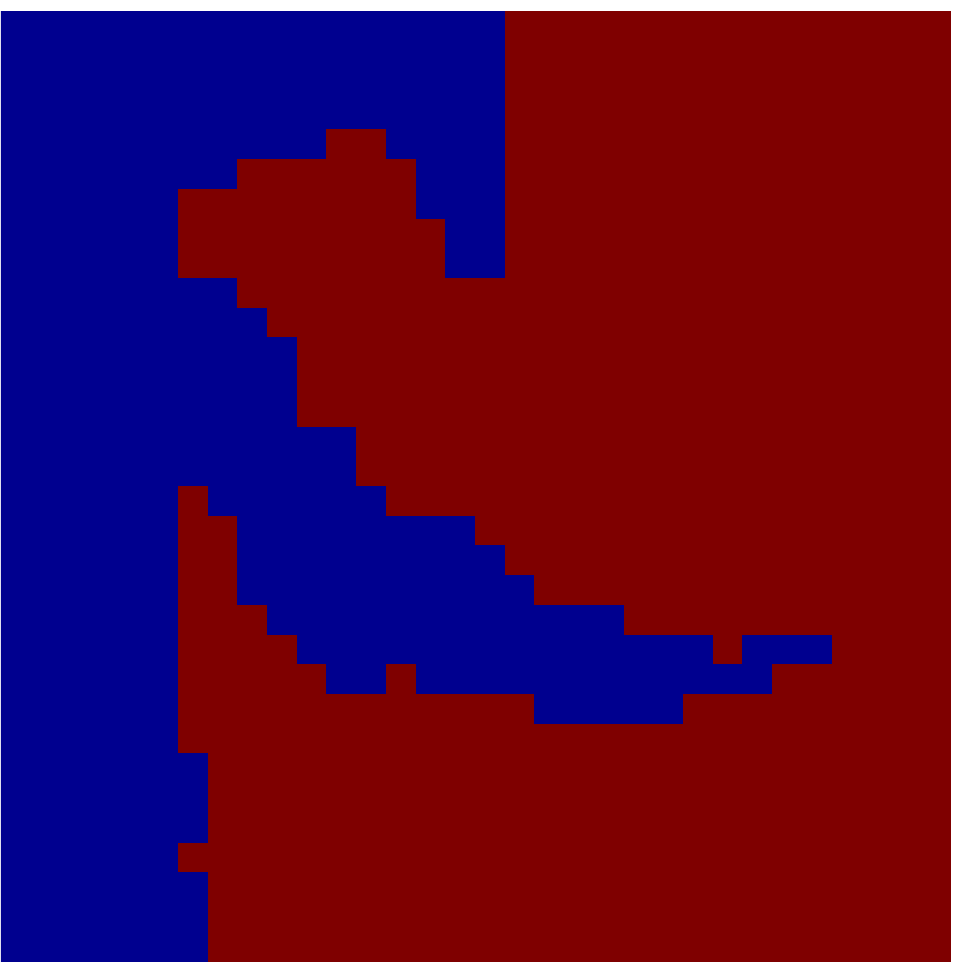} &
\includegraphics[width=1.50cm]{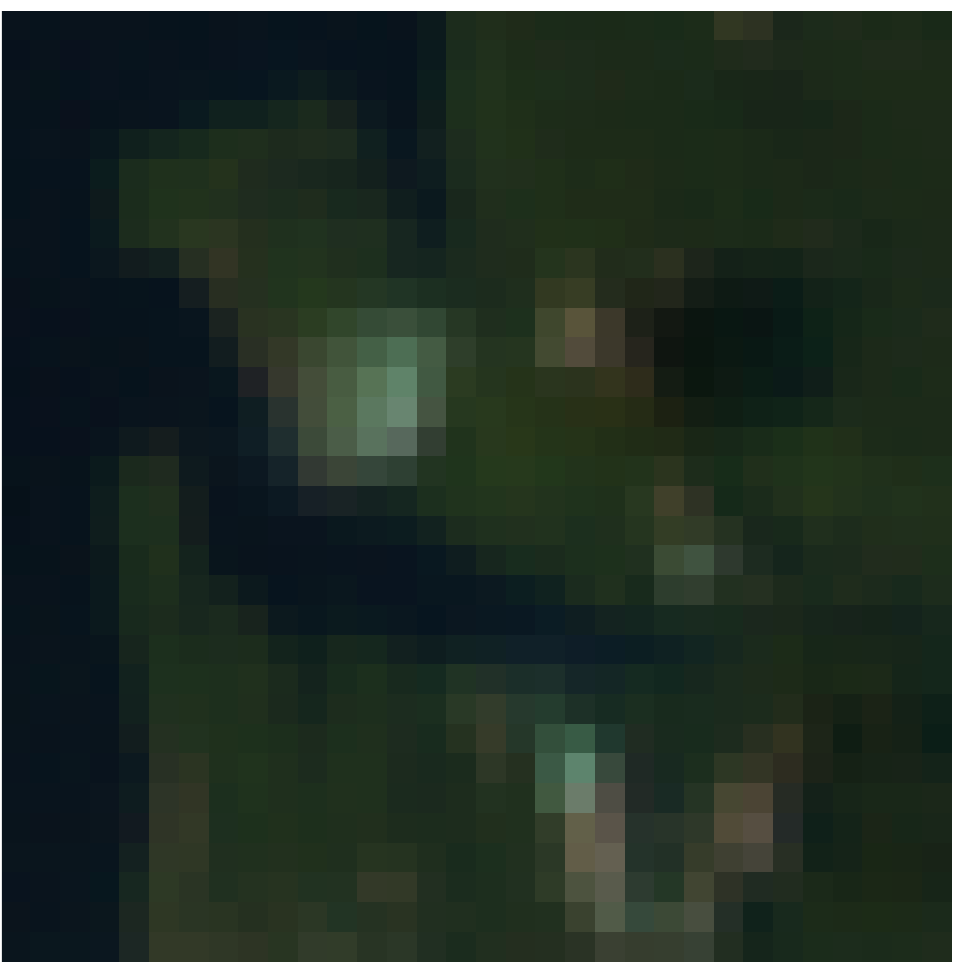} &
\includegraphics[width=1.50cm]{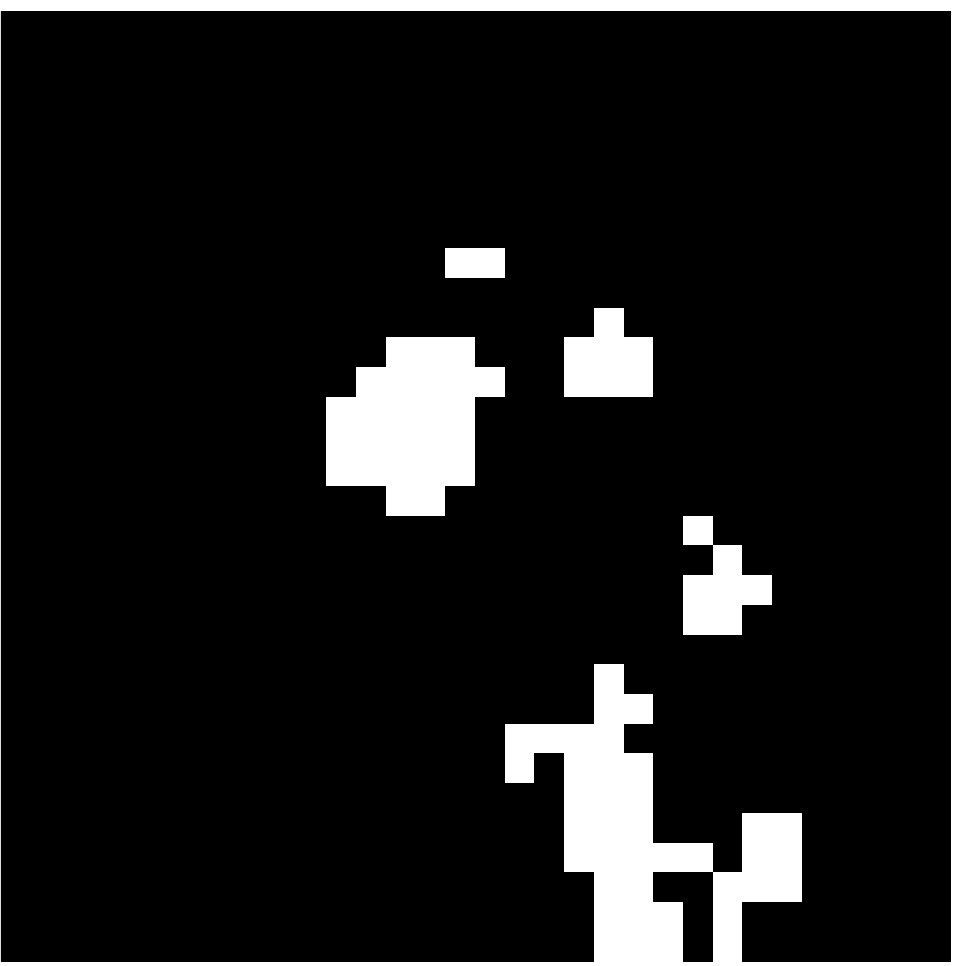} & 
\includegraphics[width=1.50cm]{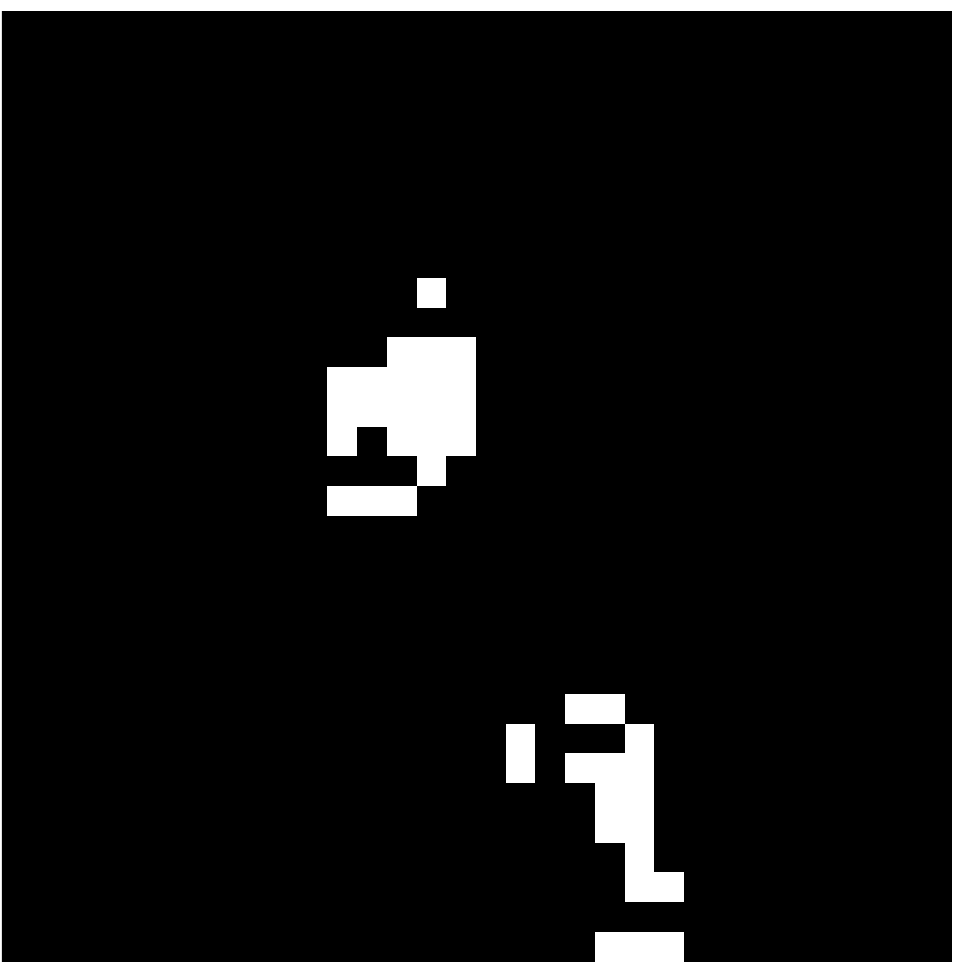}
\\
\bf{120} &
\includegraphics[width=1.50cm]{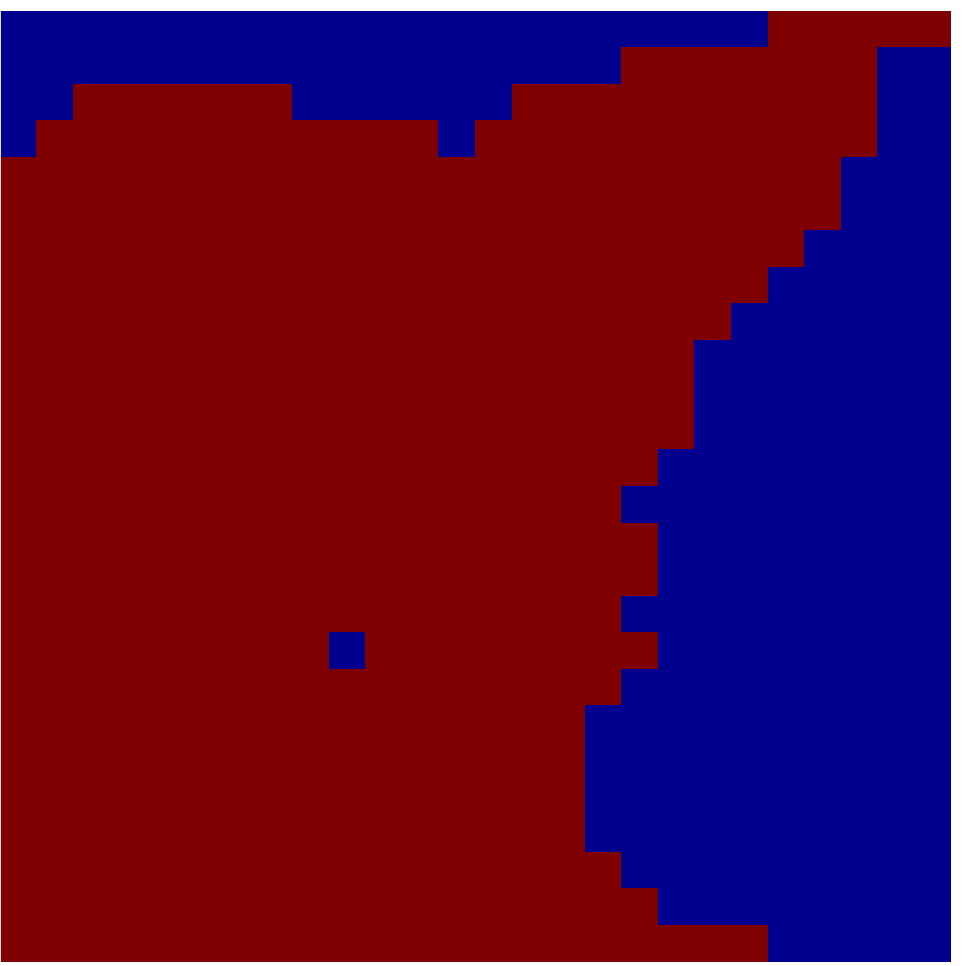} &
\includegraphics[width=1.50cm]{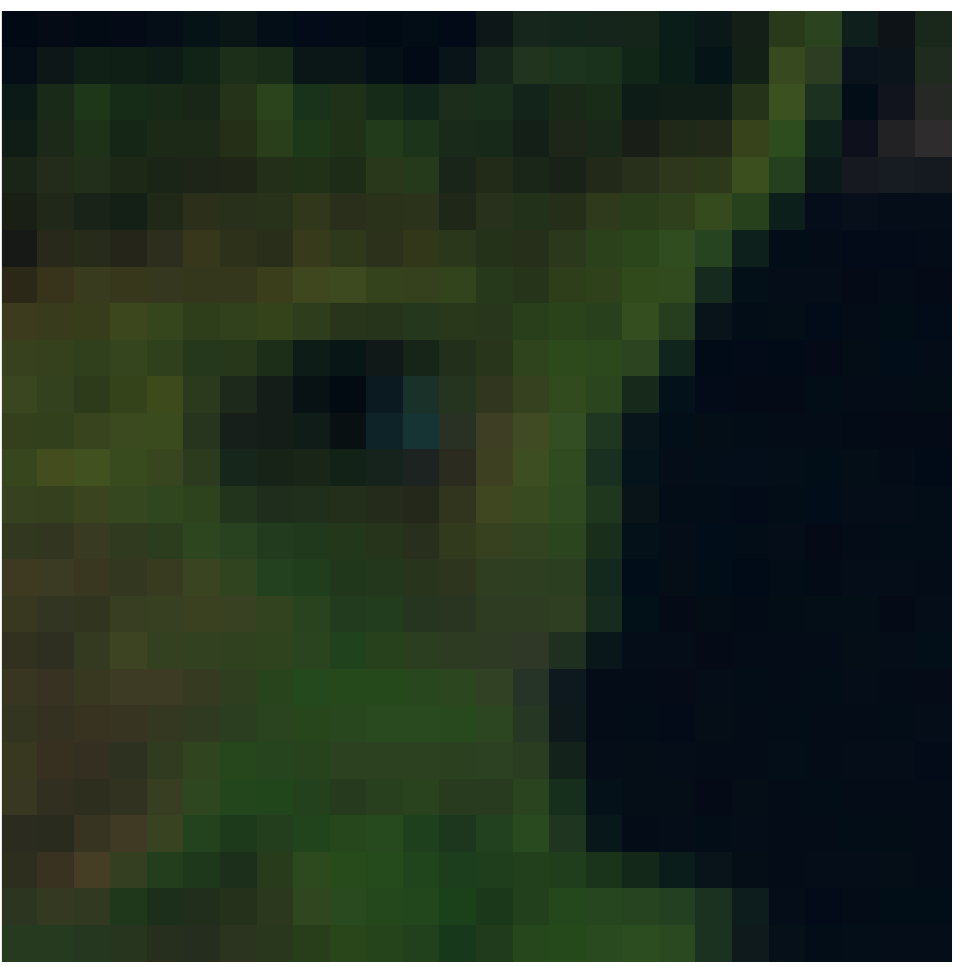} &
\includegraphics[width=1.50cm]{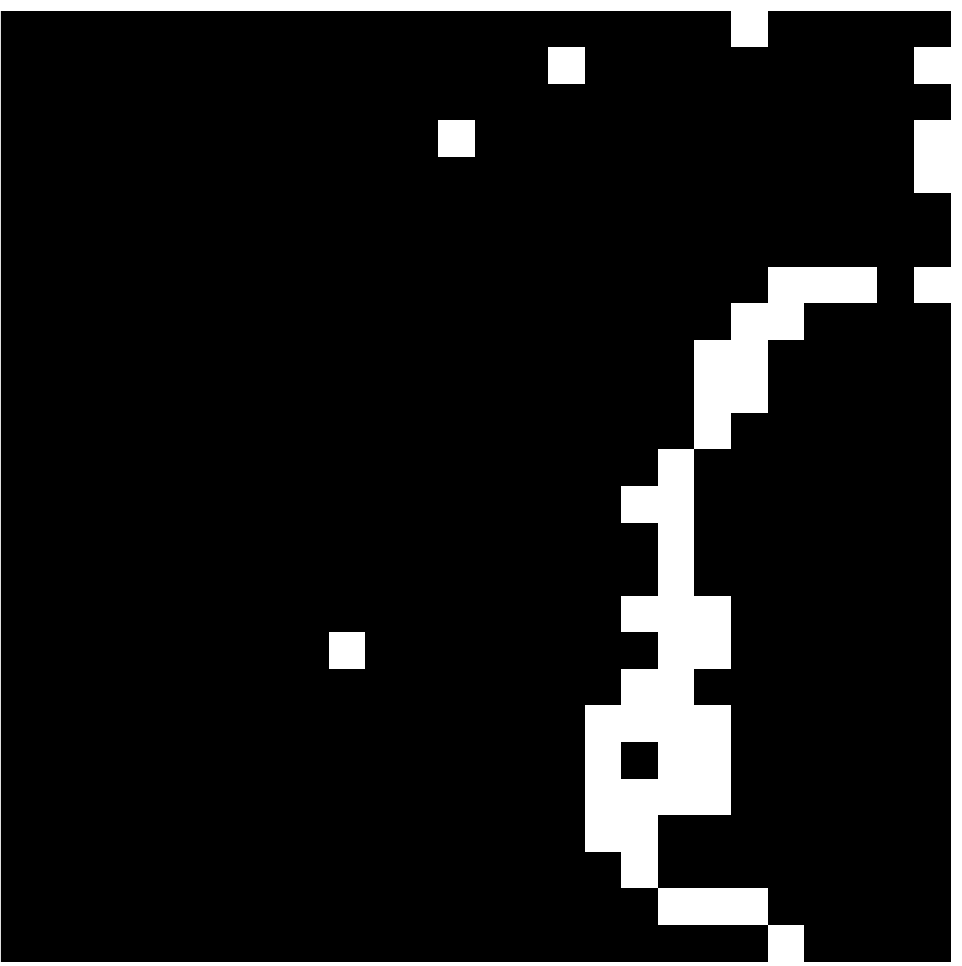} & 
\includegraphics[width=1.50cm]{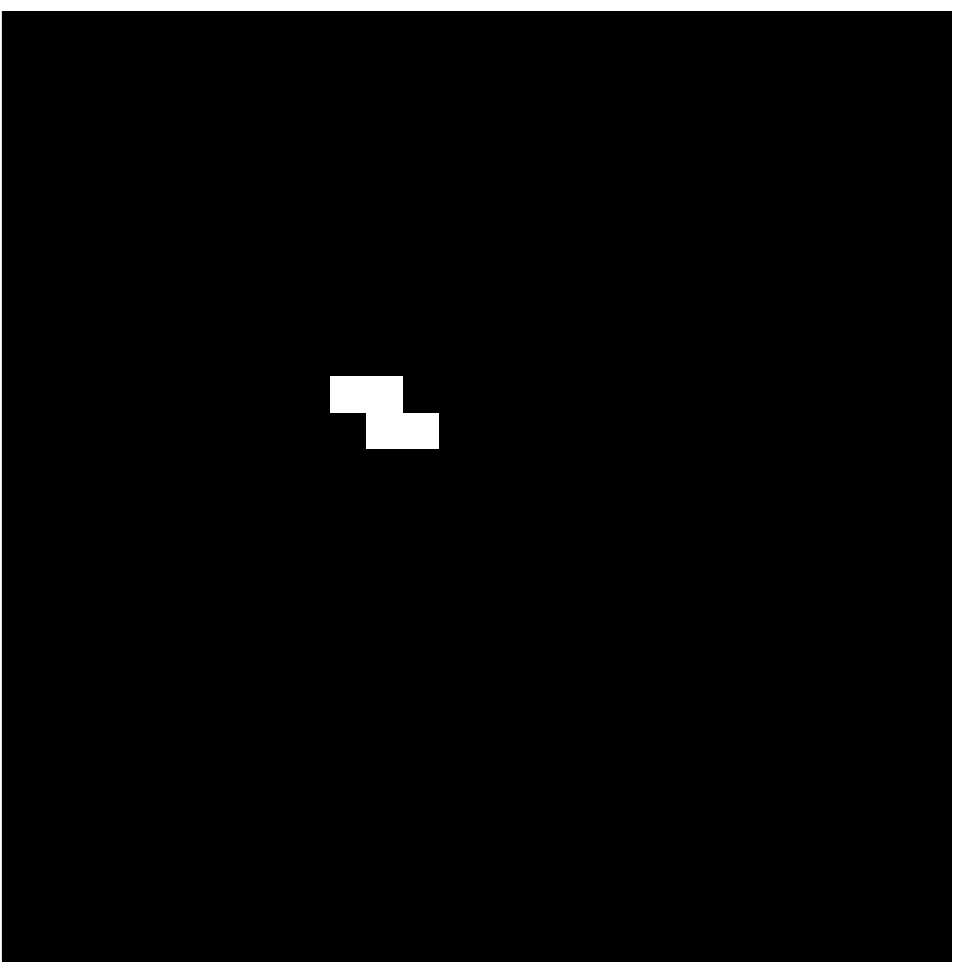}
\\
\bf{131} &
\includegraphics[width=1.50cm]{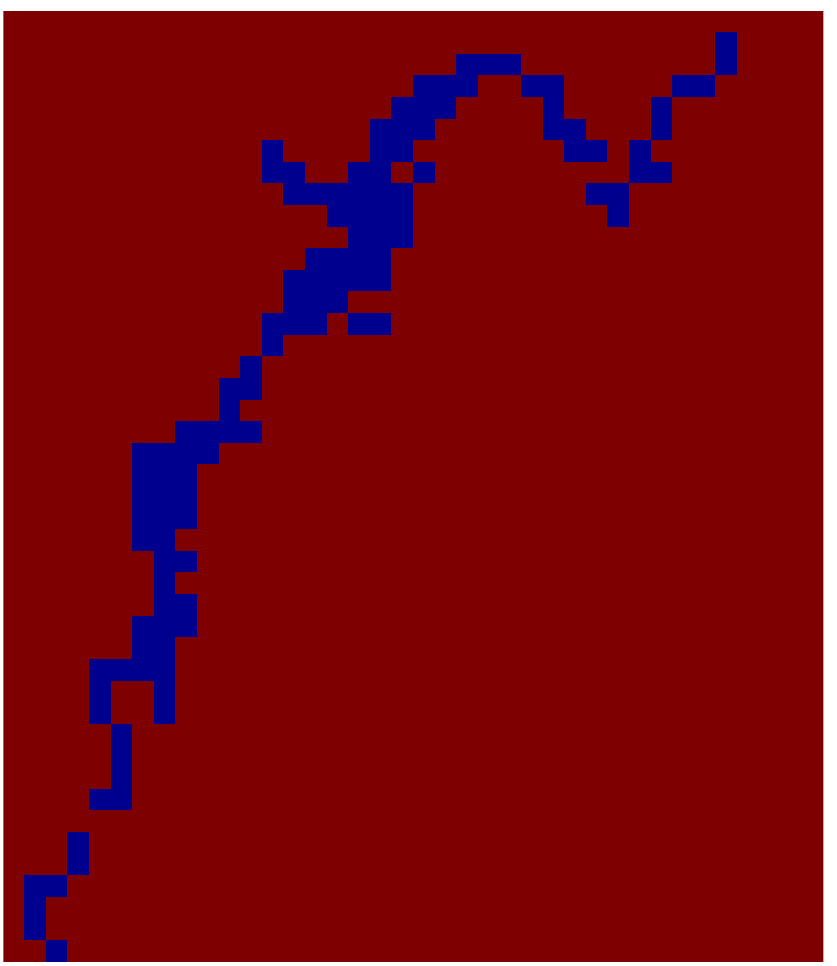} &
\includegraphics[width=1.50cm]{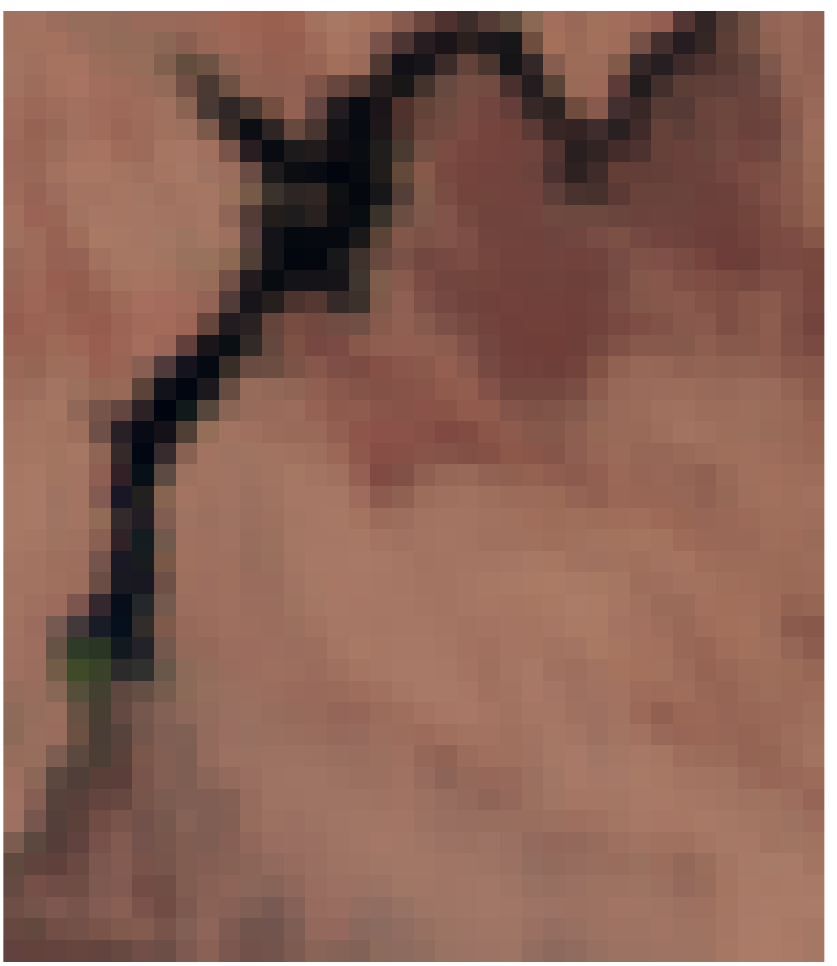} &
\includegraphics[width=1.50cm]{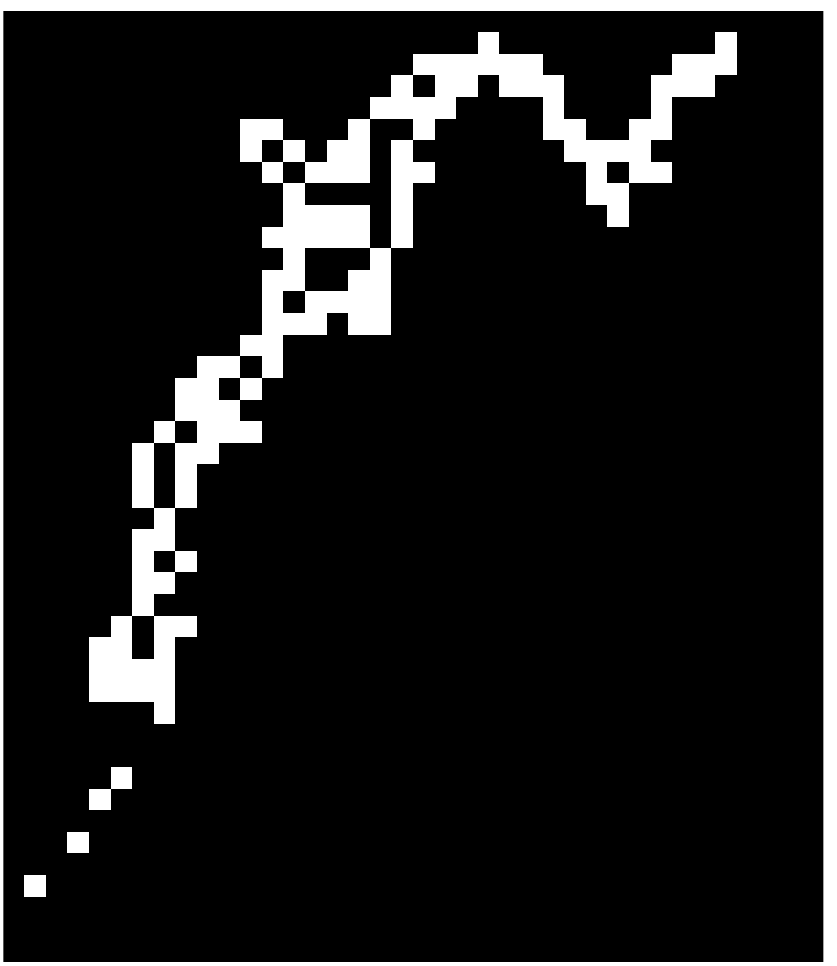} & 
\includegraphics[width=1.50cm]{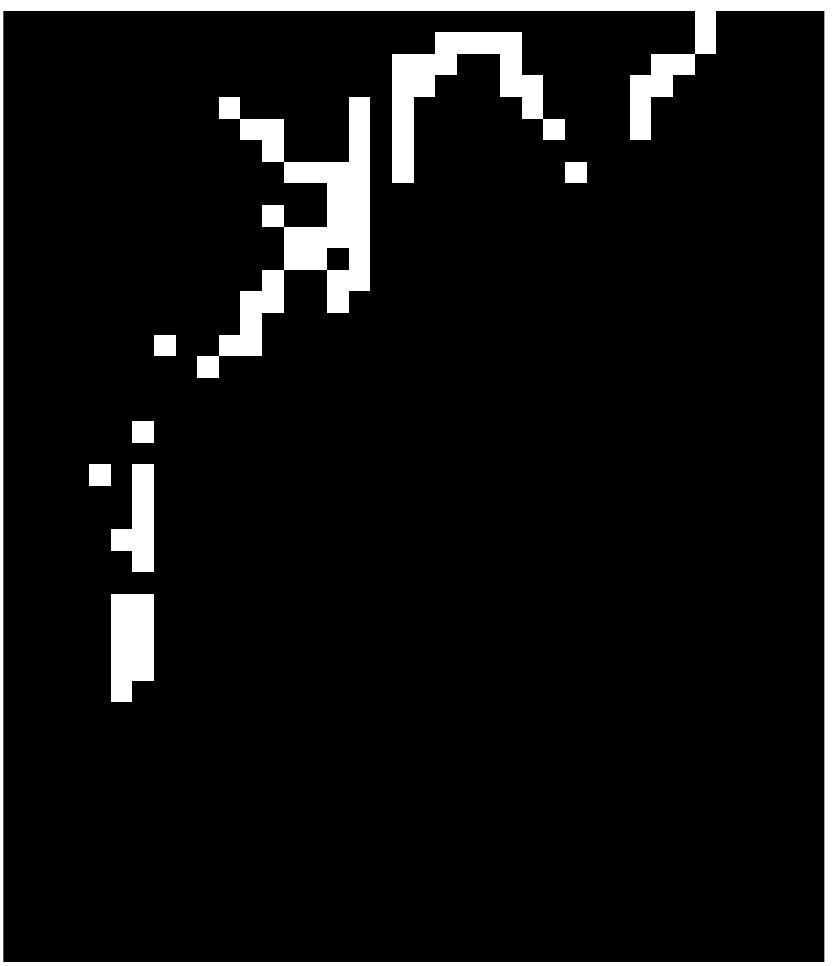}
\vspace{0.2cm} 
\\
\bf{154} &
\includegraphics[width=1.50cm]{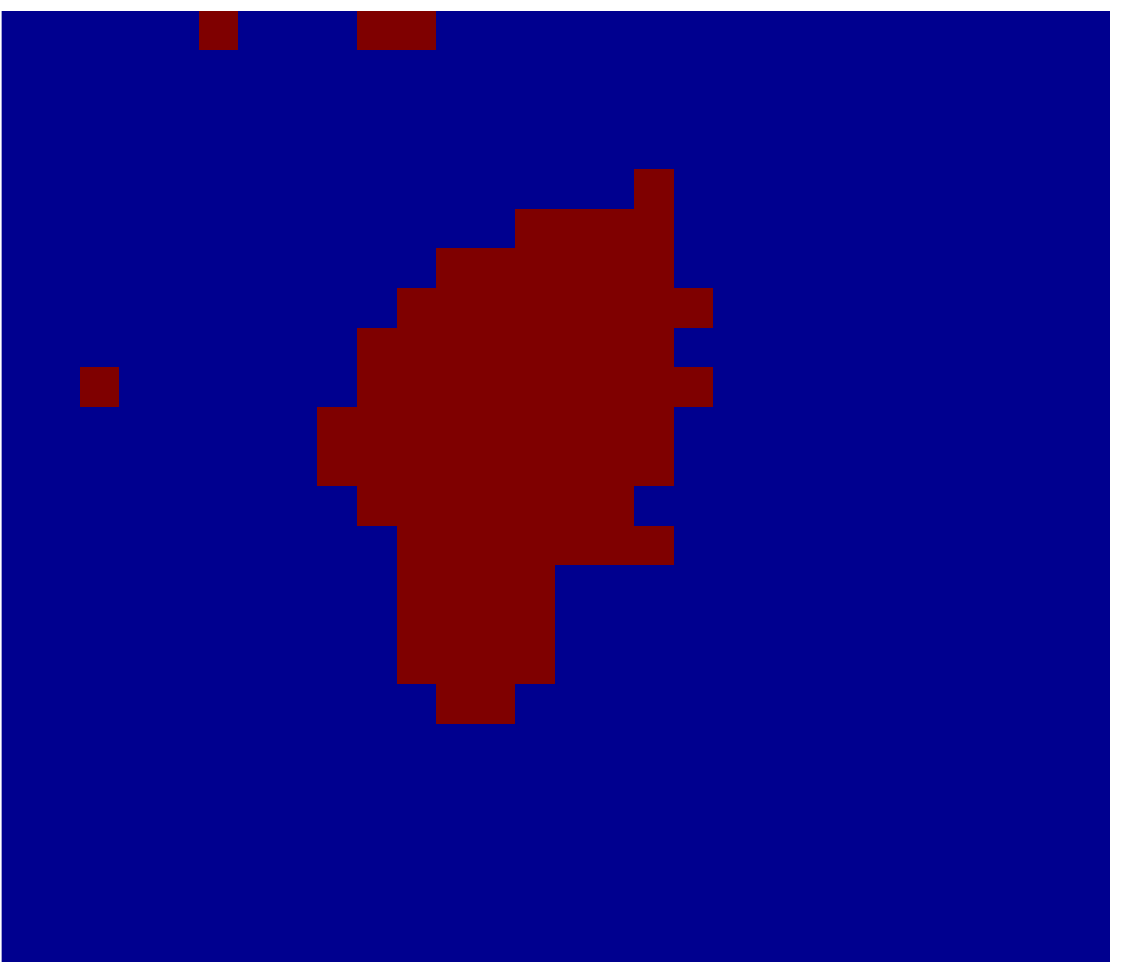} &
\includegraphics[trim=0 0 0 0.8cm, clip, width=1.50cm]{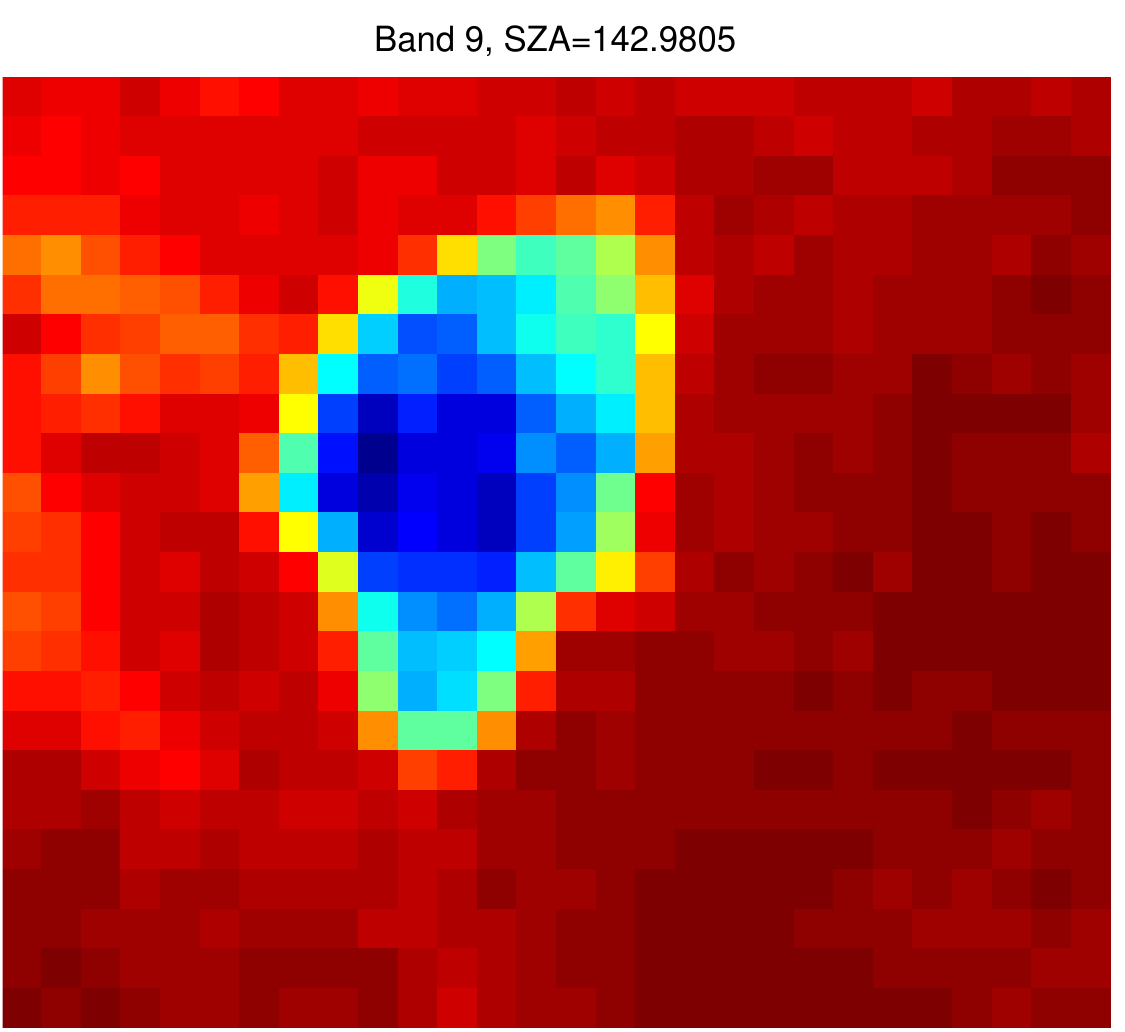} &
\includegraphics[width=1.50cm]{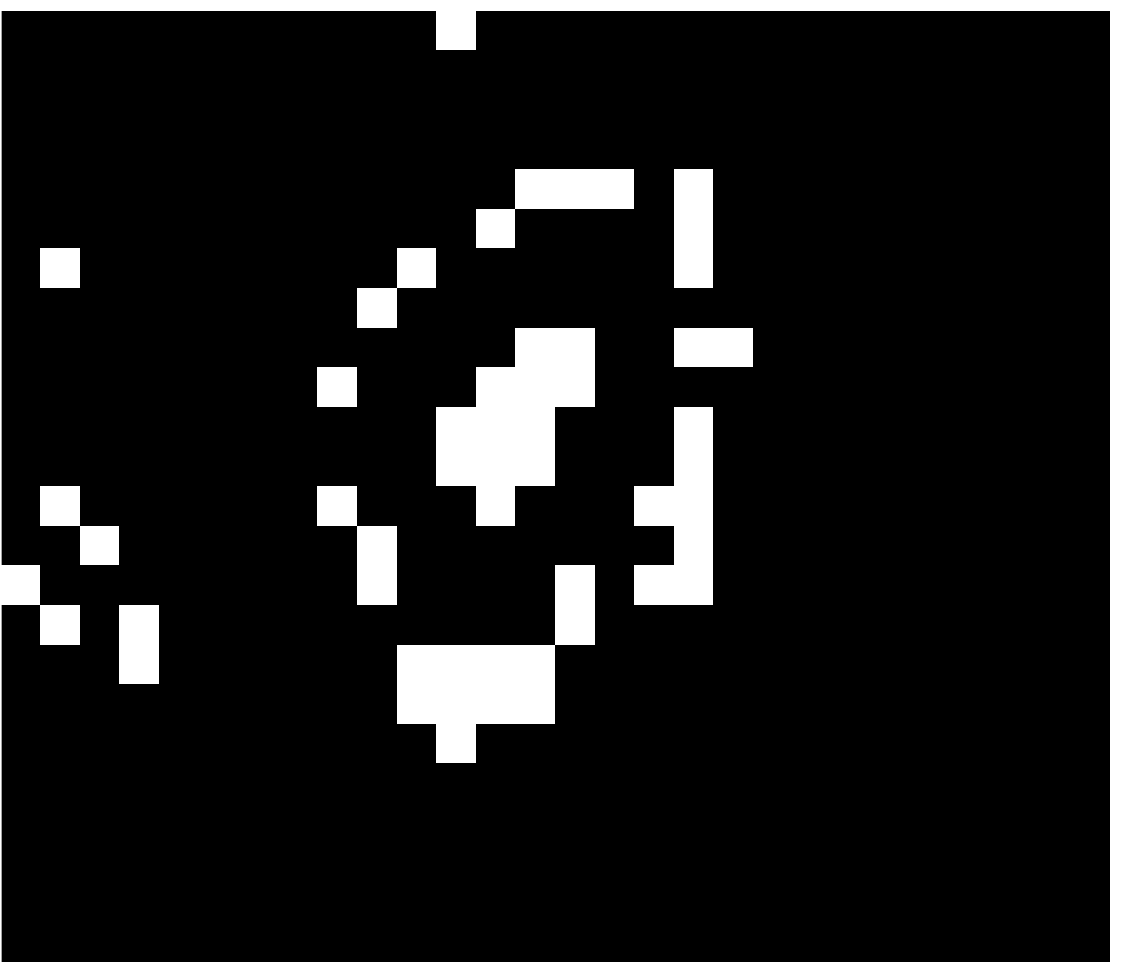} & 
\includegraphics[width=1.50cm]{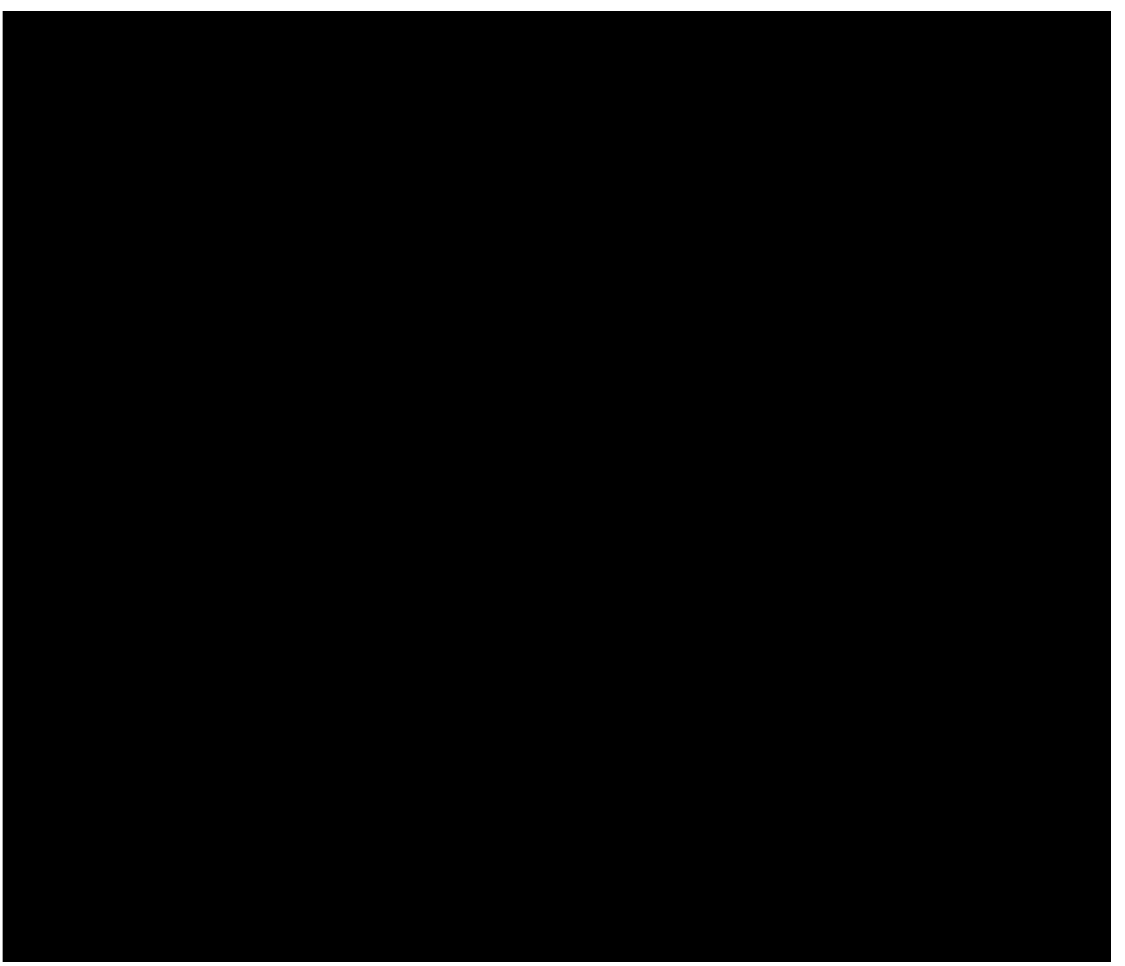}
\vspace{0.2cm} 
\\
\bf{177} &
\includegraphics[width=1.50cm]{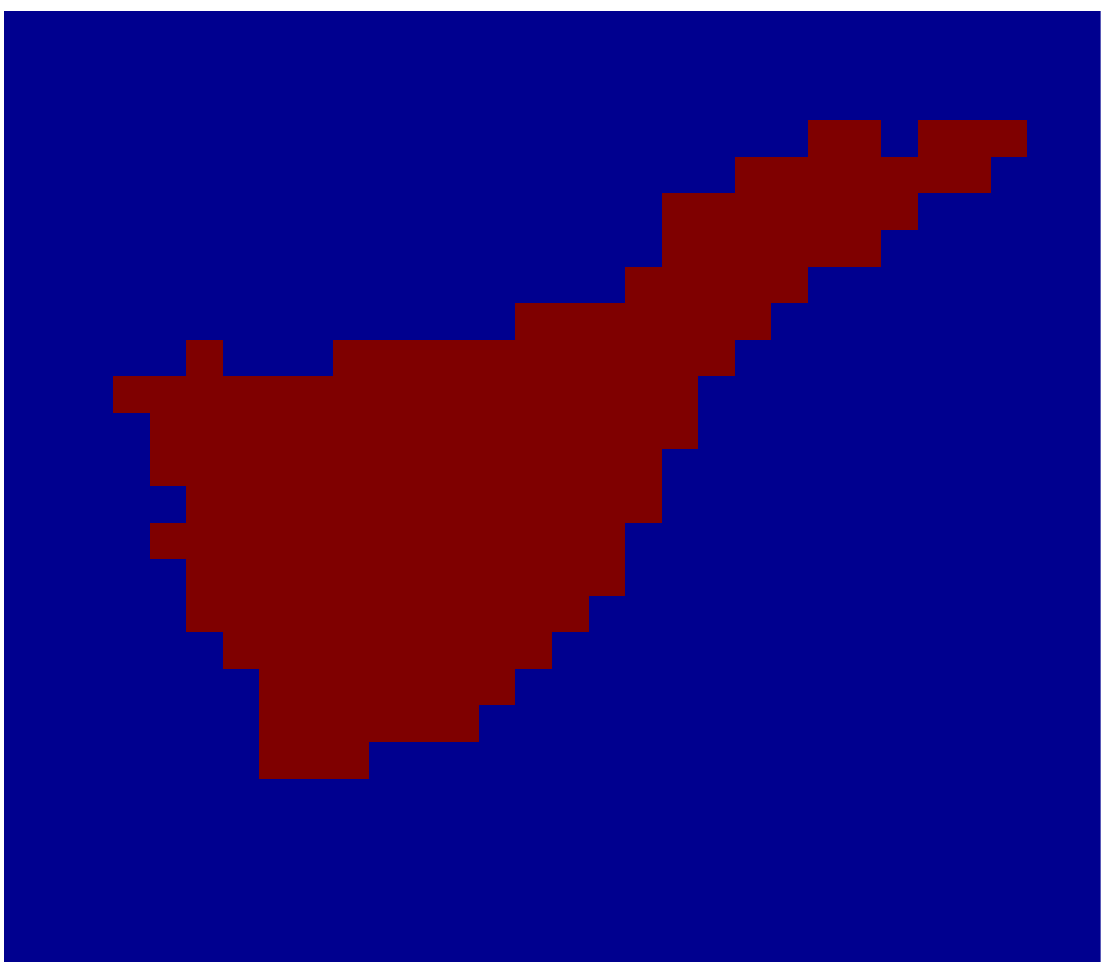} &
\includegraphics[trim=0 0 0 0.8cm, clip, width=1.50cm]{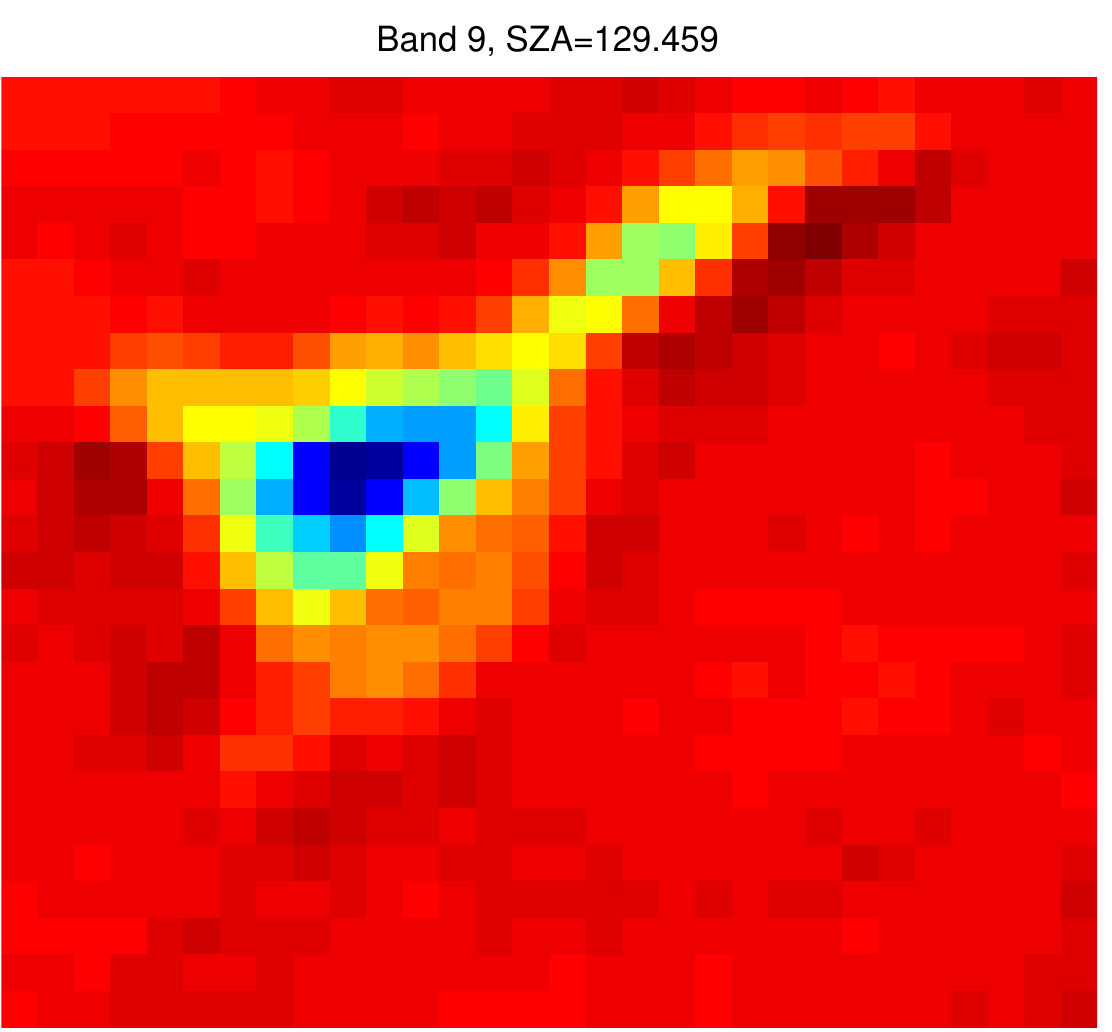} &
\includegraphics[width=1.50cm]{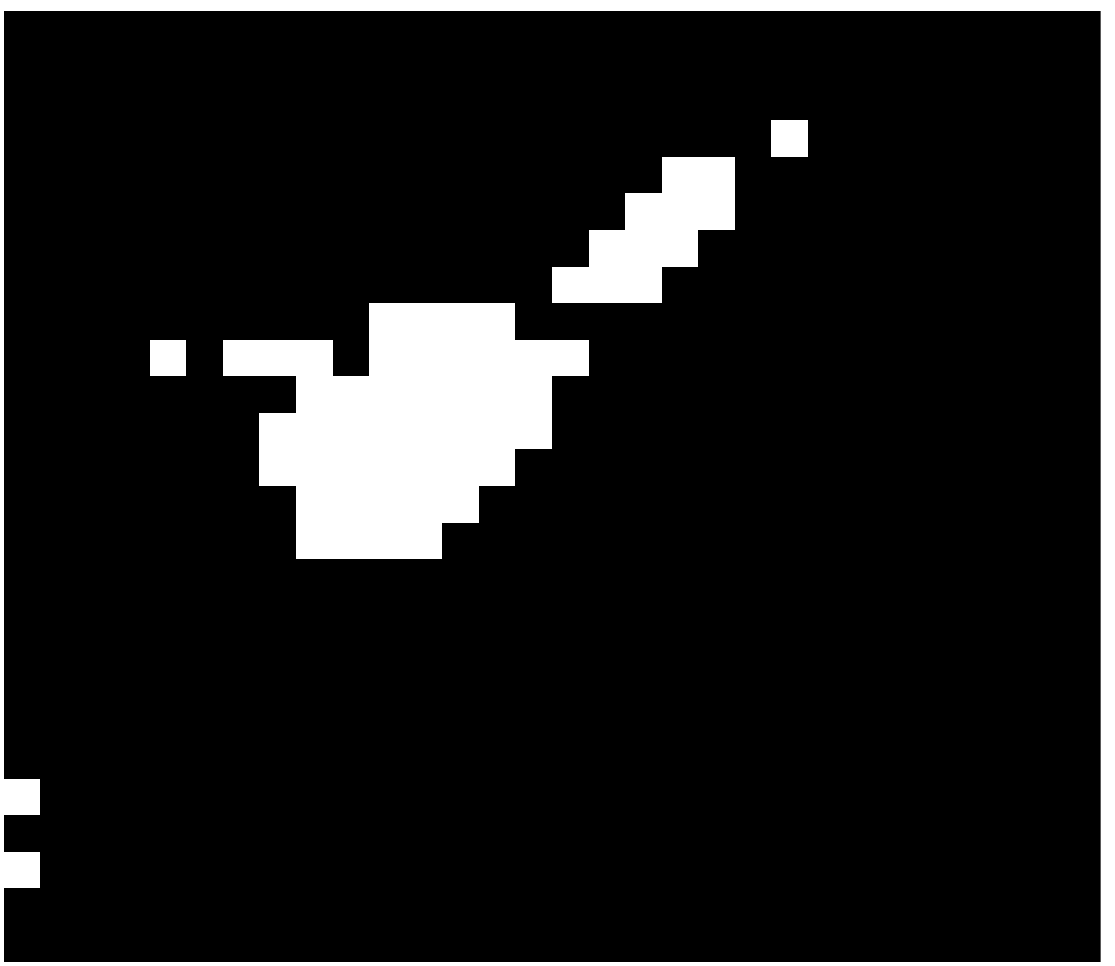} & 
\includegraphics[width=1.50cm]{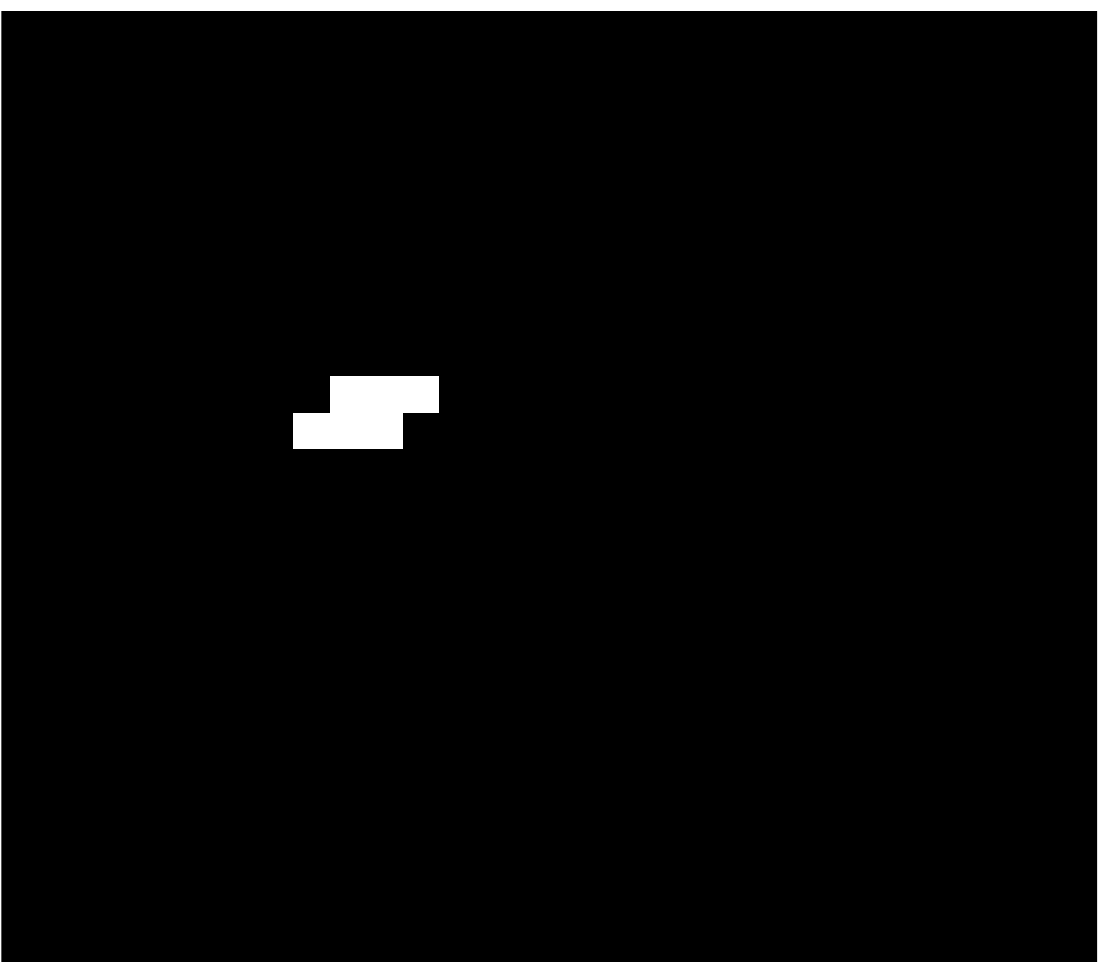}
\\
\bf{190} &
\includegraphics[width=1.50cm]{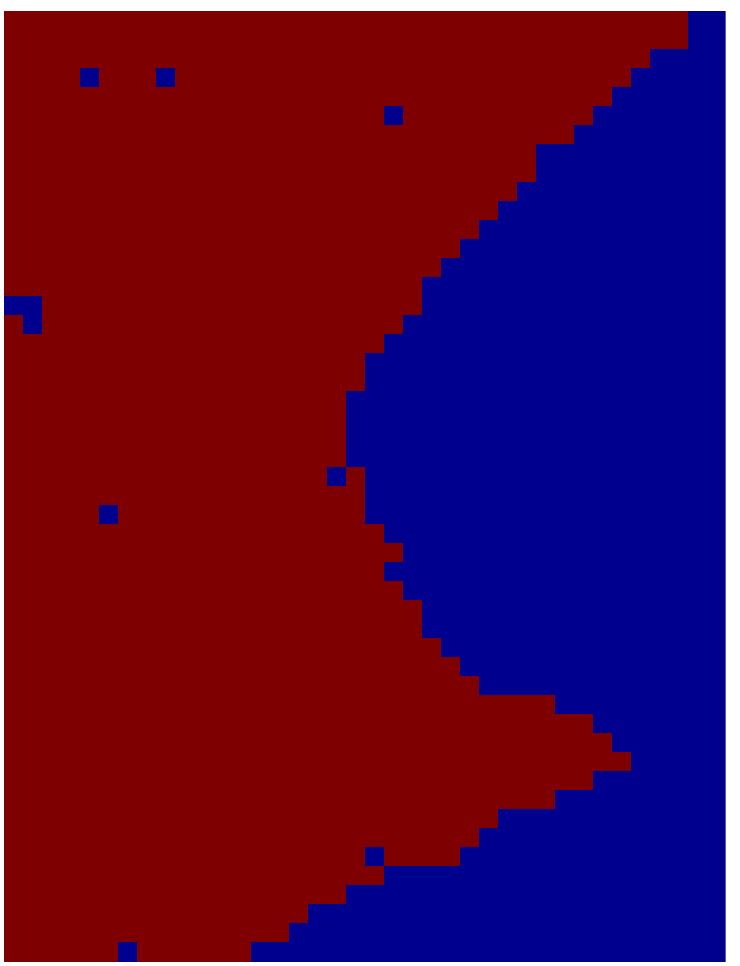} &
\includegraphics[width=1.50cm]{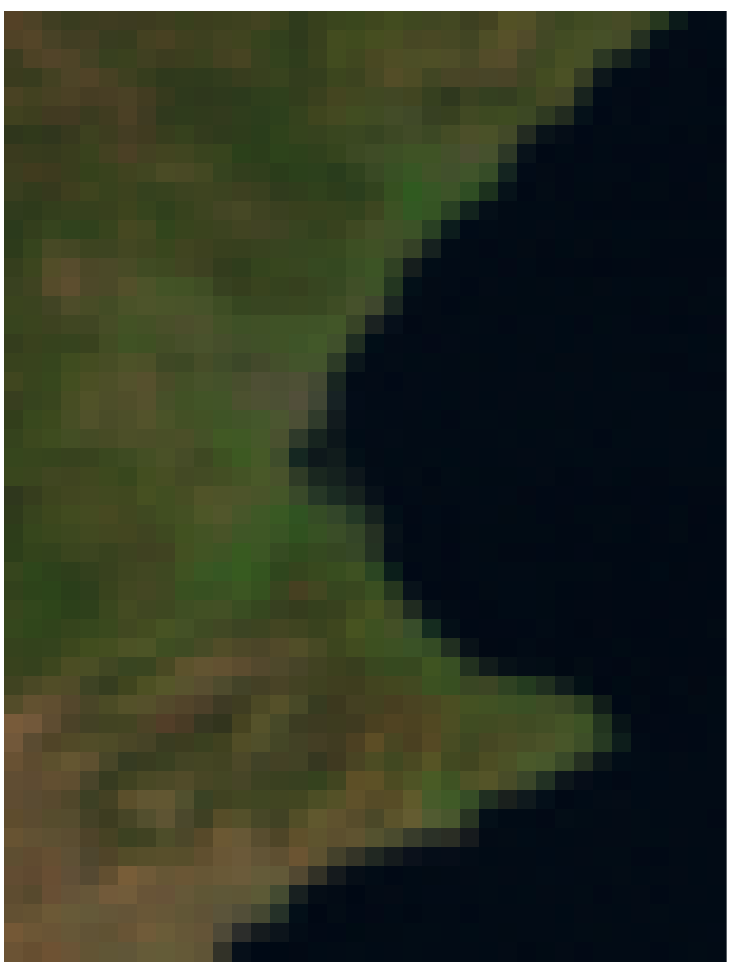} &
\includegraphics[width=1.50cm]{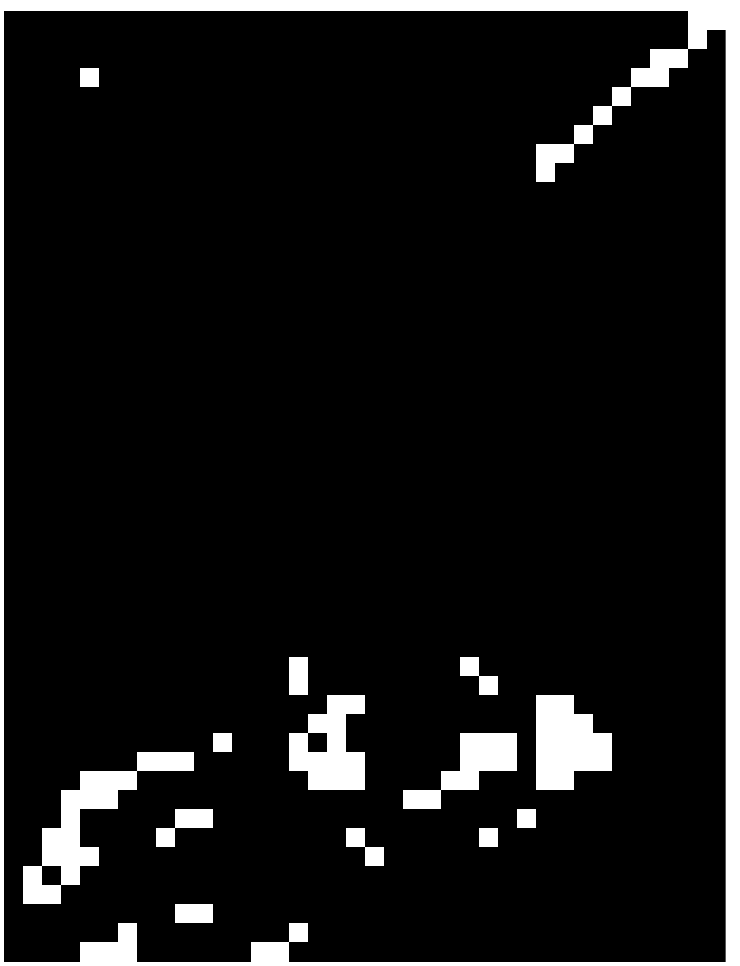} & 
\includegraphics[width=1.50cm]{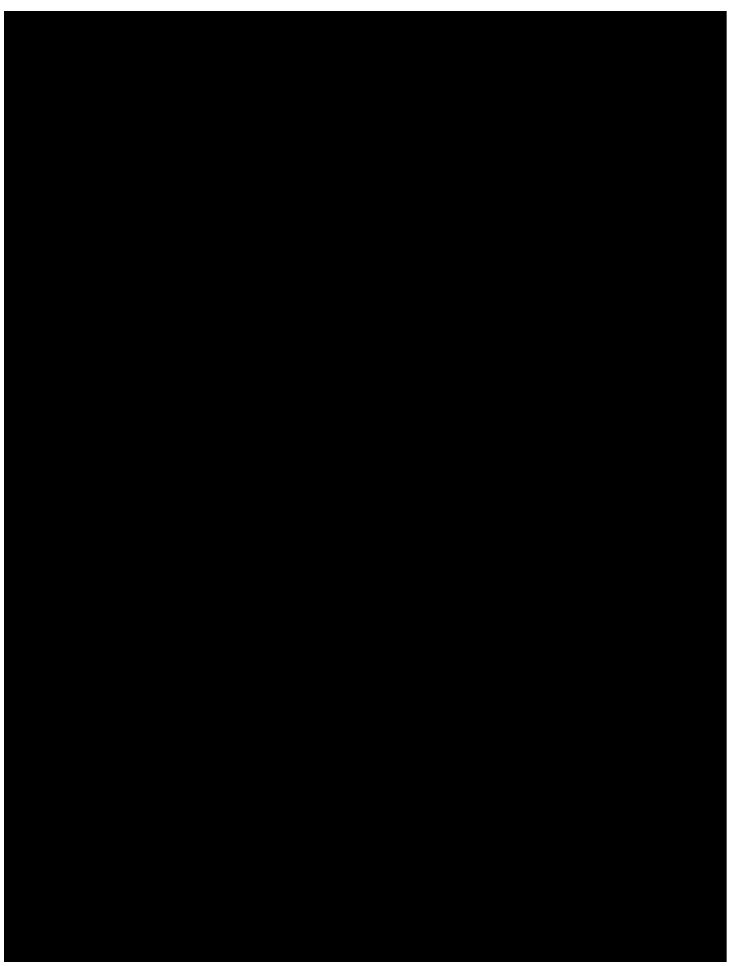}
\\
\end{tabular}
\end{center}
\vspace{-0.2cm}
\caption{Classification maps for particular chips with low cloud coverage, potentially over coastline. Mask colour code: `cloud-free' in black and `cloudy' in white.}\label{fig:predCoast190}
\end{figure}

Here we analyze the issue of consistency of the L2 mask, paying special attention to the performance of the classification scheme in the case of very low rate of clouds over coastlines. 
Figure~\ref{fig:predCoast190} shows the land cover type, the RGB composite (or band 9 for night acquisitions), the L2 cloud masks used as ground truth (best available proxy), and the predictions from the proposed scheme based on SVMs.   
We can essentially conclude that the L2 cloud mask contains clear mistakes, especially over coastlines (as was observed before). This can be clearly noted by comparing the L2 mask with the corresponding RGB (or band 9 at night). See for example results for LMs 0, 14, 48, 63, 83, 120, 131, 154, 177 and 190 where there are no clouds in the scenes.

In some of the previous cases, the SVM classifiers generally commit less errors (e.g. LMs 0, 17, 63, 83, 120, 131, 154, 177 and 190), thus confirming the suitability of the proposed approach.
In other cases, the SVM scheme commits larger misclassifications (e.g. LMs 14, 48 and 131), yet probably due to the fact that SVMs are learning the L2 mask errors in these LMs.
Furthermore, note that the SVM classifier shows less false positives over islands or isolated land masses in some cases (e.g. 120, 154, and 177). \blue{Particularly, the few pixels detected wrongly as clouds on the predicted mask agree with the volcano peaks (LMs 120 and 177).}

\subsection{Evaluation of global results}

\begin{figure}[t]
\centerline{\includegraphics[width=0.5\textwidth]{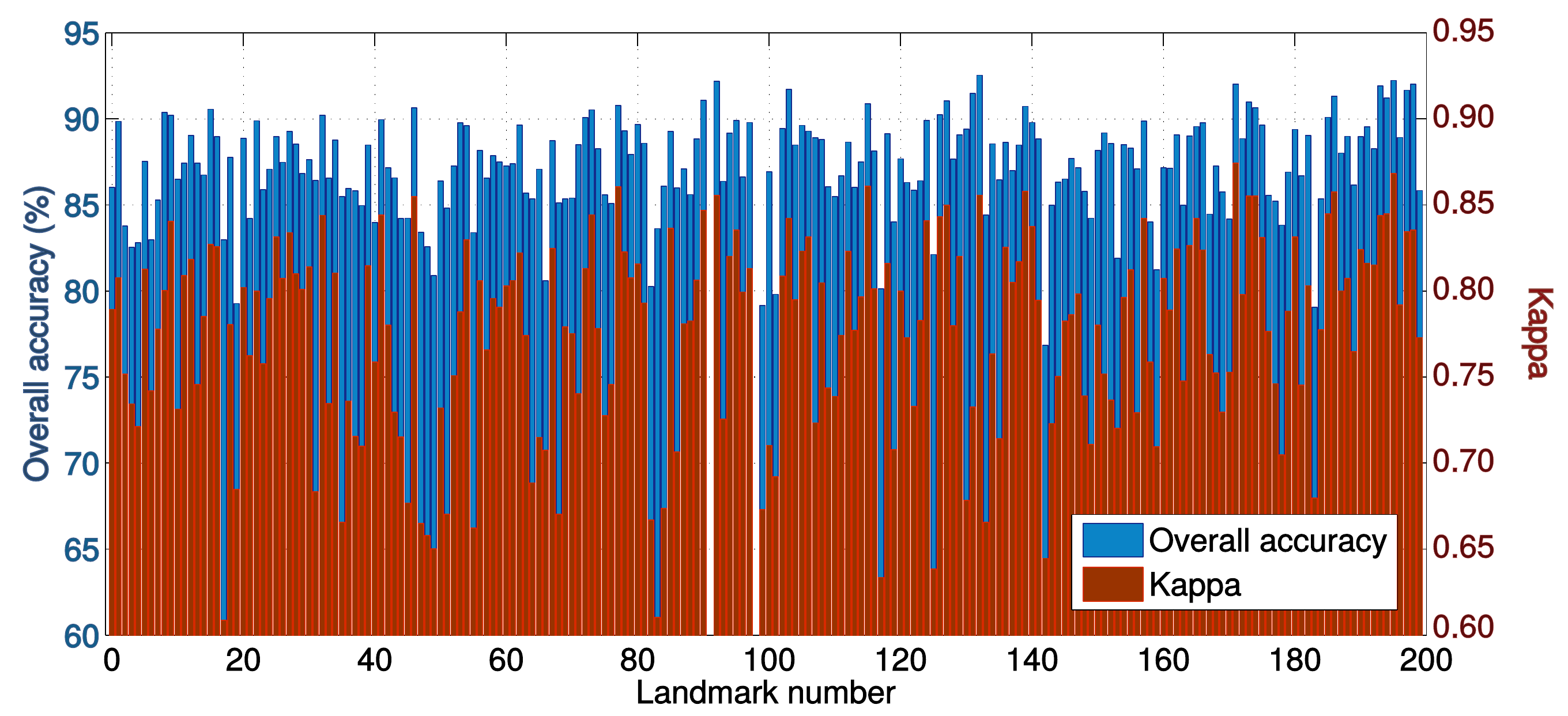}}
\caption{Overall accuracy and Kappa over the 200 landmarks (LM) of the database. Landmarks LM91 and LM98 were excluded of the analysis due to wrong labels on the L2 cloud mask.\label{fig:global_analysis}}
\end{figure}

\begin{figure}[t]
\centerline{\includegraphics[width=0.25\textwidth]{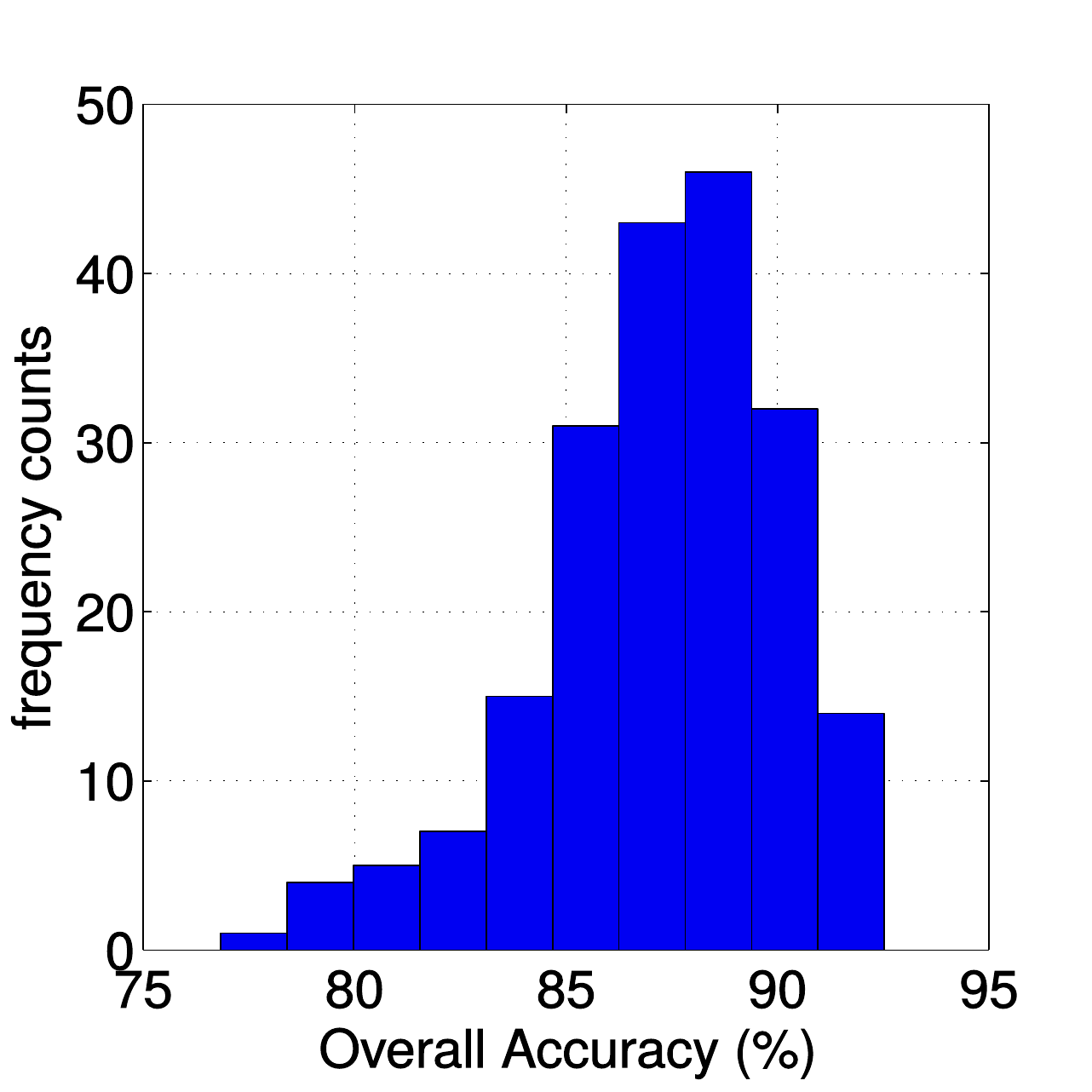}\includegraphics[width=0.25\textwidth]{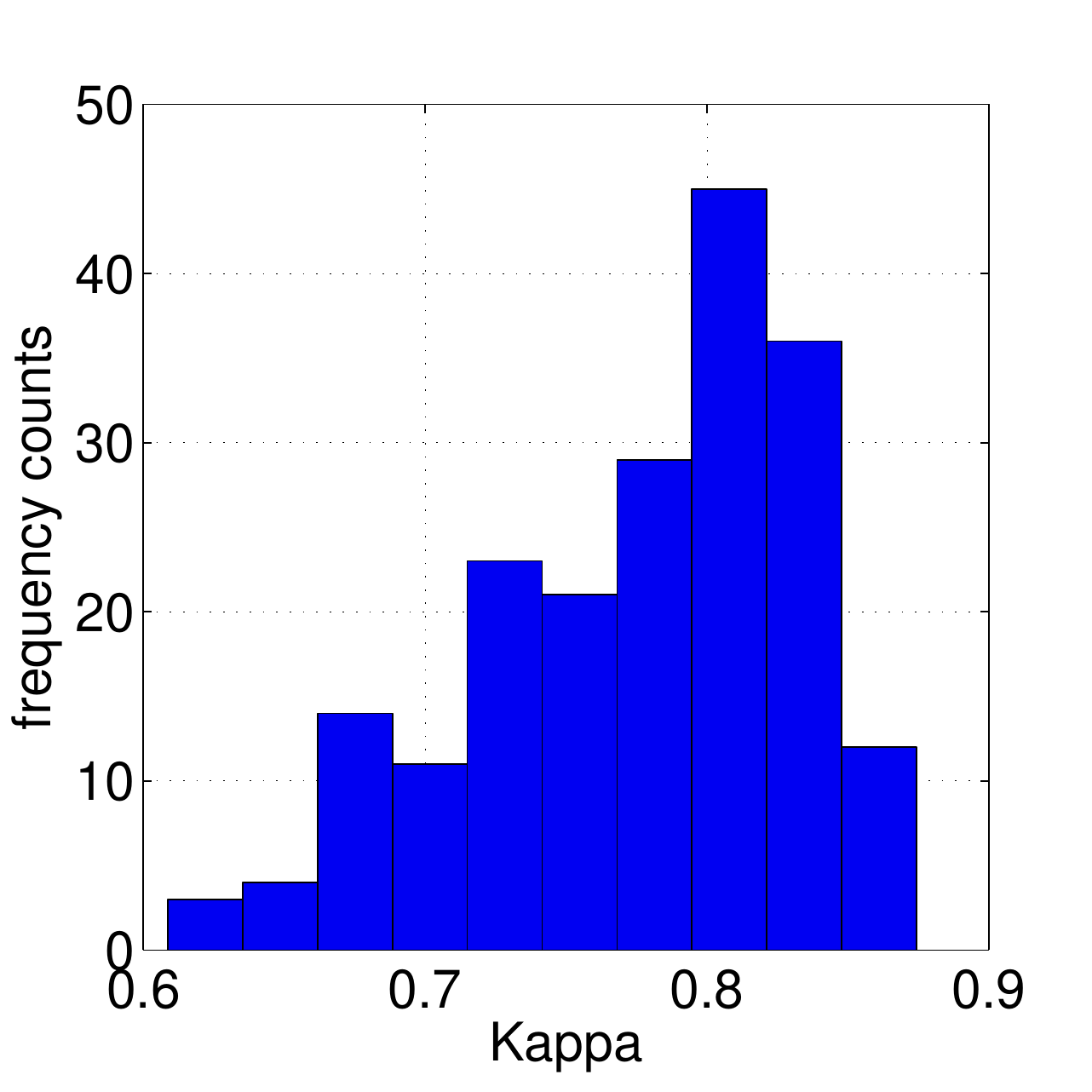}}
\caption{Left: overall accuracy (\%), and right: Kappa statistic. These plots are the histograms of the global classification results over the 200 landmarks (LM) of the database. Landmarks LM91 and LM98 were excluded of the analysis due to wrong labels on the L2 cloud mask.
\label{fig:global_analysis_hist}}
\end{figure}

The proposed classification processing chain for cloud detection was applied to the whole landmarks database. Internally, the proposed scheme is built on four classification models that are trained for each landmark depending on the illumination conditions. In order to obtain new predictions, the system loads a chip image, extracts the required features, and, depending on the SZA for the landmark, applies one of the four trained SVM classifiers for this landmark. 
\blue{Global classification results for the 200 landmarks are shown in Fig.~\ref{fig:global_analysis}. We can observe we obtained more than 85\% OA in more than 82\% of the landmarks, and approximately 70\% of them exceeded 0.75 of Kappa's statistic. 
Fig.~\ref{fig:global_analysis_hist} shows the histograms of the global classification results over these 200 landmarks. As we can see, the designed system has obtained good results in both OA and Kappa statistic. As we have mentioned, results are concentrated around $87\%$ of global accuracy and 0.8 of Kappa.}

\section{Conclusions}\label{sec:conclusions}


In this paper we introduced a general scheme for the identification of cloudy pixels over landmarks in MSG SEVIRI images, which is based on a pattern recognition scheme especially taylored to the problem. This automatic machine learning scheme achieves an improved performance in landmark recognition and matching, which are critical steps in Image Navigation and Registration (INR) models, as well as to maintain the geometric quality assessment (GQA) in the instrument data processing.

The proposed methodology is based on the combination of dedicated SVMs for particular landmarks and illumination conditions. This divide and conquer strategy is revealed as highly efficient to tackle this large scale problem with millions of MSG multispectral image chips. The results were analyzed quantitatively (in terms of detection accuracy) and qualitatively by visual inspection of the predicted cloud masks.  
The scheme can be actually extended in several ways. First, one may be interested in detecting cloudy chips rather than individual predictions per pixel, so we intend to pursue this strategy in the future as a complementary module for the pixel-based cloudy detector. The most evident advantage is the computational cost, which comes at the price of extracting relevant and discriminant features at chip level. Second, the module has to be improved for predictions over the coastlines. 
Last but not least, the recognition scheme can be made adaptive over time: this would avoid the need for retraining, and would adapt to seasonal and year changes. 
All these are matters of on-going research.


\bibliographystyle{IEEEtran}
\bibliography{landmarks,sensyf}
\begin{IEEEbiography}[{\includegraphics[width=1in,height=1.25in,keepaspectratio]{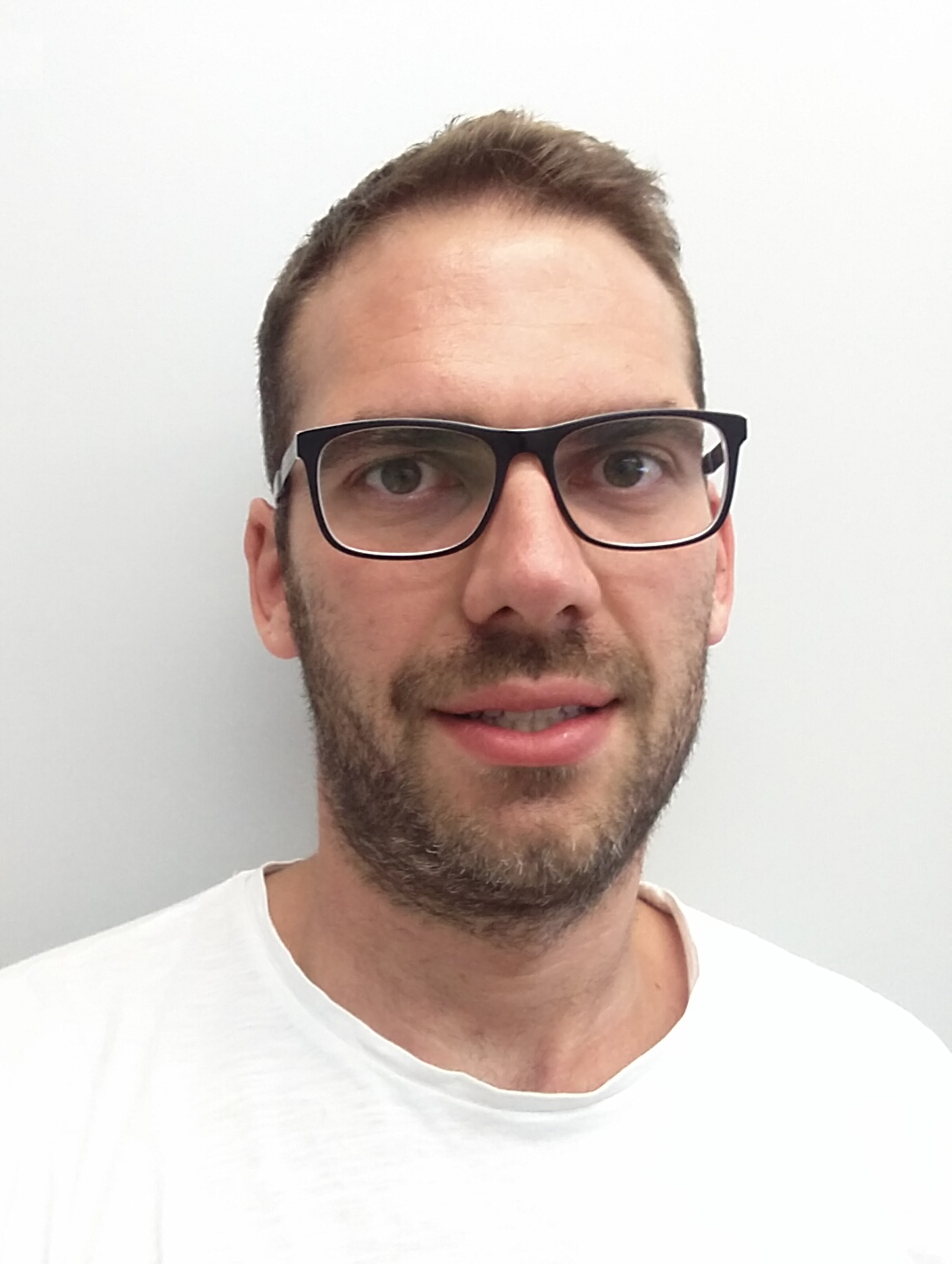}}]%
{Adri\'{a}n P\'{e}rez-Suay}
obtained his B.Sc. degree in Mathematics (2007), a Master degree in Advanced Computing and Intelligent Systems (2010) and a Ph.D. degree in Computational Mathematics and Computer Science (2015), all from the Universitat de Val\`encia. He is assistant professor in the Dept. of Mathematics in the Universitat de Val\`encia. He is currently a Postdoctoral Researcher at the Image and Signal Processing (ISP) working on dependence estimation, kernel methods and causal inference for remote sensing data analysis.
\end{IEEEbiography}
\begin{IEEEbiography}[{\includegraphics[width=1in,height=1.25in,clip,keepaspectratio]{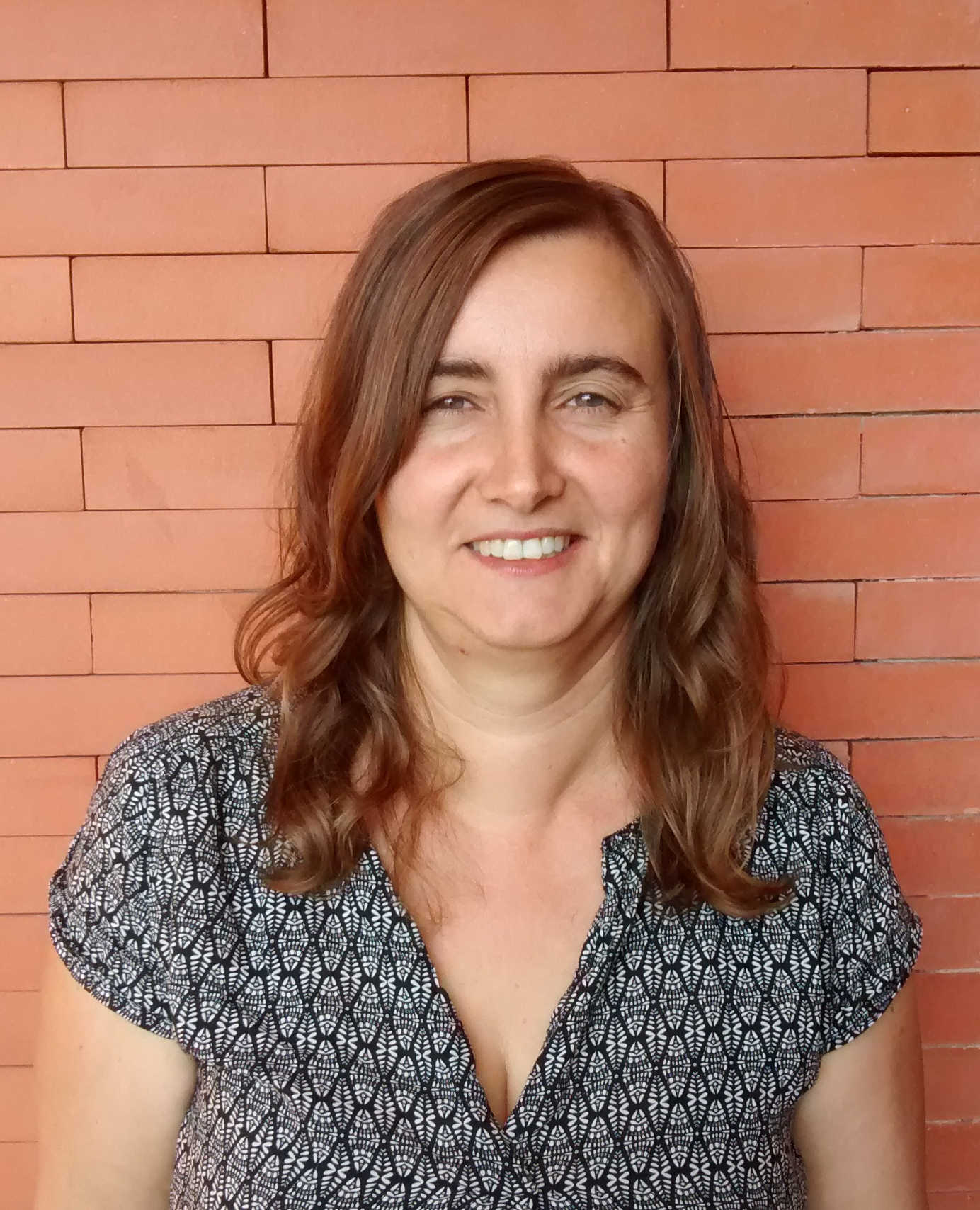}}]{Julia Amor\'os-L\'opez}
Julia Amor\'os L\'opez received the B.Sc. degree in Technical Engineering in Telecommunication and the M.Sc. degrees in Electronics Engineering from the Universitat de Val\`encia in 2000 and 2002, respectively. She is currently Associate Professor in the Department of Electronic Engineering at the Universitat de Val\`encia and researcher at the ISP group where her work is mainly related to machine learning and image processing in remote sensing applications.
\end{IEEEbiography} 
\begin{IEEEbiography}[{\includegraphics[width=1in,height=1.25in,clip,keepaspectratio]{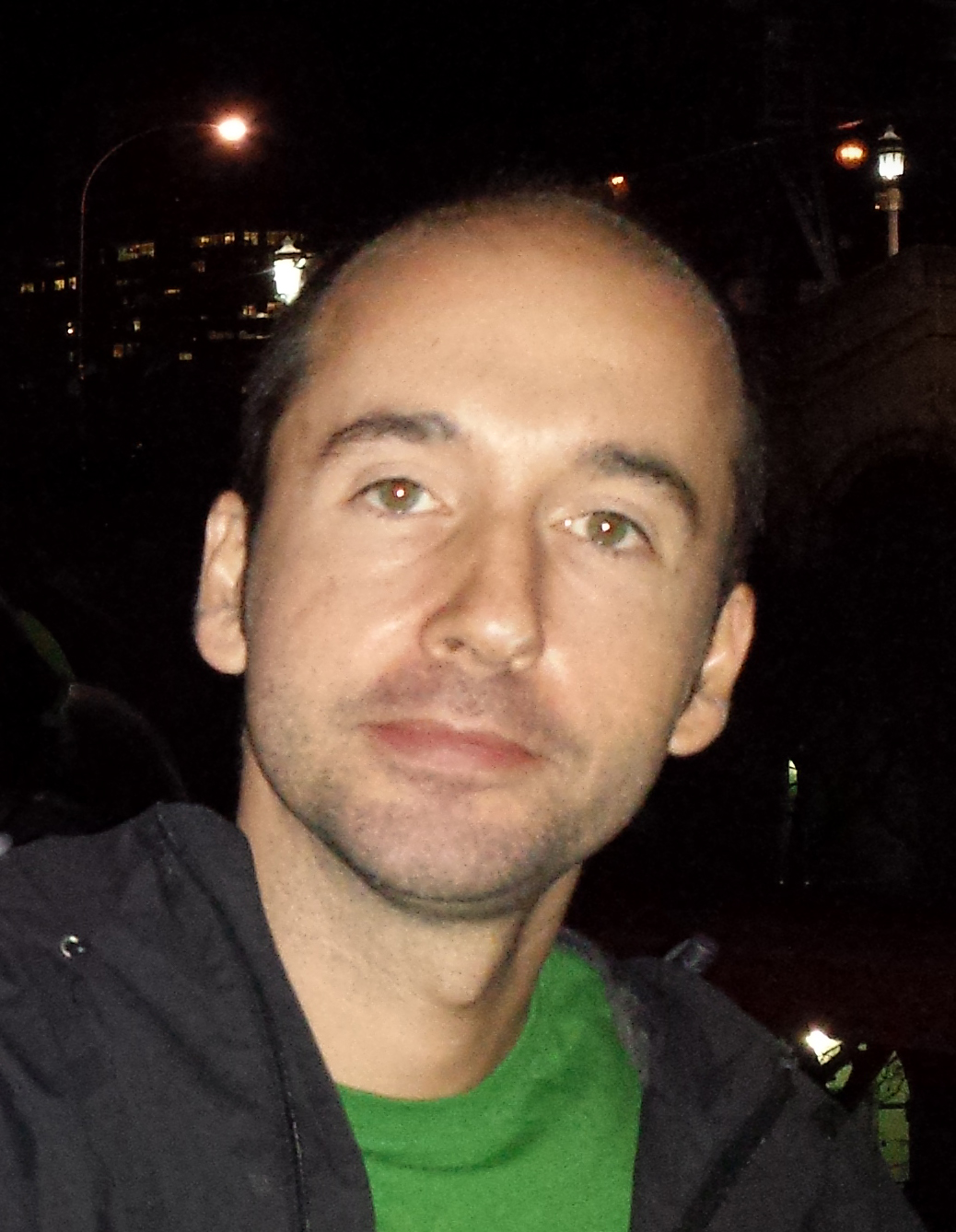}}]{Luis G{\'o}mez-Chova}
\textbf{Luis G{\'o}mez-Chova} (S'08-M'09-SM'15) received his PhD degree in electronics engineering from the Universitat de Val\`encia, Spain, in 2008. He is currently Full Professor in the Department of Electronic Engineering at the Universitat de Val\`encia. He is also a researcher at the Image and Signal Processing (ISP) Group where his work is mainly related to pattern recognition and machine learning applied to remote sensing multispectral images and cloud screening (h-index 30). 
\end{IEEEbiography} 
\begin{IEEEbiography}[{\includegraphics[width=1in,height=1.25in,clip,keepaspectratio]{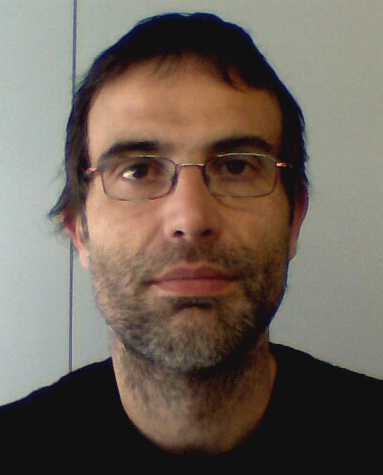}}]{Jordi Mu\~noz-Mar\'i}
Jordi Mu\~noz-Mar\'i was born in Val\`encia, Spain in 1970, and received a B.Sc. degree in Physics (1993), a B.Sc. degree in Electronics Engineering (1996), and a Ph.D. degree in Electronics Engineering (2003) from the Universitat de Val\`encia. He is currently an Associate Professor in the Electronics Engineering Department at the Universitat de Val\`encia, where he teaches Electronic Circuits and Digital Signal Processing. He also teaches the subjects of Machine Learning, Active
Learning and Big Data in a Data Science Master. He is a research member of the Image and Signal Processing (ISP) group. His research activity is tied to the study and development of machine learning algorithms for signal, image processing and remote sensing. 
\end{IEEEbiography} 
\begin{IEEEbiography}[{\includegraphics[width=1in,height=1.25in,clip,keepaspectratio]{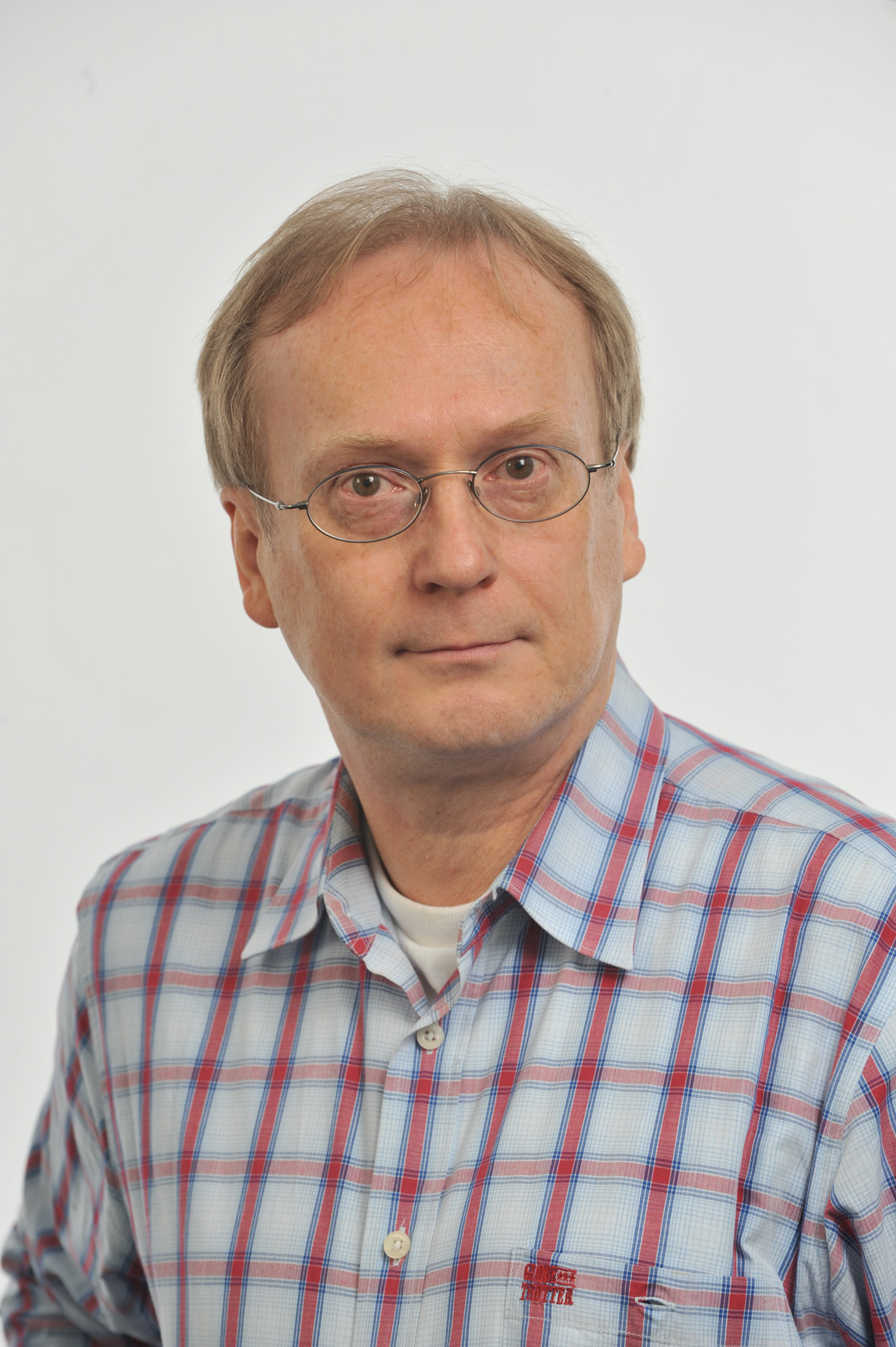}}]{Dieter Just}Dieter Just received a Ph.D. degree in physics from
the University of Essen, Germany, in 1989. In 1988, he joined the IBM T. J. Watson
Research Center, Yorktown Heights, NY, where he participated in research on 
nonlinear optics, image processing, and neural networks. Between 1990 and 1995, 
he was with the German Remote Sensing Data Center, German Aerospace Center (DLR), 
Oberpfaffenhofen, where he worked on SAR algorithms and software development for synthetic
aperture radar, in particular, for the SIR-C/X-SAR mission. Since 1995, he has
been with the European Organization for the Exploitation of Meteorological
Satellites (EUMETSAT), Darmstadt, Germany. At EUMETSAT he has supported the
development of the Meteosat Second and Third Generation Programmes, as well as  the
first and second generation of the  EUMETSAT Polar System. Currently, he
is the competence area manager for Image Navigation \& Registration, 
and Calibration at EUMETSAT. Dr. Just is a member of IEEE.
\end{IEEEbiography}
\begin{IEEEbiography}[{\includegraphics[width=1in,height=1.25in,clip,keepaspectratio]{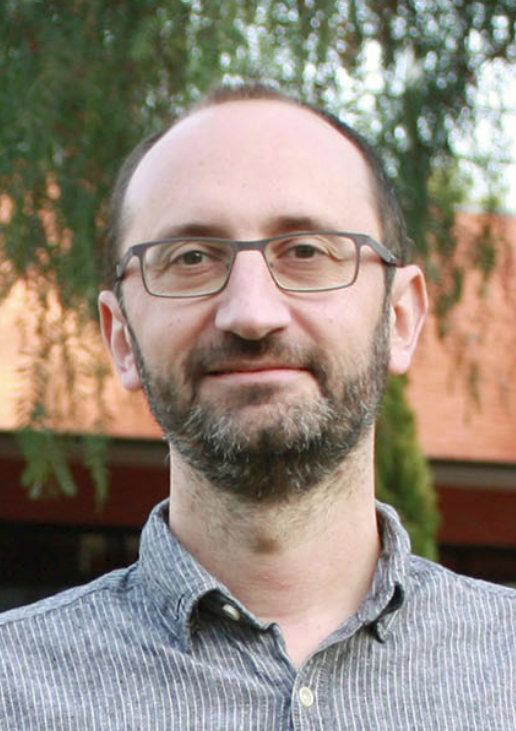}}]{Gustau Camps-Valls}
(M'04, SM'07, FM'18) received a PhD in Physics in 2002 from the Universitat de Val\`encia, and he is currently Full professor in Electrical Engineering, and coordinator of the Image and Signal Processing (ISP) group in the same university, \url{http://isp.uv.es}. He is interested in the development of machine learning algorithms for geoscience and remote sensing data analysis.
\end{IEEEbiography}

\end{document}